\documentclass[10pt]{report}

\usepackage[table]{xcolor}
\usepackage{mathrsfs,amssymb,amsmath}
\usepackage{graphicx}
\usepackage{hyperref} 
\usepackage{multirow} 
\usepackage{xspace}
\usepackage{booktabs}
\usepackage{enumitem} 
\usepackage{subcaption} 
\usepackage{longtable}

\hypersetup{colorlinks,linkcolor={red!50!black},citecolor={blue!50!black},urlcolor={blue!80!black}}

\usepackage[toc,page]{appendix}
\usepackage[top=1.5in, bottom=1.5in, left=1.41in, right=1.41in]{geometry}
\usepackage{titlesec}
\newcommand{\chapnumfont}{\usefont{T1}{pnc}{b}{n}\fontsize{100}{100}\selectfont}
\colorlet{chapnumcol}{gray!75}  
\titleformat{\chapter}[display]{\filleft\bfseries}{\filleft\chapnumfont\textcolor{chapnumcol}{\thechapter}}{-24pt}{\Huge}
\setlist{topsep=2.2pt,itemsep=0.5pt} 
\usepackage{etoolbox}
\makeatletter
\patchcmd{\ttlh@hang}{\parindent\z@}{\parindent\z@\leavevmode}{}{}
\patchcmd{\ttlh@hang}{\noindent}{}{}{}
\makeatother

\usepackage{tikz}
\usetikzlibrary{shapes,calc,positioning,automata,arrows,trees}
\usepackage[tikz]{bclogo}
\renewcommand\logowidth{15pt}
\newcommand\bcpen{\includegraphics[width=\logowidth]{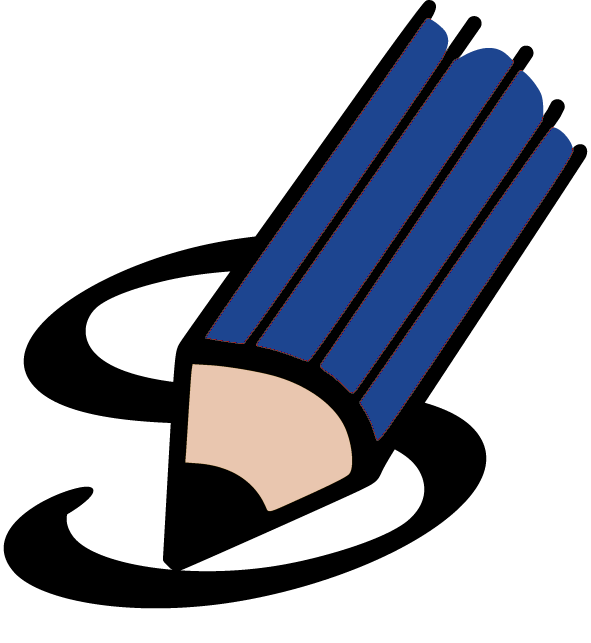}} 
\newcommand\bcdico{\includegraphics[width=\logowidth]{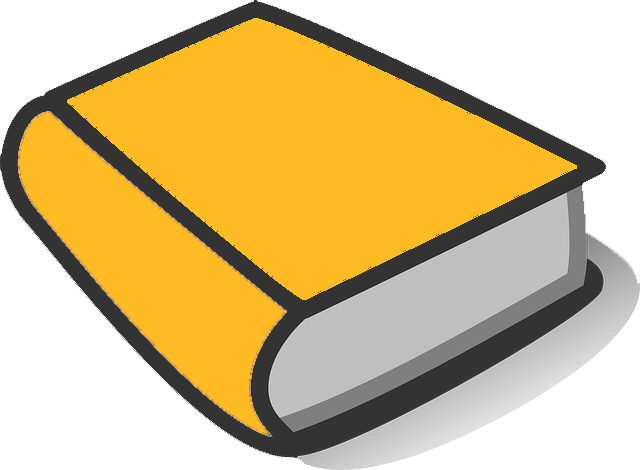}} 
\newcommand\bcroue{\includegraphics[width=\logowidth]{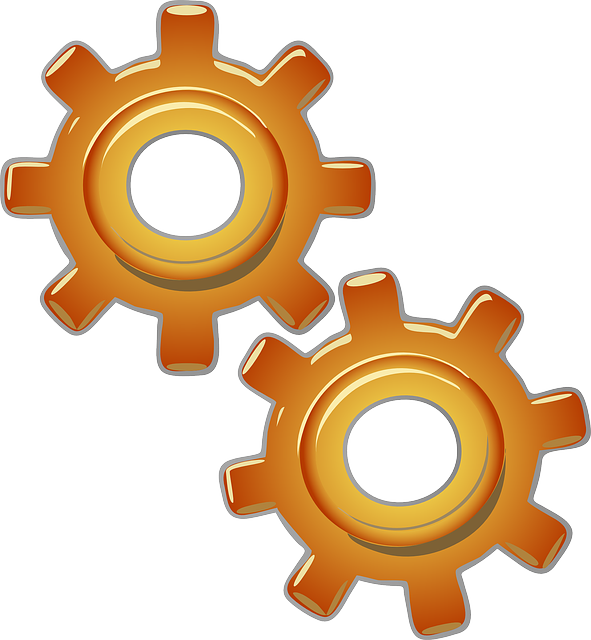}} 
\usepackage[skins,breakable,xparse]{tcolorbox}
\tikzset{
  dirtree/.style={
    grow via three points={one child at (0.8,-0.7) and two children at (0.8,-0.7) and (0.8,-1.45)}, 
    edge from parent path={($(\tikzparentnode\tikzparentanchor)+(.4cm,0cm)$) |- (\tikzchildnode\tikzchildanchor)}, growth parent anchor=west, parent anchor=south west},
}

\usepackage{listingsutf8}

\lstdefinelanguage{syntax}{
  keywords={realExpr,rankType,orderedType,iaBaseRelation,paBaseRelation,rcc8BaseRelation,measureType,combinationType,blockType,decimal,branchingType,restartsType,consistencyType,filteringType,varhType,valhType,intCtr,integer,unsignedInteger,constraint,metaConstraint,intExpr,boolExpr,boolExprSet,boolExprReal,wspace,dimensions,realVar,realVal,qualVar,graphVar,node,varType,setVar,setVal,frameworkType,proba,intIntvl,realIntvl,number,intVar,intVal,intValEx,symbol,operator,operand,boolean,identifier,state,restrictionList},
  keywordstyle=\color{dblue}\slshape,
  basewidth  = {.5em,0.5em},
  escapechar=@,
  xleftmargin=1pt,xrightmargin=1pt,
  breaklines=true,basicstyle=\ttfamily\linespread{1.0}\small,backgroundcolor=\color{colorsy},inputencoding=utf8/latin9,texcl
}

\lstdefinelanguage{semantics}{
  keywords={identifier}, 
  basewidth  = {.5em,0.5em},
  escapechar=@,
  xleftmargin=1pt,xrightmargin=1pt,
  breaklines=true,basicstyle=\ttfamily\linespread{1.20}\small,backgroundcolor=\color{colorse},inputencoding=utf8/latin9,texcl,mathescape
}

\lstdefinelanguage{xcsp}{
  keywords={region,instance,variables,literals,var,domain,constraints,objectives,annotations,annotation,array,extension,intension, tuples, supports,conflicts,args,quantification,fuzzy,relation,interval,point,agents,agent,allDifferent,count,ordered,vars,ctrs,comm,stages,decision, stochastic,required,possible,nodes,edges,arcs,minimize,maximize,linear,operator,value,allEqual,among,atLeast,atMost,exactly,limit,cumulative,circuit,regular,mdd,hamming,channel,element,values,transitions,origins,durations,indexes,permutation,coeffs,nValues,clause,cube,instantiation,sort,lex,precedence,cardPath,slide,slidingAmong,slidingSum,sequence,gsc,seqbin,binPacking,maximum,minimum, maximumArg, minimumArg, hardTemplate,softTemplate,capacities,loads,intervals,costMatrix,knapsack,allDisjoint,overlap,exists,forall,aggregate,dbd,output,min,max,varHeuristic,valHeuristic,lastConflict,vals,BC, AC, FC, prepro, search, filtering, restarts,group,width, widths, stretch,lengths,block,start,final,balance,path,tree, root, succs, range, image, allDistant, roots, capacity, bins, sizes, heights,and, or, except, function, matrix, sumCosts, increasing, index, list,lists,set,mset,terminal,rules,grammar,mapping,cards, ranges, rowCards, colCards, indexes, not, allIntersecting, costMeasure, accept, limits, relation, sum, comparison, operand, condition, conditions, nCircuits, nPaths, nTrees, rowOccurs, colOccurs, occurs, noOverlap,cost, total, number, cardinality, scope, static, xor, iff, ifThen, ifThenElse, row, adhoc, form, note
},
  basewidth  = {.6em,0.6em},
  keywordstyle=\color{mblue}\bfseries,
  ndkeywords={note,class,of,consistency,branching,lc,combination,defaultDegree,id, size, startIndex, format, type, reifiedBy, hreifiedFrom, hreifiedTo, as, measure, degree, threshold, cutoff, factor, varsModel, ctrsModel, commModel, for, case, closed, rank, restriction, circular, offset, collect, window, violable, order , defaultCost, solution, optimum, violationCost, violationMeasure},
  ndkeywordstyle=\color{dviolet}\bfseries,
  identifierstyle=\color{black},
  sensitive=false,
  comment=[l]{//},
  morecomment=[s]{<!--}{-->},
  commentstyle=\color{dred}\ttfamily,
  stringstyle=\color{dgreen}\ttfamily,
  morestring=[b]',
  morestring=[b]",
  escapechar=@,
  showstringspaces=false,
  xleftmargin=1pt,xrightmargin=1pt,
  breaklines=true,basicstyle=\ttfamily\small,backgroundcolor=\color{colorex},inputencoding=utf8/latin9,texcl
}

\lstdefinelanguage{json}{
    basewidth  = {.6em,0.6em},
    basicstyle=\normalfont\ttfamily,
    breaklines=true,
    morestring=[b]',
    morestring=[b]", 
    sensitive=false,
    stringstyle=\color{dgreen}\ttfamily, 
    escapechar=!,
    showstringspaces=false,
    xleftmargin=1pt,xrightmargin=1pt,
    breaklines=true,basicstyle=\ttfamily\small,backgroundcolor=\color{colorex},inputencoding=utf8/latin9,texcl
}

\newcommand{\core}[1]{ 
  \medskip \begin{tcolorbox}[
    enhanced,breakable,
    boxsep=0pt,top=0pt,bottom=0pt,left=6mm,right=1mm,
    toprule=0.1mm,leftrule=0.1mm,rightrule=0.25mm,bottomrule=0.25mm,shadow={0.2mm}{-0.2mm}{0mm}{dgray},
    overlay unbroken and first={\node (logo) at ([xshift=4mm,yshift=-5mm]frame.north west) {#1}; \draw[black,line width=1.5pt] (logo) -- ([xshift=4mm,yshift=1.5mm]frame.south west);  },
    colframe=dgray,titlerule=-0.2mm,toptitle=3mm,coltitle=black,fonttitle=\bfseries,
    lines before break=6, pad at break*=10pt
}
\newcounter{cntSy}
\newcounter{cntSe}
\newcounter{cntEx}

\ifx\bw\undefined
  \lstnewenvironment{syntax}{\lstset{language=syntax}}{}
  \lstnewenvironment{semantics}{\lstset{language=semantics}}{}
  \lstnewenvironment{xcsp}{\lstset{language=xcsp}}{} 
  \lstnewenvironment{json}{\lstset{language=json}}{}
  \newenvironment{boxsy}
    {\stepcounter{cntSy} \core{\bcpen} ,colback=colorsy,title style={color=colorsy},title=~ Syntax \thecntSy]}
    {\end{tcolorbox}} 
  \newenvironment{boxse}
    {\stepcounter{cntSe} \core{\bcdico} ,colback=colorse,title style={color=colorse},title=~ Semantics \thecntSe]}
    {\end{tcolorbox}} 
  \newenvironment{boxex}
    {\stepcounter{cntEx} \core{\bcroue} ,colback=colorex,title style={color=colorex},title=~ Example \thecntEx]}
    {\end{tcolorbox}} 
  \newenvironment{simplex}
    {\medskip \begin{bclogo}[couleur = colorex,couleurBord=dgray,epBord=0.01,arrondi=0.2,couleurOmbre=dgray,ombre=true,epOmbre=0.05,barre=none,logo=]{}\vspace{-0.5cm}}
    {\vspace{-0.1cm}\end{bclogo} \medskip}
\else
  \lstnewenvironment{syntax}{\lstset{language=syntax,backgroundcolor=\color{white}}}{}
  \lstnewenvironment{semantics}{\lstset{language=semantics,backgroundcolor=\color{white}}}{}
  \lstnewenvironment{xcsp}{\lstset{language=xcsp,backgroundcolor=\color{white}}}{}
  \newenvironment{boxsy}
    {\medskip \stepcounter{cntSy} \begin{bclogo}[barre=none,logo=]{ Syntax \thecntSy}\vspace{-0.1cm}}
    {\vspace{-0.1cm}\end{bclogo} \medskip ~ \vspace{-0.3cm}}
  \newenvironment{boxse}
    {\medskip \stepcounter{cntSe} \begin{bclogo}[barre=none,logo=]{ Semantics \thecntSe}\vspace{-0.1cm}}
    {\vspace{-0.1cm}\end{bclogo} \medskip}
  \newenvironment{boxex}
    {\medskip \stepcounter{cntEx} \begin{bclogo}[barre=none,logo=]{ Example \thecntEx}\vspace{-0.1cm}}
    {\vspace{-0.1cm}\end{bclogo} \medskip}
  \newenvironment{simplex}
    {\medskip \begin{bclogo}[barre=none,logo=]{}\vspace{-0.5cm}}
    {\vspace{-0.1cm}\end{bclogo} \medskip}
\fi

\newtheorem{remark}{Remark}

\definecolor{v2lgray}{gray}{0.85}
\definecolor{dgray}{rgb}{0.4,0.4,0.4}
\definecolor{dblue}{RGB}{0,0,99}
\definecolor{dred}{RGB}{150,6,54}
\definecolor{dgreen}{RGB}{47,135,7}
\definecolor{dviolet}{RGB}{102,0,153}
\definecolor{mblue}{RGB}{0,0,180}
\definecolor{colorse}{RGB}{255,248,220}
\definecolor{colorsy}{HTML}{F2F2F2}
\definecolor{colorex}{HTML}{FFE3BE}
\definecolor{grey1}{rgb}{0.9,0.9,0.9}
\definecolor{grey}{rgb}{0.75,0.75,0.75}

\def\N{\mathbb{N}}
\def\Q{\mathbb{Q}}
\def\D{{\mathsf{D}}}
\def\B{{\mathsf{B}}}

\def\DINT{{\mathsf{D}_\mathsf{int}}}
\def\BINT{{\mathsf B}_\mathsf{int}}

\def\ti{\textrm{-}}
\def\tr{\;\!\triangleright}
\def\st{\!:\!}

\def\gecode{Gecode\xspace}
\def\choco{Choco3\xspace}

\def\mzinc{MiniZinc\xspace}

\def\cat{Global Constraint Catalog\xspace}
\def\ace{ACE\xspace}

\def\x3{{XCSP$^3$}\xspace}
\def\j3{JvCSP$^3$\xspace}
\def\p3{{PyCSP$^3$}\xspace}

\newcommand{\xml}[1]{{\tt <#1>}} 
\newcommand{\att}[1]{{\tt #1}} 
\newcommand{\val}[1]{{\tt "#1"}} 

\newcommand{\bnf}[1]{\textsl{\color{dblue}{#1}}}
\newcommand{\bnfX}[1]{\texttt{<}\bnf{#1}\texttt{.../>}}
\newcommand{\norX}[1]{\texttt{<#1.../>}}

\newcommand{\gb}[1]{{\tt #1}} 
\newcommand{\gbc}[1]{\textcolor{dblue}{{\mathit #1}}} 
\newcommand{\nn}[1]{{\tt #1}} 
\newcommand{\nm}[1]{\mathit{#1}} 
\newcommand{\sy}[1]{{\ttfamily {\slshape #1}}}  
\newcommand{\ns}[1]{{\mathcal #1}}  

\newcommand*{\com}[1]{\hfill \textcolor{dgray}{// #1}} 
\newcommand*{\jsn}[1]{\textcolor{mblue}{#1}} 
\newcommand{\violet}[1]{{\small \textcolor{dviolet}{#1}}}

\newcommand{\va}[1]{{\boldsymbol #1}} 

\usepackage{version}
\includeversion{xl} \excludeversion{xc}

\begin{xl}
\title{\includegraphics[scale=0.5]{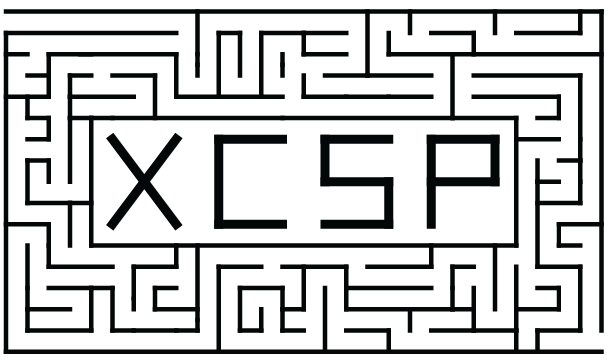}  \\~ \\ \textcolor{dred}{{\bf \x3\\ An Integrated Format for Benchmarking \\Combinatorial Constrained Problems}}}
\end{xl}

\begin{xc}
\title{\includegraphics[scale=0.5]{xcsp.png}  \\~ \\ \textcolor{dred}{{\bf \x3-core\\ A Format for Representing Constraint Satisfaction/Optimization Problems}}}
\end{xc}

\author{
  Fr\'ed\'eric Boussemart \and Christophe Lecoutre \\
  \and Gilles Audemard \and C\'edric Piette\\
~ \\
CRIL- CNRS, UMR 8188, University of Artois \\
Rue de l'universit\'e, SP 16, 62307 Lens, France \\
~ \\
\href{https://www.pycsp.org}{www.pycsp.org} ~ ~ \href{https://www.xcsp.org}{www.xcsp.org}
}

\begin{xl}
\date{{\vspace{2cm} \Large \hrule \smallskip  {\bf \textcolor{black}{\x3 Specifications -- Version 3.2 \\ August 28, 2024}} \\ \hrule}}
\end{xl}

\begin{xc}
\date{{\vspace{2cm} \Large \hrule \smallskip  {\bf \textcolor{black}{\x3-core Specifications -- Version 3.1 \\  August 28, 2024}} \\ \hrule}}
\end{xc}

\begin{document}
\maketitle
\thispagestyle{empty}


{\bf Licence}. \href{https://creativecommons.org/licenses/by-sa/4.0/}{Creative commons (CC BY-SA 4.0)} for specifications and \href{https://en.wikipedia.org/wiki/MIT_License}{MIT Licence} for any derived piece of software and benchmarks. 

\vfill
\begin{xl}
A version of this document, limited to \x3-core, is available \cite{xcsp3core}.  
\end{xl}
\begin{xc}
A version of this document, for complete \x3 specifications, is available \cite{xcsp3}.  
\end{xc}
\x3 is the format used in the competition that is organized annually, notably in 2022 \cite{compet22}, 2023 \cite{compet23} and 2024 \cite{compet24}.
Please, do not hesitate to contact us for any kind of feedback. Send an email to:
\begin{quote}
  \{lecoutre,audemard\}@cril.fr
\end{quote}

\begin{xl}
\begin{abstract}
\x3 is an integrated format for representing combinatorial constrained problems, which can deal with mono/multi optimization, many types of variables, cost functions, reification, views, annotations, variable quantification, distributed, probabilistic and qualitative reasoning.
It is also compact, and easy to read and to parse. 
Interestingly, it captures the structure of models (problems), by the possibility of declaring arrays of variables, and identifying syntactic and semantic groups of constraints.
The number of constraints is kept under control by introducing a limited set of basic constraint forms, and producing almost automatically some of their variations through lifting, restriction, sliding, logical combination and relaxation mechanisms.
As a result, \x3 encompasses practically all constraints that can be found in major constraint solvers developed in the CP (Constraint Programming) community.
A website, which is developed conjointly with the format, contains many models and series of instances.
The objective of \x3 is to ease the effort required to test and compare different algorithms by providing a common test-bed of combinatorial constrained instances. 
\end{abstract}
\end{xl}

\begin{xc}
\begin{abstract}
In this document, we introduce \x3-core, a subset of \x3 that allows us to represent constraint satisfaction/optimization problems.
The interest of \x3-core is multiple:
(i) focusing on the most popular frameworks (CSP and COP) and constraints,
(ii) facilitating the parsing process by means of dedicated \x3-core parsers written in Java, C++ and Python (using callback functions),
(iii) and defining a core format for comparisons (competitions) of constraint solvers.
\end{abstract}
\end{xc}

\tableofcontents

\chapter{\textcolor{gray!95}{Introduction}}\label{cha:intro}

We propose a standard Constraint Programming (CP) format, called \x3, to represent various forms of combinatorial problems subject to constraints to be satisfied and objectives to be optimized.
Compared to its predecessor XCSP 2.1 \cite{RL_xml}, \x3 is a major extension that allows us to build integrated representations of combinatorial constrained problems.
It lets the possibility of generating and exchanging files containing precise descriptions of problem instances, so that fair comparisons of problem-solving approaches can be made in good conditions, and experiments can be reproduced easily.
\x3 can be seen as an {\em intermediate Constraint Programming format preserving the structure of models}.
In other words, \x3 is neither a flat format, such as XCSP 2.1 \cite{RL_xml} or FlatZinc \cite{B_FZ}, nor a modeling language such as MiniZinc \cite{NSBBDT_minizinc} or ESSENCE \cite{FGJMM_essence}.
While modeling languages provide facilities to model parameterized classes of problem instances, and flat formats basically enumerate variables and constraints in sequence for separate problem instances, \x3,
as we shall see, stands between these two forms of representation (or abstraction).
Interestingly, it is possible to use a Python library called \p3 \cite{pycsp3} 
for modeling constrained problems and compiling them into \x3 instances.

As a solid basis for \x3, we employ the Extensible Markup Language (XML) \cite{XML}, which is a widely used flexible text format. 
XML is used as a support for the overall architecture of \x3, facilitating the issue of parsing through common tools such as DOM 
and SAX. 
Nowadays, tags and attributes that make up XML are commonly employed conventions for structuring information, as they are the main components of HTML.
Accordingly, this choice makes our format \x3 easily understandable for anybody who has some knowledge in Information Technology. 
It is important to note that within the global architecture of \x3, there are various XML elements that are not completely decomposed into finer XML terms.
For example, each integer is not put within its own element.
This choice we {\em deliberately} made allows compact, readable and easily modifiable problem representations, without placing too much burden on parser developers because internal data that needs to be analyzed is rather elementary.
We believe that adopting a ``complete'' XML approach would render \x3 instances very verbose, which would lessen readability (which is one of our primary goals).


\begin{xl}
\section{Features of \x3}\label{sec:features}

Here are the main features of \x3:
\begin{itemize}
\item {\bf Large Range of Frameworks}. \x3 allows us to represent many forms of combinatorial constrained problems since it can deal with:

\begin{itemize}
\item constraint satisfaction: CSP (Constraint Satisfaction Problem)
\item mono and multi-objective optimization: COP (Constraint Optimization Problem)
\item preferences and costs: WCSP (Weighted Constraint Satisfaction Problem) and FCSP (Fuzzy Constraint Satisfaction Problem)
\item variable quantification: QCSP(+) (Quantified Constraint Satisfaction Problem) and QCOP(+) (Quantified Constraint Optimization Problem)
\item probabilistic constraint reasoning: SCSP (Stochastic Constraint Satisfaction Problem) and SCOP (Stochastic Constraint Optimization Problem) 
\item qualitative reasoning: QSTR (Qualitative Spatial and Temporal Reasoning) and TCSP (Temporal Constraint Satisfaction Problem)
\item continuous constraint solving: NCSP (Numerical Constraint Satisfaction Problem) and NCOP (Numerical Constraint Optimization Problem) 
\item distributed constraint reasoning: DisCSP (Distributed Constraint Satisfaction Problem) and DCOP (Distributed Constraint Optimization Problem)
\end{itemize}
\item {\bf Large Range of Constraints}. A very large range of constraints is available, encompassing practically (i.e., to a very large extent) all constraints that can be found in major constraint solvers such as, e.g., Choco and Gecode.  
\item {\bf Limited Number of Concepts}. We paid attention to control the number of concepts and basic constraint forms, advisedly exploiting automatic variations of these forms through lifting, restriction, sliding, combination and relaxation mechanisms, so as to facilitate global understanding. 
\item {\bf Readability}. The new format is more compact, and less redundant, than XCSP 2.1, making it very easy to read and understand, especially as variables and constraints can be handled under the form of arrays and groups. Also, anyone can modify very easily an instance by hand.
  We do believe that a rookie in CP can easily manage (read, write, and update) small instances in \x3.
\item {\bf Flexibility}. It will be very easy to extend the format, if necessary, in the future, for example by adding new kind of global constraints, or by adding a few XML attributes in order to handle new concepts.
\item {\bf Ease of Parsing}. Thanks to the XML architecture of the format, basically, it is easy to parse instance files at a coarse-grain level. Besides, parsers written in {\tt Java} and {\tt C++} are available.
\item {\bf Dedicated Website}. A website, companion of \x3, is available, with many downloadable models/series/instances. 
\end{itemize}

Before starting the description of the new format, let us briefly introduce what are the main novelties of \x3 with respect to XCSP 2.1:

\begin{itemize}
\item {\bf Optimization}. \x3 can manage both mono-objective and multi-objective optimization.
\item {\bf New Types of Variables}. It is possible to define 0/1, integer, symbolic, real, stochastic, set, qualitative, and graph variables, in \x3. 
\item {\bf Lifted and Restricted forms of Constraints}. It is natural to extend basic forms of constraints over lists (tuples), sets and multi-sets. It is simple to build restricted forms of constraints by considering some properties of lists.
\item {\bf Meta-constraints}. It is possible to exploit sliding and logical mechanisms over variables and constraints.
\item {\bf Soft constraints}. Cost-based relaxed constraints and cost functions can be defined easily.
\item {\bf Reification}. Half and full reification is easy, and made possible  by letting the user associate a 0/1 variable with any constraint of the problem through a dedicated XML attribute.
\item {\bf Generalized Forms (Views)}. In \x3, it is possible to post constraints with arguments that are not limited to simple variables or constants, thus, avoiding in some situations the necessity of introducing, at modeling time, auxiliary variables and constraints, and permitting solvers that can handle variable views to do it.
\item {\bf Preservation of Structure}. It is possible to post variables under the form of arrays (of any dimension) and to post constraints in (semantic or syntactic) groups, thereby preserving the structure of the models. 
\item {\bf No Redundancy}. In \x3, redundancy is very limited, making instance representations more compact and less error-prone.
\item {\bf Compactness}. Posting variables and constraints in arrays/groups, while avoiding redundancy, makes instances in \x3 more compact than similar instances in XCSP 2.1.
\item {\bf Annotations}. It is possible to add annotations to the instances, for indicating for example search guidance (heuristics) and filtering advices.
\end{itemize}

As you may certainly imagine from this description, \x3 is a major rethinking of XCSP 2.1.
Because some problems inherent to the way XCSP 2.1 was developed had to be fixed, we have chosen not to make \x3 backward-compatible.
However, do not feel concerned too much about this issue because all series of instances available in XCSP 2.1 have been translated into \x3, and as said above, parsers are made available.
\end{xl}

\section{Complete Tool Chain}\label{sec:chain}

\begin{figure}[p]
\begin{center}
\begin{tikzpicture}[scale=0.9, every node/.style={scale=0.9}]
\tikzstyle{sn}=[draw,rounded corners,minimum height=12mm,fill=blue!30,text width=6cm,text centered]
\node[draw=none,text centered,text width=3cm] (1) at (0,0) {Modeling\\ Languages/Libraries}; 
\node[draw=none,text centered,text width=3cm] (2) at (0,-2.5) {Intermediate\\ Format}; 
\node[draw=none,text centered,text width=3cm] (2) at (0,-5) {Flat\\ Formats}; 
\node[draw=none] (l) at (0,-1.25) {};
\node[draw=none] (r) at (9.5,-1.25) {};
\node[draw=none] (l2) at (0,-3.75) {};
\node[draw=none] (r2) at (9.5,-3.75) {};
\node[draw=none] (a1) at (10,0.5) {$+$};
\node[draw=none] (a2) at (10,-5.5) {$-$};
\node[sn] (a) at (5,0) {OPL, MiniZinc, Essence, \p3, \j3, ...}; 
\node[sn] (b) at (5,-2.5) {\x3}; 
\node[sn] (c) at (5,-5) {XCSP 2.1, FlatZinc, wcsp}; 
\draw[->,>=latex] (a1) -- (a2);
\node[draw=none,rotate=-90] at (10.6,-2.2) {Abstraction};
\draw[dotted] (l) -- (r);
\draw[dotted] (l2) -- (r2);
\end{tikzpicture}
\end{center}
\caption{Modeling Languages, Intermediate and Flat Formats.\label{fig:modfor}}
 \end{figure}
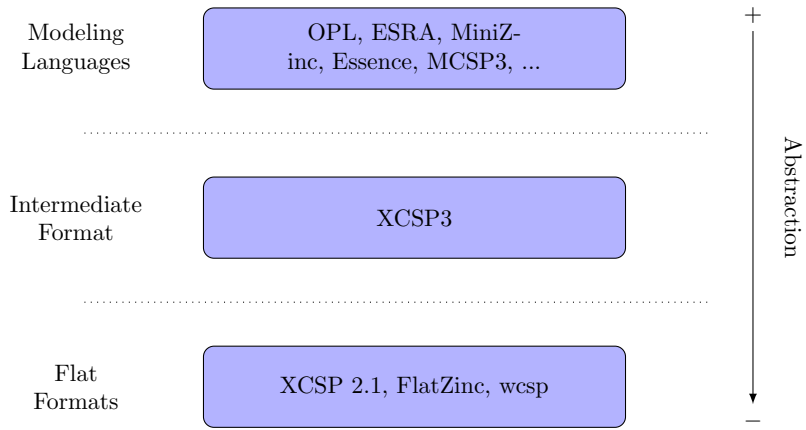

It is important to understand that \x3 is not a modeling language: \x3 aims at being a format to be handled easily by both humans and machines (parsers/solvers).
We show this graphically in Figure \ref{fig:modfor}. \x3 can be considered as an intermediate format. It is not as flat as FlatZinc and XCSP 2.1 since \x3 allows us to specify arrays of variables and groups/blocks of constraints. 
To make things clear, observe that the representation of combinatorial constrained problems can be handled at three different levels of abstraction:
\begin{enumerate}
\item high level of abstraction. At this level, modeling languages (or libraries) are used. They permit to use control structures (such as loops) and separate models from data. It means that for a given problem, you typically write a model which is a kind of problem abstraction parameterized by some formal data.
  For a specific instance, you need to provide some effective data. In practice, you have a file for the model, and for each instance you have an additional file for the data corresponding to this instance. 
\item intermediate level. At this level, there is no separation between the model and the data, although the structure remains visible. To the best of our knowledge, \x3 is the only intermediate format proposed in the literature. For each instance, you only have a file.
\item low level. At this level, we find flat formats. Each component (variable or constraint) of a problem instance is represented independently. As a consequence, the initial structure of the problem is lost (or, deteriorated). 
\end{enumerate}

\begin{figure}[p]
\begin{center}
\begin{tikzpicture}[scale=1, every node/.style={scale=1}]
\tikzstyle{sn}=[draw,rounded corners,minimum height=10mm,fill=blue!30,text width=5cm,text centered]
\node (user) at (4.95,2.4) {\includegraphics[scale=0.25]{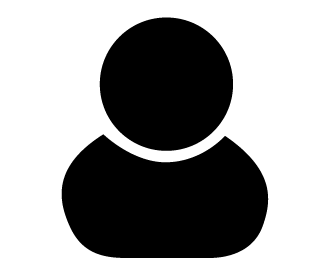}};
\node[draw=none] (root) at (5,1.5) {};
\node[draw=none] (l) at (0,-1.25) {};
\node[draw=none] (r) at (10.5,-1.25) {};
\node[draw=none] (l2) at (10,-3.75) {};
\node[draw=none] (r2) at (10.5,-3.75) {};
\node[draw,minimum height=7mm,fill=grey!30,text width=4cm,text centered] (model) at (2.5,0.5) {\textcolor{dgreen}{Model} \p3 \\{\scriptsize (Python 3)}}; 
\node[draw,minimum height=7mm,fill=grey!30,text width=4cm,text centered] (data) at (7.5,0.5) {\textcolor{dgreen}{Data} \\{\scriptsize (JSON)}}; 

\node[sn] (a) at (5,-1) {Compiler}; 
\node[draw,minimum height=7mm,fill=grey!30,text width=4cm,text centered] (b) at (5,-2.7) {\x3 \textcolor{dgreen}{Instance} \\{\scriptsize (XML)}}; 
\node[sn,text width=1.5cm] (s1) at (0,-4.8) {ACE}; 
\node[sn,text width=1.5cm] (s2) at (2,-4.8) {Choco}; 
\node[sn,text width=1.5cm] (s3) at (4,-4.8) {Mistral};
\node[sn,text width=1.5cm] (s4) at (6,-4.8) {Picat}; 
\node[sn,text width=1.5cm] (s5) at (8,-4.8) {...};
\node[sn,text width=1.5cm] (s6) at (10,-4.8) {OR-Tools};

\draw[->,>=latex] (root) -- (model);
\draw[->,>=latex] (root) -- (data);
\draw[->,>=latex] (model) -- (a);
\draw[->,>=latex] (data) -- (a);
\draw[->,>=latex] (a) -- (b);

\draw[->,>=latex] (b) -- (s1);
\draw[->,>=latex] (b) -- (s2);
\draw[->,>=latex] (b) -- (s3);
\draw[->,>=latex] (b) -- (s4);
\draw[->,>=latex] (b) -- (s5);
\draw[->,>=latex] (b) -- (s6);
\node (computer) at (4.98,-6.3) {\includegraphics[scale=0.02]{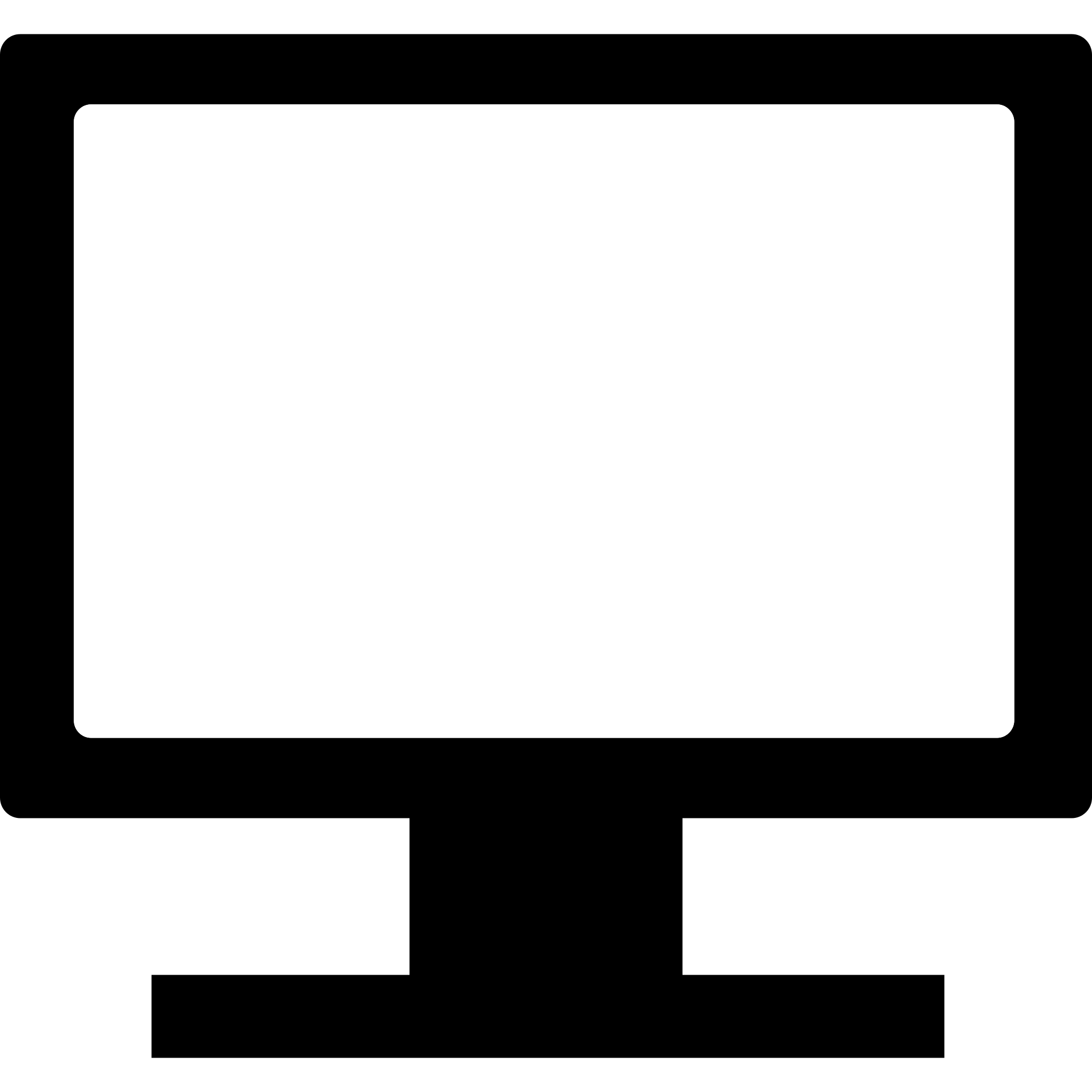}};
\end{tikzpicture}
\end{center}
\caption{Complete process for modeling and solving combinatorial constrained problems.\label{fig:mcsp3}}
\end{figure}

Although this document is devoted to \begin{xl}\x3\end{xl}\begin{xc}\x3-core\end{xc}, note that we propose a complete tool chain for handling combinatorial constrained problems.
The main ingredients of this complete tool chain are:
\begin{itemize}
\item \p3, \href{https://www.pycsp.org}{www.pycsp.org}: a Python library for modeling constrained problems (or equivalently, \j3, a Java-based API); see \href{https://github.com/xcsp3team/pycsp3}{github.com/xcsp3team/pycsp3} 
\item \x3, \href{https://www.xcsp.org}{www.xcsp.org}: an intermediate format used to represent problem instances while preserving the structure of models
\end{itemize}

Hence, for modeling problems in the \x3 ecosystem, the user can choose between two well-known languages (Python and Java).
As shown in Figure \ref{fig:mcsp3}, when a user has to solve a given problem such as e.g., a school timetabling problem, he can: 
\begin{enumerate}
\item write a model of this problem using either the Python library \p3 (i.e., write a Python file) or the Java modeling API \j3 (i.e., write a Java file) 
\item provide the data, in JSON format, to specify the precise problem instance to be solved (e.g., the data about the courses, rooms and teachers corresponding to the first week of the semester)
\item combine model and data by running an operation of compilation that generates an \x3 instance (file) 
\item solve the generated \x3 instance (file) using solvers like for example Choco \cite{CT_choco} or OscaR \cite{oscar}
\end{enumerate}

\noindent This approach has many advantages:
\begin{itemize}
\item Python (and Java), JSON, and XML are robust mainstream technologies
\item specifying data with JSON guarantees a unified notation, easy to read for both humans and machines 
\item writing models with Python 3 (or Java 8) avoids the user learning a new programming language
\item representing problem instances with coarse-grained XML guarantees compactness and readability; note that using JSON instead of XML for representing instances would have been possible but has some drawbacks, as explained in an appendix  of \begin{xl}this document\end{xl}\begin{xc}\cite{xcsp3}\end{xc}
\end{itemize}


In the following sections of this chapter, we provide some basic information concerning the syntax and the semantics of \x3, which will be useful for reading the rest of the document.
More precisely, we present the skeleton of typical \begin{xl}\x3\end{xl}\begin{xc}\x3-core\end{xc} instances, and introduce a few important notes related to \x3 syntax and semantics.

\section{Skeleton of \x3 Problem Instances}\label{sec:skeleton}

\begin{xl}
To begin with, you will find below the syntactic form of the skeleton of a {\em typical} \x3 problem instance, i.e., the skeleton used by most\footnote{Syntactic details for various frameworks are given in Chapter \ref{cha:frameworks}.} of the frameworks recognized by \x3. 
In our syntax, as shown in Appendix \ref{cha:bnf}, Page \pageref{cha:bnf}, we combine XML and BNF. XML is used for describing the main elements and attributes of \x3, whereas BNF is typically used for describing the textual contents of \x3 elements and attributes, with 
in particular \verb!|!, \verb![ ]!, \verb!( )!, \verb!*! and \verb!+! respectively standing for alternation, optional character, grouping and repetitions (0 or more, and 1 or more).
\end{xl}
\begin{xc}
To begin with, you will find below the syntactic form of the skeleton of a {\em typical} \x3-core problem instance, i.e., the skeleton used for an instance of either framework CSP or framework COP.
In our syntax, as shown in Appendix \ref{cha:bnf}, we combine XML and BNF. XML is used for describing the main elements and attributes of \x3, whereas BNF is typically used for describing the textual contents of \x3 elements and attributes, with 
in particular \verb!|!, \verb![ ]!, \verb!( )!, \verb!*! and \verb!+! respectively standing for alternation, optional character, grouping and repetitions (0 or more, and 1 or more).
\end{xc}

Note that \norX{elt} denotes an XML element of name \verb!elt! whose description is given later in the document (this is the meaning of \verb!...!). 
Strings written in dark blue italic form refer to BNF non-terminals defined in Appendix \ref{cha:bnf}.\begin{xl}
For example, \bnf{frameworkType} corresponds to a token chosen among CSP, COP, WCSP, $\ldots$ 
 Also, for the sake of simplicity, we write \bnfX{constraint} and \bnfX{metaConstraint} for respectively denoting any constraint element such as e.g., \norX{sum} or \norX{allDifferent}, and any meta-constraint such as e.g., \norX{slide} or \norX{not}.

\begin{boxsy}
\begin{syntax} 
<instance format="XCSP3" type="frameworkType">
  <variables>
    (   <var.../> 
      | <array.../>
    )+ 
  </variables>
  <constraints>
    (   <constraint.../> 
      | <metaConstraint.../> 
      | <group.../> 
      | <block.../>
    )* 
  </constraints>
  [<objectives [combination="combinationType"]>
    (   <minimize.../> 
      | <maximize.../>
    )+ 
   </objectives>]
  [<annotations.../>]
</instance>
\end{syntax}
\end{boxsy}

As you can observe, a typical instance is composed of variables, constraints, and possibly objectives (and annotations).
Variables are declared stand-alone (element \norX{var}) or under the form of arrays (element \norX{array}).
Constraints can be elementary (element \bnfX{constraint} or more complex (element \bnfX{metaConstraint}).
Constraints can also be posted in groups and/or declared in blocks; \norX{group} and \norX{block} correspond to structural mechanisms. 
Finally, any objective boils down to minimizing or maximizing a certain function, and is represented by an element \norX{minimize} or \norX{maximize}.
\end{xl}
\begin{xc}
  For example, \bnf{frameworkType} corresponds to the token CSP or the token COP.
 Also, for the sake of simplicity, we write \bnfX{constraint} for  denoting any constraint element such as e.g., \norX{sum} or \norX{allDifferent}.

\begin{boxsy}
\begin{syntax} 
<instance format="XCSP3" type="frameworkType">
  <variables>
    (   <var.../> 
      | <array.../>
    )+ 
  </variables>
  <constraints>
    (   <constraint.../> 
      | <group.../> 
      | <block.../>
    )* 
  </constraints>
  [<objectives>
    (   <minimize.../> 
      | <maximize.../>
    ) 
   </objectives>]
  [<annotations.../>]
</instance>
\end{syntax}
\end{boxsy}

As you can observe, a typical instance is composed of variables, constraints, and possibly one objective (note that for \x3-core, only one element can be present in \norX{objectives}).
Variables are declared stand-alone (element \norX{var}) or under the form of arrays (element \norX{array}).
Constraints are denoted by \bnfX{constraint}, and can be posted in groups and/or declared in blocks; \norX{group} and \norX{block} correspond to structural mechanisms. 
Finally, any objective boils down to minimizing or maximizing a certain function, and is represented by an element \norX{minimize} or \norX{maximize}.
\end{xc}
It is important to note that \norX{objectives} and \norX{annotations} are optional.

\begin{figure}[h!]
  \begin{center}
    \begin{tikzpicture}[node distance=1cm, >=stealth, auto] 
      \tikzstyle{node}=[state,minimum size=6mm,fill=white,draw,font=\sffamily\normalsize]
      \tikzstyle{arete}=[thick,-,>=stealth]
      \def\xbase{0}
      \def\ybase{2.5}
      
      \node[node] (xa) at (\xbase+0.5,\ybase+0.5) {$0$};
      \node[node] (xb) at (\xbase,\ybase) {$1$};
      \draw (\xbase-0.4,\ybase+0.8) node {$x$};
      
      \node[node] (ya) at (2,1.85) {$0$};
      \node[node] (yb) at (2,1.15) {$1$};
      \draw (2.8,1.5) node {$y$};
      
      \node[node] (za) at (0.5,0) {$0$};
      \node[node] (zb) at (0,0.5) {$1$};
      \draw (-0.4,-0.3) node {$z$};
      
      \draw[rotate around={45:(0.25,2.75)}] (0.25,2.75) ellipse (1cm and 0.5cm); 
      \draw (2,1.5) ellipse (0.5cm and 1cm); 
      \draw[rotate around={-45:(0.25,0.25)}] (0.25,0.25) ellipse (1cm and 0.5cm); 
      
      \draw (xa) -- (ya) ;
      \draw (xb) -- (yb) ;  
      \draw (xa) -- (za) ;
      \draw (xb) -- (zb) ;
      \draw (ya) -- (zb) ;
      \draw (yb) -- (za) ;
    \end{tikzpicture}
  \end{center}
  \caption{A toy constraint network.\label{fig:firstEx}}
\end{figure}
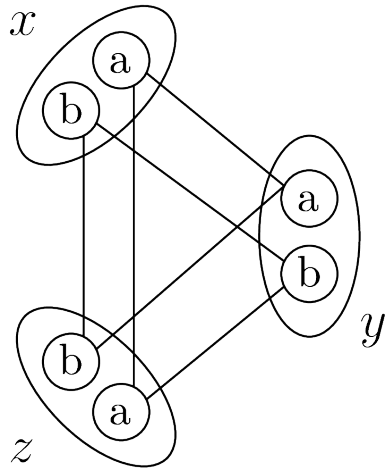

\medskip
For a first illustration of \x3, let us consider the small constraint network depicted in Figure \ref{fig:firstEx}.
We have here three integer variables, $x$, $y$, $z$ of domain $\{0,1\}$ and three binary constraints, one per pair of variables.
An edge in this graph means a compatibility between two values of two different variables.

In \x3, we obtain:

\begin{boxex}
\begin{xcsp}
<instance format="XCSP3" type="CSP">
  <variables>
    <var id="x"> 0 1 </var>
    <var id="y"> 0 1 </var>
    <var id="z"> 0 1 </var>
  </variables>
  <constraints>
    <extension>
      <list> x y </list>
      <supports> (0,0)(1,1) </supports>
    </extension>
    <extension>
      <list> x z </list>
      <supports> (0,0)(1,1) </supports>
    </extension>
    <extension>
      <list> y z </list>
      <supports> (0,1)(1,0) </supports>
    </extension>
  </constraints>
</instance>
\end{xcsp} 
\end{boxex}


Of course, we shall give all details of this representation in the next chapters, but certainly you are comfortable to make the correspondence between the figure and the \x3 code.
If you pay attention to the constraints, you can detect that two of them are similar.
Indeed, the list of supports (compatible pairs) between $x$ and $y$ is the same as that between $x$ and $z$. 
In \x3, it is possible to avoid such redundancy by using \begin{xl}either the concept of constraint group or the attribute \att{as}\end{xl}\begin{xc}the concept of constraint group\end{xc}.  
Also, because the three variables share the same type and the same domain, it is possible to avoid redundancy by using either the concept of variable array or the attribute \att{as}.
We shall describe all these concepts in the next chapters. 

As a second illustration, we consider the arithmetic optimization example introduced in the \mzinc Tutorial: ``A banana cake which takes 250g of self-raising flour, 2 mashed bananas, 75g sugar and 100g of butter, and a chocolate cake
which takes 200g of self-raising flour, 75g of cocoa, 150g sugar and 150g of butter. We can
sell a chocolate cake for $\$4.50$ and a banana cake for $\$4.00$. And we have 4kg self-raising
flour, 6 bananas, 2kg of sugar, 500g of butter and 500g of cocoa. The question is how many
of each sort of cake should we bake for the fete to maximize the profit?''
Here is how this small problem can be encoded in \x3. 
Surely, you can analyze this piece of code (if you are told that \verb!le! stands for ``less than or equal to''). Notice that any \x3 element can be given the optional attribute \att{note}, which is used as a short comment. 

\begin{boxex}
\begin{xcsp}
<instance format="XCSP3" type="COP">
  <variables>
    <var id="b" note="number of banana cakes"> 0..100 </var>
    <var id="c" note="number of chocolate cakes"> 0..100 </var>
  </variables>
  <constraints>
    <intension> le(add(mul(250,b),mul(200,c)),4000) </intension>
    <intension> le(mul(2,b),6) </intension>
    <intension> le(add(mul(75,b),mul(150,c)),2000) </intension>
    <intension> le(add(mul(100,b),mul(150,c)),500) </intension>
    <intension> le(mul(75,c),500) </intension>
  </constraints>
  <objectives>
    <maximize> add(mul(b,400),mul(c,450)) </maximize>
  </objectives>
</instance>
\end{xcsp} 
\end{boxex}

As we shall see, we can group together the constraints that share the same syntax (by introducing so-called constraint templates, which are abstract forms of constraints with parameters written \%i) and we can describe specialized forms of objectives.
An equivalent representation of the previous instance is:

\begin{boxex}
\begin{xcsp}
<instance format="XCSP3" type="COP">
  <variables>
    <var id="b" note="number of banana cakes"> 0..99 </var>
    <var id="c" note="number of chocolate cakes"> 0..99 </var>
  </variables>
  <constraints>
    <group>
      <intension> le(add(mul(%0,%1),mul(%2,%3)),%4) </intension>
      <args> 250 b 200 c 4000 </args>
      <args> 75 b 150 c 2000 </args>
      <args> 100 b 150 c 500 </args>
    </group>
    <group>
      <intension> le(mul(%0,%1),%2) </intension>
      <args> 2 b 6 </args>
      <args> 75 c 500 </args>
    </group>
  </constraints>
  <objectives>
    <maximize type="sum">
      <list> b c </list>
      <coeffs> 400 450 </coeffs>
    </maximize>
  </objectives>
</instance>
\end{xcsp} 
\end{boxex}

Alternatively, we could even use the global constraint \gb{sum}, so as to obtain: 

\begin{boxex}
\begin{xcsp}
<instance format="XCSP3" type="COP">
  <variables>
    <var id="b" note="number of banana cakes"> 0..99 </var>
    <var id="c" note="number of chocolate cakes"> 0..99 </var>
  </variables>
  <constraints>
    <sum note="using the 4000 grams of flour">
      <list> b c </list>
      <coeffs> 250 200 </coeffs>
      <condition> (le,4000) </condition>
    </sum>
    <sum note="using the 6 bananas">
      <list> b </list>
      <coeffs> 2 </coeffs>
      <condition> (le,6) </condition>
    </sum>
    <sum note="using the 2000 grams of sugar">
      <list> b c </list>
      <coeffs> 75 150 </coeffs>
      <condition> (le,2000) </condition>
    </sum>
    <sum note="using the 500 grams of butter">
      <list> b c </list>
      <coeffs> 100 150 </coeffs>
      <condition> (le,500) </condition>
    </sum>
    <sum note="using the 500 grams of cocoa">
      <list> c </list>
      <coeffs> 75 </coeffs>
      <condition> (le,500) </condition>
    </sum>
   </constraints>
  <objectives>
    <maximize type="sum" note="maximizing the profit (400 and 450 cents for each banana and chocolate cake, respectively)"> 
      <list> b c </list>
      <coeffs> 400 450 </coeffs>
    </maximize>
  </objectives>
</instance>
\end{xcsp} 
\end{boxex}

Since we have just mentioned \mzinc, let us recall again that \x3 is not a modeling language.
\x3 has no advanced control structures and no mechanism for separating model and data,
but aims at being an intermediate format, simple to be used by both the human and the machine (parser/solver) while keeping the structure of the problems.
However, the user can use \p3 \cite{pycsp3} as modeling libraries for generating \x3 instances.

\begin{xl}
\section{XML versus JSON}\label{sec:json}

For various reasons, some people prefer working with JSON, instead of XML.
As shown in Appendix \ref{cha:json}, we can easily convert \x3 instances from XML to JSON by following a few rules. 
We illustrate this on the previous examples. Their JSON representations are shown below.

\begin{boxex}
\begin{json}
{
  !\jsn{"@format"}!:"XCSP3",
  !\jsn{"@type"}!:"CSP",
  !\jsn{"variables"}!:{
    !\jsn{"var"}!:[{
      !\jsn{"@id"}!:"x",
      !\jsn{"domain"}!:"a b"
    }, {
      !\jsn{"@id"}!:"y",
      !\jsn{"domain"}!:"a b"
    }, {
      !\jsn{"@id"}!:"z",
      !\jsn{"domain"}!:"a b"
    }]
  },
  !\jsn{"constraints"}!:{
    !\jsn{"extension"}!:[{
      !\jsn{"list"}!:"x y",
      !\jsn{"supports"}!:"(a,a)(b,b)"
    }, {
      !\jsn{"list"}!:"x z",
      !\jsn{"supports"}!:"(a,a)(b,b)"
    }, {
      !\jsn{"list"}!:"y z",
      !\jsn{"supports"}!:"(a,b)(b,a)"
    }]
  }
}
\end{json} 
\end{boxex}

\begin{boxex}
\begin{json}
{
  !\jsn{"@format"}!:"XCSP3",
  !\jsn{"@type"}!:"COP",
  !\jsn{"variables"}!:{
    !\jsn{"var"}!:[{
      !\jsn{"@id"}!:"b",
      !\jsn{"domain"}!:"0..100"
    }, {
      !\jsn{"@id"}!:"c",
      !\jsn{"domain"}!:"0..100"
    }]
  },
  !\jsn{"constraints"}!:{
    !\jsn{"intension"}!:[
    "le(add(mul(250,b),mul(200,c)),4000)",
    "le(mul(2,b),6)",
    "le(add(mul(75,b),mul(150,c)),2000)",
    "le(add(mul(100,b),mul(150,c)),500)",
    "le(mul(75,c),500)"
    ]
  },
  !\jsn{"objectives"}!:{
    !\jsn{"maximize"}!:"add(mul(b,400),mul(c,450))"
  }
}
\end{json} 
\end{boxex}

\begin{remark}
Although JSON looks attractive, there exist some limitations that make us preferring XML. These limitations are discussed in Appendix \ref{cha:json}. 
\end{remark}
\end{xl}

\section{\x3-core}\label{sec:core}

\x3-core is a subset of \x3, with what can be considered as the main concepts of Constraint Programming (although, of course, this is quite subjective).
The interest of \x3-core is multiple:
\begin{itemize}
\item focusing on the most popular frameworks (CSP and COP) and constraints,
\item facilitating the parsing process by means of available parsers written in Java and C++ (which use callback functions),
\item and defining a core format for comparisons and competitions of constraint solvers.
\end{itemize}

\x3-core, targeted for CSP (Constraint Satisfaction Problem) and COP (Constraint Optimization problem), handles integer variables, mono-objective optimization, and 25 important constraints, described in the next chapters of this document:
\begin{itemize}
\item generic constraints
  \begin{itemize}
    \item \gb{intension} 
    \item \gb{extension}
  \end{itemize}
\item language-based constraints
  \begin{itemize}
  \item \gb{regular} 
  \item \gb{mdd}
  \end{itemize}
\item comparison-based constraints
  \begin{itemize}
  \item \gb{allDifferent} (and \gb{allDifferentList})
  \item \gb{allEqual}
  \item \gb{ordered} (and \gb{lex})
  \item \gb{precedence}
  \end{itemize}
\item counting constraints
  \begin{itemize}
  \item \gb{sum} 
  \item \gb{count} (capturing \gb{atLeast}, \gb{atMost}, \gb{exactly}, \gb{among})
  \item \gb{nValues}
  \item \gb{cardinality}
  \end{itemize}
\item connection constraints
  \begin{itemize}
  \item \gb{maximum} 
  \item \gb{minimum} 
  \item \gb{element} 
  \item \gb{channel}
  \end{itemize}
\item packing and scheduling constraints
  \begin{itemize}
  \item \gb{noOverlap} (capturing \gb{disjunctive} and \gb{diffn}) 
  \item \gb{cumulative}
  \item \gb{binPacking}
  \item \gb{knapsack}
  \end{itemize}
\item other constraints
  \begin{itemize}
  \item \gb{circuit}
  \item \gb{instantiation}
  \item \gb{slide}
  \end{itemize}
\end{itemize}

Importantly, as for full \x3, the structure of problems (models) can be preserved in \x3-core thanks to the following mechanisms:
\begin{itemize}
\item arrays of variables
\item groups of constraints
\item blocks of constraints
\item the meta-constraint \gb{slide}
\end{itemize}

Some on-line information about \x3-core can be found at \href{www.xcsp.org}{www.xcsp.org}.
Finally, note that the following parsers are available:
\begin{itemize}
\item a Java 8 parser for \x3-core (and also for a large part of full \x3)
\item a C++ parser for \x3-core
\end{itemize}

\section{Important Notes about Syntax and Semantics}\label{sec:syni}

As already indicated, for the syntax, we invite the reader to consult Appendix B. 
In this section, we discuss some important points. 


\paragraph{Constraint Parameters and Numerical Conditions.\label{sec:how}}
One important decision concerns the way constraint parameters are defined.
We have adopted the following rules:
\begin{enumerate}
\item Most of the time, when defining a constraint, there is a main list of variables (not necessarily, its full scope) that must be handled. We have chosen to call\footnote{Exceptions are for some scheduling constraints, where \xml{origins} is more appropriate.} it \verb;<list>;. Identifying the main list of variables (using its type/form) permits in a very simple and natural way to derive variants of constraints, by using alternative forms\begin{xl}such as as \verb;<set>; and \verb;<mset>; as shown in Chapter \ref{cha:lifted}\end{xl}.
\item Also, quite often, we need to introduce numerical conditions (comparisons) composed of an operator $\odot$ and a right-hand side operand $k$ that is a value, a variable, an interval or a set; the left-hand side being indirectly defined by the constraint.
The numerical condition is a kind of terminal operation to be applied after the constraint has ``performed some computation''.
We propose to represent each numerical condition, i.e., each pair $(\odot,k)$ by a single \x3 element \xml{condition} containing both the operator and the right operand, between parentheses and with comma as separator.
\item All other parameters are given by their natural names, as for example \verb;<transitions>; for the constraint \gb{regular}, or \verb;<index>; for  the constraint \gb{element}.
\end{enumerate}

\bigskip
It is important to note that each numerical condition $(\odot,k)$ semantically stands for ``$\odot\; k$'', where: 
\begin{itemize}
\item either $\odot \in \{<,\leq,>,\geq,=, \neq\}$ and $k$ is a value or a variable; in \x3, we thus have to choose a symbol in \verb!{lt,le,gt,ge,eq,ne}! for $\odot$,
\item or $\odot \in \{\in,\notin\}$ and $k$ is an integer interval of the form $l..u$, or an integer set of the form $\{a_1,\ldots,a_p\}$ with $a_1,\ldots,a_p$ integer values; in \x3, we thus have to choose a symbol in \verb!{in,notin}! for $\odot$.
\end{itemize}

For example, we may have $(\odot,k)$  denoting ``$> 10$'', ``$\neq z$'' or ``$\in 1..5$'' by respectively writing \verb!(gt,10)!, \verb!(ne,z)! and \verb!(in,1..5)!.
In \x3, we shall syntactically represent a numerical condition as follows:

\begin{boxsy}
\begin{syntax} 
<condition> "(" operator "," operand ")" </condition>
\end{syntax}
\end{boxsy}

where \bnf{operator} and \bnf{operand} are two BNF non-terminals defined in Appendix \ref{cha:bnf}.

\begin{xl}
\paragraph{Attributes \att{id}, \att{as} and \att{reifiedBy}.}
\end{xl}
\begin{xc}
\medskip\noindent {\bf Attributes \att{id} and \att{as}.}  
\end{xc}
In the next chapters, we shall present the syntax as precisely as possible.
However, we shall make the following simplifications.

Any XML element in \x3 can be given a value for the attribute \att{id}.
This attribute is required for the elements \norX{var} and \norX{array}, but it is optional for all other elements.
When presenting the syntax, we shall avoid to systematically write \verb![id="identifier"]!, each time we put an element for which \att{id} is optional.
Note that, as for HTML5, each value of an attribute \att{id} must be document-wide unique.

Any XML element in \x3 can be given a value for the attribute \att{as}.
This attribute is always optional.
Again, when presenting the syntax, we shall avoid to systematically write \verb![as="identifier"]!, each time we introduce an element.
This form of aliasing is discussed in Chapter \ref{cha:groups}.

\begin{xl}
Any constraint and meta-constraint can be given a value for one of the three following attributes \att{reifiedBy}, \att{hreifiedFrom} and \att{hreifiedTo}.
These attributes are optional and mutually exclusive.
When presenting the syntax in this document, we shall avoid to systematically write  \verb![reifiedBy="!\bnf{01Var}\verb!"|hreifiedFrom="!\bnf{01Var}\verb!"|hreifiedTo="!\bnf{01Var}\verb!"]! each time we introduce a constraint or meta-constraint.
Reification is discussed in Chapter \ref{cha:groups}.
\end{xl}

\paragraph{Generalized Forms of Constraints and Objectives (Views).}
For clarity, we always give the syntax of constraints in a rather usual restricted form.
For example, we shall write \verb!(intVar wspace)*! when a list of integer variables is naturally expected.
However, most of time, it is possible to replace variables by more general expressions.
This is typically the case for the element \xml{list} when it is involved in a constraint (or objective).
When expressions are given instead of variables, we speak about the {\em generalized} forms of constraints (and objectives); this is related to the concept of (variable) view \cite{ST_viewscp}.
This is acceptable (the XML schema that we provide for \x3 allows such forms) although we employ a more restrictive syntax in this document.
Generalized forms of constraints are introduced for the most important cases all along the paper, and discussed in Chapter \ref{cha:groups}.
The tools, modeling API and parsers, that we develop together with \x3 can deal with (some of) these generalized forms.
At the time of writing, it concerns the constraints \gb{allDifferent}, \gb{allEqual}, \gb{sum}, \gb{count}, \gb{nValues}, \gb{maximum} and \gb{minimum}, as well as the specialized forms of \gb{minimize} and \gb{maximize}.

\paragraph{Whitespaces.}
The tolerance with respect to whitespaces is the following:
\begin{itemize}
\item leading and trailing whitespaces for values of XML attributes are not tolerated. For example,  \verb!id=" x"! is not valid. 
\item leading and trailing whitespaces for textual contents of XML elements are tolerated, and not required. Although, for the sake of simplicity, we prefer writing: \begin{quote} \verb!<list> (intVar wspace)* </list>! \end{quote} 
instead of the heavier and more precise form:
\begin{quote} \verb!<list> [intVar (wspace intVar)*] </list>! \end{quote}
it must be clear that no trailing whitespace is actually required.
\item numerical conditions must not contain any whitespace. For example, \verb!(lt, 10)! is not valid.
\item functional expressions must not contain any whitespace. For example, \verb!add(x,y )! is not valid.
\item whitespaces are tolerated between vectors (tuples) but not within vectors, where a vector (tuple) is a sequence of elements put between brackets with comma as separator. For example, \verb!<supports> (1,3)( 2,4) </supports>! is not valid, whereas: 
\begin{quote}
\begin{verbatim}
<supports>
   (1,3)
   (2,4) 
</supports>
\end{verbatim}
\end{quote}
is.
\item whitespaces are not tolerated at all when expressing decimals, rationals, integer intervals, real intervals, and probabilities.  For example, \verb!3. 14!, \verb!3 /2!, \verb!1 .. 10!,  \verb![1.2, 5]!, and \verb!2: 1/6! are clearly not valid expressions.
\end{itemize}

\paragraph{Semantics.}
Concerning the semantics, here are a few important remarks:

\begin{itemize}
\item when presenting the semantics, we distinguish between a variable $x$ and its assigned value $\va{x}$ (note the bold face on the symbol $x$). 
\item Concerning the semantics of a numerical condition $(\odot,k)$, and depending on the form of $k$ (a value, a variable, an interval or a set), we shall indiscriminately use $\va{k}$ to denote the value $k$, the value of the variable $k$, the interval $l..u$ represented by $k$, or the set $\{a_1,\ldots,a_p\}$ represented by $k$.
\end{itemize}

\section{Notes (Short Comments) and Classes}\label{sec:notes}

In \x3, it is possible to associate a note (short comment) with any element.
It suffices to use the attribute \att{note} whose value can be any string.
Of course, for simplicity, when presenting the syntax, we shall never write \verb![note="string"]!, each time an element is introduced.
Here is a small instance with two variables and one constraint, each one accompanied by a note.

\begin{boxex}
\begin{xcsp}
<instance format="XCSP3" type="CSP">
  <variables>
    <var id="x" note="x is a number between 1 and 10"> 1..10 </var>
    <var id="y" note="y denotes the square of x"> 1..100 </var>
  </variables>
  <constraints>
    <intension note="this constraint links x and y"> 
      eq(y,mul(x,x)) 
    </intension>
  </constraints>
</instance>
\end{xcsp} 
\end{boxex}

In \x3, it is also possible to associate tags with any element.
As in HTML, these tags are introduced by the attribute \att{class}, whose value is a sequence of identifiers, with whitespace used as a separator.
Again, for simplicity, when presenting the syntax, we shall never write \verb![class="(identifier wspace)+"]!, each time an element is introduced.
As we shall see in Section \ref{sec:blocks}, these tags can be predefined or user-defined.

\section{Structure of the Document}

\begin{xl}
The document is organized in four parts.
In the first part, we show how to define different types of variables (Chapter \ref{cha:variables}) and objectives (Chapter \ref{cha:objectives}).
In the second part, we describe basic forms of constraints over simple discrete variables (Chapter \ref{cha:constraints}), complex discrete variables (Chapter \ref{cha:sets}) and continuous variables (Chapter \ref{cha:real}).
Advanced (i.e., non-basic) forms of constraints are presented in the third part of the document: they correspond to lifted and restricted constraints (Chapter \ref{cha:lifted}), meta-constraints (Chapter \ref{cha:meta}) and soft constraints (Chapter \ref{cha:cost}).
Finally, in the fourth part, we introduce groups, blocks, reification, views and aliases (Chapter \ref{cha:groups}), frameworks (Chapter \ref{cha:frameworks}) and annotations (Chapter \ref{cha:annotations}).
\end{xl}
\begin{xc}
The document is organized in four parts.
In the first part, we show how to define different types of variables (Chapter \ref{cha:variables}) and objectives (Chapter \ref{cha:objectives}).
In the second part, we describe basic forms of constraints over simple discrete variables (Chapter \ref{cha:constraints}).
Advanced (i.e., non-basic) forms of constraints are presented in the third part of the document: they correspond to lifted constraints (Chapter \ref{cha:lifted}) and meta-constraints (Chapter \ref{cha:meta}). 
Finally, in the fourth part, we introduce groups, blocks, views and aliases (Chapter \ref{cha:groups}), frameworks (Chapter \ref{cha:frameworks}) and annotations (Chapter \ref{cha:annotations}).
\end{xc}

\part{Variables and Objectives}

\begin{xl}
In this part, we show how variables and objectives are represented in \x3.
\end{xl}
\begin{xc}
In this part, we show how variables and objectives are represented in \x3-core.
\end{xc}

\chapter{\textcolor{gray!95}{Variables}}\label{cha:variables}

\begin{xl}
\begin{figure}[h!]
\begin{tikzpicture}[dirtree, every node/.style={draw=black,thick,anchor=west}]
\tikzstyle{selected}=[fill=colorex]
\tikzstyle{optional}=[fill=gray!10]
  \node [fill=gray!20] {{\bf Variables}}
    child { node [selected] {Discrete Variables}
      child { node {Simple Discrete Variables}
        child { node [optional] {Integer variables}}
        child { node [optional] {Symbolic Variables}}
      }
      child [missing] {}				
      child [missing] {}	
      child { node {Complex Discrete Variables}
        child { node [optional] {Set variables}}
        child { node [optional] {Graph Variables}}
      } 
      child [missing] {}	
      child [missing] {}	
      child { node [optional] {Stochastic Variables}}
    }		
    child [missing] {}	
    child [missing] {}				
    child [missing] {}
    child [missing] {}				
    child [missing] {}
    child [missing] {}				
    child [missing] {}				
    child { node [selected] {Continuous Variables}
      child { node [optional] {Real Variables}}
      child { node [optional] {Qualitative Variables}}
    }
    child [missing] {}				
    child [missing] {} 
    child { node [dotted,selected] {Arrays of Variables}}
    ;
\end{tikzpicture}
\caption{The different types of variables\label{fig:typeVar}}
\end{figure}
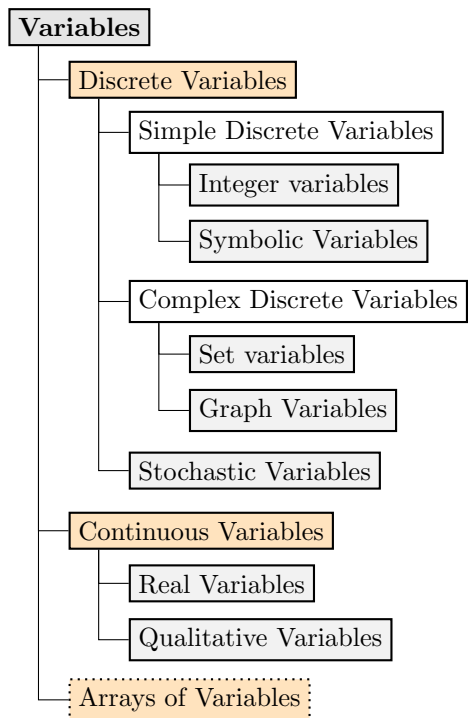

Variables are the basic components of combinatorial problems.
In \x3, as shown in Figure \ref{fig:typeVar}, you can declare:

\begin{itemize}
\item {\em discrete} variables, that come in three categories:
  \begin{itemize}
  \item {\em simple} discrete variables
    \begin{itemize}
    \item integer variables, including 0/1 variables that can be used to represent Boolean variables too
    \item symbolic variables 
    \end{itemize}
  \item {\em complex} discrete variables
    \begin{itemize}
    \item set variables
    \item graph variables
    \end{itemize} 
  \item {\em stochastic} discrete variables
  \end{itemize}
\item {\em continuous} variables
 \begin{itemize}
 \item {\em real} variables
 \item {\em qualitative} variables 
 \end{itemize} 
\end{itemize}

You can also declare:
\begin{itemize}
\item $k$-dimensional arrays of variables, with $k \geq 1$
\end{itemize}
\end{xl}

\begin{xc}
Variables are the basic components of combinatorial problems.
In \x3-core, you can declare:
\begin{itemize}
\item integer variables, including 0/1 variables that can be used to represent Boolean variables too
\item $k$-dimensional arrays of integer variables, with $k \geq 1$
\end{itemize}
\end{xc}

Recall that stand-alone variables as well as arrays of variables must be put inside the element \xml{variables}, as follows:

\begin{boxsy}
\begin{syntax} 
<variables>
  (<var.../> | <array.../>)+ 
</variables>
\end{syntax}
\end{boxsy}

Also as already mentioned, for the syntax, we combine XML and BNF, with BNF non-terminals written in dark blue italic form (see  Appendix \ref{cha:bnf}).   
The general syntax of an element \xml{var} is:


\begin{xl}
\begin{boxsy}
\begin{syntax} 
<var id="identifier" [type="varType"]> 
  ... @\com{Domain description}@ 
</var>
\end{syntax}
\end{boxsy}
\end{xl}
\begin{xc}
\begin{boxsy}
\begin{syntax} 
<var id="identifier"> 
  ... @\com{Domain description}@ 
</var>
\end{syntax}
\end{boxsy}
\end{xc}

A stand-alone variable can thus be declared by means of an element \xml{var}, and its identification is given by the value of the required attribute \att{id}.
\begin{xl}In accordance with the value (\val{integer} by default) of the attribute \att{type}, this\end{xl}\begin{xc}This\end{xc} element contains the description of its domain, i.e., the values that can possibly be assigned to it.

Sometimes, however, an element \xml{var} has no content at all.\begin{xl}
Actually, there are three such situations:
\begin{itemize}
\item when a variable is qualitative, its domain is implicit, as we shall see in Section \ref{sec:qualVar},
\item when a (non-qualitative) variable has an empty domain, as discussed in Section \ref{sec:empty},
\item when the attribute \att{as} is present.
\end{itemize}

For the last case, we have:
\end{xl}\begin{xc}
  For \x3-core, there is one such situation. This is when  the attribute \att{as} is present.
  We then have:
\end{xc}

\begin{xl}
\begin{boxsy}
\begin{syntax} 
<var id="identifier" [type="varType"] as="identifier" /> 
\end{syntax}
\end{boxsy}
\end{xl}
\begin{xc}
\begin{boxsy}
\begin{syntax} 
<var id="identifier" as="identifier" /> 
\end{syntax}
\end{boxsy}
\end{xc}

The value of \att{as} must correspond to the value of the attribute \att{id} of another element.
When this attribute is present, the content of the element is exactly the same as the one that is referred to (as if we applied a copy and paste operation), as discussed in Section \ref{sec:as}. 
From now on, for simplicity, we shall systematically omit the optional attribute \att{as}, although its role will be illustrated in the section about integer variables.

\medskip
The general syntax of an element \xml{array} is:


\begin{xl}
\begin{boxsy}
\begin{syntax} 
<array id="identifier" [type="varType"] size="dimensions" [startIndex="integer"]> 
   ... @\com{Domain description}@ 
</array>
\end{syntax}
\end{boxsy}
\end{xl}
\begin{xc}
\begin{boxsy}
\begin{syntax} 
<array id="identifier" size="dimensions"> 
   ... @\com{Domain description}@ 
</array>
\end{syntax}
\end{boxsy}
\end{xc}

\medskip
The chapter is organized as follows.\begin{xl}
From Section \ref{sec:var01} to Section \ref{sec:qualVar}, we develop the syntax for the different types of stand-alone variables, namely, 0/1 variables, integer variables, symbolic variables, real variables, set variables, graph variables, stochastic variables and qualitative variables. 
\end{xl}\begin{xc}
First, we develop in Section \ref{sec:var01} and \ref{sec:varInteger} the syntax for 0/1 variables, and integer variables.\end{xc} Then, in Section \ref{sec:array}, we describe arrays of variables, and in Section \ref{sec:empty} we discuss about variables with empty domains as well as about undefined and useless variables.
In the last section of this chapter, Section \ref{sec:solutions}, we show how solutions can be represented.


\section{Zero/One Variables}\label{sec:var01}

There is no specific language keyword for denoting a 0/1 variable. Instead, a 0/1 variable is defined as an integer variable (see next section). 

\begin{remark}
In \x3, there is no difference between a 0/1 variable and a Boolean variable: when appropriate, 0 stands for $\nm{false}$ and 1 stands for $\nm{true}$.
\end{remark}

\section{Integer Variables}\label{sec:varInteger}

Integer variables are given a domain of values by listing them, using whitespace as separator.
More precisely, the content of an element \xml{var} that represents an integer variable is an ordered sequence of integer values and integer intervals.
We have for example: 
\begin{itemize}
\item 1 5 10 that corresponds to the set $\{1,5,10\}$.
\item 1..3 7 10..14 that corresponds to the set $\{1,2,3,7,10,11,12,13,14\}$.
\end{itemize}

\begin{xl}
\begin{boxsy}
\begin{syntax} 
<var id="identifier" [type="@integer@"]>
  ((intVal | intIntvl) wspace)*
</var>
\end{syntax}
\end{boxsy}
\end{xl}
\begin{xc}
\begin{boxsy}
\begin{syntax} 
<var id="identifier">
  ((intVal | intIntvl) wspace)*
</var>
\end{syntax}
\end{boxsy}
\end{xc}

As an illustration, below, both variables \nn{foo} and \nn{bar} exhibit the same domain, whereas \nn{qux} mixes integer and integer intervals:

\begin{boxex}
\begin{xcsp}
<var id="foo"> 0 1 2 3 4 5 6 </var>
<var id="bar"> 0..6 </var>
<var id="qux"> -6..-2 0 1..3 4 7 8..11 </var>
\end{xcsp}
\end{boxex}

As mentioned earlier, declaring a 0/1 variable is made explicitly, i.e, with the same syntax as an ``ordinary'' integer variable.
For example, below, \nn{b1} and \nn{b2} are two 0/1 variables that can be served as Boolean variables in logical expressions:

\begin{boxex}
\begin{xcsp}
<var id="b1"> 0 1 </var>
<var id="b2"> 0 1 </var>
\end{xcsp}
\end{boxex}

\begin{xl}
Not all domains of integer variables are necessarily finite.
Indeed, it is possible to use the special value $\nm{infinity}$, preceded by the mandatory sign $+$ or $-$ (necessarily, as a bound for an integer interval).
For example:

\begin{boxex}
\begin{xcsp}
<var id="x"> 0..+infinity </var>
<var id="y"> -infinity..+infinity </var>
\end{xcsp}
\end{boxex}
\end{xl}
\begin{xc}
In \x3-core, all domains are necessarily finite.\bigskip
\end{xc}

Finally, when domains of certain variables are similar, and it appears that declaring array(s) is not appropriate, we can use the optional attribute \att{as} to indicate that an element has the same content as another one; the value of \att{as} must be the value of an attribute \att{id}, as explained in Section \ref{sec:as} of Chapter \ref{cha:groups}. Below, $v_2$ is a variable with the same domain as $v_1$.

\begin{boxex}
\begin{xcsp}
<var id="v1"> 2 5 8 9 12 15 22 25 30 50 </var>
<var id="v2" as="v1" /> 
\end{xcsp}
\end{boxex}

\begin{xl}
\begin{remark}
As shown by the syntax, for an integer variable, the attribute \att{type} is optional: if present, its value must be \val{integer}.
\end{remark}
\end{xl}

\begin{remark}
The integer values and intervals listed in the domain of an integer variable must always be in increasing order, without several occurrences of the same value. For example, \verb|0..10 10| is forbidden to be the content of an element \xml{var}.
\end{remark}

\begin{xl}
\section{Symbolic Variables}\label{sec:varSymbolic}

A symbolic variable is defined from a finite domain containing a sequence of symbols as possible values (whitespace as separator).
These symbols are identifiers, and so, must start with a letter.
For a symbolic variable, the attribute \att{type} for element \xml{var} is required, and its value must be \val{symbolic}.

\begin{boxsy}
\begin{syntax} 
<var id="identifier" type="symbolic">
  (symbol wspace)* 
</var>
\end{syntax}
\end{boxsy}

An example is given by variables \nn{light}, whose domain is $\{\nm{green},\nm{orange},\nm{red}\}$, and \nn{person}, whose domain is $\{\nm{tom},\nm{oliver},\nm{john},\nm{marc}\}$.

\begin{boxex}
\begin{xcsp}
<var id="light" type="symbolic"> green orange red </var>
<var id="person" type="symbolic"> tom oliver john marc </var>
\end{xcsp}
\end{boxex}

\begin{remark}
Integer values and symbolic values cannot be mixed inside the same domain. 
\end{remark}

\begin{remark}
The values listed in the domain of a symbolic variable are not necessarily given in increasing (lexicographic) order, but they must be all distinct.
\end{remark}
\end{xl}

\begin{xl}
\section{Real Variables}\label{sec:varReal}

Real variables are given a domain of values by listing real intervals, 
using whitespace as separator.
More precisely, the content of an element \xml{var} that represents a real variable contains an ordered sequence of real intervals (two bounds separated by a comma and enclosed between open or closed square brackets), whose bounds are integer, decimal or rational values.
In practice, most of the time, only one interval is given.
For a real variable, the attribute \att{type} for element \xml{var} is required, and its value must be \val{real}.

\begin{boxsy}
\begin{syntax} 
<var id="identifier" type="real">
  (realIntvl wspace)*
</var>
\end{syntax}
\end{boxsy}

We have for example: 
\begin{boxex}
\begin{xcsp}
<var id="w" type="real"> [0,+infinity[ </var>
<var id="x" type="real"> [-4,4] </var>
<var id="y" type="real"> [2/3,8.355] </var>
\end{xcsp}
\end{boxex}

\begin{remark}
Whenever infinity is involved, the associated square bracket(s) must be open, so that we shall always write \verb|]-infinity| and \verb|+infinity[|.
\end{remark}

\begin{remark}
The intervals listed in the domain of a real variable must always be in increasing order, without overlapping. For example, \verb|[0,10] [8,20]| is forbidden as content of any element \xml{var}.
\end{remark}

\section{Set Variables}\label{sec:varSet}

Set variables have associated set domains.
To begin with, we focus on integer set variables, i.e., on variables that must be assigned a set of integer values.
It is usual\footnote{There exist alternatives to represent domains, as the length-lex representation \cite{GH_length}. We might introduce them in the future if they become more popular in solvers.} that a set domain is approximated by a set interval specified by its upper and lower bounds (subset-bound representation), defined by some appropriate ordering on the domain values \cite{G_interval,G_constraints}.
The core idea is to approximate the domain of a set variable $s$ by a closed interval denoted $[s_{min}, s_{max}]$, specified by its unique least upper bound $s_{min}$, and unique greatest lower bound
$s_{max}$, under set inclusion.
We have $s_{min}$ that contains the {\em required} elements of $s$ and $s_{max}$ that contains in
addition the {\em possible} elements of $s$.
For example, if $[s_{min}=\{1, 5\}, s_{max}=\{1, 3, 5, 6\}]$ is the domain of an integer set variable $s$, then it means that the elements 1 and 5 necessarily belong
to $s$ and that 3 and 6 are possible elements of $s$.
The variable $s$ can then be assigned any set in the lattice defined from $s_{min}$ and $s_{max}$.

To define an integer set variable, we just have to introduce two elements \xml{required} and \xml{possible} within the element \xml{var}.
Such elements contain an ordered sequence of integers and integer intervals, as for ordinary integer variables.
For an integer set variable, the attribute \att{type} for \xml{var} is required, and its value must be \val{set}.

\begin{boxsy}
\begin{syntax} 
<var id="identifier" type="set">
  [<required> ((intVal | intIntvl) wspace)* </required> 
   <possible> ((intVal | intIntvl) wspace)* </possible>]
</var>
\end{syntax}
\end{boxsy}

An integer set variable $s$ of domain $[\{1, 5\}, \{1, 3, 5, 6\}]$ is thus represented by:

\begin{boxex}
\begin{xcsp}
<var id="s" type="set"> 
  <required> 1 5 </required> 
  <possible> 3 6 </possible>
</var>
\end{xcsp}
\end{boxex}

It is also possible to define symbolic set variables. In that case, elements \xml{required} and \xml{possible} must each contain a sequence of symbols (identifiers). 
For a symbolic set variable, the attribute \att{type} for \xml{var} is required, and its value must be \val{symbolic set}.

\begin{boxsy}
\begin{syntax} 
<var id="identifier" type="symbolic set">
  [<required> (symbol wspace)* </required> 
   <possible> (symbol wspace)* </possible>]
</var>
\end{syntax}
\end{boxsy}

The following example exhibits a symbolic set variable named \nn{team}, which must contain elements \nn{Bob} and \nn{Paul}, and can additionally contain elements \nn{Emily}, \nn{Luke} and \nn{Susan}.

\begin{boxex}
\begin{xcsp}
<var id="team" type="symbolic set"> 
  <required> Bob Paul </required> 
  <possible> Emily Luke Susan </possible>
</var>
\end{xcsp}
\end{boxex}

\begin{remark}
It is forbidden to have a value present both in \xml{required} and \xml{possible}.
\end{remark}

\section{Graph Variables}\label{sec:varGraph}

Graph variables have associated graph domains.
More precisely, graph variables have domains that are approximated by the lattice of graphs included between two bounds: the greatest lower bound and the least upper bound of the lattice \cite{DDD_cpgraph}.
Graph variables can be directed or undirected, meaning that either edges or arcs are handled.
For a graph variable $g$, the greatest lower bound $g_{min}$ defines the set of nodes and edges/arcs which are known to be part of $g$ while the least upper bound graph $g_{max}$ defines the set of possible nodes and edges/arcs in $g$.
For example, if $[g_{min}=(\{a,b\}, \{(a,b)\}), g_{max}=(\{a,b,c\},\{(a,b),(a,c),(b,c)\})]$ is the domain of an undirected graph variable $g$, then it means that the nodes $a$ and $b$, as well as the edge $(a,b)$ necessarily belong to $g$ and that the node $c$ as well as the edges $(a,c)$ and $(b,c)$ are possible elements of $g$.

To define a graph variable, we have to introduce two elements \xml{required} and \xml{possible} within the element \xml{var}.
Inside these elements, we find the elements \xml{nodes} and \xml{edges} for an undirected graph variable, and the elements \xml{nodes} and \xml{arcs} for a directed graph variable.
For a graph variable, the attribute \att{type} for \xml{var} is required, and its value must be either \val{undirected graph} or \val{directed graph}.

\begin{boxsy}
\begin{syntax} 
<var id="identifier" type="undirected graph">
 [<required> 
    <nodes> (symbol wspace)* </nodes> 
    <edges> ("(" symbol "," symbol ")")* </edges>
  </required> 
  <possible> 
    <nodes> (symbol wspace)* </nodes> 
    <edges> ("(" symbol "," symbol ")")* </edges>
  </possible>]
</var>
\end{syntax}
\end{boxsy}

\begin{boxsy}
\begin{syntax} 
<var id="identifier" type="directed graph">
 [<required> 
    <nodes> (symbol wspace)* </nodes> 
    <arcs> ("(" symbol "," symbol ")")* </edges>
  </required> 
  <possible> 
    <nodes> (symbol wspace)* </nodes> 
    <arcs> ("(" symbol "," symbol ")")* </edges>
  </possible>]
</var>
\end{syntax}
\end{boxsy}

The example given by Figure \ref{fig:graph}, for an undirected graph variable $g$ whose domain is $[(\{a,b\}, \{(a,b)\}), (\{a,b,c\},\{(a,b),(a,c),(b,c)\})]$, is thus represented by:

\begin{boxex}
\begin{xcsp}
<var id="g" type="undirected graph"> 
  <required> 
    <nodes> a b </nodes>
    <edges> (a,b) </edges>
  </required>
  <possible> 
    <nodes> c </nodes>
    <edges> (a,c)(b,c) </edges>
  </possible>
</var>
\end{xcsp}
\end{boxex}

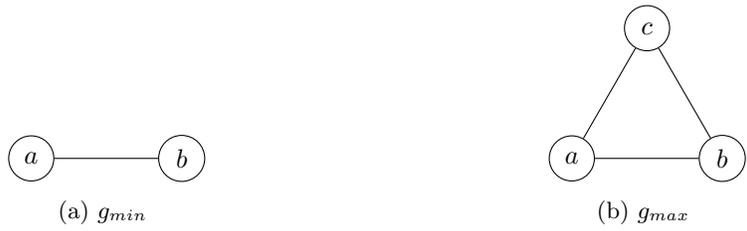
\begin{figure}[h]
  \centering
  \begin{subfigure}[b]{0.49\textwidth}
   \centering
    \begin{tikzpicture}
      \tikzstyle{sn}=[draw,circle,minimum size=6mm]
      \node[sn,opacity=0] (c) at (0,0) {$c$}; 
      \node[sn] (a) at (240:2) {$a$}; 
      \node[sn] (b) at (-60:2) {$b$}; 
      \draw (a) -- (b);
    \end{tikzpicture}
    \caption{$g_{min}$ \label{fig:mingraph}} 
  \end{subfigure}
  \begin{subfigure}[b]{0.49\textwidth}
    \centering
    \begin{tikzpicture}
      \tikzstyle{sn}=[draw,circle,minimum size=6mm]
      \node[sn] (c) at (0,0) {$c$};
      \node[sn] (a) at (240:2) {$a$};
      \node[sn] (b) at (-60:2) {$b$};
      \draw (a) -- (b) -- (c) -- (a);
    \end{tikzpicture}    
    \caption{$g_{max}$ \label{fig:maxgraph}}
  \end{subfigure}
\caption{The domain of an undirected graph variable $g$\label{fig:graph}}
\end{figure}

\begin{remark}
It is forbidden to have an element (node, edge, or arc) present both in \xml{required} and \xml{possible}.
\end{remark}

\section{Stochastic Variables}\label{sec:varStochastic}

In some situations, when modeling a problem, it is useful to associate a probability distribution with the values that are present in the domain of an integer variable.
Such variables are typically uncontrollable (i.e., not decision variables).
This is for example the case with the Stochastic CSP (SCSP) framework introduced by T. Walsh \cite{W_stochastic} to capture combinatorial decision problems involving uncertainty.
In \x3, the domain of an integer stochastic variable is defined as usual by a sequence of integers and integer intervals, but each element of the sequence is given a probability preceded by the symbol ":".
The value of a probability can be 0, 1, a rational value or a decimal value in $]0,1[$.
Note that it is also possible to declare symbolic stochastic variables.
For an integer stochastic variable, the attribute \att{type} for \xml{var} is required, and its value must be \val{stochastic}. 

\begin{boxsy}
\begin{syntax} 
<var id="identifier" type="stochastic">
  ((intVal | intIntvl) ":" proba wspace)+ 
</var>
\end{syntax}
\end{boxsy}

For a symbolic stochastic variable, the attribute \att{type} for \xml{var} is required, and its value must be \val{symbolic stochastic}. 

\begin{boxsy}
\begin{syntax} 
<var id="identifier" type="symbolic stochastic">
  (symbol ":" proba wspace)+
</var>
\end{syntax}
\end{boxsy}

Of course, the sum of probabilities associated with the different possible values of a stochastic variable must be equal to 1, and consequently, it is not possible to build a stochastic variable with an empty domain. 

An illustration is given by: 

\begin{boxex}
\begin{xcsp}
<var id="foo" type="stochastic"> 
   5:1/6 
  15:1/3 
  25:1/2 
</var>
<var id="dice" type="stochastic"> 
  1..6:1/6 
</var>
<var id="coin" type="symbolic stochastic"> 
  heads:0.5 
  tails:0.5 
</var>
\end{xcsp}
\end{boxex}

\begin{remark}
When the probability is associated with an integer interval, it applies independently to every value of the interval.
\end{remark}

\section{Qualitative Variables}\label{sec:qualVar}

In qualitative spatial and temporal reasoning (QSTR) \cite{K_temporal}, one has to reason with entities that corresponds to points, intervals, regions, ...
The variables that are introduced represent such entities, and their domains, being continuous, cannot be described extensionally.
This is the reason why we simply use the attribute \att{type} to refer to the implicit domain of qualitative variables.

In \x3, we can currently refer to the following types:
\begin{itemize}
\item \val{point}, when referring to the possible time points (or equivalently, points of the line) 
\item \val{interval}, when referring to the possible time intervals (or equivalently, intervals of the line) 
\item \val{region},  when referring to possible regions in Euclidean space (or in a topological space) 
\end{itemize}

\begin{boxsy}
\begin{syntax}
<var id="identifier" type="point" />
<var id="identifier" type="interval" />
<var id="identifier" type="region" />
\end{syntax}
\end{boxsy}

An illustration is given by:

\begin{boxex}
\begin{xcsp}
<var id="foo" type="point" /> 
<var id="bar" type="interval" /> 
<var id="qux" type="region" /> 
\end{xcsp}
\end{boxex}

\begin{remark}
Other domains for qualitative variables might be introduced in the future.
\end{remark}
\end{xl}

\section{Arrays of Variables}\label{sec:array}

Interestingly, \x3 allows us to declare $k$-dimensional arrays of variables, with $k \geq 1$, by introducing elements \xml{array} inside \xml{variables}.
We recall the general (simplified) syntax for arrays:

\begin{xl}
\begin{boxsy}
\begin{syntax} 
<array id="identifier" [type="varType"] size="dimensions" [startIndex="integer"]> 
  ...
</array>
\end{syntax}
\end{boxsy}
\end{xl}
\begin{xc}
\begin{boxsy}
\begin{syntax} 
<array id="identifier" size="dimensions"> 
  ...
</array>
\end{syntax}
\end{boxsy}
\end{xc}

Hence, for each such element, there is a required attribute \att{id} and a required attribute \att{size} whose value gives the structure of the array under the form "[nb$_1$]...[nb$_p$]" with nb$_1$, \ldots, nb$_p$ being strictly positive integers.
The number of dimensions of an array is the number of pairs of opening/closing square brackets, and the size of each dimension is given by its bracketed value. 
For example, if $x$ is an array of 10 variables, you just write \val{[10]}, and if $y$ is a 2-dimensional array, 5 rows by 8 columns, you write \val{[5][8]}.\begin{xl}
Indexing used for any dimension of an array starts at $0$, unless the (optional) attribute \att{startIndex} gives another value.
Of course, it is possible to define arrays of any kind of variables, as e.g., arrays of symbolic variables, arrays of set variables, etc. by using the attribute \att{type}: all variables of an array have the same type.
\end{xl}\begin{xc}
Indexing used for any dimension of an array starts at $0$.\end{xc}
The content of an element \xml{array} of a specified type is defined similarly to the content of an element \xml{var} of the same type, and basically, all variables of an array have  the same domain, except if mixed domains are introduced (as we shall see in subsection \ref{sub:mixed}).

\begin{xl}
To define an array $x$ of 10 integer variables with domain $1..100$, a 2-dimensional array $y$ ($5 \times 8$) of 0/1 variables, an array $\mathit{diceYathzee}$ of 5 stochastic variables, and an array $z$ of 12 symbolic set variables with domain $[\{a,b\},\{a,b,c,d\}]$, we write:  

\begin{boxex}
\begin{xcsp}
<array id="x" size="[10]"> 1..100 </array> 
<array id="y" size="[5][8]"> 0 1 </array> 
<array id="diceYathzee" size="[5]" type="stochastic"> 
  1..6:1/6 
</array>
<array id="z" size="[12]" type="symbolic set"> 
  <required> a b </required> 
  <possible> c d </possible> 
</array> 
\end{xcsp}
\end{boxex}

Importantly, it is necessary to be able to identify variables in arrays. We simply use the classical ``[]'' notation, with indexing starting at 0 (unless another value is given by \att{startIndex}).
For example, assuming that 0 is the ``starting index'', $x[0]$ is the first variable of the array $x$, and $y[4][7]$ the last variable of the array $y$.
\end{xl}
\begin{xc}
To define an array $x$ of 10 integer variables with domain $1..100$, and a 2-dimensional array $y$ ($5 \times 8$) of 0/1 variables, we write:  

\begin{boxex}
\begin{xcsp}
<array id="x" size="[10]"> 1..100 </array> 
<array id="y" size="[5][8]"> 0 1 </array> 
\end{xcsp}
\end{boxex}

Importantly, it is necessary to be able to identify variables in arrays. We simply use the classical ``[]'' notation, with indexing starting at 0.
For example, $x[0]$ is the first variable of the array $x$, and $y[4][7]$ the last variable of the array $y$.
\end{xc}

\subsection{Using Compact Forms}\label{sub:compactForms}

Sometimes, one is interested in selecting some variables from an array, for example, the variables in the first row of a 2-dimensional array.
We use integer intervals for that purpose, as in $x[3..5]$ and $y[2..3][0..1]$, and we refer to such expressions as {\em compact lists of array variables}.
In a context where a list of variables is expected, it is possible to use this kind of notation, and the result is then considered to be a list of variables, ordered according to a lexicographic order $\prec$ on index tuples (for example $y[2][4]$ is before $y[2][5]$ since $(2,4) \prec (2,5)$). 
On our previous example, in a context where a list of variables is expected, $x[3..5]$ denotes the list of variables $x[3]$, $x[4]$ and $x[5]$, while $y[2..3][0..1]$ denotes the list of variables $y[2][0]$, $y[2][1]$, $y[3][0]$ and $y[3][1]$.
It is also possible to omit an index, with a meaning similar to using $0..s$ where $s$ denotes the largest possible index. For example, $y[2][]$ is equivalent to  $y[2][0..7]$. 

Finally, one may wonder how compact lists of array variables (such as $x[3..5]$, $y[2..3][0..1]$, $x[]$, $y[][]$) precisely expand in the context of the \x3 elements that will be presented in the next chapters.
The rule is the following:
\begin{enumerate}
\item if a 2-dimensional compact list (such as $y[][]$) appears in an element \xml{matrix}, \xml{origins}, \xml{lengths},  \xml{rowOccurs} or \xml{colOccurs}, the compact list expands as a sequence of tuples (one tuple per row, with variables of the row separated by a comma, enclosed between parentheses). For example, 
\begin{simplex}
\begin{xcsp}
<matrix>
  y[][]
</matrix>
\end{xcsp}
\end{simplex}
expands as:
\begin{simplex}
\begin{xcsp}
<matrix> 
   (y[0][0],y[0][1],...,y[0][7]) 
   (y[1][0],y[1][1],...,y[1][7])
   ...
   (y[4][0],y[4][1],...,y[4][7])
</matrix>
\end{xcsp}
\end{simplex}
\item in all other situations, the compact list expands as a list of variables, with whitespace as a separator.  For example, we have: 
\begin{simplex}
\begin{xcsp}
<list>
  y[2..3][0..1]
</list>
\end{xcsp}
\end{simplex}
that expands as:
\begin{simplex}
\begin{xcsp}
<list>
  y[2][0] y[2][1] y[3][0] y[3][1]
</list>
\end{xcsp}
\end{simplex}
\end{enumerate}

Using this rule, it is always possible to expand all compact lists of array variables in order to get a form of the problem instance with only references to simple variables.

\subsection{Dealing With Mixed Domains}\label{sub:mixed}

Sometimes, the variables from the same array naturally have mixed domains. 
One solution is to build a large domain that can be suitable for all variables, and then to post unary domain constraints.
But this not very satisfactory.

Another solution with \x3 is to insert the different definitions of domains inside the \xml{array} element.
When several subsets of variables of an array have different domains, we simply have to put an element \xml{domain} for each of these subsets.
An attribute \att{for} indicates the list of variables to which the domain definition applies.
The value of \att{for} can be a sequence (whitespace as separator) of variable identifiers (with compact forms authorized), or the special value \val{others}, which is used to declare a default domain.
Only one element \xml{domain} can have its attribute \att{for} set to \val{others}, and if it is present, it is necessary at the last position.
The syntax for arrays that involve variables with different domains is:

\begin{xl}
\begin{boxsy}
\begin{syntax} 
<array id="identifier" [type="varType"] size="dimensions" [startIndex="integer"]> 
 (<domain for="(intVar wspace)+"> ... </domain>)+ 
 [<domain for="others"> ... </domain>]
</array>
\end{syntax}
\end{boxsy}
\end{xl}
\begin{xc}
\begin{boxsy}
\begin{syntax} 
<array id="identifier" size="dimensions"> 
 (<domain for="(intVar wspace)+"> ... </domain>)+ 
 [<domain for="others"> ... </domain>]
</array>
\end{syntax}
\end{boxsy}
\end{xc}

As an illustration, the 2-dimensional array $x$ below is such that the variables of the first, second and third rows have $1..10$, $1..20$ and $1..15$ as domains, respectively.
Also, all variables of array $y$ have $\{2,4,6\}$ as domain except for $y[4]$ whose domain is $\{0,1\}$.
Finally, all variables of array $z$ have $\{0,1\}$ as domain except for the variables that belong to the lists $z[][0..1][]$ and $z[][2][2..4]$.

\begin{boxex}
\begin{xcsp}
<array id="x" size="[3][5]"> 
  <domain for="x[0][]"> 1..10 </domain>
  <domain for="x[1][]"> 1..20 </domain>
  <domain for="x[2][]"> 1..15 </domain>
</array> 
<array id="y" size="[10]"> 
  <domain for="y[4]"> 0 1 </domain> 
  <domain for="others"> 2 4 6 </domain>
</array> 
<array id="z" size="[5][5][5]"> 
  <domain for="z[][0..1][] z[][2][2..4]"> 0..10 </domain> 
  <domain for="others"> 0 1 </domain>
</array>
\end{xcsp}
\end{boxex}


\begin{xl}
\section{Empty Domains, Undefined and Useless Variables}\label{sec:empty}

In some (rare) cases, one may want to represent variables with empty domains.
For example, if an inconsistency is detected during an inference process, it may be interesting to represent the state of the variable domains (including those that became empty).
For all types of variables, except for qualitative ones, an empty content for an element \xml{var} (or \xml{array}) means that the domain is empty (if, of course, the attribute \att{as} is not present).

In our illustration, below, we have an integer variable $x$ and a set variable $y$, with both an empty domain.

\begin{boxex}
\begin{xcsp}
<var id="x" />
<var id="y" type="set" />
\end{xcsp}
\end{boxex}
\end{xl}
\begin{xc}
\section{Undefined and Useless Variables}\label{sec:empty}
\end{xc}

We conclude this chapter with a short discussion about the concept of undefined and useless variables.
An {\em undefined} (domain) variable is a variable with no domain definition\begin{xc}.\end{xc}\begin{xl} (note that this is different from a variable having an empty domain).\end{xl}
Actually, undefined variables can only belong to arrays showing different domains for subsets of variables: when a variable does not match any value of attributes \att{for}, then it is undefined, and so must be completely ignored (by solvers).

As an illustration, below, the variable $z[5]$ is undefined. 

\begin{boxex}
\begin{xcsp}
<array id="z" size="[10]"> 
  <domain for="z[0..4]"> 1..10 </domain>
  <domain for="z[6..9]"> 1..20 </domain>
</array>
\end{xcsp}
\end{boxex}

A variable is said {\em useless} if it is involved nowhere (neither in constraints nor in objective functions); otherwise, it is said {\em useful}.
On the one hand, a useless variable can be the result of a reformulation/simplification process (and one may wish to keep such variables, for various reasons).
On the other hand, modeling can introduce useless variables, typically for symmetrical reasons, which may happen when introducing arrays of variables.

For the example below, suppose that only variables $[i][j]$ of $t$ such that $i>j$ are involved in constraints.
This means that all variables $[i][j]$ with $i \leq j$ are useless.

\begin{boxex}
\begin{xcsp}
<array id="t" size="[8][8]"> 1..10 </array> 
\end{xcsp}
\end{boxex}
 
\begin{remark}
It is not valid to have a variable both undefined and useful. In other words, it is not permitted in \x3 to have an undefined variable that is involved in a constraint or an objective. 
\end{remark}

To summarize, undefined variables must be ignored by solvers, and useless variables can be discarded.
We might consider introducing in the future new elements, attributes or annotations for clearly specifying which variables are useless, 
but note that it can be easily identified by parsers.
The tool for checking solutions (and their costs) that we have developped are able to handle the presence or the absence of useless variables.
So, users are given the possibility to submit complete instantiations or partial instantiations only involving useful variables.

\section{Solutions}\label{sec:solutions}

In \x3, we can also represent solutions, i.e., \x3 output.
We simply use an element \xml{instantiation} that gives a value for each (useful) variable of the instance.
More precisely, to define a solution, we introduce two elements \xml{list} and \xml{values} inside the element \xml{instantiation}: the ith variable of \xml{list} is assigned the ith value of \xml{values}. Obviously, it is not possible to have several occurrences of the same variable inside \xml{list}.

\begin{xl}
There is a required attribute \att{type}. 
For decision problems, e.g., CSP or SCSP, its value is necessarily \val{solution}.
For optimization problems, e.g., COP or WCSP, its value is either \val{solution} or \val{optimum}, and another required attribute \att{cost} gives the cost of the (optimal) solution.
The syntax is given below for integer variables only.
\end{xl}
\begin{xc}
There is a required attribute \att{type}. 
For decision problems, i.e., CSP, its value is necessarily \val{solution}.
For optimization problems, i.e., COP, its value is either \val{solution} or \val{optimum}, and another required attribute \att{cost} gives the cost of the (optimal) solution.
The syntax is given below for integer variables only.
\end{xc}
  
\begin{boxsy}
\begin{syntax} 
<instantiation type="solution|optimum" [cost="integer"] >
   <list> (intVar wspace)+ </list>
   <values> (intVal wspace)+ </values>
</instantiation>
\end{syntax}
\end{boxsy}

As an illustration, let us consider the optimization problem from Chapter \ref{cha:intro}.
The optimal solution can be represented by:

\begin{boxex}
\begin{xcsp} 
<instantiation type="optimum" cost="1700">
   <list> b c </list>
   <values> 2 2 </values>
</instantiation>
\end{xcsp}
\end{boxex}

\begin{xl}
Of course, although the syntax is not developed here (because it is immediate), for solutions, it is possible to deal with other types of variables:
\begin{itemize}
\item for symbolic variables, values are represented by symbols as e.g., \nn{green}, \nn{lucie} or \nn{b},
\item for real variables, values are represented by real numbers or intervals as e.g., $7.2$ or $[3.01,3.02]$,
\item for set variables, values are represented by set values, as e.g., $\{2,3,7\}$ or $\{\}$.
\end{itemize}
\end{xl}

Importantly, when some useless variables are present in an instance, the user is offered three possibilities to deal with them:
\begin{itemize}
\item the useless variables (and of course their values) are not listed,
\item the useless variables are given the special value '*',
\item the useless variables are given any value from their domains.
\end{itemize}

Let us illustrate this with a small CSP example.
The following instance involves an array of 4 variables, but one of them, $x[3]$, is useless. 

\begin{boxex}
\begin{xcsp}
<instance format="XCSP3" type="CSP">
  <variables>
    <array id="x" size="4"> 1..3 </array>
  </variables>
  <constraints>
    <intension> eq(add(x[0],x[1]),x[2]) </intension>
  </constraints>
</instance>
\end{xcsp} 
\end{boxex}

To represent the solution $x[0]=1, x[1]=1, x[2]=2$, one can choose between the three following representations:

\begin{boxex}
\begin{xcsp} 
<instantiation type="solution">
   <list> x[0] x[1] x[2] </list>
   <values> 1 1 2 </values>
</instantiation>
\end{xcsp}
\end{boxex}

\begin{boxex}
\begin{xcsp} 
<instantiation type="solution">
   <list> x[] </list>
   <values> 1 1 2 * </values>
</instantiation>
\end{xcsp}
\end{boxex}

\begin{boxex}
\begin{xcsp} 
<instantiation type="solution">
   <list> x[] </list>
   <values> 1 1 2 1 </values>
</instantiation>
\end{xcsp}
\end{boxex}

For the last representation, we could have chosen the value 2 or the value 3 for the last variable $x[3]$.
Note that the interest of these different forms clearly depends on the context.  

\begin{remark}
Interestingly, a solution can be perceived as a constraint; see Section \ref{ctr:instantiation}.
The attributes are thus simply ignored.
\end{remark}


Since Specifications 3.0.7, one may use compact forms of integer sequences.
This is the case for:
\begin{itemize}
\item \xml{lengths} in \xml{ordered}
\item \xml{coeffs} in \xml{sum}
\item \xml{lengths} in \xml{noOverlap}
\item \xml{lengths} and \xml{heights} in \xml{cumulative}
\item \xml{sizes} in \xml{binPacking}
\item \xml{weights} and \xml{profits} in \xml{knapsack}
\item \xml{weights} and \xml{balance} in \xml{flow}
\item \xml{values} in \xml{instantiation}
\end{itemize}

We can then write $v$x$k$ for the integer $v$ occurring $k$ times in sequence.
The last example above can then be equivalently written:
\begin{boxex}
\begin{xcsp} 
<instantiation type="solution">
   <list> x[] </list>
   <values> 1x2 2 1 </values>
</instantiation>
\end{xcsp}
\end{boxex}

\chapter{\textcolor{gray!95}{Objectives}}\label{cha:objectives}

There are two kinds of elements that can be used for representing objectives.
You can use:
\begin{itemize}
\item an element \xml{minimize}
\item or an element \xml{maximize}
\end{itemize}

These elements must be put inside \xml{objectives}.
\begin{xl} The role of the attribute \att{combination} will be explained at the end of this chapter.\end{xl}
\begin{xc}For \x3-core, only mono-objective optimization is considered, meaning that only one element can be present inside \xml{objectives}.\end{xc}

\begin{xl}
\begin{boxsy}
\begin{syntax} 
<objectives [combination="combinationType"]>
  (<minimize.../> | <maximize.../>)+ 
</objectives>
\end{syntax}
\end{boxsy}
\end{xl}
\begin{xc}
\begin{boxsy}
\begin{syntax} 
<objectives>
  (<minimize.../> | <maximize.../>)+ 
</objectives>
\end{syntax}
\end{boxsy}
\end{xc}

Each element \xml{minimize} and \xml{maximize} has an optional attribute \att{id} and an optional attribute \att{type}, whose value can currently be:
\begin{itemize}
\item \val{expression}
\item \val{sum} 
\item \val{minimum} 
\item \val{maximum}
\item \val{nValues}
\item \val{lex}
\end{itemize}

\section{Objectives in Functional Forms}

\begin{xl}
The default value for \att{type} is \val{expression}, meaning that the content of the element \xml{minimize} or \xml{maximize} is necessarily a numerical functional expression (of course, possibly just a variable) built from operators described in Tables \ref{tab:semanticsi} and \ref{tab:semanticss} for COP instances (that deal with integer values only), and Table \ref{tab:semanticsf} for NCOP instances (that deal with real values).
\end{xl}
\begin{xc}
The default value for \att{type} is \val{expression}, meaning that the content of the element \xml{minimize} or \xml{maximize} is necessarily a numerical functional expression (of course, possibly just a variable) built from operators described in Table \ref{tab:semanticsi}.
\end{xc}

An objective in functional form is thus defined by an element \xml{minimize} or \xml{maximize}.
We only give the syntax of \xml{minimize} (as the syntax for \xml{maximize} is quite similar) for COP instances.
Here, an integer functional expression is referred to as \bnf{intExpr} in the syntax box below; its precise syntax is given in Appendix \ref{cha:bnf}.

\begin{boxsy}
\begin{syntax} 
<minimize [id="identifier"] [type="expression"]>
  intExpr
</minimize>
\end{syntax}
\end{boxsy}

\begin{xl}
For NCOP instances, just replace above \verb!intExpr! by \verb!realExpr!. 
\end{xl}

An example is given below for an objective \nn{obj1} that consists in minimizing the value of the variable $z$, and an objective \nn{obj2} that consists in maximizing the value of the expression $x+2y$. 

\begin{boxex}
\begin{xcsp}
<minimize id="obj1"> z </minimize>
<maximize id="obj2"> add(x,mul(y,2)) </maximize>
\end{xcsp}
\end{boxex}

This way of representing objectives is generic, but when possible, it is better to use specialized forms in order to simplify the \x3 code and also to inform directly solvers of the nature of the objective(s).

\section{Objectives in Specialized Forms}

Whatever is the type among \val{sum}, \val{minimum}, \val{maximum}, \val{nValues} and \val{lex}, two forms are possible.
We show this for the element \xml{minimize}, but of course, this is quite similar for the element \xml{maximize}. 
Here, we give the syntax for COP instances:

\begin{boxsy}
\begin{syntax} 
<minimize [id="identifier"] type="sum|minimum|maximum|nValues|lex">
   <list> (intVar wspace)2+ </list>
   [<coeffs> (intVal wspace)2+ | (intVar wspace)2+ </coeffs>]
</minimize>
\end{syntax}
\end{boxsy}

\begin{xl}
For NCOP instances, you just have to replace \verb!(intVar wspace)2+! and \verb!(intVal wspace)2+! by \verb!(realVar wspace)2+! and \verb!(realVal wspace)2+!.
\end{xl}

There is one possible coefficient per variable.
When the element \xml{coeffs} is absent, coefficients are all assumed to be equal to 1, and the opening/closing tags for \xml{list} become optional, which gives:

\begin{boxsy}
\begin{syntax} 
<minimize [id="identifier"] type="sum|minimum|maximum|nValues|lex">
   (intVar wspace)2+ @\com{Simplified Form}@
</minimize>
\end{syntax}
\end{boxsy}

For the semantics, we consider that $X=\langle x_1,x_2,\ldots,x_k\rangle$ and $C=\langle c_1,c_2,\ldots,x_k\rangle$ denote the lists of variables and coefficients.
Also, minimize$_{lex}$ denotes minimization over tuples when considering the lexicographic order.
Finally, the type is given as third argument of elements $\gb{minimize}$ below.

\begin{boxse}
\begin{semantics}
$\gb{minimize}(X,C,\nm{sum})$  : minimize $\sum_{i=1}^{|X|} c_i \times x_i$
$\gb{minimize}(X,C,\nm{minimum})$ : minimize $\min_{i=1}^{|X|} c_i \times x_i$
$\gb{minimize}(X,C,\nm{maximum})$ : minimize $\max_{i=1}^{|X|} c_i \times x_i$
$\gb{minimize}(X,C,\nm{nValues})$ : minimize $|\{c_i \times x_i : 1 \leq i \leq |X|\}|$
$\gb{minimize}(X,C,\nm{lex})$ : minimize$_{lex}$ $\langle c_1 \times x_1, c_2 \times x_2, \ldots, c_k \times x_k \rangle$ 
\end{semantics}
\end{boxse}

The following example shows an objective \texttt{obj1} that consists in minimizing $2x_1 + 4x_2 + x_3 + 4x_4 +8x_5$, and an objective \texttt{obj2} that consists in minimizing the highest value among those taken by variables $y_1,y_2,y_3,y_4$.

\begin{boxex}
\begin{xcsp}
<minimize id="obj1" type="sum"> 
  <list> x1 x2 x3 x4 x5 </list> 
  <coeffs> 2 4 1 4 8 </coeffs>
</minimize>
<minimize id="obj2" type="maximum"> 
  <list> y1 y2 y3 y4 </list> 
</minimize>
\end{xcsp}
\end{boxex}

Because the opening and closing tags of \xml{list} are optional here (as there is no element \xml{coeffs}), the objective \val{obj2} can be simply written:

\begin{boxex}
\begin{xcsp}
<minimize id="obj2" type="maximum"> y1 y2 y3 y4 </minimize>
\end{xcsp}
\end{boxex}


\paragraph{Generalized Forms of Objectives (Views).}
Any objective (in specialized form) defined with \xml{minimize} or \xml{maximize} always contains an element \xml{list}.
Instead of listing variables, it is also possible to list integer expressions:
\begin{quote}
\verb!<list> (intExpr wspace)2+ </list>!
\end{quote}

as for example in:

\begin{boxex}
\begin{xcsp}
<minimize type="maximum"> 
  <list> add(s[0],61) add(s[1],9) add(s[2],87) </list> 
</minimize>
\end{xcsp}
\end{boxex}

which can be simplified as:

\begin{boxex}
\begin{xcsp}
<minimize type="maximum">
  add(s[0],61) add(s[1],9) add(s[2],87)
</minimize>
\end{xcsp}
\end{boxex}

Similarly, instead of listing integers or variables, it is also possible to list integer expressions in the element \xml{coeffs}:

\begin{quote}
  \verb!<coeffs> (intExpr wspace)2+ </coeffs>!
\end{quote}

\begin{xl}
\section{Multi-objective Optimization}

When dealing with several objectives, it is possible to indicate how these objectives must be combined.
The element \xml{objectives} has an optional attribute \att{combination}.
Current possible values for this attribute are:
\begin{itemize}
\item \val{lexico}: the objectives are lexicographically ordered (according to their positions in \xml{objectives}).
\item \val{pareto}: no objective is more important than another one. 
\end{itemize}

Note that this attribute is forbidden when there is only one objective.
The default value is \val{pareto}. More values might be introduced in the future.

For the following example, we try first to minimize the value of $z$, and in case of equality, we try to maximize the value of the expression $x+(y-10)$.

\begin{boxex}
\begin{xcsp}
<objectives combination="lexico">
  <minimize id="obj1"> z </minimize>
  <maximize id="obj2"> add(x,sub(y,10)) </maximize>
</objectives>
\end{xcsp}
\end{boxex}
\end{xl}

\part{Constraints}

\begin{xl}
  In this part of the document, we show how basic forms of constraints are represented in \x3.
This part contains three chapters, as follows: 
\bigskip

\begin{tikzpicture}[dirtree, every node/.style={draw=black,thick,anchor=west}]
\tikzstyle{selected}=[fill=colorex]
\tikzstyle{optional}=[fill=gray!10]
 \node  [fill=gray!20] {{\bf Constraints}}
    child { node [selected] {Constraints over Simple Discrete Variables}
      child { node [optional] {Constraints over Integer Variables}}
      child { node [optional] {Constraints over Symbolic Variables}}
    }		
    child [missing] {}	
    child [missing] {}	
    child { node [selected] {Constraints over Complex Discrete Variables}
      child { node [optional] {Constraints over Set Variables}}
      child { node [optional] {Constraints over Graph Variables}}
    }		
    child [missing] {}	
    child [missing] {}	
    child { node [selected] {Constraints over Continuous Variables}
      child { node [optional] {Constraints over Real Variables}}
      child { node [optional] {Constraints over Qualitative Variables}}
    }
    ;
\end{tikzpicture}
\end{xl}
\begin{xc}
  In this part, we introduce the popular constraints identified in \x3-core. 
In the document describing the full \x3 specifications \cite{xcsp3}, you will find additional constraints.  
\end{xc}

\begin{xl}
\chapter{\textcolor{gray!95}{Constraints over Simple Discrete Variables}}\label{cha:constraints}
\end{xl}
\begin{xc}
\chapter{\textcolor{gray!95}{Constraints of \x3-core}}\label{cha:constraints}
\end{xc}

The element \xml{constraints} may contain many kinds of elements, as \bnfX{constraint} denotes any kind of stand-alone constraint as shown in Appendix \ref{cha:bnf}.
For example, this can be \norX{sum},  \norX{intension}, \norX{regular}, \norX{allDifferent}, and so on.\begin{xl} In this chapter, we are interested in constraints over simple discrete variables that are either integer or symbolic variables.
  First, we focus on (elementary) constraints that involve integer variables.\end{xl}\begin{xc} In this chapter, we are interested in constraints over integer variables.\end{xc} Of course, this includes constraints involving 0/1 variables, which also serve in our format as Boolean variables.
\begin{xl} Some of these constraints are also appropriate for symbolic variables, as we shall see at the end of this chapter.
Constraints involving set, graph and continuous variables will be introduced in next chapters.
\end{xl}

Each constraint corresponds to an XML element 
that is put inside the element \xml{constraints}.
Each constraint has an optional attribute \att{id}, and contains one or several XML elements that can be seen as the parameters of the constraint.
A constraint parameter is an XML element that usually contains a simple term like a numerical value (possibly, an interval), a variable id, or a more complex term like a list of values, a list of (ids of) variables or a list of tuples. 

\begin{xl}
Note that many global constraints have been introduced in CP; see \cite{HK_global,R_globalsurvey,BCR_global}.
For a well-detailed documentation, we solicit the reader to consult the \href{https://sofdem.github.io/gccat/}{global constraint catalog}.
Below, we introduce constraints per family as shown by Figure \ref{fig:intCtrs}.

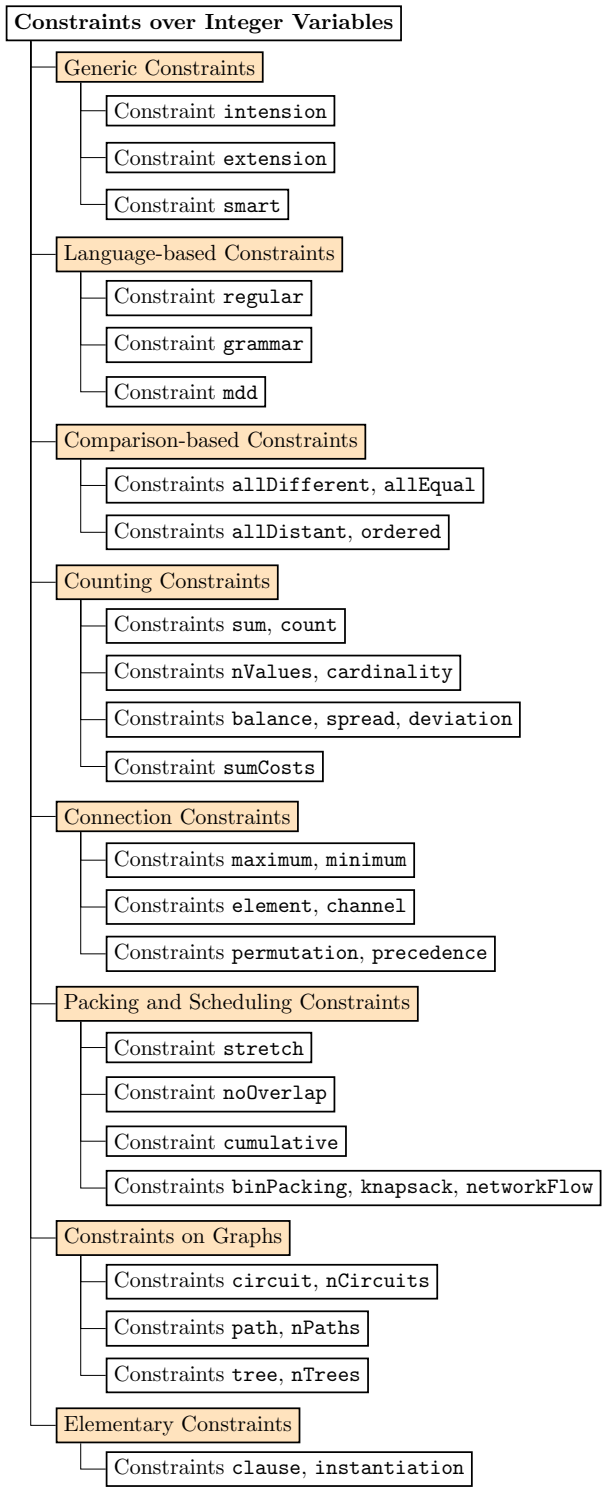
\begin{figure}[p]
\resizebox{!}{0.97\textheight}{
\begin{tikzpicture}[dirtree,every node/.style={draw=black,thick,anchor=west}] 
\tikzstyle{selected}=[fill=colorex]
\tikzstyle{optional}=[fill=gray!12]
\node {{\bf Constraints over Integer Variables}}
    child { node [selected] {Generic Constraints}
      child { node {Constraint \gb{intension}}}
      child { node {Constraint \gb{extension}}}
    }		
    child [missing] {}	
    child [missing] {}				
    child { node [selected] {Language-based Constraints}
      child { node {Constraint \gb{regular}}}
      child { node {Constraint \gb{grammar}}}
      child { node {Constraint \gb{mdd}}}
    }		
    child [missing] {}	
    child [missing] {}	
    child [missing] {}		
    child { node [selected] {Comparison-based Constraints}
      child { node {Constraints \gb{allDifferent}, \gb{allEqual}}}
      child { node {Constraints \gb{allDistant}, \gb{ordered}}}
      child { node {Constraint \gb{precedence}}}
    }
    child [missing] {}	
    child [missing] {}	
    child [missing] {}		
    child { node [selected] {Counting Constraints}
      child { node {Constraints \gb{sum}, \gb{count}}}
      child { node {Constraints \gb{nValues}, \gb{cardinality}}}
      child { node {Constraints \gb{balance}, \gb{spread}, \gb{deviation}}}
      child { node {Constraint \gb{sumCosts}}}
    } 
    child [missing] {}	
    child [missing] {}
    child [missing] {}
    child [missing] {}	
    child { node [selected] {Connection Constraints}
      child { node {Constraints \gb{maximum}, \gb{minimum}, \gb{maximumArg}, \gb{minimumArg}}}
      child { node {Constraints \gb{element}, \gb{channel}}} 
      child { node {Constraint \gb{permutation}}}
    } 
    child [missing] {}
    child [missing] {}	
    child [missing] {}
    child { node [selected] {Packing and Scheduling Constraints}
      child { node {Constraint \gb{stretch}}}
      child { node {Constraint \gb{noOverlap}}} 
      child { node {Constraint \gb{cumulative}}}
      child { node {Constraints \gb{binPacking}, \gb{knapsack}, \gb{flow} }}
    } 
    child [missing] {}
    child [missing] {}	
    child [missing] {}	
    child [missing] {} 
    child { node [selected] {Constraints on Graphs}
      child { node {Constraints \gb{circuit}, \gb{nCircuits}}}
      child { node {Constraints \gb{path}, \gb{nPaths}}} 
      child { node {Constraints \gb{tree}, \gb{nTrees}}}
    } 
    child [missing] {}	
    child [missing] {}	
    child [missing] {} 
    child { node [selected] {Elementary Constraints}
     child { node {Constraints \gb{clause}, \gb{instantiation}}}
    }
    ;
\end{tikzpicture}
}
\caption{The different types of constraints over integer variables.\label{fig:intCtrs}} 
\end{figure}
\end{xl}

\medskip
For the syntax, we invite the reader to consult Chapter \ref{cha:intro} and Appendix \ref{cha:bnf}.
Recall that BNF non-terminals are written in dark blue italic form, such as for example \bnf{intVar} and \bnf{intVal} that respectively denote an integer variable and an integer value.  
In \x3, an \bnf{intVar} corresponds to the id of a variable declared in \xml{variables} and an \bnf{intVal} corresponds to a value in $\N$.
Note that when the value of a parameter can be an \bnf{intVal} or an \bnf{intVar}, we only give the semantics for \bnf{intVar}.
To simplify the presentation, we omit to specify \verb![id="identifier"]! when introducing the various types of constraints.

\section{Constraints over Integer Variables}

\subsection{Generic Constraints}

In this section, we present general ways of representing constraints, namely, intentional and extensional forms of constraints.
\begin{xl}
  We also include so-called hybrid forms of extensional constraints (initially called smart constraints \cite{MDL_smart}) that represent a kind of hybrization between intentional and extensional forms.
\end{xl}
The elements we introduce are:
\begin{enumerate}
\item \gb{intension}
\item \gb{extension}
\end{enumerate}

\subsubsection{Constraint \gb{intension}}\label{ctr:intension}

Intentional constraints form an important type of constraints.
They are defined from Boolean expressions, usually called predicates.
For example, the constraint $x+y=z$ corresponds to an equation that is an expression evaluated to $\nm{true}$ or $\nm{false}$ according to the values assigned to the variables $x$, $y$ and $z$.
Predicates are represented under functional form in \x3: the function name appears first, followed by the operands between parenthesis (with comma as a separator). 
The \x3 representation of the constraint $x+y=z$ is $eq(add(x,y),z)$. 
Operators on integers (including Booleans since we assume that $\nm{false} = 0$ and $\nm{true} = 1$) that can be used to build predicates are presented in Table \ref{tab:semanticsi}.
When we refer to a Boolean operand or result, we mean either the integer value 0 or the integer value 1.
This allows us to combine Boolean expressions with arithmetic operators (for example, addition) without requiring any type conversions.
For example, it is valid to write $eq(add(lt(x,5),lt(y,z)),1)$ for stating that exactly one of the Boolean expressions $x<5$ and $y<z$ must be true, although it may be possible (and more relevant) to write it differently.

\begin{remark}
Everytime we are referring to (the result of) a Boolean expression, think about either the integer value 0 (false) or the integer value 1 (true).
\end{remark}

An intensional constraint is defined by an element \xml{intension}, containing an element \xml{function} that describes the functional representation of the predicate, referred to as \bnf{boolExpr} in the syntax box below, and whose precise syntax is given in Appendix \ref{cha:bnf}. 

\begin{boxsy}
\begin{syntax} 
<intension>
  <function> boolExpr </function>
</intension>
\end{syntax}
\end{boxsy}

Note that the opening and closing tags for \xml{function} are optional, which gives:

\begin{boxsy}
\begin{syntax} 
<intension> boolExpr </intension> @\com{Simplified Form}@
\end{syntax}
\end{boxsy}

\begin{table}[h!]
\begin{center}
{\small
\begin{tabular}{cccc} 
\rowcolor{v2lgray}{} {\textcolor{dred}{\bf Operation}} &  {\textcolor{dred}{\bf Arity}} &  {\textcolor{dred}{\bf Syntax}} &  {\textcolor{dred}{\bf Semantics}} \\
\multicolumn{2}{c}{ } \\
\multicolumn{4}{l}{\textcolor{dred}{Arithmetic (integer operands and integer result)}} \\
\midrule
Opposite  & 1 & neg($x$) & $-x$ \\
Absolute Value &  1 & abs($x$) & $| x |$ \\
Addition    & $r \geq 2$ & add($x_1,\ldots,x_r$) & $x_1 + \ldots + x_r$ \\
Subtraction & 2 & sub($x,y$) & $x - y$ \\
Multiplication & $r \geq 2$ & mul($x_1,\ldots,x_r$) & $x_1 * \ldots * x_r$ \\
Integer Division & 2 & div($x,y$) & $x / y$ \\
Remainder  & 2 & mod($x,y$) & $x \% y$ \\ 
Square  & 1 & sqr($x$) & $x^2$ \\
Power  & 2 & pow($x,y$) & $x^{y}$ \\
Minimum & $r \geq 2$ & min($x_1,\ldots,x_r$) & $\min \{x_1,\ldots,x_r\}$ \\
Maximum & $r \geq 2$ & max($x_1,\ldots,x_r$) & $\max \{x_1,\ldots,x_r\}$ \\
Distance  & 2 & dist($x,y$) & $| x - y |$ \\
\midrule
\multicolumn{2}{c}{ } \\
\multicolumn{4}{l}{\textcolor{dred}{Relational (integer operands and Boolean result)}} \\
\midrule
Less than     & 2 & lt(x,y) & $x < y$ \\
Less than or equal     & 2 & le($x,y$) & $x \leq y$ \\
Greater than or equal     & 2 & ge($x,y$) & $x \geq y$ \\
Greater than     & 2 & gt($x,y$) & $x > y$ \\
Different from    & 2 &  ne($x,y$) & $x \neq y$ \\
Equal to    & $r \geq 2$ & eq($x_1,\ldots,x_r$) & $x_1 = \ldots = x_r$ \\
\midrule
\multicolumn{2}{c}{ } \\
\multicolumn{4}{l}{\textcolor{dred}{Set ($a_i$: integers, $s$: set of integers (no variable permitted), $x$: integer operand)}} \\
\midrule
Empty set & 0 & set() & $\emptyset$ \\
Non-empty set & $r > 0$ & set$(a_1,\ldots,a_r)$ & $\{a_1,\ldots,a_r\}$ \\
Membership & 2 & in($x,s$) & $x \in s$ \\
\midrule
\multicolumn{2}{c}{ } \\
\multicolumn{4}{l}{\textcolor{dred}{Logic (Boolean operands and Boolean result)}} \\
\midrule
Logical not     & 1 & not($x$) & $\lnot x$ \\
Logical and    & $r \geq 2$ &  and($x_1,\ldots,x_r$) & $x_1 \land \ldots \land x_r$ \\
Logical or    & $r \geq 2$ & or($x_1,\ldots,x_r$) & $x_1 \lor \ldots \lor x_r$ \\
Logical xor    & $r \geq 2$ & xor($x_1,\ldots,x_r$) & $x_1 \oplus \ldots \oplus x_r$ \\
Logical equivalence& $r \geq 2$ & iff($x_1,\ldots,x_r$) & $x_1 \Leftrightarrow \ldots \Leftrightarrow x_r$ \\
Logical implication & 2 & imp($x,y$) & $x \Rightarrow y$ \\
\midrule
\multicolumn{2}{c}{ } \\
\multicolumn{4}{l}{\textcolor{dred}{Control}} \\
\midrule
Alternative     & 3 & if($b,x,y$) & value of $x$, if $b$ is true, \\
                &   &           & value of $y$, otherwise     \\
\bottomrule
\end{tabular}
}
\end{center}
\caption{Operators on integers that can be used to build predicates.
  As Boolean values $\nm{false}$ and $\nm{true}$ are represented by integer values $0$ and $1$, when an integer (operand/result) is expected, one can provide a Boolean value.
  In other words, ``integer'' encompasses ``Boolean''. For example, for arithmetic operators, operands can simply be 0/1 (Boolean). 
\label{tab:semanticsi}}
\end{table}


Below, $P$ denotes a predicate expression with $r$ formal parameters (not shown here, for simplicity), $X=\langle x_1,x_2,\ldots,x_r \rangle$ a sequence of $r$ variables, the scope of the constraint, and $P(\va{x}_1,\va{x}_2,\ldots,\va{x}_r)$ the value (0 or 1) returned by $P$ for any instantiation of the variables of $X$. 

\begin{boxse}
\begin{semantics}
  $\gb{intension}(X,P)$, with $X=\langle x_1,x_2,\ldots,x_r \rangle$ and $P$ a predicate iff 
  $P(\va{x}_1,\va{x}_2,\ldots,\va{x}_r) = 1$  @\com{recall that 1 stands for true}@
\end{semantics}
\end{boxse}

For example, for constraints $c_1: x+y=z$ and $c_2: w \geq z$, we have: 

\begin{boxex}
\begin{xcsp}
<intension id="c1"> 
  <function> eq(add(x,y),z) </function>
</intension> 
<intension id="c2"> 
  <function> ge(w,z) </function>
</intension>
\end{xcsp}
\end{boxex}

or, equivalently, in simplified form:

\begin{boxex}
\begin{xcsp}
<intension id="c1"> eq(add(x,y),z) </intension> 
<intension id="c2"> ge(w,z) </intension>
\end{xcsp}
\end{boxex}

Most of the time, intensional constraints correspond to primitive constraints which admit specialized filtering algorithms, such as the binary operators $=, \neq, <, \ldots$ 
When parsing, it is rather easy to identify those primitive constraints.

\begin{remark}
It is not possible to use compact lists of array variables (see Chapter \ref{cha:variables})\begin{xc}.\end{xc}\begin{xl} and  \val{+/-infinity} values in predicates.\end{xl}
\end{remark}

\begin{remark}\label{rem:divneg}
Building expressions that involve integer division (with operator \texttt{div} or \texttt{mod}) where either operand can be negative is {\bf strongly discouraged}.
In case of such a situation, the rule is ``rounding towards 0'' (as in C or Java).
Do note that language designers had to choose if their language will round towards zero, negative infinity, or positive infinity when doing integer division. 
\end{remark}

\subsubsection{Constraint \gb{extension}}\label{ctr:extension}

Extensional constraints, also called table constraints form another important type of constraints.
They are defined by enumerating the list of tuples that are allowed (supports) or forbidden (conflicts).
Some algorithms for binary extensional constraints are AC3 \cite{M_AC3}, AC4 \cite{MH_AC4}, AC6 \cite{B_AC6}, AC2001 \cite{BRYZ_optimal}, AC3$^{rm}$ \cite{LH_study} and AC3$^{bit+rm}$ \cite{LV_enforcing}. 
Some algorithms for non-binary extensional constraints are GAC-schema \cite{BR_GAC7}, GAC-nextIn \cite{LR_fast}, GAC-nextDiff \cite{GJMN_data}, GAC-va \cite{LS_generalized}, STR2 \cite{L_str2}, AC5TC-Tr \cite{MHD_optimal}, STR3 \cite{LLY_path} and GAC4r \cite{PR_GAC4}.
The state-of-the-art algorithm is CT (Compact-Table), as described in \cite{DHLPPRS_efficiently,VLS_extending}; a filtering algorithm based on similar principles has been independently proposed in \cite{WXYL_optimizing}.

An extensional constraint contains then two elements.
The first element is an element \xml{list} that indicates the scope of the constraint (necessarily a list of variable ids).
The second element is either an element \xml{supports} or an element \xml{conflicts}, depending on the semantics of the {\em ordinary} tuples that are listed in lexicographic order within the element.

For classical non-unary {\em positive} table constraints, we have:

\begin{boxsy}
\begin{syntax} 
<extension>
  <list> (intVar wspace)2+ </list>
  <supports> ("(" intVal ("," intVal)+ ")")* </supports>
</extension>
\end{syntax}
\end{boxsy}

\begin{boxse}
\begin{semantics}
$\gb{extension}(X,S)$, with $X=\langle x_1,x_2,\ldots,x_r \rangle$ and $S$ the set of supports, iff 
  $\langle \va{x}_1,\va{x}_2,\ldots,\va{x}_r  \rangle \in S$ 

$\gbc{Prerequisite}: \forall \tau \in S, |\tau|=|X| \geq 2$
\end{semantics}
\end{boxse}

For classical non-unary {\em negative} table constraints, we have:

\begin{boxsy}
\begin{syntax} 
<extension>
  <list> (intVar wspace)2+ </list>
  <conflicts> ("(" intVal ("," intVal)+ ")")* </conflicts>
</extension>
\end{syntax}
\end{boxsy}

\begin{boxse}
\begin{semantics}
$\gb{extension}(X,C)$, with $X=\langle x_1,x_2,\ldots,x_r \rangle$ and $C$ the set of conflicts, iff
  $\langle \va{x}_1,\va{x}_2,\ldots,\va{x}_r \rangle \notin C$ 

$\gbc{Prerequisite}: \forall \tau \in C, |\tau|=|X| \geq 2$
\end{semantics}
\end{boxse}

Here is an example with a ternary constraint $c_1$ defined on scope $\langle x_1,x_2,x_3 \rangle$ with supports $\{\langle 0,1,0\rangle,\langle 1,0,0\rangle,\langle 1,1,0\rangle,\langle 1,1,1\rangle\}$ and a quaternary constraint $c_2$ defined on scope $\langle y_1,y_2,y_3,y_4 \rangle$ with conflicts $\{\langle 1,2,3,4\rangle,\langle 3,1,3,4 \rangle\}$. 

\begin{boxex}
\begin{xcsp}
<extension id="c1">
  <list> x1 x2 x3 </list>
  <supports> (0,1,0)(1,0,0)(1,1,0)(1,1,1) </supports> 
</extension>
<extension id="c2">   
  <list> y1 y2 y3 y4 </list>
  <conflicts> (1,2,3,4)(3,1,3,4) </conflicts> 
</extension>
 \end{xcsp}
\end{boxex}
 
Quite often, when modeling, there are groups of extensional constraints that share the same relations.\begin{xl}
It is then interesting to exploit the concept of syntactic group, presented in Section \ref{sec:group}, or if not appropriate, the attribute \att{as} described in Section \ref{sec:as}.
\end{xl}\begin{xc}
It is then interesting to exploit the concept of syntactic group, presented in Section \ref{sec:group}.
\end{xc}
  
For unary table constraints, one uses a simplified form: supports and conflicts are indeed defined exactly as the domains of integer variables.
It means that we directly put integer values and intervals in increasing order within \xml{supports} and \xml{conflicts}.

For unary positive table constraints, we have:

\begin{boxsy}
\begin{syntax} 
<extension>
  <list> intVar </list>
  <supports>  ((intVal | intIntvl) wspace)* </supports>
</extension>
\end{syntax}
\end{boxsy}

For unary negative table constraints, we have:

\begin{boxsy}
\begin{syntax} 
<extension>
  <list> intVar </list>
  <conflicts>  ((intVal | intIntvl) wspace)* </conflicts>
</extension>
\end{syntax}
\end{boxsy}

The semantics is immediate.
As an illustration, the constraint $c_3$ corresponds to $x \in \{1,2,4,8,9,10\}$.
\begin{boxex}
\begin{xcsp}
<extension id="c3">
  <list> x </list>
  <supports> 1 2 4 8..10 </supports>
</extension>
 \end{xcsp}
\end{boxex} 

We have just considered ordinary tables that are tables containing {\em ordinary} tuples, i.e., classical sequences of values as in $(1,2,0)$.
\begin{xl}
Short tables \cite{NGJM_short} can additionally contain {\em short} tuples, which are tuples involving the special symbol $*$ as in $(0,*,2)$, and compressed tables \cite{KW_compression,XY_optimizing} can additionally contain {\em compressed} tuples, which are tuples involving sets of values as in $(0,\{1,2\},3)$. 
We now introduce the \x3 representation of these extended forms of (non-unary) extensional constraints. 
\end{xl}
\begin{xc}
Short tables \cite{NGJM_short} can additionally contain {\em short} tuples, which are tuples involving the special symbol $*$ as in $(0,*,2)$.
\end{xc}
For short table constraints, the syntax is obtained from that of non-unary ordinary table constraints given at the beginning of this section, by replacing  \bnf{intVal} by \bnf{intValShort} in elements \xml{supports} and \xml{conflicts}; \bnf{intValShort} is defined by \verb!intVal | "*"! in Appendix \ref{cha:bnf}.
The semantics is such that everytime the special symbol "*" occurs in a tuple any possible value is accepted for the associated variable(s). 
For example, the following constraint $c_4$ has two short tuples, $\langle 1,*,1,2\rangle$ and $\langle 2,1,*,*\rangle$.
Assuming that the domain of all variables is $\{1,2\}$, the first short tuple is equivalent to the two ordinary tuples $\langle 1,1,1,2\rangle$ and $\langle 1,2,1,2\rangle$ whereas the second short tuple is equivalent to the four ordinary tuples $\langle 2,1,1,1\rangle$, $\langle 2,1,1,2\rangle$, $\langle 2,1,2,1\rangle$ and $\langle 2,1,2,2\rangle$.

\begin{boxex}
\begin{xcsp}
<extension id="c4">
  <list> z1 z2 z3 z4 </list>
  <supports> (1,*,1,2)(2,1,*,*) </supports>
</extension>
 \end{xcsp}
\end{boxex} 

\begin{xl}
For compressed table constraints, the syntax is obtained from that of non-unary ordinary table constraints given at the beginning of this section, by replacing  \bnf{intVal} by \bnf{intValCompressed} in elements \xml{supports} and \xml{conflicts}; \bnf{intValCompressed} is defined by \verb!intVal | "*" | setVal! in Appendix \ref{cha:bnf}.
The semantics is such that everytime a set of values occurs in a tuple any value from this set is accepted for the associated variable(s). 
The set of ordinary tuples represented by a compressed tuple $\tau$ is then the Cartesian product built from the components of $\tau$.
For example, in constraint $c_5$, $\langle 0,\{1,2\},0,\{0,2\}\rangle$ is a compressed tuple representing the set of ordinary tuples in $\{0\}\times\{1,2\}\times\{0\}\times\{0,2\}$, which contains $\langle 0,1,0,0 \rangle$, $\langle 0,1,0,2 \rangle$, $\langle 0,2,0,0 \rangle$ and $\langle 0,2,0,2 \rangle$.

\begin{boxex}
\begin{xcsp}
<extension id="c5">
  <list> w1 w2 w3 w4 </list>
  <supports> (0,{1,2},0,{0,2})(1,0,1,2)(2,{0,2},0,2) </supports>
</extension>
 \end{xcsp}
\end{boxex} 
\end{xl}

\begin{remark}
In any table, ordinary tuples must always be given in lexicographic order, without any repetitions.
\end{remark}

\begin{xl}
\subsubsection{Hybrid Forms of Constraint \gb{extension}}\label{ctr:hybrid}

An hybrid constraint \gb{extension} (also called constraint \gb{smart} \cite{MDL_smart}) is defined from an hybrid table, authorizing entries to contain simple arithmetic restrictions (which can be seen as intern constraints).
An hybrid table contains then some hybrid tuples (at least, one), which involve restrictions of the form: 
\begin{enumerate}
\item $x \odot a$   
\item $x \in S$,
\item $x \in a..b$,
\item $x \not\in S$
\item $x \not\in a..b$  
\item $x \odot y$ 
\item $x \odot y + a$
\item $x \odot y - a$
\item $x \odot y + z$
\item $x \odot y - z$
\end{enumerate}
where $x$, $y$ and $z$ are variables in the scope of the constraint, $a$ and $b$ are integers, $S$ is a set of integers, $a..b$ is an integer interval and $\odot$ is a relational operator in the set $\{<,\leq,\geq,>, =,\neq\}$.

\medskip\noindent The semantics is the following: an ordinary tuple $\tau$ is allowed by an hybrid constraint \gb{extension} $c$ iff there exists at least one tuple in the table of $c$ such that $\tau$ satisfies that tuple, including all restrictions (if any) imposed on it.
Currently, there are two levels of hybridisation that can be subject to a filtering algorithm (enforcing generalized arc-consistency):
\begin{itemize}
\item level 1 of hybridisation, where restrictions correspond to unary constraints only; referred to as basic smart constraints in \cite{VLDS_extendingBasic};
\item level 2 of hybridisation, where some restrictions correspond to binary constraints; referred to as acyclic smart constraints in \cite{MDL_smart}.
\end{itemize}

\medskip\noindent Here are the rules for representing in \x3 such restrictions in an hybrid tuple:
\begin{itemize}
\item the left operand ($x$) is never represented as it corresponds to the variable associated with the position in the tuple where the restriction is put
\item the operator $\in$ is never represented
\item the operator $\not\in$ is represented by the UTF-16 character whose hexadecimal value is 2201 
\item $S$ is naturally represented as a set, between brackets, as e.g., in $\{2,3,8\}$
\item $a..b$ is naturally represented as an interval, as e.g., in $2..10$  
\item we use the following UTF-16 characters (values in hexadecimal here) to represent the relational operators:
  \begin{itemize}
    \item $\neq$ : 2260
    \item $<$ : FE64
    \item $\leq$ : 2264
    \item $\geq$ : 2265
    \item $>$ : FE65
  \end{itemize}
\item the variable $y$ (resp., $z$) occurring in the right operand of binary restrictions is represented by the character \% followed by the index of $y$ (resp., $z$) in the scope of the constraint. 
\item an attribute \att{type} is added to the element \xml{extension}; its value is either \val{hybrid-1} or \val{hybrid-2}, depending on the level of hybridisation of the table.
\end{itemize}

\medskip\noindent Sometimes, it is possible to transform ordinary tables into hybrid tables \cite{LKLD_automatic}.
Hybrid tables can be useful for modeling, permitting compact and structured representation of constraints. 
They also facilitate the reformulation of constraints, including well-known global constraints.

As a first illustration, assuming that $dom(x_i)=\{1,2,3\}, \forall i \in 1..3$, the following table constraint:
\begin{quote}
  \textcolor{dred}{$\langle x_1,x_2,x_3 \rangle \in \{(1,2,1), (1,3,1), (2,2,2), (2,3,2), (3,2,3), (3,3,3)\}$}
\end{quote}
can be represented by the following hybrid constraint $c_1$: 
\begin{center}
  \begin{tabular}{| >{\centering\arraybackslash}p{1.2cm} >{\centering\arraybackslash}p{1.2cm} >{\centering\arraybackslash}p{1.2cm} |}
    \multicolumn{1}{c}{$x_1$} & \multicolumn{1}{c}{$x_2$} & \multicolumn{1}{c}{$x_3$}\\			
    \hline
    $*$ & $\geq 2$ & $= x_1$ \\
    \hline
  \end{tabular}
\end{center}

or in another equivalent form:
\begin{quote}
  \textcolor{dred}{$\langle x_1,x_2,x_3 \rangle \in \{(*,\geq 2, \%0)\}$}
\end{quote}

which gives in \x3:

\begin{boxex}
\begin{xcsp}
<extension type="hybrid-2">
  <list> x1 x2 x3 </list>
  <supports> (*,@$\geq 2$@,%0) </supports> 
</extension>
\end{xcsp}
\end{boxex}

As a second illustration, let us consider a car configuration problem.
We assume that the cars to be configured have 2 colors (one for the body, $colB$, and the other for the roof, $\mathit{colR}$), a model number $mod$, an option pack $pck$, and an onboard computer $cmp$.
A configuration rule states that, for a particular model number $a=2$ and some fancy body color set $S=\{3,5\}$, an option pack less than a certain pack $b=4$ implies that the onboard computer cannot be the most powerful one, $c=3$, and that the roof color has to be the same as the body color. 
We obtain:
\begin{quote}
  \textcolor{dred}{$(mod = a \land colB \in S \land pck < b) \Rightarrow (cmp \neq c \land colR = colB)$}
\end{quote}
that can be represented under a disjunctive form:
\begin{quote}
  \textcolor{dred}{$mod \neq a \lor colB \not\in S \lor pck \geq b \lor (cmp \neq c \land colR = colB)$}
\end{quote}
which gives the following hybrid constraint $c_2$:
\begin{center}
  \begin{tabular}{| >{\centering\arraybackslash}p{1.2cm} >{\centering\arraybackslash}p{1.2cm} >{\centering\arraybackslash}p{1.2cm} >{\centering\arraybackslash}p{1.2cm} >{\centering\arraybackslash}p{1.2cm} |}
    \multicolumn{1}{c}{$mod$} & \multicolumn{1}{c}{$colB$} & \multicolumn{1}{c}{$colR$} & \multicolumn{1}{c}{$pck$} & \multicolumn{1}{c}{$cmp$} \\   				
    \hline
    $\neq 2$ & $\ast$ & $\ast$ & $\ast$ & $\ast$ \\
    $\ast$ & $\not\in \{3,5\}$ & $\ast$ & $\ast$ & $\ast$ \\
    $\ast$ & $\ast$ & $\ast$ & $\geq 4$ & $\ast$ \\
    $\ast$ & $\ast$ & $= colB$ & $\ast$ & $\neq 3$ \\
    \hline
  \end{tabular}
\end{center}

or in another equivalent form:
\begin{quote}
\textcolor{dred}{$\langle mod,colB,colR,pck,cmp \rangle \in \{  \\
 \textcolor{white}{white} (\neq 2,*,*,*,*), (*,\complement\{3,5\},*,*,*), (*,*,*,\geq 4,*), (*,*,\%1,*,\neq 3) \\\}$}
\end{quote}

which gives in \x3:

\begin{boxex}
\begin{xcsp}
<extension type="hybrid-2">
  <list> mod colB colR pck cmp </list>
  <supports>
    (@$\neq 2$@,*,*,*,*)(*,@$\complement\{3,5\}$@,*,*,*)(*,*,*,@$\geq 4$@,*)(*,*,%1,*,@$\neq 3$@)
  </supports> 
</extension>
\end{xcsp}
\end{boxex} 

Finally, as an illustration of a global constraint that can be reformulated as an hybrid table constraint, we have: 
\begin{quote}
  \textcolor{dred}{\gb{maximum}$(\langle x_1, x_2, x_3, x_4 \rangle)=y$}
\end{quote}
which gives the following hybrid constraint $c_3$:
\begin{center}
\begin{tabular}{| >{\centering\arraybackslash}p{1.2cm} >{\centering\arraybackslash}p{1.2cm} >{\centering\arraybackslash}p{1.2cm} >{\centering\arraybackslash}p{1.2cm} >{\centering\arraybackslash}p{1.2cm} |}
  \multicolumn{1}{c}{$x_1$} & \multicolumn{1}{c}{$x_2$} & \multicolumn{1}{c}{$x_3$} & \multicolumn{1}{c}{$x_4$} & \multicolumn{1}{c}{$y$} \\   				
  \hline
  $\ast$ & $\leq x_1$ & $\leq x_1$ & $\leq x_1$ & $= x_1$ \\
  $\leq x_2$ & $\ast$ &  $\leq x_2$ & $\leq x_2$ & $= x_2$ \\
  $\leq x_3$ & $\leq x_2$ & $\ast$ & $\leq x_2$ & $= x_3$ \\
  $\leq x_4$ & $\leq x_4$ & $\leq x_4$ & $\ast$ & $= x_4$ \\
  \hline
\end{tabular}
\end{center}

\end{xl}

\subsection{Constraints defined from Languages}

In this section, we present constraints that are defined from advanced data structures such as automatas and diagrams, which are structures exhibiting languages. More specifically, we introduce:
\begin{enumerate}
\item \gb{regular} 
\begin{xl} \item \gb{grammar}
\end{xl}
\item \gb{mdd}
\end{enumerate}

\subsubsection{Constraint \gb{regular}}\label{ctr:regular}

The constraint \gb{regular} \cite{CB_constraints,P_regular} ensures that the sequence of values assigned to the variables it involves forms a word that can be recognized by a deterministic (or non-deterministic) finite automaton.
The scope of the constraint is given by the element \xml{list}, and three elements, \xml{transitions}, \xml{start} and \xml{final}, are introduced for representing respectively the transitions, the start state and the final (accept) states of the automaton.
Note that the set of states and the alphabet can be inferred from \xml{transitions}.
When the automaton is non-deterministic, we can find two transitions $(q_i,a,q_j)$ and $(q_i,a,q_k)$ such that $q_j \neq q_k$. 
In what follows, \sy{state} stands for any identifier and \sy{states} for a sequence of identifiers (whitespace as separator). 

\begin{boxsy}
\begin{syntax} 
<regular> 
  <list> (intVar wspace)+ </list>
  <transitions> ("(" state "," intVal "," state ")")+ </transitions>
  <start> state </start>
  <final> (state wspace)+ </final>
</regular>
\end{syntax}
\end{boxsy}

Below, $L(A)$ denotes the language recognized by a deterministic (or non-deterministic) finite automaton $A$.

\begin{boxse}
\begin{semantics}
$\gb{regular}(X,A)$, with $X=\langle x_1,x_2,\ldots,x_r \rangle$ and $A$ a finite automaton, iff 
  $\va{x}_1\va{x}_2\ldots\va{x}_r \in L(A)$ 
\end{semantics}
\end{boxse}

\begin{center}
\begin{tikzpicture}[node distance=1cm, >=stealth, auto] 
 \tikzstyle{snode}=[state,minimum size=6mm]
 \node[snode, initial]            (a)                       {$a$};
 \node[snode]                     (b)[right=of a]          {$b$};
 \node[snode]                     (c)[right=of b]          {$c$};
 \node[snode]                     (d)[right=of c]          {$d$};
 \node[snode, accepting]          (e)[right=of d]          {$e$};
 \path[->] 
 (a)  edge[loop above]  node{0}  (a)
 (a)  edge node{1}   (b)
 (b)  edge node{1}   (c)
 (c)  edge node{0}   (d)
 (d)  edge[loop above]  node{0}  ()
 (d)  edge node{1}   (e)
 (e)  edge[loop above]  node{0}  ();
\end{tikzpicture}
\end{center}

As an example, the constraint defined on scope $\langle x_1,x_2,\ldots,x_7 \rangle$ from the simple automation depicted above  is:

\begin{boxex}
\begin{xcsp}  
<regular>
   <list> x1 x2 x3 x4 x5 x6 x7 </list>
   <transitions> 
     (a,0,a)(a,1,b)(b,1,c)(c,0,d)(d,0,d)(d,1,e)(e,0,e) 
   </transitions>
   <start> a </start>
   <final> e </final>
</regular>
 \end{xcsp}
\end{boxex}

\begin{xl}
\subsubsection{Constraint \gb{grammar}}\label{ctr:grammar}

The constraint \gb{grammar} \cite{S_theory,QW_decomposing,KS_efficient,QW_decompositions,KS_grammar} ensures that the sequence of values assigned to the variables it involves belongs to the language defined by a formal grammar.
The scope of the constraint is given by the element \xml{list}, and three elements, \xml{terminal}, \xml{rules} and \xml{start}, are introduced for representing respectively the terminal symbols (in our case, integer values), the production rules and the start non-terminal symbol of the grammar.
Each rule is composed of a ``head'' word (on the left) containing an arbitrary number of symbols (provided that at least one of them is a non-terminal symbol) and a ``body'' word (on the right) containing an arbitrary number of symbols (possibly, none).
It is important to note that both the head and body words are represented by a sequence of symbols, using whitespace as a separator between each pair of consecutive symbols.
This avoids ambiguity, and besides, this allows us to infer the set of non-terminal symbols from \xml{terminal} and \xml{rules}.
In what follows, \sy{symbol} stands for any identifier. 

\begin{boxsy}
\begin{syntax} 
<grammar>
  <list> (intVar wspace)+ </list>
  <terminal> (intVal wspace)+ </terminal>
  <rules> 
    ("(" (intVal | symbol) (wspace (intVal | symbol))* "," 
        [(intVal | symbol) (wspace (intVal | symbol))*] ")")+ 
  </rules>
  <start> symbol </start>
</grammar>
\end{syntax}
\end{boxsy}

Below, $L(G)$ denotes the language recognized by a formal grammar $G$.

\begin{boxse}
\begin{semantics}
$\gb{grammar}(X,G)$, with $X=\langle x_1,x_2,\ldots,x_r \rangle$ and $G$ a grammar, iff 
  $\va{x}_1\va{x}_2\ldots\va{x}_r \in L(G)$ 
\end{semantics}
\end{boxse}

As an example, let us consider the grammar defined from the set of non-terminal symbols $N=\{A,S\}$ ($S$ being the start symbol), the set of terminal symbols $\Sigma=\{0,1,2\}$, and the following rules (with $\epsilon$ denoting the empty string):
\begin{itemize}
\item $S \rightarrow 0S$
\item $S \rightarrow 1S$
\item $A \rightarrow \epsilon$
\item $A \rightarrow 2A$
\end{itemize}

This grammar describes the same language as the regular expression $0^*12^*$.
Assuming that $x_1$, $x_2$ and $x_3$ are three integer variables with domain $\{0,1,2\}$, we could build a constraint \gb{grammar} as follows:

\begin{boxex}
\begin{xcsp}  
<grammar>
  <list> x1 x2 x3 </list>
  <terminal> 0 1 2 </terminal>
  <rules> (S,0 S)(S,1 S)(A,)(A,2 A) </rules>    
  <start> S </start>
</grammar>
 \end{xcsp}
\end{boxex}
\end{xl}

\subsubsection{Constraint \gb{mdd}}\label{ctr:mdd}

The constraint \gb{mdd} \cite{CY_maintaining,CY_maintainingr,CY_mdd,PR_GAC4} ensures that the sequence of values assigned to the variables it involves follows a path going from the root of the described MDD (Multi-valued Decision Diagram) to the unique terminal node.
Because the graph is directed, acyclic, with only one root node and only one terminal node, we just need to introduce \xml{transitions}.

\begin{boxsy}
\begin{syntax} 
<mdd>
  <list> (intVar wspace)+ </list>
  <transitions> ("(" state "," intVal "," state ")")+ </transitions>
</mdd>
\end{syntax}
\end{boxsy}

Below, $L(M)$ denotes the language recognized by a MDD $M$.

\begin{boxse}
\begin{semantics}
$\gb{mdd}(X,M)$, with  $X=\langle x_1,x_2,\ldots,x_r \rangle$ and $M$ a MDD, iff 
  $\va{x}_1\va{x}_2\ldots\va{x}_r \in L(M)$ 
\end{semantics}
\end{boxse}

\begin{center}
\begin{tikzpicture}[>=stealth]
  \tikzstyle{snode}=[draw,circle,minimum size=6mm]
  \tikzstyle{sedge}=[draw,->,>=latex]
  \tikzstyle{slabel}=[midway,scale=1.2]
  \node[snode] (0) at (0,4.5) {$r$};
  \node[snode] (1) at (-1.5,3)  {$n_1$};
  \node[snode] (2) at (0,3) {$n_2$};
  \node[snode] (3) at (1.5,3) {$n_3$};
  \node[snode] (4) at (-0.75,1.5)  {$n_4$};
  \node[snode] (5) at (0.75,1.5) {$n_5$};
  \node[snode] (6) at (0,0) {$t$};
  \node[minimum size=6mm] (x1) at (4,3.75)  {$x_1$};
  \node[minimum size=6mm] (x2) at (4,2.25) {$x_2$};
  \node[minimum size=6mm] (x3) at (4,0.75) {$x_3$};

  \draw[sedge] (0) -- node[slabel,left]{0} (1);
  \path[sedge] (0) edge node[slabel,right]{$1$} (2);
  \path[sedge] (0) edge node[slabel,right]{$2$} (3);
  \path[sedge] (1) edge node[slabel,left]{$2$} (4);
  \path[sedge] (2) edge node[slabel,right]{$2$} (4);
  \path[sedge] (3) edge node[slabel,right]{$0$} (5);
  \path[sedge] (4) edge node[slabel,left]{$0$} (6);
  \path[sedge] (5) edge node[slabel,right]{$0$} (6);
\end{tikzpicture}
\end{center}

As an example, the constraint of scope $\langle x_1,x_2,x_3 \rangle$ is defined from the simple MDD depicted above (with root node $r$ and terminal node $t$) as:

\begin{boxex}
\begin{xcsp}  
<mdd>
   <list> x1 x2 x3 </list>
   <transitions>
     (r,0,n1)(r,1,n2)(r,2,n3)
     (n1,2,n4)(n2,2,n4)(n3,0,n5)
     (n4,0,t)(n5,0,t) 
   </transitions>
</mdd>
 \end{xcsp}
\end{boxex}

\subsection{Comparison-based Constraints}

In this section, we present constraints that are based on comparisons between pairs of variables. More specifically, we introduce:
\begin{enumerate}
\item \gb{allDifferent} (capturing \gb{allDifferentExcept})
\item \gb{allEqual} (capturing \gb{allEqualExcept})
\begin{xl} \item \gb{allDistant}
\end{xl}
\item \gb{ordered}
\item \gb{precedence}
\end{enumerate}

\subsubsection{Constraint \gb{allDifferent}}\label{ctr:allDifferent}

The constraint \gb{allDifferent}, see \cite{R_filtering,H_alldiff,GMN_generalized}, ensures that the variables in \xml{list} must all take different values.
A variant, called \gb{allDifferentExcept} in the literature \cite{BCR_global,C_dulmage}, enforces variables to take distinct values, except those that are assigned to some specified values (often, the single value 0).
This is the reason why we introduce an optional element \xml{except}. 

\begin{boxsy}
\begin{syntax} 
<allDifferent>
  <list> (intVar wspace)2+ </list>
  [<except> (intVal wspace)+ </except>]
</allDifferent> 
\end{syntax}
\end{boxsy}

Note that the opening and closing tags of \xml{list} are optional if \xml{list} is the unique parameter of the constraint, which gives:

\begin{boxsy}
\begin{syntax} 
<allDifferent> (intVar wspace)2+ </allDifferent> @\com{Simplified Form}@
\end{syntax}
\end{boxsy}

\begin{boxse}
\begin{semantics}
$\gb{allDifferent}(X,E)$, with $X=\langle x_1,x_2,\ldots \rangle$, iff 
  $\forall  (i,j) : 1 \leq i < j \leq |X|, \va{x}_i \neq \va{x}_j \lor \va{x}_i \in E \lor \va{x}_j \in E$
$\gb{allDifferent}(X)$ iff $\gb{allDifferent}(X,\emptyset)$ 
\end{semantics}
\end{boxse}

For example, below, the constraint $c_1$ forces all variables $x_1,x_2,x_3,x_4,x_5$ to take different values, and the constraint $c_2$ forces each of the variables of the 1-dimensional array $y$ to take either the value 0 or a value different from the other variables.

\begin{boxex}
\begin{xcsp}
<allDifferent id="c1"> 
  x1 x2 x3 x4 x5 
</allDifferent> 
<allDifferent id="c2">
  <list> y[] </list>
  <except> 0 </except>
</allDifferent>
\end{xcsp}
\end{boxex}

\begin{xl}
  \begin{remark}
Possible restricted forms of \gb{allDifferent} are possible by forcing variables to be, for example, \gb{symmetric} and/or \gb{irreflexive}.
For more details, see Section \ref{sec:restricted}.
\end{remark}
\end{xl}

\paragraph{Generalized Form of \gb{allDifferent} (View).}
Remember that, for simplicity, the syntax and semantics in this document are given for rather strict forms, but the format can easily accept more sophisticated forms.
This is related to the concept of view, as discussed in Section \ref{sec:views}.
Any \gb{allDifferent} constraint contains an element \xml{list}, where, instead of listing variables, one can list integer expressions:
\begin{quote}
  \verb!<list> (intExpr wspace)2+ </list>!
\end{quote}
as for example in:

\begin{boxex}
\begin{xcsp}
<allDifferent> 
  add(x1,1) add(x2,2) add(x3,3) 
</allDifferent> 
\end{xcsp}
\end{boxex}

\subsubsection{Constraint \gb{allEqual}}\label{ctr:allEqual}

The constraint \gb{allEqual} ensures that all involved variables take the same value.
This is mainly an ease of modeling.
A variant called \gb{allEqualExcept} enforces variables to take the same value, except those that are assigned to some specified values (often, the single value 0).
This is the reason why we introduce an optional element \xml{except}. 

\begin{boxsy}
\begin{syntax} 
<allEqual>
  <list> (intVar wspace)2+ </list>
  [<except> (intVal wspace)+ </except>]
</allEqual> 
\end{syntax}
\end{boxsy}

Note that the opening and closing tags of \xml{list} are optional if \xml{list} is the unique parameter of the constraint, which gives:

\begin{boxsy}
\begin{syntax} 
<allEqual> (intVar wspace)2+ </allEqual> @\com{Simplified Form}@
\end{syntax}
\end{boxsy}

\begin{boxse}
\begin{semantics}
$\gb{allEqual}(X,E)$, with $X=\langle x_1,x_2,\ldots \rangle$, iff 
  $\forall  (i,j) : 1 \leq i < j \leq |X|, \va{x}_i = \va{x}_j \lor \va{x}_i \in E \lor \va{x}_j \in E$
$\gb{allEqual}(X)$ iff $\gb{allEqual}(X,\emptyset)$
\end{semantics}
\end{boxse}

As an example, we have:
\begin{boxex}
\begin{xcsp}
<allEqual> 
  x1 x2 x3 x4 x5
</allEqual> 
\end{xcsp}
\end{boxex}

\paragraph{Generalized Form of \gb{allEqual} (View).}
Any \gb{allEqual} constraint contains an element \xml{list}, where, instead of listing variables, one can list integer expressions:
\begin{quote}
  \verb!<list> (intExpr wspace)2+ </list>!
\end{quote}
as for example in:

\begin{boxex}
\begin{xcsp}
<allEqual> 
  add(x1,1) add(x2,2) add(x3,3) 
</allEqual> 
\end{xcsp}
\end{boxex}

\begin{xl}
\subsubsection{Constraint \gb{allDistant}}\label{ctr:allDistant}

The constraint \gb{allDistant}  ensures that the distance between each pair of variables of \xml{list} is subject to a numerical condition.
In \cite{R_interdistance,LQLP_quadratic}, this constraint is studied with respect to the operator $\geq$, and called \gb{interDistance}.

\begin{boxsy}
\begin{syntax} 
<allDistant>
  <list> (intVar wspace)2+ </list>
  <condition> "(" operator "," operand ")" </condition> 
</allDistant> 
\end{syntax}
\end{boxsy}

\begin{boxse}
\begin{semantics}
$\gb{allDistant}(X,(\odot,k))$,  with $X=\langle x_1,x_2,\ldots \rangle$, iff 
  $\forall (i, j) : 1 \leq i < j \leq |X|, |\va{x}_i - \va{x}_{j}| \odot \va{v}$
\end{semantics}
\end{boxse}

Below, the constraint $c_1$ corresponds to $|x_1 - x_2| \geq 2 \land |x_1 - x_3| \geq 2 \land |x_2 - x_3| \geq 2$, whereas the constraint $c_2$ guarantees that $\forall (i,j) : 1 \leq i < j \leq 4$, we have $2 \leq |y_i - y_j| \leq 4$

\begin{boxex}
\begin{xcsp}
<allDistant id="c1"> 
  <list> x1 x2 x3 </list> 
  <condition> (ge,2) </condition>
</allDistant> 
<allDistant id="c2"> 
  <list> y1 y2 y3 y4 </list> 
  <condition> (in,2..4) </condition>
</allDistant> 

\end{xcsp}
\end{boxex}
\end{xl}

\subsubsection{Constraint \gb{ordered}}\label{ctr:ordered}

The constraint \gb{ordered} ensures that the variables of \xml{list} are ordered in sequence according to a relational operator specified in \xml{operator}, whose value must be in $\{<,\leq,\geq,>\}$.
The optional element \xml{lengths} indicates the minimum distances between any two successive variables of \xml{list}.

\begin{boxsy}
\begin{syntax} 
<ordered>
  <list> (intVar wspace)2+ </list>
  [<lengths> (intVal wspace)2+ | (intVar wspace)2+ </lengths>] 
  <@operator@> "lt" | "le" | "ge" | "gt" </@operator@>
</ordered> 
\end{syntax}
\end{boxsy}

\begin{boxse}
\begin{semantics}
$\gb{ordered}(X,L,\odot)$, with $X=\langle x_1,x_2,\ldots \rangle$, $L=\langle l_1,l_2,\ldots \rangle$ and $\odot \in \{<,\leq,\geq,>\}$, iff 
  $\forall i : 1 \leq i < |X|, \va{x}_i + l_i \odot \va{x}_{i+1}$
$\gb{ordered}(X,\odot)$, with $X=\langle x_1,x_2,\ldots \rangle$ and $\odot \in \{<,\leq,\geq,>\}$, iff 
  $\forall i : 1 \leq i < |X|, \va{x}_i \odot \va{x}_{i+1}$

$\gbc{Prerequisite}: |X| = |L| + 1$
\end{semantics}
\end{boxse}

Below, the constraint $c_1$ is equivalent to $x_1 < x_2 < x_3 < x_4$, and the constraint $c_2$ is equivalent to  $y_1 + 5 \geq y_2 \land y_2 + 3 \geq y_3$.
\begin{boxex}
\begin{xcsp}
<ordered id="c1"> 
  <list> x1 x2 x3 x4 </list> 
  <operator> lt </operator>
</ordered> 
<ordered id="c2"> 
  <list> y1 y2 y3  </list>
  <lengths> 5 3 </lengths>
  <operator> ge </operator>
</ordered>
\end{xcsp}
\end{boxex}

The constraints from the \cat that are captured by \gb{ordered} are:
\begin{itemize}
\item \gb{increasing}, \gb{strictly\_increasing}
\item \gb{decreasing}, \gb{strictly\_decreasing} 
\end{itemize}

\begin{xl}
  As a matter of fact, if the element \xml{lengths} is absent, one can define a constraint \gb{ordered} without using the element \xml{operator}: it suffices to add a special attribute \att{case} to \xml{ordered}.
This way, the opening and closing tags of \xml{list} become optional, making the \x3 representation more compact.

\begin{boxsy}
\begin{syntax} 
<ordered case="orderedType"> (intVar wspace)2+ </ordered> @\com{Simplified Form}@
\end{syntax}
\end{boxsy}

The possible values for the attribute \att{case} are \val{increasing}, \val{strictlyIncreasing}, \val{decreasing} and \val{strictlyDecreasing}.
The constraint $c_1$, introduced above, can then be written: 
\begin{boxex}
\begin{xcsp}
<ordered id="c1" case="strictlyIncreasing"> x1 x2 x3 x4 </ordered>
\end{xcsp}
\end{boxex}

\begin{remark}
In the future, the format might be extended to include an element specifying the order (preferences between values) to follow.
\end{remark}
\end{xl}

\subsubsection{Constraint \gb{precedence}}\label{ctr:precedence}

The constraint \gb{precedence}, see \cite{LL_precedence,W_precedence}, ensures that if a variable $x$ of \xml{list} is assigned the $i+1th$ value of \xml{values}, then another variable of \xml{list}, that precedes $x$, is assigned the $ith$ value of \xml{values}
The optional attribute \att{covered} indicates whether each value of \xml{values} must be assigned by at least one variable in \xml{list} (\val{false}, by default).

\begin{boxsy}
\begin{syntax} 
<precedence>
   <list> (intVar wspace)2+ </list>
   [<values [covered="boolean"]> (intVal wspace)2+ </values>]
</precedence>
\end{syntax}
\end{boxsy}

However, note that the parameter \xml{values} is optional: when this is the case, \xml{values} is assumed to be the ordered set of values that can be collected over the domains of variables in \xml{list}. 
Also, note that the opening and closing tags of \xml{list} are optional if \xml{list} is the unique parameter of the constraint, which then gives:

\begin{boxsy}
\begin{syntax} 
<precedence> (intVar wspace)2+ </precedence> @\com{Simplified Form}@
\end{syntax}
\end{boxsy}

For the semantics, $V^{\nm{cv}}$ means \att{covered}=\val{true}.

\begin{boxse}
\begin{semantics}
$\gb{precedence}(X,V)$, with $X=\langle x_1,x_2,\ldots \rangle$ and $V=\langle v_1,v_2,\ldots \rangle$ iff  
 $\forall i : 1 \leq i < |V|, v_{i+1} \in \{\va{x}_i : 1 \leq i \leq |X|\} \Rightarrow  v_{i} \in \{\va{x}_i : 1 \leq i \leq |X|\}$
 $\forall i : 1 \leq i < |V| \land v_{i+1} \in \{\va{x}_i : 1 \leq i \leq |X|\}$, 
   $\min\{j : 1 \leq j \leq |X| \land \va{x}_j = v_i\} < \min\{j :  1 \leq j \leq |X| \land \va{x}_j = v_{i+1}\}$
$\gb{precedence}(X)$ iff $\gb{precedence}(X,\cup \{dom(x_i) : x_i \in X\})$  @\com{ordered set being assumed}@
$\gb{precedence}(X,V^{\nm{cv}})$ iff  $\gb{precedence}(X,V) \wedge v_{|V|} \in \{\va{x}_i : 1 \leq i \leq |X|\}$
\end{semantics}
\end{boxse}

\begin{boxex}
\begin{xcsp}  
<precedence>
   <list> x1 x2 x3 x4 </list>
   <values> 4 0 1 </values>
</precedence>
 \end{xcsp}
\end{boxex}
 
The constraint \gb{precedence} captures \gb{int\_value\_precede} and \gb{int\_value\_precede\_chain} in the \cat.
Note that 
\gb{precedence} belongs now to \x3-core.

\subsection{Counting and Summing Constraints}

In this section, we present constraints that are based on counting the number of times variables or values satisfy a certain condition, and also those that are based on summations. More specifically, we introduce:
\begin{enumerate}
\item \gb{sum} (sometimes called \gb{linear} in the literature)
\item \gb{count} (capturing \gb{among},  \gb{atLeast}, \gb{atMost} and \gb{exactly})
\item \gb{nValues} (capturing \gb{nValuesExcept})
\item \gb{cardinality}
\begin{xl}\item \gb{balance}
\item \gb{spread}
\item \gb{deviation}
\item \gb{sumCosts}
\end{xl}
\end{enumerate}

\subsubsection{Constraint \gb{sum}}\label{ctr:sum}

The constraint \gb{sum} is one of the most important constraint. 
When the optional element \xml{coeffs} is missing, it is assumed that all coefficients are equal to 1.
The constraint is subject to a numerical condition.

\begin{boxsy}
\begin{syntax} 
<sum>
   <list> (intVar wspace)2+ </list>
   [ <coeffs> (intVal wspace)2+ | (intVar wspace)2+ </coeffs> ] 
   <condition> "(" operator "," operand ")" </condition> 
</sum>
\end{syntax}
\end{boxsy}

Although in practice, coefficients are most the time integer values, we introduce the semantics with variable coefficients.

\begin{boxse}
\begin{semantics}
$\gb{sum}(X,C,(\odot,k))$, with $X=\langle x_1,x_2,\ldots \rangle$, and $C=\langle c_1,c_2,\ldots\rangle$, iff
  $(\sum_{i=1}^{|X|} \va{c}_i \times \va{x}_i) \odot \va{k}$ 

$\gbc{Prerequisite}: |X| = |C| \geq 2$
\end{semantics}
\end{boxse}

The following constraint $c_1$ states that the values taken by variables $x_1 ,x_2, x_3$ and $y$ must respect the linear function $x_1 \times 1 + x_2 \times 2 + x_3 \times 3 >y$.

\begin{boxex}
\begin{xcsp}  
<sum id="c1">
  <list> x1 x2 x3 </list>
  <coeffs> 1 2 3 </coeffs>
  <condition> (gt,y) </condition>  
</sum>
 \end{xcsp}
\end{boxex}

A form of \gb{sum}, sometimes called \gb{subset-sum} or \gb{knapsack} \cite{T_dynamic,PQ_counting} involves the operator ``in'', and ensures that the obtained sum belongs (or not) to a specified interval.
The following constraint $c_2$ states that the values taken by variables $y_1,y_2,y_3,y_4$ must respect $2 \leq y_1 \times 4 + y_2 \times 2 + y_3 \times 3 + y_4 \times 1 \leq 5$.

\begin{boxex}
\begin{xcsp}  
<sum id="c2">
  <list> y1 y2 y3 y4 </list>
  <coeffs> 4 2 3 1 </coeffs>
  <condition> (in,2..5) </condition>  
</sum>
\end{xcsp}
\end{boxex}

Since Specifications 3.0.7, one may use compact forms of integer sequences (in elements \xml{values} of \xml{instantiation} and \xml{coeffs} of \xml{sum}) by writting $v$x$k$ for standing that the integer $v$ occurs $k$ times in sequence.
This means that, assuming that $w$ is an array of 6 integer variables, one can write:
\begin{boxex}
\begin{xcsp}  
<sum>
  <list> w[] </list>
  <coeffs> 1x4 2x2 </coeffs>
  <condition> (le,10) </condition>  
</sum>
\end{xcsp}
\end{boxex}
instead of: 
\begin{boxex}
\begin{xcsp}  
<sum>
  <list> w[] </list>
  <coeffs> 1 1 1 1 2 2 </coeffs>
  <condition> (le,10) </condition>  
</sum>
\end{xcsp}
\end{boxex}

\paragraph{Generalized Form of \gb{sum} (View).}
Any \gb{sum} constraint contains an element \xml{list}, where, instead of listing variables, one can list integer expressions:
\begin{quote}
  \verb!<list> (intExpr wspace)2+ </list>!
\end{quote}
As an illustration for \gb{sum}, the constraint $(x_1 = 1) + (x_2 > 2) + (x_3 = 1) \leq 2$ can be written as:

\begin{boxex}
\begin{xcsp}
<sum> 
  <list> eq(x1,1) gt(x2,2) eq(x3,1) </list>
  <condition> (le,2) </condition> 
</sum>
\end{xcsp}
\end{boxex}

Similarly, instead of listing variables (or integers), it is also possible to list integer expressions in the element \xml{coeffs}:
\begin{quote}
  \verb!<coeffs> (intExpr wspace)2+ </coeffs>!
\end{quote}

\subsubsection{Constraint \gb{count}}\label{ctr:count}

The constraint \gb{count}, introduced in CHIP \cite{BC_chip} and Sicstus \cite{COC_open}, ensures that the number of variables in \xml{list} which are assigned a value in \xml{values} respects a numerical condition. It is also present in Gecode with the same name and in the \cat where it is called \gb{counts}.
This constraint captures known constraints \gb{atLeast}, \gb{atMost}, \gb{exactly} and \gb{among}.

\begin{boxsy}
\begin{syntax} 
<count>
   <list> (intVar wspace)2+ </list>
   <values> (intVal wspace)+ | (intVar wspace)+ </values> 
   <condition> "(" operator "," operand ")" </condition>
</count>
\end{syntax}
\end{boxsy}

To simplify, we assume for the semantics that $V$ is a set of integer values.

\begin{boxse}
\begin{semantics}
$\gb{count}(X,V,(\odot,k))$, with $X=\langle x_1,x_2,\ldots \rangle$, iff 
  $|\{i : 1 \leq i \leq |X| \land \va{x}_i \in V\}| \odot \va{k}$ 
\end{semantics}
\end{boxse}

Below, $c_1$ enforces that the number of variables in $\{v_1,v_2,v_3,v_4\}$ that take the value of variable $v$ must be different from the value of $k_1$.
Constraints $c_2$, $c_3$, $c_4$ and $c_5$ illustrate how to represent \gb{atLeast}, \gb{atMost}, \gb{exactly} and \gb{among}.
\begin{itemize}
\item $c_2$ represents $\gb{among}(k_2,\{w_1,w_2,w_3,w_4\},\{1,5,8\})$;
\item $c_3$ represents $\gb{atLeast}(k_3,\{x_1,x_2,x_3,x_4,x_5\},1)$;
\item $c_4$ represents $\gb{atMost}(2,\{y_1,y_2,y_3,y_4\},0)$;
\item $c_5$ represents $\gb{exactly}(k_5,\{z_1,z_2,z_3\},z)$.
\end{itemize}

\begin{boxex}
\begin{xcsp}  
<count id="c1">
  <list> v1 v2 v3 v4 </list>
  <values> v </values>
  <condition> (ne,k1) </condition>
</count>
<count id="c2">   // among
  <list> w1 w2 w3 w4 </list>
  <values> 1 5 8 </values>
  <condition> (eq,k2) </condition>
</count>
<count id="c3" >  // atLeast
  <list> x1 x2 x3 x4 x5 </list>
  <values> 1 </values> 
  <condition> (ge,k3) </condition>
</count>
<count id="c4" > // atMost
  <list> y1 y2 y3 y4 </list>
  <values> 0 </values>
  <condition> (le,2) </condition>
</count>
<count id="c5"> // exactly
  <list> z1 z3 z3 </list>
  <values> z </values>  
  <condition> (eq,k5) </condition>
</count>
\end{xcsp}
\end{boxex}

\paragraph{Generalized Form of \gb{count} (View).}
Any \gb{count} constraint contains an element \xml{list}, where, instead of listing variables, one can list integer expressions:
\begin{quote}
  \verb!<list> (intExpr wspace)2+ </list>!
\end{quote}

\subsubsection{Constraint \gb{nValues}}\label{ctr:nValues}

The constraint \gb{nValues} \cite{BHHKW_nvalue}, ensures that the number of distinct values taken by variables in \xml{list} respects a numerical condition.
A variant, called \gb{nValuesExcept} \cite{BHHKW_nvalue} discards some specified values (often, the single value 0).
This is the reason why we introduce an optional element \xml{except}.

\begin{boxsy}
\begin{syntax} 
<nValues>
   <list> (intVar wspace)2+ </list>
   [<except> (intVal wspace)+ </except>]
   <condition> "(" operator "," operand ")" </condition>
</nValues>
\end{syntax}
\end{boxsy}

\begin{boxse}
\begin{semantics}
$\gb{nValues}(X,E,(\odot,k))$, with $X=\langle x_1,x_2,\ldots \rangle$, iff 
  $|\{\va{x}_i : 1 \leq i \leq |X|\} \setminus E| \odot \va{k}$ 
$\gb{nValues}(X,(\odot,k))$ iff $\gb{nValues}(X,\emptyset,(\odot,k))$
\end{semantics}
\end{boxse}

\begin{remark}
This element captures \gb{atLeastNValues} and \gb{atMostNValues}, since it is possible to specify the relational operator in \xml{condition}. 
\end{remark}

In the following example, the constraint $c_1$ states that there must be exactly three distinct values taken by variables $x_1, \ldots, x_4$, whereas $c_2$ states at most $w$ distinct values must be taken by variables $y_1, \ldots, y_5$. The constraint $c_3$ ensures that two different values are taken by variables $z_1, \ldots, z_4$, considering that the value 0 must be ignored. 

\begin{boxex}
\begin{xcsp}  
<nValues id="c1">
   <list> x1 x2 x3 x4 </list>
   <condition> (eq,3) </condition> 
</nValues>
<nValues id="c2">
   <list> y1 y2 y3 y4 y5 </list>
   <condition> (le,w) </condition>  
</nValues>
<nValues id="c3">
   <list> z1 z2 z3 z4 </list>
   <except> 0 </except>
   <condition> (eq,2) </condition>  
</nValues>
 \end{xcsp}
\end{boxex}

\begin{xl}
\begin{remark}
The constraint \gb{increasingNValues} can be built by adding restriction \gb{increasing} to \xml{list}.
For more details about restricted constraints, see section \ref{sec:restricted}.
\end{remark}
\end{xl}

\paragraph{Generalized Form of \gb{nValues} (View).}
Any \gb{nValues} constraint contains an element \xml{list}, where, instead of listing variables, one can list integer expressions:
\begin{quote}
  \verb!<list> (intExpr wspace)2+ </list>!
\end{quote}

\subsubsection{Constraint \gb{cardinality}}\label{ctr:cardinality}

The constraint \gb{cardinality}, also called \gb{globalCardinality} or \gb{gcc} in the literature, see \cite{R_gcc, H_integrated}, ensures that the number of occurrences of each value in \xml{values}, taken by variables of \xml{list}, is related to a corresponding element (value, variable or interval) in \xml{occurs}.
The element \xml{values} has an optional attribute  \att{closed} (\val{false}, by default): when \att{closed}=\val{true}, this means that all variables of \xml{list} must be assigned a value from \xml{values}.

\begin{boxsy}
\begin{syntax} 
<cardinality>
   <list> (intVar wspace)2+ </list>
   <values [closed="boolean"]> (intVal wspace)+ | (intVar wspace)+ </values>
   <occurs> (intVal wspace)+ | (intVar wspace)+ | (intIntvl wspace)+ </occurs>
</cardinality>
\end{syntax}
\end{boxsy}

For simplicity, for the semantics, we assume that $V$ only contains values and $O$ only contains variables. Note that $V^{\nm{cl}}$ means that \att{closed}=\val{true}.

\begin{boxse}
\begin{semantics}
$\gb{cardinality}(X,V,O)$, with $X=\langle x_1,x_2,\ldots \rangle$, $V=\langle v_1,v_2,\ldots \rangle$, $O=\langle o_1, o_2,\ldots \rangle$,
  iff $\forall j : 1 \leq j \leq |V|, |\{i : 1 \leq i \leq |X| \land \va{x}_i =v_j\}| = \va{o}_j$
$\gb{cardinality}(X,V^{\nm{cl}},O)$ iff $\gb{cardinality}(X,V,O) \land \forall i : 1 \leq i \leq |X|, \va{x}_i \in V$

$\gbc{Prerequisite}: |X| \geq 2 \land |V| = |O| \geq 1$
\end{semantics}
\end{boxse}

In the following example, $c_1$ states that the value 2 must be assigned to 0 or 1 variable (from $\{x_1,x_2,x_3,x_4\}$), the value 5 must be assigned to at least 1 and at most 3 variables, and the value 10 must be assiged to at least 2 and at most 3 variables.
Note that it is possible for a variable of $c_1$ to be assigned a value not present in $\{2,5,10\}$ since by default we have \att{closed}=\val{false}.
For $c_2$, the number of variables from $\{y_1,y_2,y_3,y_4,y_5\}$ that take value 0 must be equal to $z_0$, and so on.
Note that it is not possible for a variable $y_i$ to be assigned a value not present in $\{0,1,2,3\}$ since \att{closed}=\val{true}.

\begin{boxex}
\begin{xcsp}  
<cardinality id="c1">
   <list> x1 x2 x3 x4 </list>
   <values> 2 5 10 </values>
   <occurs> 0..1 1..3 2..3 </occurs>
</cardinality>
<cardinality id="c2">
   <list> y1 y2 y3 y4 y5 </list>
   <values closed="true"> 0 1 2 3 </values>
   <occurs> z0 z1 z2 z3 </occurs>
</cardinality>
 \end{xcsp}
\end{boxex}

The form of the constraint obtained by only considering variables in all contained elements is called \gb{distribute} in \mzinc.
In that case, for the semantics, we must additionally guarantee:
\begin{quote}
$\forall (i,j) : 1 \leq i < j \leq |V|, \va{v}_i \neq \va{v}_j$.
\end{quote}

\begin{boxex}
\begin{xcsp}  
<cardinality id="c3">
  <list> w1 w2 w3 w4 </list>
  <values> v1 v2 </values>
  <occurs> n1 n2 </occurs>
</cardinality>
\end{xcsp}
\end{boxex}

\begin{xl}
  The constraint \gb{cardinalityWithCosts} \cite{R_cost} will be discussed in Section \ref{ctr:gcccosts}.

\begin{remark}
The constraint \gb{increasing\_global\_ardinality} defined in the \cat can be built by adding restriction \gb{increasing} to \xml{list}.
For more details about restricted constraints, see section \ref{sec:restricted}.
\end{remark}
\end{xl}

\begin{xl}
\subsubsection{Constraint \gb{balance}}\label{ctr:balance}

The constraint \gb{balance} \cite{BCR_global,BHKKPQW_balance} ensures that the difference between the maximum number of occurrences and the minimum number of occurrences among the values assigned to the variables in \xml{list} respects a numerical condition.
If the optional element \xml{values} is present, then all variables must be assigned to a value from this set; see  \gb{balance$^*$} in \cite{BHKKPQW_balance}.

\begin{boxsy}
\begin{syntax} 
<balance>
   <list> (intVar wspace)2+ </list>
   [<values> (intVal wspace)+ </values>]
   <condition> "(" operator "," operand ")" </condition>
</balance>
\end{syntax}
\end{boxsy}

\begin{boxse}
\begin{semantics}
$\gb{balance}(X,(\odot,k))$, with $X=\langle x_1,x_2,\ldots \rangle$, iff 
  $\max_{v \in V} |\{i : 1 \leq i \leq |X| \land \va{x}_i=v\}|- \min_{v \in V} |\{i : 1 \leq i \leq |X| \land \va{x}_i=v\}| \odot \va{k}$ 
    with $V = \{\va{x}_i : 1 \leq i \leq |X|\}$
$\gb{balance}(X,V,(\odot,k))$ iff $\gb{balance}(X,(\odot,k)) \land \forall x_i \in X, \va{x}_i \in V$
\end{semantics}
\end{boxse}


\begin{boxex}
\begin{xcsp}  
<balance id="c1">
   <list> x1 x2 x3 x4 </list>
   <condition> (eq,k) </condition>
</balance>
<balance id="c2">
   <list> y1 y2 y3 y4 y5 </list>
   <values> 0 1 2 </values>
   <condition> (lt,2) </condition>
</balance>
\end{xcsp}
\end{boxex}

\subsubsection{Constraint \gb{spread}}\label{ctr:spread}

The constraint \gb{spread}  \cite{PR_spread,SDDR_spread,S_solving} ensures that the variance of values taken by variables of \xml{list} respects a numerical condition.
The element \xml{total} is optional; if present, its value must be equal to the sum of values of variables in \xml{list}.

\begin{boxsy}
\begin{syntax} 
<spread>
   <list> (intVar wspace)2+ </list>
   [<total> intVar </total>]
   <condition> "(" operator "," operand ")" </condition>
</spread>
\end{syntax}
\end{boxsy}

\begin{boxse}
\begin{semantics}
$\gb{spread}(X,s,(\odot,k))$, with $X=\langle x_1,x_2,\ldots x_r \rangle$,  iff  
  $\va{s} = \sum_{i=1}^{r} \va{x}_i$
  $(r \sum_{i=1}^{r} \va{x}_i^2 - \va{s}^2) \odot \va{k}$
\end{semantics}
\end{boxse}


\subsubsection{Constraint \gb{deviation}}\label{ctr:deviation}

The constraint \gb{deviation}  \cite{SDDR_deviation,S_solving} ensures that the deviation of values taken by variables of \xml{list} respects a numerical condition. 
The element \xml{total} is optional; if present, its value must be equal to the sum of values of variables in \xml{list}.

\begin{boxsy}
\begin{syntax} 
<deviation>
   <list> (intVar wspace)2+ </list>
   [<total> intVar </total>]
   <condition> "(" operator "," operand ")" </condition>
</deviation>
\end{syntax}
\end{boxsy}

\begin{boxse}
\begin{semantics}
$\gb{deviation}(X,s,(\odot,v))$, with $X=\{x_1,x_2,\ldots,x_r\}$,  iff  
  $\va{s} = \sum_{i=1}^{r} \va{x}_i$
  $\sum_{i=1}^{r} |r\va{x}_i - \va{s}| \odot \va{v}$
\end{semantics}
\end{boxse}


\subsubsection{Constraint \gb{sumCosts}}\label{ctr:sumCosts}

The constraint \gb{sumCosts} ensures that the sum of integer costs for values taken by variables in \xml{list} respects a numerical condition.  
This constraint is particularly useful, when combined with other constraints, to express preferences (as shown in Section \ref{sec:andUse}). 

A first form of \gb{sumCosts} allows us to define a cost matrix, where an integer cost is associated with each variable-value pair that we call {\em literal}.
The matrix contains tuples of integer values; there is one tuple per variable, containing as many costs as values present in the domain of the variable.

\begin{boxsy}
\begin{syntax} 
<sumCosts>
   <list> (intVar wspace)2+ </list>
   <costMatrix> ("(" intVal ( "," intVal)* ")")2+ </costMatrix>
   <condition> "(" operator "," operand ")" </condition> 
</sumCosts>
\end{syntax}
\end{boxsy}

For the semantics, we assume that $M$ is a matrix that gives the cost $M[x_i][a_i]$ for any value $a_i=\va{x}_i$ assigned to the $ith$ variable $x_i$ of \xml{list}.

\begin{boxse}
\begin{semantics}
$\gb{sumCosts}(X,M,(\odot,k))$, with $X=\langle x_1,x_2,\ldots\rangle$, iff 
  $\sum \{M[x_i][\va{x}_i] : 1 \leq i \leq |X|\} \odot \va{k}$
\end{semantics}
\end{boxse}

The following example shows a constraint $c_1$ that involves 3 variables $y_1,y_2,y_3$ in its main list, with four values in each variable domain (let us say $\{0,1,2,3\}$).
The assignment costs are as follows: 10 for $(y_1,0)$, 0 for $(y_1,1)$, 5 for $(y_1,2)$, 0 for $(y_1,3)$, 0 for $(y_2,0)$, 5 for $(y_2,1)$, and so on.
For $c_1$, the sum of assignment costs must be less than or equal to 12.

\begin{boxex}
\begin{xcsp}  
<sumCosts id="c1">
   <list> y1 y2 y3 </list> 
   <costMatrix> 
     (10,0,5,0)   // costs for y1
     (0,5,0,0)    // costs for y2
     (5,10,0,0)   // costs for y3
   </costMatrix>
   <condition> (le,12) </condition>
</sumCosts>
\end{xcsp}
\end{boxex} 

A second form of \gb{sumCosts} allows us to define the cost matrix under a dual point of view.
We replace the element \xml{costMatrix} by a sequence of elements \xml{literals}, where each such element has a required attribute \att{cost} that gives the common cost of all literals (variable-value pairs) contained inside the element.
It is possible to use the optional attribute \att{defaultCost} for \xml{sumCosts} to specify the cost of  all implicit literals (i.e., those not explicitly listed). 

\begin{boxsy}
\begin{syntax} 
<sumCosts [defaultCost="integer"]>
   <list> (intVar wspace)2+ </list>
   (<literals cost="integer"> ("(" intVar "," intVal ")")+ </literals>)+
   <condition> "(" operator "," operand ")" </condition> 
</sumCosts>
\end{syntax}
\end{boxsy}

For the semantics, we assume, for any pair $(x_i,a_i)$, that $cost^{\ns{A}}(x_i,a_i)$ denotes the cost $cost(A_j)$ of the set $A_j$ such that $(x_i,a_i) \in A_j$.

\begin{boxse}
\begin{semantics}
$\gb{sumCosts}(X,\ns{A},(\odot,k))$, with $X=\langle x_1,x_2,\ldots\rangle$, and $\ns{A}=\langle A_1,A_2,\ldots \rangle$  where $A_i$ is a set of literals of cost $cost(P_i)$, iff $\sum \{cost^{\ns{A}}(x_i,\va{x}_i) : 1 \leq i \leq [X|\} \odot \va{k}$

$\gbc{Prerequisite}:$
  $\forall (i,j) : 1 \leq i < j \leq |\ns{A}|, A_i \cap A_j = \emptyset$
  $\forall (x,a) : x \in X \land a \in dom(x), \exists A_i \in \ns{A} : (x,a) \in A_i$
\end{semantics}
\end{boxse}

When a value $a$ always admits the same cost, whatever the variable $x_i$ is, we can simplify $(x_1,a),(x_2,a),\ldots$ by $(*,a)$.

We give an illustration.
The constraint $c_{1b}$ is equivalent to constraint $c_1$ introduced above.
The constraint $c_2$ involves 4 variables $w_1,w_2,w_3,w_4$ in its main list, with 3 values in each domain (let us say $\{0,1,2\}$).
The assignment costs are as follows: 10 for $(w_1,0)$, $(w_2,0)$, $(w_3,0)$, $(w_4,0)$ and $(w_3,1)$, and 0 for the other literals (variable-value pairs).

\bigskip
\begin{boxex}
\begin{xcsp}  
<sumCosts id="c1b" defaultCost="0">
   <list> y1 y2 y3 </list> 
   <literals @{\violet{cost}@="10"> (y1,0)(y3,1) </literals>
   <literals @{\violet{cost}@="5"> (y1,2)(y2,1)(y3,0) </literals>
   <condition> (le,12) </condition> 
</sumCosts>
<sumCosts id="c2" defaultCost="0">
   <list> w1 w2 w3 w4 </list> 
   <literals @{\violet{cost}@="10"> (*,0)(w3,1) </literals>
   <condition> (eq,z) </condition>
</sumCosts>
\end{xcsp}
\end{boxex}

A related form of \gb{sumCosts} is defined by using a sequence of elements \xml{values}, where each such element has a required attribute \att{cost} that gives the common cost of all values contained inside the element, i.e., whatever the variables are.
The optional attribute \att{defaultCost} for \xml{sumCosts} is useful for specifying the cost of  all implicit  values (i.e., those not explicitly listed).  

\begin{boxsy}
\begin{syntax} 
<sumCosts [defaultCost="integer"]>
   <list> (intVar wspace)2+ </list>
   (<values cost="integer"> (intVal wspace)+ </values>)+
   <condition> "(" operator "," operand ")" </condition> 
</sumCosts>
\end{syntax}
\end{boxsy}

For the semantics, we assume, for any value $a_i$ in the domain of a variable of $X$, that $cost^{\ns{V}}(a_i)$ denotes the cost $cost(V_j)$ of the set $V_j$ such that $a_i \in V_j$.

\begin{boxse}
\begin{semantics}
$\gb{sumCosts}(X,\ns{V},(\odot,k))$, with $X=\langle x_1,x_2,\ldots\rangle$, and $\ns{V}=\langle V_1,V_2,\ldots \rangle$ where $V_i$ is a set of values of cost $cost(V_i)$, iff 
  $\sum \{cost^{\ns{V}}(\va{x}_i) : 1 \leq i \leq [X|\} \odot \va{k}$

$\gbc{Prerequisite}:$
  $\forall (i,j) : 1 \leq i < j \leq |\ns{V}|, V_i \cap V_j = \emptyset$
  $\forall a \in \cup_{x \in X} dom(x), \exists V_i \in \ns{V} : a \in V_i$
\end{semantics}
\end{boxse}

The constraint $c_3$ involves 5 variables $v_1,v_2,v_3,v_4,v_5$ in its main list, with 4 values in each domain (let us say $\{1,2,3,4\}$).
The assignment costs are as follows: 10 for $(v_1,2)$, $(v_2,2)$, $\ldots$, $(v_1,4)$, $(v_2,4)$, $\ldots$, and 0 for the other literals.
The constraint $c_3$ forces $z$ to be equal to the sum of the assignment costs.

\bigskip
\begin{boxex}
\begin{xcsp}  
<sumCosts id="c3" defaultCost="0">
   <list> v1 v2 v3 v4 v5 </list> 
   <values @{\violet{cost}@="10"> 2 4 </values>
   <condition> (eq,z) </condition>
</sumCosts>
\end{xcsp}
\end{boxex}

Finally, by replacing the element \xml{list} by an element \xml{set}, we can represent the constraint \gb{sum\_of\_weights\_of\_distinct\_values} \cite{BCT_cost}, where any cost is defined per value and can be taken into account only once. 

\begin{boxsy}
\begin{syntax} 
<sumCosts [defaultCost="integer"]>
   <set> (intVar wspace)2+ </set>
   (<values cost="integer"> (intVal wspace)+ </values>)+
   <condition> "(" operator "," operand ")" </condition> 
</sumCosts>
\end{syntax}
\end{boxsy}

For the semantics of this constraint variant, that we refer to as \gb{sumCosts$^{set}$}, we assume, for any value $a_i$ in the domain of a variable of $X$, that $cost^{\ns{V}}(a_i)$ denotes the cost $cost(V_j)$ of the set $V_j$ such that $a_i \in V_j$.

\begin{boxse}
\begin{semantics}
$\gb{sumCosts^{set}}(X,\ns{V},(\odot,k))$, with $X=\langle x_1,x_2,\ldots\rangle$, and $\ns{V}=\langle V_1,V_2,\ldots \rangle$ where $V_i$ is a set of values of cost $cost(V_i)$, iff 
  $\sum \{cost^{\ns{V}}(a) : a \in \cup_{x \in X} dom(x) \land \exists i: 1 \leq i \leq [X| \land \va{x}_i = a\} \odot \va{k}$

$\gbc{Prerequisite}:$
  $\forall (i,j) : 1 \leq i < j \leq |\ns{V}|, V_i \cap V_j = \emptyset$
  $\forall a \in \cup_{x \in X} dom(x), \exists V_i \in \ns{V} : a \in V_i$
\end{semantics}
\end{boxse}

\begin{remark}
We have just seen that turning an element \xml{list} into an element \xml{set} allows us to define a variant for the constraint  \gb{sumCosts}. This principle will be generalized in Chapter \ref{cha:lifted}.
\end{remark}
\end{xl}

\subsection{Connection Constraints}

In this section, we present constraints that establish a connection between different variables. More specifically, we introduce:
\begin{enumerate}
\item \gb{maximum} (and \gb{maximumArg})
\item \gb{minimum} (and \gb{minimumArg})
\item \gb{element}
\item \gb{channel}
\end{enumerate}

\begin{xl}
\subsubsection{Constraints \gb{maximum} and \gb{maximumArg}}\label{ctr:maximum}\label{ctr:maximumArg}
\end{xl}
\begin{xc}
\subsubsection{Constraints \gb{maximum}}\label{ctr:maximum}

\end{xc}

The constraint \gb{maximum} ensures that the maximum value among those assigned to variables of \xml{list} respects a numerical condition.

\begin{boxsy}
\begin{syntax} 
<maximum>
   <list> (intVar wspace)2+ </list>
   <condition> "(" operator "," operand ")" </condition> 
</maximum>
\end{syntax}
\end{boxsy}

\begin{boxse}
\begin{semantics}
$\gb{maximum}(X,(\odot,k))$, with $X=\langle x_1,x_2,\ldots\rangle$, iff 
  $\max\{\va{x}_i : 1 \leq i \leq |X|\} \odot \va{k}$
\end{semantics}
\end{boxse}

In the following example, the constraint $c_1$ states that $\max\{x_1,x_2,x_3,x_4\} = y$ whereas $c_2$ states that $\max\{z_1,z_2,z_3,z_4,z_5\} < w$.

\begin{boxex}
\begin{xcsp}  
<maximum id="c1">
   <list> x1 x2 x3 x4 </list>
   <condition> (eq,y) </condition>
</maximum>
<maximum id="c2">
   <list> z1 z2 z3 z4 z5 </list>
   <condition> (lt,w) </condition>
</maximum>
 \end{xcsp}
\end{boxex}
 
\begin{xl}
Another related form is the constraint \gb{maximumArg}, sometimes called \gb{arg\_max}, which ensures that the index of a maximum variable (i.e., a variable with a maximal value) in \xml{list} respects a numerical condition.
The optional attribute \att{startIndex} of \xml{list} gives the number used for indexing the first variable in \xml{list} (0, by default). 
The optional attribute \att{rank} of \xml{maximumArg} indicates if we refer to the first index, the last index or any index of a maximum variable of \xml{list}; it must respect the numerical condition (\val{any}, by default).

\begin{boxsy}
\begin{syntax} 
<maximumArg [rank="rankType"]>
   <list [startIndex="integer"]> (intVar wspace)2+ </list>
   <condition> "(" operator "," operand ")" </condition>
</maximumArg>
\end{syntax}
\end{boxsy}

We give the semantics for \att{rank}=\val{any}.

\begin{boxse}
\begin{semantics}
$\gb{maximum}(X,i)$, with $X=\langle x_1,x_2,\ldots \rangle$, iff @\com{indexing assumed to start at 1}@
  $\va{i} \in \{j : 1 \leq j \leq |X| \land \va{x}_j = \max\{\va{x}_k : 1 \leq k \leq |X|\}\}$
$\gb{maximumArg}(X,(\odot,k))$ iff $\exists i : \gb{maximum}(X,i) \land i \odot \va{k}$
\end{semantics}
\end{boxse}
\end{xl}

\paragraph{Generalized Form of \gb{maximum} (View).}
Any \gb{maximum} or \gb{maximumArg} constraint contains an element \xml{list}, where, instead of listing variables, one can list integer expressions:
\begin{quote}
  \verb!<list> (intExpr wspace)2+ </list>!
\end{quote}

\begin{xl}
\subsubsection{Constraints \gb{minimum} and \gb{minimumArg}}\label{ctr:minimum}\label{ctr:minimumArg}
\end{xl}
\begin{xc}
\subsubsection{Constraints \gb{minimum}}\label{ctr:minimum}
\end{xc}

The constraint \gb{minimum} ensures that the minimum value among the values assigned to variables in \xml{list} respects a numerical condition.

\begin{boxsy}
\begin{syntax} 
<minimum>
   <list> (intVar wspace)2+ </list>
   <condition> "(" operator "," operand ")" </condition> 
</minimum>
\end{syntax}
\end{boxsy}

\begin{boxse}
\begin{semantics}
$\gb{minimum}(X,(\odot,k))$,  with $X=\langle x_1,x_2,\ldots\rangle$, iff 
   $\min\{\va{x}_i : 1 \leq i \leq |X|\} \odot \va{k}$
\end{semantics}
\end{boxse}

In the following example, the constraint $c_1$ states that $\min\{x_1,x_2,x_3,x_4\} = y$ whereas $c_2$ states that $\min\{z_1,z_2,z_3,z_4,z_5\} \neq w$.

\begin{boxex}
\begin{xcsp}  
<minimum id="c1">
   <list> x1 x2 x3 x4 </list>
   <condition> (eq,y) </condition>
</minimum>
<minimum id="c2">
   <list> z1 z2 z3 z4 z5 </list>
   <condition> (ne,w) </condition>
</minimum>
 \end{xcsp}
\end{boxex}
 
\begin{xl}
Another related form is the constraint \gb{minimumArg}, sometimes called \gb{arg\_min}, which ensures that the index of a minium variable (i.e., a variable with a minimal value) in \xml{list} respects a numerical condition.
The optional attribute \att{startIndex} of \xml{list} gives the number used for indexing the first variable in \xml{list} (0, by default). 
The optional attribute \att{rank} of \xml{minimumArg} indicates if we refer to the first index, the last index or any index of a minimum variable of \xml{list}; it must respect the numerical condition (\val{any}, by default).

\begin{boxsy}
\begin{syntax} 
<minimumArg [rank="rankType"]>
   <list [startIndex="integer"]> (intVar wspace)2+ </list>
   <condition> "(" operator "," operand ")" </condition>
</minimumArg>
\end{syntax}
\end{boxsy}

We give the semantics for \att{rank}=\val{any}.

\begin{boxse}
\begin{semantics}
$\gb{minimum}(X,i)$, with $X=\langle x_1,x_2,\ldots \rangle$, iff @\com{indexing assumed to start at 1}@
  $\va{i} \in \{j : 1 \leq j \leq |X| \land \va{x}_j = \min\{\va{x}_k : 1 \leq k \leq |X|\}\}$
$\gb{minimumArg}(X,(\odot,k))$ iff $\exists i : \gb{minimum}(X,i) \land i \odot \va{k}$
\end{semantics}
\end{boxse}
\end{xl}

\paragraph{Generalized Form of \gb{minimum} (View).}
Any \gb{minimum} or \gb{minimumArg} constraint contains an element \xml{list}, where, instead of listing variables, one can list integer expressions:
\begin{quote}
  \verb!<list> (intExpr wspace)2+ </list>!
\end{quote}

\subsubsection{Constraint \gb{element}}\label{ctr:element}

The first form of the constraint \gb{element} \cite{HC_generality} simply ensures that \xml{value} is element of \xml{list}, i.e., equal to one value among those assigned to variables of \xml{list}.

\begin{boxsy}
\begin{syntax} 
<element>
  <list> (intVar wspace)2+ </list>
  <value> intVal | intVar </value>
</element>
\end{syntax}
\end{boxsy}

The semantics is: 

\begin{boxse}
\begin{semantics}
$\gb{element}(X,v)$, with $X=\langle x_0,x_1,\ldots \rangle$, iff 
  $\exists i : 0 \leq i < |X| \land \va{x}_i = \va{v}$ 
\end{semantics}
\end{boxse}

The second (very usual) form of the constraint \gb{element} ensures that the value of \xml{list} at position \xml{index} respects a numerical condition.
The optional attribute \att{startIndex} gives the number used for indexing the first variable in \xml{list} (0, by default). 
For \x3-core, the number used for indexing the first variable in \xml{list} is necessarily 0.

\begin{boxsy}
\begin{syntax} 
<element>
  <list [startIndex="integer"]> (intVar wspace)2+ | (intVal wspace)2+ </list>
  <index> intVar </index> 
  <condition> "(" operator "," operand ")" </condition>  
</element>
\end{syntax}
\end{boxsy}

The semantics is:

\begin{boxse}
\begin{semantics}
$\gb{element}(X,i,(\odot,k))$, with $X=\langle x_0,x_1,\ldots \rangle$, iff 
  $\va{x}_{\va{i}} \odot \va{k}$ 
\end{semantics}
\end{boxse}

The following constraint $c_1$ states that the ith variable of $\langle x_1,x_2,x_3,x_4 \rangle$ must be equal to $v$. For example, if $i$ is equal to 2, then $x_2$ must be equal to $v$.
The constraint $c_2$ ensures that $z$ is element (member) of the 1-dimensional array $y$.

\begin{boxex}
\begin{xcsp}  
<element id="c1">
  <list startIndex="1"> x1 x2 x3 x4 </list>
  <index> i </index>
  <condition> (eq,v) </condition>
</element>
<element id="c2">
  <list> y[] </list>
  <value> z </value>
</element>
\end{xcsp}
\end{boxex}

A {\em deprecated form} for representing $c_1$ is obtained by replacing \xml{condition} by \xml{value} (this is possible because we have the operator 'eq').
This gives:
\begin{boxex}
\begin{xcsp}  
<element id="c1">
  <list startIndex="1"> x1 x2 x3 x4 </list>
  <index> i </index>
  <value> v </value>
</element>
\end{xcsp}
\end{boxex}

Note that the element \xml{list} may contain values (instead of variables).
In that case, \xml{condition} must contain a variable.
Although any such constraint can be reformulated as a binary extensional constraint,
this variant is often used when modeling.  

As an example, the constraint below ensures that, indexing starting at 0, if $i$ is equal to 0 then $v$ must be equal to 10, if $i$ is equal to 1 then $v$ must be equal to 4, and so on.   

\begin{boxex}
\begin{xcsp}  
<element>
  <list> 10 4 7 2 3 </list>
  <index> i </index>
  <condition> (eq,v) </condition>
</element>
\end{xcsp}
\end{boxex}

\subsubsection{Constraint \gb{channel}}\label{ctr:channel}

\begin{xl}
The constraint \gb{channel} ensures that if the $ith$ variable is assigned the value $j$, then the $jth$ variable must be assigned the value $i$.
The optional attribute \att{startIndex} of \xml{list} gives the number used for indexing the first variable in this list (0, by default). 

\begin{boxsy}
\begin{syntax} 
<channel>
   <list [startIndex="integer"]> (intVar wspace)2+ </list>
</channel>
\end{syntax}
\end{boxsy}

Note that for that form, the opening and closing tags of  \xml{list} are optional when, of course, the attribute \att{startIndex} is not required, which gives:
\end{xl}
\begin{xc}
The constraint \gb{channel} ensures that if the $ith$ variable is assigned the value $j$, then the $jth$ variable must be assigned the value $i$.
For \x3-core, the number used for indexing the first variable in \xml{list} is necessarily 0.

\begin{boxsy}
\begin{syntax} 
<channel>
   <list> (intVar wspace)2+ </list>
</channel>
\end{syntax}
\end{boxsy}

Note that for that form the opening and closing tags of \xml{list} are optional, which gives:

\end{xc}

\begin{boxsy}
\begin{syntax} 
<channel> (intVar wspace)2+ </channel> @\com{Simplified Form}@
\end{syntax}
\end{boxsy}

\begin{boxse}
\begin{semantics}
$\gb{channel}(X)$, with $X=\langle x_0,x_1,\ldots \rangle$, iff
  $\forall i : 0 \leq i < |X|, \va{x}_i =j \Rightarrow \va{x}_j=i$
\end{semantics}
\end{boxse}

\begin{xl}
  Note that it is possible to enforce a stronger form through the restriction \val{irreflexive}; see Section \ref{sec:restricted}.
\end{xl}

Another classical form of \gb{channel}, sometimes called \gb{inverse} or \gb{assignment} in the literature, is defined from two separate lists of variables (that must be of same size). 
It ensures that the value assigned to the $ith$ variable of the first element \xml{list} gives the position of the variable of the second element \xml{list} that is assigned to $i$, and vice versa.
\begin{xl}For each list, the optional attribute \att{startIndex} gives the number used for indexing the first variable in this list (0, by default). 

\begin{boxsy}
\begin{syntax} 
<channel>
  <list [startIndex="integer"]> (intVar wspace)2+ </list>
  <list [startIndex="integer"]> (intVar wspace)2+ </list>
</channel>
\end{syntax}
\end{boxsy}
\end{xl}
\begin{xc}
For \x3-core, the number used for indexing the first variable in each list is necessarily 0.

\begin{boxsy}
\begin{syntax} 
<channel>
  <list> (intVar wspace)2+ </list>
  <list> (intVar wspace)2+ </list>
</channel>
\end{syntax}
\end{boxsy}
\end{xc}

\begin{boxse}
\begin{semantics}
$\gb{channel}(X,Y)$, with $X=\langle x_0,x_1,\ldots \rangle$ and $Y=\langle y_0,y_1,\ldots \rangle$, iff 
  $\forall i : 0 \leq i < |X|, \va{x}_i = j \Leftrightarrow \va{y}_j = i$ 

@{\em Prerequisite}@: $2 \leq |X| = |Y|$
\end{semantics}
\end{boxse}


\begin{boxex}
\begin{xcsp}  
<channel id="c1">
   <list> x0 x1 x2 x3 x4 </list>
   <list> y0 y1 y2 y3 y4 </list>
</channel>
 \end{xcsp}
\end{boxex}

It is possible to use this form of \gb{channel}, with two lists of different sizes.
The constraint then imposes restrictions on all variables of the first list, but not on all variables of the second list.
The syntax is the same, but the semantics is the following (note that the equivalence has been replaced by an implication):

\begin{boxse}
\begin{semantics}
$\gb{channel}(X,Y)$, with $X=\langle x_0,x_1,\ldots \rangle$ and $Y=\langle y_0,y_1,\ldots \rangle$, iff 
  $\forall i : 0 \leq i < |X|, \va{x}_i = j \Rightarrow \va{y}_j = i$ 

@{\em Prerequisite}@: $2 \leq |X| < |Y|$
\end{semantics}
\end{boxse}

Another form is obtained by considering a list of 0/1 variables to be channeled with an integer variable.
This third form of constraint \gb{channel} ensures that the only variable of \xml{list} that is assigned to 1 is at an index (position) that corresponds to the value assigned to the variable in \xml{value}.

\begin{xl}
\begin{boxsy}
\begin{syntax} 
<channel>
  <list [startIndex="integer"]> (@{\bnf{01Var}@ wspace)2+ </list>
  <value> intVar </value>
</channel>
\end{syntax}
\end{boxsy}
\end{xl}
\begin{xc}
\begin{boxsy}
\begin{syntax} 
<channel>
  <list> (@{\bnf{01Var}@ wspace)2+ </list>
  <value> intVar </value>
</channel>
\end{syntax}
\end{boxsy}
\end{xc}

\begin{boxse}
\begin{semantics}
$\gb{channel}(X,v)$, with $X=\langle x_0,x_1,\ldots\rangle$, iff 
  $\forall i : 0 \leq i < |X|, \va{x}_i = 1 \Leftrightarrow \va{v} = i$
  $\exists i : 0 \leq i < |X| \land \va{x}_i = 1$
\end{semantics}
\end{boxse}

\begin{boxex}
\begin{xcsp}  
<channel id="c2">
   <list> z0 z1 z2 z3 z4 z5 </list>
   <value> v </value>
</channel>
 \end{xcsp}
\end{boxex}

\begin{xl}
\subsubsection{Constraint \gb{permutation}}\label{ctr:permutation}

The constraint \gb{permutation} ensures that the tuple of values taken by variables of the second element \xml{list} is a permutation of the tuple of values taken by variables of the first element \xml{list}.
In other words, both lists represent the same multi-set.
The optional element \xml{mapping} gives for each value of the first tuple its position in the second tuple.
If the element \xml{mapping} is present, then it is possible to introduce the optional attribute \att{startIndex} that gives the number used for indexing the first variable in the second element \xml{list} (0, by default). 

\begin{boxsy}
\begin{syntax} 
<permutation>
  <list> (intVar wspace)2+ </list>
  <list [startIndex="integer"]> (intVar wspace)2+ </list>
  [<mapping> (intVar wspace)2+ </mapping>]
</permutation>
\end{syntax}
\end{boxsy}

\begin{boxse}
\begin{semantics}
$\gb{permutation}(X,Y)$, with $X=\langle x_1,x_2,\ldots \rangle$ and $Y=\langle y_1,y_2,\ldots \rangle$, iff 
  $\{\!\{\va{x}_i : 1 \leq i \leq |X|\}\!\} = \{\!\{\va{y}_i : 1 \leq i \leq |Y|\}\!\}$
$\gb{permutation}(X,Y,M)$, with $M=\langle m_1,m_2,\ldots \rangle$, iff 
  $\gb{permutation}(X,Y) \land \forall i : 1 \leq i \leq |X|, \va{x}_i = \va{y}_{\va{m}_i} \land \forall (i,j) : 1 \leq i < j \leq |M|, \va{m}_i \neq \va{m}_j$

$\gbc{Prerequisite}: |X|=|Y|=|M| \geq 2$
\end{semantics}
\end{boxse}

\begin{boxex}
\begin{xcsp}  
<permutation id="c">
   <list> x1 x2 x3 x4 </list>
   <list> y1 y2 y3 y4 </list>
</permutation>
 \end{xcsp}
\end{boxex}
 
The constraint \gb{permutation} captures \gb{correspondence} in the \cat.
For a generalization of \gb{permutation}, see \gb{allEqual} on multisets in Section \ref{sec:allEqualLifted}.
For a specialization of \gb{permutation}, where the second list is increasingly sorted, see Section \ref{sec:restricted}.
\end{xl}

\subsection{Packing and Scheduling Constraints}

In this section, we present constraints that are useful in packing and scheduling problems. More specifically, we introduce:
\begin{enumerate}
\begin{xl}\item \gb{stretch}
\end{xl}
\item \gb{noOverlap} (capturing \gb{disjunctive} and \gb{diffn})
\item \gb{cumulative} \begin{xl}(capturing \gb{cumulatives} too)\end{xl}
\item \gb{binPacking}
\item \gb{knapsack}
  \begin{xl}
  \item \gb{flow}
  \end{xl}
\end{enumerate}

\begin{xl}
\subsubsection{Constraint \gb{stretch}}\label{ctr:stretch}

The constraint \gb{stretch} \cite{P_filtering} aims at grouping values in sequences.
Each stretch is a maximal subsequence with the same value over the list of variables in \xml{list}.
Each stretch must have a width compatible with the corresponding interval specified in \xml{widths}.
The optional element \xml{patterns} gives the possible successive values between stretches.

\begin{boxsy}
\begin{syntax} 
<stretch>
   <list> (intVar wspace)2+ </list>
   <values> (intVal wspace)+ </values>
   <widths> (intIntvl wspace)+ </widths>
   [<patterns> ("(" intVal "," intVal ")")+ </patterns>]
</stretch>
\end{syntax}
\end{boxsy}

Given a sequence of variables $X=\langle x_1,x_2,\ldots \rangle$, and a value assigned to each of these variables, we define as $\mathtt{stretches}(\va{X})$ the set of stretches over $X$, each one being identified by a pair $(v_k,i..j)$ composed of the value $v_k$ of the stretch and the interval $i..j$ of the indexes of the variables composing the stretch.

\begin{boxse}
\begin{semantics}
$\gb{stretch}(X,V,W)$, with $X=\langle x_1,x_2,\ldots \rangle$, $V=\langle v_1,v_2,\ldots \rangle$, $W=\langle l_1..u_1,l_2..u_2,\ldots\rangle$, iff  
  $\forall (v_k,i..j) \in \mathtt{stretches}(\va{X}), l_k \leq j-i+1 \leq u_k$ 
$\gb{stretch}(X,V,W,P)$ iff 
  $\gb{stretch}(X,V,W) \land \forall i : 1 \leq i < |X|, \va{x}_i \neq \va{x}_{i+1} \Rightarrow (\va{x}_i,\va{x}_{i+1}) \in P$

$\gbc{Prerequisite}: |V| = |W|$
\end{semantics}
\end{boxse}

Although the section is devoted to integer variables, we give an example with symbolic values, as this is a straightforward adaptation. 

\begin{boxex}
\begin{xcsp}  
<stretch>
   <list> x1 x2 x3 x4 x5 x6 x7 </list>
   <values> morning afternoon night off </values>
   <widths> 1..3 1..3 2..3 2..4 </widths>
</stretch>
 \end{xcsp}
\end{boxex}

\begin{remark}
In the future, we might authorize longer forms of patterns.
\end{remark}
\end{xl}

\subsubsection{Constraint \gb{noOverlap}}\label{ctr:noOverlap}


We start with the one dimensional form of \gb{noOverlap} \cite{H_integrated} that corresponds to \gb{disjunctive} \cite{C_one} and ensures that some tasks, defined by their origins and durations (lengths), must not overlap.
The attribute \att{zeroIgnored} is optional (\val{true}, by default); when set to \val{false}, it indicates that zero-length tasks cannot be packed anywhere (cannot overlap with other tasks). 

\begin{boxsy}
\begin{syntax}  
<noOverlap [zeroIgnored="boolean"] >
   <origins> (intVar wspace)2+ </origins>
   <lengths> ((intVal | intVar) wspace)2+ </lengths>
</noOverlap>
 \end{syntax}
\end{boxsy} 

The semantics is given for \att{zeroIgnored}=\val{false}.

\begin{boxse}
\begin{semantics}
$\gb{noOverlap}(X,L)$, with $X=\langle x_0,x_1,\ldots \rangle$ and $L=\langle l_0,l_1,\ldots \rangle$, iff  
  $\forall (i,j) : 0 \leq i < j < |X|,  \va{x}_{i} + \va{l}_{i} \leq \va{x}_{j} \vee \va{x}_{j} + \va{l}_{j} \leq \va{x}_{i}$

$\gbc{Prerequisite}: |X| = |L| \geq 2$
\end{semantics}
\end{boxse}

\begin{boxex}
\begin{xcsp}  
<noOverlap>
   <origins> x0 x1 x2 x3 </origins>
   <lengths> l0 l1 l2 l3 </lengths>
</noOverlap>
 \end{xcsp}
\end{boxex}
 

The k-dimensional form of \gb{noOverlap} corresponds to \gb{diffn} \cite{BC_chip} and ensures that, given a set of $n$-dimensional boxes; for any pair of such boxes, there exists at least one dimension where one box is after the other, i.e., the boxes do not overlap.
The attribute \att{zeroIgnored} is optional (\val{true}, by default); when set to \val{false}, it indicates that zero-width boxes cannot be packed anywhere (cannot overlap with other boxes). 

\begin{boxsy}
\begin{syntax}  
<noOverlap [zeroIgnored="boolean"]>
   <origins> ("(" intVar ("," intVar)+ ")")2+ </origins>
   <lengths> 
     ("(" (intVal | intVar) ("," (intVal | intVar))+ ")")2+ 
   </lengths>
</noOverlap>
 \end{syntax}
\end{boxsy} 

The semantics is given for \att{zeroIgnored}=\val{false}; for readability reasons, we consider below $n+1$-dimensional boxes.

\begin{boxse}
\begin{semantics}
$\gb{noOverlap}(\ns{X},\ns{L})$, with $\ns{X}=\langle (x_{0,0},\ldots,x_{0,n}),(x_{1,0},\ldots,x_{1,n}),\ldots \rangle$ and $\ns{L}=\langle (l_{0,0},\ldots,l_{0,n}),(l_{1,0},\ldots,l_{1,n}),\ldots \rangle$, iff  
  $\forall (i,j) : 0 \leq i < j < |\ns{X}|,  \exists k \in 0..n :  \va{x}_{i,k} + \va{l}_{i,k} \leq \va{x}_{j,k} \vee \va{x}_{j,k} + \va{l}_{j,k} \leq \va{x}_{i,k}$

$\gbc{Prerequisite}: |\ns{X}| = |\ns{L}| \geq 2$
\end{semantics}
\end{boxse}

The following constraint enforces that all four 3-dimensional specified boxes do not overlap.
The first box has origin $(x_0,y_0,z_0)$ and length $(2,4,1)$, meaning that this box is situated from $x_0$ to $x_0+2$ on x-axis, from $y_0$ to $y_0+4$ on y-axis, and from $z_0$ to $z_0+1$ on z-axis.

\begin{boxex}
\begin{xcsp}
<noOverlap>
   <origins> (x0,y0,z0)(x1,y1,z1)(x2,y2,z2)(x3,y3,z3) </origins>  
   <lengths> (2,4,1)(4,2,3)(5,1,2)(3,3,2) </lengths>  
</noOverlap>
 \end{xcsp}
\end{boxex}

\subsubsection{Constraint \gb{cumulative}}\label{ctr:cumulative}

\begin{xl}
Here, we have a collection of tasks, each one described by 4 attributes: origin, length, end and height.
In \x3, the element \xml{ends} is optional.
The constraint \gb{cumulative} \cite{AB_extending} enforces that at each point in time, the cumulated height of tasks that overlap that point, respects a numerical condition.
When the operator ``le'' is used, this corresponds to not exceeding a given limit.

\begin{boxsy}
\begin{syntax}  
<cumulative>
   <origins> (intVar wspace)2+ </origins>
   <lengths> (intVal wspace)2+ | (intVar wspace)2+ </lengths>
   [<ends> (intVar wspace)2+ </ends>]
   <heights> (intVal wspace)2+ | (intVar wspace)2+ </heights>
   <condition> "(" operator "," operand ")" </condition> 
</cumulative>
\end{syntax}
\end{boxsy} 

\begin{boxse}
\begin{semantics}
$\gb{cumulative}(X,L,H,(\odot,k))$, with $X=\langle x_0,x_1,\ldots\rangle$, $L=\langle l_0,l_1,\ldots\rangle$, $H=\langle h_0,h_1,\ldots\rangle$, iff 
  $\forall t \in \N, \sum \{\va{h}_i : 0 \leq i < |X| \land \va{x}_i \leq t < \va{x}_i + \va{l}_i\} \odot \va{k}$

$\gbc{Prerequisite}: |X| = |L| = |H| \geq 2$
\end{semantics}
\end{boxse}
\end{xl}
\begin{xc}
Here, we have a collection of tasks, each one described by 3 attributes: origin, length, and height.
The constraint \gb{cumulative} \cite{AB_extending} enforces that at each point in time, the cumulated height of tasks that overlap that point, respects a numerical condition.
When the operator ``le'' is used, this corresponds to not exceeding a given limit.

\begin{boxsy}
\begin{syntax}  
<cumulative>
   <origins> (intVar wspace)2+ </origins>
   <lengths> (intVal wspace)2+ | (intVar wspace)2+ </lengths>
   <heights> (intVal wspace)2+ | (intVar wspace)2+ </heights>
   <condition> "(" operator "," operand ")" </condition> 
</cumulative>
\end{syntax}
\end{boxsy} 

\begin{boxse}
\begin{semantics}
$\gb{cumulative}(X,L,H,(\odot,k))$, with $X=\langle x_0,x_1,\ldots\rangle$, $L=\langle l_0,l_1,\ldots\rangle$, $H=\langle h_0,h_1,\ldots\rangle$, iff 
  $\forall t \in \N, \sum \{\va{h}_i : 0 \leq i < |X| \land \va{x}_i \leq t < \va{x}_i + \va{l}_i\} \odot \va{k}$

$\gbc{Prerequisite}: |X| = |L| = |H| \geq 2$
\end{semantics}
\end{boxse}
\end{xc}

\begin{xl}
If the element \xml{ends} is present, for the semantics, we have additionally that:
\begin{quote}
$\forall i : 0 \leq i < |X| , \va{x}_i + \va{l}_i = \va{e}_i$
\end{quote}
\end{xl}

An example is given below for five tasks; the first task starts at $s_0$, and its length and height (resource consumption level) are respectively 3 and 1.
At any time point, it is not possible to exceed 4 (resource consumption units).

\begin{boxex}
\begin{xcsp}
<cumulative>
   <origins> s0 s1 s2 s3 s4 </origins>
   <lengths> 3 2 5 4 2 </lengths>
   <heights> 1 2 1 1 3 </heights>
   <condition> (le,4) </condition>
</cumulative>
\end{xcsp}
\end{boxex}
 
\begin{xl}
A refined form of \gb{cumulative} corresponds to the constraint sometimes called \gb{cumulatives} \cite{BC_cumulatives}.
We have a set of available machines (resources), with their respective numerical conditions in \xml{conditions}, which replaces \xml{condition}, through a sequence of pairs composed of an operator and an operand.
Each task is thus performed by a machine, which is given in an element \xml{machines}.
For each machine, the cumulated height of tasks that overlap a time point for this machine must respect the numerical (capacity) condition associated with the machine.
The optional attribute \att{startIndex} gives the number used for indexing the first numerical condition (and so, the first machine) in this list (0, by default). 

\begin{boxsy}
\begin{syntax}  
<cumulative>
   <origins> (intVar wspace)2+ </origins>
   <lengths> (intVal wspace)2+ | (intVar wspace)2+ </lengths>
   [ <ends> (intVar wspace)2+ </ends> ]
   <heights> (intVal wspace)2+ | (intVar wspace)2+ </heights>
   <machines> (intVar wspace)2+ </machines>
   <conditions [startIndex="integer"]> 
     ("(" operator "," operand ")")2+ 
   </conditions>
</cumulative>
 \end{syntax}
\end{boxsy} 

\begin{boxse}
\begin{semantics}
$\gb{cumulative}(X,L,H,M,C)$, with $X=\langle x_1,x_2,\ldots\rangle$,  $L=\langle l_1,l_2,\ldots\rangle$, $H=\langle h_1,h_2,\ldots\rangle$, $M=\langle m_1,m_2,\ldots\rangle$, and $C=\langle (\odot_1,k_1),(\odot_2,k_2),\ldots\rangle$ iff
  $\forall m \in \{\va{m}_i : 1 \leq i \leq |M|\}, \forall t \in \N : \exists 1 \leq i \leq |H| : \va{x}_i \leq t < \va{x}_i + \va{l}_i \land \va{m}_i=m$, 
    $\sum \{\va{h}_i : 1 \leq i \leq |H| \land \va{x}_i \leq t < \va{x}_i + \va{l}_i \land \va{m}_i=m\} \odot_m k_m$

$\gbc{Prerequisite}: |X| = |L| = |H| = |M| \geq 2$
\end{semantics}
\end{boxse}
\end{xl}

\subsubsection{Constraint \gb{binPacking}}\label{ctr:binPacking}

The first form of the constraint \gb{binPacking} \cite{S_constraint,S_solving,CMOS_bin} ensures that a list of items, whose sizes are given, are put in different bins in such a way that the total size of the items in each bin
respects a numerical condition (always the same, because the capacity is assumed to be the same for all bins). 
When the operator ``le'' is used, this corresponds to not exceeding the capacity of each bin. 

\begin{boxsy}
\begin{syntax}  
<binPacking>
   <list> (intVar wspace)2+ </list>
   <sizes> (intVal wspace)2+ </sizes>
   <condition> "(" operator "," operand ")" </condition> 
</binPacking>
 \end{syntax}
\end{boxsy} 

\begin{boxse}
\begin{semantics}
$\gb{binPacking}(X,S,(\odot,k))$, with $X=\langle x_1,x_2,\ldots \rangle$ and $S=\langle s_1,s_2,\ldots \rangle$, iff 
  $\forall b \in \{\va{x}_i : 1 \leq i \leq |X|\}, \sum \{s_i : 1 \leq i \leq |S| \land \va{x}_i =b\} \odot \va{k}$

$\gbc{Prerequisite}: |X| = |S| \geq 2$
\end{semantics}
\end{boxse}

As an example, we have:

\begin{boxex}
\begin{xcsp} 
<binPacking id="c1">
  <list> x1 x2 x3 x4 x5 </list>
  <sizes> 25 53 38 41 32 </sizes>
  <condition> (le,100) </condition>
</binPacking>
 \end{xcsp}
\end{boxex}

The second form of the constraint \gb{binPacking} associates a specific limit (capacity) with each bin.
The limits are given either by integer values or by integer variables.

\begin{boxsy}
\begin{syntax}  
<binPacking>
   <list> (intVar wspace)2+ </list>
   <sizes> (intVal wspace)2+ </sizes>
   <limits> (intVal wspace)2+ | (intVar wspace)2+ </limits> 
</binPacking>
 \end{syntax}
\end{boxsy} 

\begin{boxse}
\begin{semantics}
$\gb{binPacking}(X,S,C)$, with $X=\langle x_1,x_2,\ldots \rangle$, $S=\langle s_1,s_2,\ldots \rangle$ and $C=\langle c_1,c_2,\ldots \rangle$ iff 
  $\forall b \in \{\va{x}_i : 1 \leq i \leq |X|\}$, $\sum \{s_i : 1 \leq i \leq |S| \land \va{x}_i =b\} \leq \va{c_b}$
  
$\gbc{Prerequisite}: |X| = |S| \geq 2$
\end{semantics}
\end{boxse}

The third form of the constraint \gb{binPacking} associates a specific load with each bin.
The loads are given either by integer values or by integer variables.

\begin{boxsy}
\begin{syntax}  
<binPacking>
   <list> (intVar wspace)2+ </list>
   <sizes> (intVal wspace)2+ </sizes>
   <loads> (intVal wspace)2+ | (intVar wspace)2+ </loads> 
</binPacking>
 \end{syntax}
\end{boxsy} 

\begin{boxse}
\begin{semantics}
$\gb{binPacking}(X,S,L)$, with $X=\langle x_1,x_2,\ldots \rangle$, $S=\langle s_1,s_2,\ldots \rangle$ and $L=\langle l_1,l_2,\ldots \rangle$ iff 
  $\forall b \in \{\va{x}_i : 1 \leq i \leq |X|\}$, $\sum \{s_i : 1 \leq i \leq |S| \land \va{x}_i =b\} = \va{l_b}$.
  
$\gbc{Prerequisite}: |X| = |S| \geq 2$
\end{semantics}
\end{boxse}

The constraint $c_2$ involves 7 items and 3 bins. Here, the variables $l_1$, $l_2$ and $l_3$ are used to denote loads.

\begin{boxex}
\begin{xcsp} 
<binPacking id="c2">
   <list>  y1 y2 y3 y4 y5 y6 y7 </list>
   <sizes> 12 7 21 36 19 22 30 </sizes>
   <loads> l1 l2 l3 </loads>
</binPacking>
 \end{xcsp}
\end{boxex}

\begin{xl}
A last very general form of \gb{binPacking} involves a specific numerical condition that is associated with each available bin.
The element \xml{condition} of the first form is replaced by an element \xml{conditions}.
The optional attribute \att{startIndex} gives the number used for indexing the first numerical condition (and so, the first bin) in this list (0, by default). 

\begin{boxsy}
\begin{syntax}  
<binPacking>
   <list> (intVar wspace)2+ </list>
   <sizes> (intVal wspace)2+ </sizes> 
   <conditions [startIndex="integer"]> 
     ("(" operator "," operand ")")2+ 
   </conditions>
</binPacking>
 \end{syntax}
\end{boxsy} 

For the semantics, we assume that \att{startIndex}=\val{1}.

\begin{boxse}
\begin{semantics}
$\gb{binPacking}(X,S,C)$, with $X=\langle x_1,x_2,\ldots \rangle$, $S=\langle s_1,s_2,\ldots \rangle$, and $C=\langle (\odot_1,k_1),(\odot_2,k_2),\ldots\rangle$, iff $\forall b : 1 \leq b \leq |C|, \sum \{s_i : 1 \leq i \leq |S| \land \va{x}_i=b\} \odot_b \va{k}_b$

$\gbc{Prerequisite}: |X| = |S| \geq 2 \wedge |C| \geq 2$
\end{semantics}
\end{boxse}

The constraint $c_3$ involves 7 items and 3 bins.
The first bin has 20 as capacity, the load of second bin is (must be) exactly 8, and the load of the third bin is given by the variable $l$.   

\begin{boxex}
\begin{xcsp} 
<binPacking id="c3">
   <list>  y1 y2 y3 y4 y5 y6 y7 </list>
   <sizes> 12 7 21 36 19 22 30 </sizes>
   <conditions> (le,20)(eq,8)(lt,l) </conditions>
</binPacking>
 \end{xcsp}
\end{boxex}
\end{xl}

\subsubsection{Constraint \gb{knapsack}}\label{ctr:knapsack}

The constraint \gb{knapsack} \cite{FS_cost,S_approximated,MSS_filtering} ensures that some items are packed in a knapsack with certain weight and profit restrictions.
 
\begin{boxsy}
\begin{syntax}  
<knapsack>
   <list> (intVar wspace)2+ </list> 
   <weights> (intVal wspace)2+ </weights>
   <condition> "(" operator "," operand ")" </condition> 
   <profits> (intVal wspace)2+ </profits>
   <condition> "(" operator "," operand ")" </condition> 
</knapsack>
 \end{syntax}
\end{boxsy} 

The first element \xml{condition} is related to weights whereas the second element \xml{condition} is related to profits.
The operator of the first condition is expected to be in $\{<,\leq,=\}$ whereas the operator of the second condition is expected to be in $\{>,\geq,=\}.$ 

\begin{boxse}
\begin{semantics}
  $\gb{knapsack}(X,W,(\odot_w,k_w),P,(\odot_p,k_p))$, with $X=\langle x_1,x_2,\ldots \rangle$, $W=\langle w_1,w_2,\ldots\rangle$, and $P=\langle p_1,p_2,\ldots \rangle$, iff
  $\sum_{i=1}^{|X|} (w_i \times \va{x}_i) \odot_w \va{k_w}$
  $\sum_{i=1}^{|X|} (p_i \times \va{x}_i) \odot_p \va{k_p}$

$\gbc{Prerequisite}: |X| = |W| = |P| \geq 2$
\end{semantics}
\end{boxse}

\begin{xl}
\subsubsection{Constraint \gb{flow}}\label{ctr:flow}

The constraint \gb{flow} (initially called \gb{networkFlow}) ensures that there is a valid flow in a graph.
The flow of each arc and the balance (difference between input and output flows) of each node are given by elements \xml{list} and \xml{balance}, respectively.
Optionally, a unit weight can be associated with each, arc, and the total cost must then be related to the overall cost of the flow by a numerical condition.
Also, there can be an element \xml{capacities} that give lower and upper capacities of each arc. 

\begin{boxsy}
\begin{syntax}  
<flow>
   <list> (intVar wspace)2+ </list> 
   <balance> (intVal wspace)2+ | (intVar wspace)2+ </balance>
   <arcs> ("(" node "," node ")")2+ </arcs>
   [<capacities> (intIntvl wspace)2+ </capacities>] 
   [ 
     <weights> (intVal wspace)2+ | (intVar wspace)2+ </weights>
     <condition> "(" operator "," operand ")" </condition> 
   ]
</flow>
 \end{syntax}
\end{boxsy} 

For the semantics, considering the graph defined by \xml{arcs}, for any node $v$, $\Gamma^+(v)$ denotes the set of arcs (of this graph) with head $v$ and $\Gamma^-(v)$ denotes the set of arcs with tail $v$.

\begin{boxse}
\begin{semantics}
$\gb{flow}(X,B,A)$, with $X=\langle x_1,x_2,\ldots\rangle$, $B=\langle b_1,b_2,\ldots\rangle$, $A=\langle a_1,a_2,\ldots\rangle$, iff
  for any node $v, \sum \{\va{x}_i : 1 \leq i \leq |X| \land a_i \in \Gamma^-(v)\} - \sum \{\va{x}_i : 1 \leq i \leq |X| \land a_i \in \Gamma^+(v)\} = \va{b}_i $
$\gb{flow}(X,B,A,C)$, with $C=\langle l_1..u_1,l_2..u_2,\ldots\rangle$, iff
  $\gb{flow}(X,B,A) \land \forall i : 1 \leq i \leq |X|, \va{x}_i \in l_i..u_i$  
$\gb{flow}(X,B,A,C,W,(\odot,k))$, with $W=\langle w_1,w_2,\ldots\rangle$, iff
  $\gb{flow}(X,B,A,C) \land \sum \{\va{x}_i \times w_i : 1 \leq i \leq |X|\} \odot \va{k}$

$\gbc{Prerequisite}: |X| = |A| = |C| = |W|$
\end{semantics}
\end{boxse}
\end{xl}

\subsection{Constraints on Graphs}

In this section, we present constraints that are defined on graphs using integer variables (encoding called ``successors variables'').
We introduce:

\begin{enumerate}
\item \gb{circuit}
\begin{xl}\item \gb{nCircuits}
\item \gb{path}
\item \gb{nPaths}
\item \gb{tree}
\item \gb{nTrees}
\end{xl}
\end{enumerate}

For all these constraints, we have an element \xml{list} that contains variables $x_0, x_1,\ldots$
The assumption is that each pair $(i,\va{x}_i)$ represents an arc (or edge) of the graph to be built; if $\va{x}_i=j$, then it means that the successor of node $i$ is node $j$.
Note that a {\em loop} (also called self-loop) corresponds to a variable $x_i$ such that $\va{x}_i=i$; it is {\em isolated} if there is no variable $x_j$ with $j \neq i$ such that $\va{x}_j=i$.

\subsubsection{Constraint \gb{circuit}}\label{ctr:circuit}

The constraint \gb{circuit} \cite{BC_chip} ensures that the values taken by the variables in \xml{list} forms a circuit, with the assumption that each pair $(i,\va{x}_i)$ represents an arc.
The optional attribute \att{startIndex} of \xml{list} gives the number used for indexing the first variable/node in \xml{list} (0, by default). 
The optional element \att{size} indicates that the circuit must be of a given size (strictly greater than $1$).

It is important to note that the circuit is not required to cover all nodes (the nodes that are not present in the circuit are then self-looping).
Hence \gb{circuit}, with loops being simply ignored, basically represents \gb{subcircuit} (e.g., in \mzinc).
If ever you need a full circuit (i.e., without any loop), you have three solutions:
\begin{itemize}
\item put in \att{size} a value corresponding to the number of variables in \xml{list}
\item initially define the variables without the self-looping values, 
\item post unary constraints.
\end{itemize}


\begin{boxsy}
\begin{syntax} 
<circuit>
   <list [startIndex="integer"]> (intVar wspace)2+ </list>
   [<size> intVal | intVar </size>]
</circuit>
 \end{syntax}
\end{boxsy}

Note that the opening and closing tags of \xml{list} are optional if \xml{list} is the unique parameter of the constraint and the attribute \att{startIndex} is not necessary, which gives:

\begin{boxsy}
\begin{syntax} 
<circuit> (intVar wspace)2+ </circuit> @\com{Simplified Form}@
\end{syntax}
\end{boxsy}

\begin{boxse}
\begin{semantics}
$\gb{circuit}(X)$, with $X=\langle x_0,x_1,\ldots\rangle$, iff  @\com{capture \gb{subscircuit}}@
  $\{(i,\va{x}_i) : 0 \leq i < |X| \land i \neq \va{x}_i \}$ forms a circuit of size $> 1$ 
$\gb{circuit}(X,z)$, with $X=\langle x_0,x_1,\ldots\rangle$, iff  
  $\{(i,\va{x}_i) : 0 \leq i < |X| \land i \neq \va{x}_i \}$ forms a circuit of size $\va{z} > 1$
\end{semantics}
\end{boxse}

In the following example, the constraint states that $\langle x_0,x_1,x_2,x_3 \rangle$ must form a single circuit -- for example, with $\langle \va{x}_0,\va{x}_1,\va{x}_2,\va{x}_3 \rangle = \langle 1,2,3,0\rangle$, or a subcircuit -- for example, with $\langle \va{x}_0,\va{x}_1,\va{x}_2,\va{x}_3 \rangle = \langle 1,2,0,3\rangle$.

\begin{boxex}
\begin{xcsp}  
<circuit>
   x0 x1 x2 x3 
</circuit>
 \end{xcsp}
\end{boxex}

\begin{xl}
Note that it is possible to capture the variant of \gb{circuit} in \gecode that involves a cost matrix.
It suffices to logically combine \gb{circuit} and \gb{sumCosts} as shown in Section \ref{ctr:and}. 
\end{xl}

\begin{xl}
\subsubsection{Constraint \gb{nCircuits}}\label{ctr:nCircuits}

One may be interested in building more than one circuit.
For constraint \gb{nCircuits}, the number of circuits obtained after instantiating variables of \xml{list} must respect a numerical condition.
In the literature, this variant with $k>1$ circuits is called \gb{cycle}.
The optional attribute \att{countLoops} indicates whether or not loops must be considered as circuits (\val{true}, by default).

\begin{boxsy}
\begin{syntax} 
<nCircuits [countLoops="boolean"]>
   <list [startIndex="integer"]> (intVar wspace)2+ </list>
   <condition> "(" operator "," operand ")" </condition> 
</nCircuits>
 \end{syntax}
\end{boxsy}

\begin{boxse}
\begin{semantics}
$\gb{nCircuits}(X,(\odot,k))$, with $X=\langle x_1,x_2,\ldots\rangle$, iff  
  $\{(i,\va{x}_i) : 1 \leq i \leq |X|\}$ forms $p$ node-disjoint circuits s.t. $p \odot \va{k}$
\end{semantics}
\end{boxse}

\begin{remark}
It is always possible to avoid having self-loops as circuits, by combining \gb{circuits} with a set of unary constraints.
\end{remark}

Below, the constraint states that $\langle y_0,y_1,y_2,y_3,y_4 \rangle$ must form two disjoint circuits -- for example, with $\langle \va{y}_0,\va{y}_1,\va{y}_2,\va{y}_3,\va{y}_4 \rangle = \langle 1,0,4,2,3 \rangle$.

\begin{boxex}
\begin{xcsp}  
<nCircuits>
   <list> y0 y1 y2 y3 y4 </list>
   <condition> (eq,2) </condition>
</nCircuits>
 \end{xcsp}
\end{boxex}

\subsubsection{Constraint \gb{path}}\label{ctr:path}

The constraint \gb{path} ensures that the values taken by the variables in \xml{list} forms a path from the node specified in \xml{start} to the node specified in \xml{final}, with the assumption that each pair $(i,\va{x}_i)$ represents an arc.
The optional attribute \att{startIndex} of \xml{list} gives the number used for indexing the first variable/node in \xml{list} (0, by default). 
The successor of the final node is necessary itself (self-loop). 
The optional element \att{size} indicates that the path must be of a given size.

It is important to note that the path is not required to cover all nodes (the nodes that are not present in the path are then self-looping).
Hence \gb{path}, with loops being simply ignored, basically represents \gb{subpath} (e.g., in Choco3).
If ever you need a full path (i.e., without any loop other than the final node), you have three solutions:
\begin{itemize}
\item indicate in \att{size} the number of variables
\item initially define the variables without the self-looping values, 
\item post unary constraints.
\end{itemize}

\begin{boxsy}
\begin{syntax} 
<path>
   <list [startIndex="integer"]> (intVar wspace)2+ </list>
   <start> intVal | intVar </start>
   <final> intVal | intVar </final>
   [<size> intVal | intVar </size>]
</path>
\end{syntax}
\end{boxsy}

\begin{boxse}
\begin{semantics}
$\gb{path}(X,s,f)$, with $X=\langle x_1,x_2,\ldots\rangle$, iff  @\com{capture \gb{subspath}}@
  $\{(i,\va{x}_i) : 1 \leq i \leq |X|\ \land i \neq \va{x}_i \}$ forms a path from $\va{s}$ to $\va{f}$
$\gb{path}(X,s,f,z)$, with $X=\langle x_1,x_2,\ldots\rangle$, iff  
  $\{(i,\va{x}_i) : 1 \leq i \leq |X| \land i \neq \va{x}_i \}$ forms a path from $\va{s}$ to $\va{f}$ of size $\va{z}$
\end{semantics}
\end{boxse}

In the following example, the constraint states that $\langle x_0,x_1,x_2,x_3 \rangle$ must form a path from $s$ to $f$ -- for example, $\langle \va{x}_0,\va{x}_1,\va{x}_2,\va{x}_3\rangle = \langle 0,0,3,1\rangle$ with $\va{s}=2$ and $\va{f}=0$ satisfies the constraint.

\begin{boxex}
\begin{xcsp}  
<path>
   <list> x0 x1 x2 x3 </list>
   <start> s </start>
   <final> f </final>
</path>
\end{xcsp}
\end{boxex}

\subsubsection{Constraint \gb{nPaths}}\label{ctr:nPaths}

The constraint \gb{nPaths} ensures that the number of paths formed after instantiating variables of \xml{list} must respect a numerical condition.
The optional attribute \att{countLoops} indicates whether or not isolated loops must be considered as paths (\val{true}, by default).

\begin{boxsy}
\begin{syntax} 
<nPaths [countLoops="boolean"]>
   <list [startIndex="integer"]> (intVar wspace)2+ </list>
   <condition> "(" operator "," operand ")" </condition> 
</nPaths>
 \end{syntax}
\end{boxsy}

\begin{boxse}
\begin{semantics}
$\gb{nPaths}(X,(\odot,k))$, with $X=\langle x_1,x_2,\ldots\rangle$, iff  
  $\{(i,\va{x}_i) : 1 \leq i \leq |X|\}$ forms $p$ node-disjoint paths s.t. $p \odot \va{k}$
\end{semantics}
\end{boxse}

The constraint $c$ states that $\langle y_0,y_1,\ldots,y_7 \rangle$ must form $k$ disjoints paths -- for example, $\langle \va{y}_0,\va{y}_1,\ldots,\va{y}_7 \rangle = \langle 0,2,4,6,0,5,6,5\rangle$ with $\va{k}=3$.

\begin{boxex}
\begin{xcsp}  
<nPaths id="c">
   <list> y0 y1 y2 y3 y4 y5 y6 y7 </list>
   <condition> (eq,k) </condition>
</nPaths>
\end{xcsp}
\end{boxex}
 

\subsubsection{Constraint \gb{tree}}\label{ctr:tree}

The constraint \gb{tree} \cite{BFL_tree,FL_tree} ensures that the values taken by variables in \xml{list} forms an anti-arborescence (covering all nodes) whose root is specified by \xml{root}, with the assumption that each pair $(i,\va{x}_i)$ represents an arc (with such representation, we do obtain an anti-arborescence).
The optional attribute \att{startIndex} of \xml{list} gives the number used for indexing the first variable/node in \xml{list} (0, by default). 
The successor of the root node is necessarily itself (self-loop). 
The optional element \att{size} indicates that the anti-arborescence must be of a given size.

It is important to note that the anti-arborescence is not required to cover all nodes (the nodes that are not present in the anti-arborescence are then self-looping).
Hence \gb{tree}, with loops being simply ignored, basically represents \gb{subtree}.
If ever you need a full anti-arborescence (i.e., without any loop other than the root node), you have three solutions:
\begin{itemize}
\item indicate in \att{size} the number of variables
\item initially define the variables without the self-looping values, 
\item post unary constraints.
\end{itemize}

\begin{boxsy}
\begin{syntax} 
<tree>
   <list [startIndex="integer"]> (intVar wspace)2+ </list>
   <root> intVal | intVar </root>
   [<size> intVal | intVar </size>]
</tree>
 \end{syntax}
\end{boxsy}

\begin{boxse}
\begin{semantics}
$\gb{tree}(X,r)$, with $X=\langle x_1,x_2,\ldots\rangle$, iff  
  $\{(i,\va{x}_i) : 1 \leq i \leq |X| \land i \neq \va{x}_i \}$ forms an anti-arborescence of root $\va{r}$.
$\gb{tree}(X,r,z)$, with $X=\langle x_1,x_2,\ldots\rangle$, iff  
  $\{(i,\va{x}_i) : 1 \leq i \leq |X| \land i \neq \va{x}_i \}$ forms an anti-arborescence of root $\va{r}$ and size $\va{z}$
\end{semantics}
\end{boxse}

In the following example, the constraint states that $\langle x_0,x_1,x_2,x_3\rangle$ must form an anti-arborescence of root $r$ -- for example, $\langle \va{x}_0,\va{x}_1,\va{x}_2,\va{x}_3 \rangle = \langle 1,3,3,3\rangle$ with $\va{r}=3$ satisfies the constraint.

\begin{boxex}
\begin{xcsp}  
<tree>
   <list> x0 x1 x2 x3 </list>
   <root> r </root>
</tree>
\end{xcsp}
\end{boxex}

\subsubsection{Constraint \gb{nTrees}}\label{ctr:nTrees}

The constraint \gb{nTrees} ensures that the number of anti-arborescences formed after instantiating variables in \xml{list}  must respect a numerical condition. 
The optional attribute \att{countLoops} indicates whether or not isolated loops must be considered as trees (\val{true}, by default).

\begin{boxsy}
\begin{syntax} 
<nTrees [countLoops="boolean"]>
   <list> (intVar wspace)2+ </list>
   <condition> "(" operator "," operand ")" </condition> 
</nTrees>
 \end{syntax}
\end{boxsy}

\begin{boxse}
\begin{semantics}
$\gb{nTrees}(X,(\odot,k))$, with $X=\langle x_1,x_2,\ldots\rangle$, iff  
  $\{(i,\va{x}_i) : 1 \leq i \leq |X|\}$ forms $p$ disjoint anti-arborescences s.t. $p \odot \va{k}$
\end{semantics}
\end{boxse}

The constraint $c$ states that $\langle y_0,y_1,\ldots,y_7 \rangle$ must form $k$ node-disjoints anti-arborescences -- for example, $\langle \va{y}_0,\va{y}_1,\ldots,\va{y}_7 \rangle = \langle 0,4,4,6,0,0,6,4 \rangle$ with $\va{k}=2$.

\begin{boxex}
\begin{xcsp}  
<nTrees id="c">
   <list> y0 y1 y2 y3 y4 y5 y6 y7 </list>
   <condition> (eq,k) </condition>
</nTrees>
\end{xcsp}
\end{boxex}

\end{xl}


\subsection{Elementary Constraints}

In this section, we present some elementary constraints that are frequently encountered.
We introduce:

\begin{enumerate}
\begin{xl}\item \gb{clause}
\end{xl}
\item \gb{instantiation}
\end{enumerate}

\begin{xl}
\subsubsection{Constraint \gb{clause}}\label{ctr:clause}

The constraint \gb{clause} represents a disjunction of literals, put in an element \xml{list}, where a literal is either a 0/1 variable (serving as a Boolean variable) or its negation. 
The negation of a variable $x$ is simply written $not(x)$. 

\begin{boxsy}
\begin{syntax} 
<clause>
   <list> ((@{\bnf{01Var}@ | "not("@{\bnf{01Var}@")") wspace)+ </list> 
</clause>
\end{syntax}
\end{boxsy}

Note that the opening and closing tags of \xml{list} are optional, which gives:

\begin{boxsy}
\begin{syntax} 
<clause> ((@{\bnf{01Var}@ | "not("@{\bnf{01Var}@")") wspace)+ </clause> @\com{Simplified Form}@
\end{syntax}
\end{boxsy}

\begin{boxse}
\begin{semantics}
$\gb{clause}(L)$ iff 
  $\exists x \in L : \va{x} = 1 \lor \exists \lnot x \in L : \va{x} = 0$
\end{semantics}
\end{boxse}

In the following example, the constraint states that $x_0 \vee \lnot x_1 \vee x_2 \vee x_3 \vee \lnot x4$

\begin{boxex}
\begin{xcsp}  
<clause>
   x0 @not@(x1) x2 x3 @not@(x4) 
</clause>
 \end{xcsp}
\end{boxex}
\end{xl}

\subsubsection{Constraint \gb{instantiation}}\label{ctr:instantiation}\label{ctr:cube}

The constraint \gb{instantiation} represents a set of unary constraints corresponding to variable assignments.
There are three main interests in introducing it.
First, when modeling, rather often we need to instantiate a subset of variables (for example, for taking some hints/clues into consideration, or for breaking some symmetries).
It is simpler, more natural and informative to post a constraint \gb{instantiation} than a set of unary constraints \gb{intension}.
Second, instantiations can be used to represent partial search instantiations that can be directly injected into \x3 instances.
Third, instantiations allow us to represent solutions, as explained in Section \ref{sec:solutions}, having this way the output of the format made compatible with the input.

The constraint \gb{instantiation} ensures that the ith variable $x$ of \xml{list} is assigned the ith value of \xml{values}. 
Every variable can only occur once in \xml{list}. 

\begin{boxsy}
\begin{syntax} 
<instantiation>
   <list> (intVar wspace)+ </list>
   <values> (intVal wspace)+ </values>
</instantiation>
\end{syntax}
\end{boxsy}

\begin{boxse}
\begin{semantics}
$\gb{instantiation}(X,V)$, with $X=\langle x_0,x_1\ldots \rangle$ and $V=\langle v_0,v_1,\ldots \rangle$ iff  
  $\forall i : 0 \leq i < |X|, x_i = v_i$

$\gbc{Prerequisite}: |X| = |V| > 0$
\end{semantics}
\end{boxse}

In our example below, we have a first constraint that enforces $x=12$, $y=4$ and $z=30$, and a second constraint involving an array $w$ of 6 variables such that $w[0]=1, w[1]=0, w[2]=2, w[3]=1, w[4]=3, w[5]=1$.

\begin{boxex}
\begin{xcsp}  
<instantiation>
   <list> x y z </list>
   <values> 12 4 30 </values>
</instantiation>
<instantiation>
   <list> w[] </list>
   <values> 1 0 2 1 3 1 </values>
</instantiation>
 \end{xcsp}
\end{boxex}

\begin{xl}
\begin{remark}
The constraint \gb{instantiation} is a generalization of the constraint \gb{cube} used in the SAT community when dealing with DNF (Disjunctive Normal Form) expressions.
\end{remark}
\end{xl}

\begin{remark}
The attributes coming with the element \xml{instantiation} when used to represent a solution (see Section \ref{sec:solutions})  are simply ignored (and tolerated) when a solution is injected in the input. 
\end{remark}

Since Specifications 3.0.7, one may use compact forms of integer sequences (in elements \xml{values} of \xml{instantiation} and \xml{coeffs} of \xml{sum}) by writting $v$x$k$ for standing that the integer $v$ occurs $k$ times in sequence.
This means that assuming that $w$ is an array of 6 variables where the three first variables must be assigned to 0 and the three last variables must be assigned to 1,
one can write:
\begin{boxex}
\begin{xcsp}  
<instantiation>
   <list> w[] </list>
   <values> 0 0 0 1 1 1 </values>
</instantiation>
 \end{xcsp}
\end{boxex}

or more compactly:

\begin{boxex}
\begin{xcsp}  
<instantiation>
   <list> w[] </list>
   <values> 0x3 1x3 </values>
</instantiation>
 \end{xcsp}
\end{boxex}



\begin{xl}
\section{Constraints over Symbolic Variables}

Many constraints introduced above for integer variables can be used with symbolic variables.
Basically, this is always when no arithmetic computation is involved.
In the current version of this document, we shall not list exhaustively all such cases.

We just give an illustration of an extensional constraint involving three symbolic variables of domain $\{a,b,c\}$. 


\begin{boxex}
\begin{xcsp}
<extension id="c1">
  <list> x1 x2 x3 </list>
  <supports> (a,b,a)(b,a,c)(b,c,b)(c,a,b) </supports> 
</extension>
 \end{xcsp}
\end{boxex}

\section{Adhoc Constraints} \label{ctr:adhoc}

In some situations, you may want to introduce a particular constraint by arbitrarily defining its semantics (and arguments).
This may be a (new) global constraint that some user wants to try out (and implement in a solver).
In \x3, it is possible to handle such constraints, called \gb{adhoc}.
A constraint \gb{adhoc} is given a name (form), and contain a specific number of arguments.
It may also contain a comment (note).
The syntax is:

\begin{boxsy}
\begin{syntax} 
<adhoc>
   <form> identifier </form>
   [<note> string </note>]
   ... @\com{arguments specific to the form of the adhoc constraint}@ 
</adhoc>
\end{syntax}
\end{boxsy}

The semantics is defined by the user.

For illustrating \gb{adhoc} constraints, we propose to simulate two well-known constraints (of course, there is no real interest in doing so; this is just for simplicity).
Let us consider the following \x3 code:

\begin{boxex}
\begin{xcsp}
<adhoc>
  <form> myAllDiff </form>
  <note> this is my demo simulating @alldifferent@ </note>
   <list> x[] </list>
</adhoc>
<adhoc>
  <form> mySum </form>
  <list> x[] </list>
  <coeffs> 1 2 3 4 5 6 7 8 9 10 </coeffs>
  <condition> (le,165) </condition>
</adhoc>
\end{xcsp}
\end{boxex}

Here, we have two \gb{adhoc} constraints: the first one is supposed to simulate \gb{allDifferent} and the second one \gb{sum}.
Note how the argumenst have been adapted to these two forms of constraints.

In the Java \x3 parser, the callback function is: 

\begin{json}
void buildCtrAdhoc(String id, String form, Map<String, Object> map);
\end{json}

In the constraint solver \ace \cite{ace}, we can then implement something like:

\begin{json}  
void buildCtrAdhoc(String id, String form, Map<String, Object> map) {
  if (form.equals("myAllDiff")) {
    XVarInteger[] list = (XVarInteger[]) map.get("list");
    problem.allDifferent(trVars(list));
  } else if (form.equals("mySum")) {
    XVarInteger[] list = (XVarInteger[]) map.get("list");
    int[] coeffs = (int[]) map.get("coeffs");
    Condition condition = (Condition) map.get("condition");
    problem.sum(trVars(list), coeffs, trVar(condition));
  }
}
\end{json}

It means that you can rather easily call the propagators (of possibly new constraints) you want by intercepting the right forms of adhoc constraints.

\end{xl}


\begin{xl}
\chapter{\textcolor{gray!95}{Constraints over Complex Discrete Variables}}\label{cha:sets}

In this chapter, we introduce constraints over complex discrete variables, namely, set and graph variables.
We shall use the terms \bnf{setVar} and \bnf{graphVar} to denote respectively a set and a graph variable.
These terms as well as related ones, like for example, \bnf{setVal}, are defined in Appendix \ref{cha:bnf}.

\begin{figure}[h]
\begin{tikzpicture}[dirtree, every node/.style={draw=black,thick,anchor=west}]
\tikzstyle{selected}=[fill=colorex]
\tikzstyle{optional}=[fill=gray!10]
  \node {{\bf Constraints over Set Variables}}
    child { node [selected] {Generic Constraints}
      child { node {Constraint \gb{intension}}}
    }		
    child [missing] {}		
    child { node [selected] {Comparison-based Constraints}
      child { node {Constraint \gb{allDifferent} and \gb{allEqual}}} 
      child { node {Constraint \gb{allIntersecting}}}
      child { node {Constraint \gb{ordered}}}
    }
    child [missing] {}	
    child [missing] {}	
    child [missing] {}		
    child { node [selected] {Counting and Summing Constraints}
      child { node {Constraint \gb{sum}}} 
      child { node {Constraint \gb{count}}}
      child { node {Constraints \gb{range} and \gb{roots}}}
    } 
    child [missing] {}	
    child [missing] {}
    child [missing] {}
    child { node [selected] {Connection Constraints}
      child { node {Constraints \gb{element} and \gb{channel}}} 
      child { node {Constraint \gb{partition}}} 
      child { node {Constraint \gb{precedence}}}
    } 
    ; 
\end{tikzpicture}
\caption{The different types of constraints over set variables.\label{fig:setCtrs}} 
\end{figure}
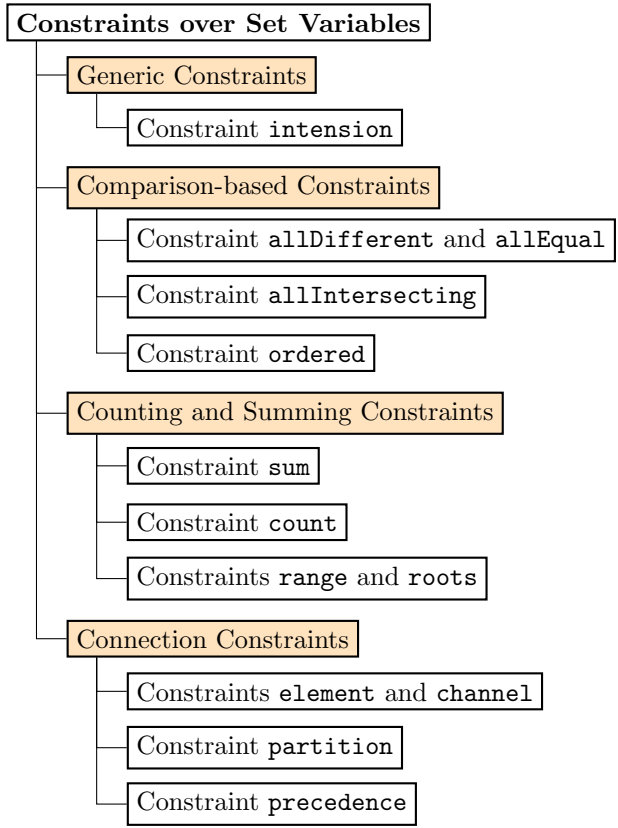

\section{Constraints over Set Variables}

We first introduce constraints over set variables per family, as indicated by Figure \ref{fig:setCtrs}.
In what follows, set variables are denoted by lowercase letters $s$, $t$, possibly subscripted.
For sequences of set variables, we use uppercase letters $S$, $T$, possibly subscripted, as e.g., $S=\langle s_1,s_2,\ldots\rangle$.
For sequences of sequences of set variables, we use uppercase letters $\ns{S}$, $\ns{T}$ in mathcal font, as e.g., $\ns{S}=\langle S_1,S_2,\ldots\rangle$.

\subsection{Generic Constraints}

\subsubsection{Constraint \gb{intension}}\label{ctr:intensionSet}

Unsurprisingly, a constraint \gb{intension} defined on set variables is built similarly to a constraint \gb{intension} defined on integer variables.
Simply, some additional operators are available. They are given by Table \ref{tab:semanticss} that shows the specific operators that can be used to build predicates with set operands.
Of course, it is possible to combine such operators with those presented for integers (and Booleans) in Table \ref{tab:semanticsi}.
Note that the operators on convexity are used in \gecode.

\begin{table}[h]
\begin{center}
{\footnotesize
\begin{tabular}{cccc} 
\rowcolor{v2lgray} \textcolor{dred}{\bf Operation} &  \textcolor{dred}{\bf Arity} &  \textcolor{dred}{\bf Syntax} &  \textcolor{dred}{\bf Semantics} \\
\multicolumn{2}{c}{ } \\
\multicolumn{4}{l}{\textcolor{dred}{Arithmetic (set operands and integer result)}} \\
\midrule
Cardinality     & 1 & card($s$) & $|s|$ \\
Smallest element & 1 & min($s$) & $\min s$ \\
Greatest element & 1 & max($s$) & $\max s$ \\
\multicolumn{2}{c}{ } \\
\multicolumn{4}{l}{\textcolor{dred}{Set (set operands, except for the operator \texttt{set}, and set result)}} \\
\midrule
set & $r > 0$ & set$(x_1,\ldots,x_r)$ & $\{x_1,\ldots,x_r\}$ \\
Union    & $r \geq 2$ &  union($s_1,\ldots,s_r$) & $s_1 \cup \ldots \cup s_r$ \\
Intersection    & $r \geq 2$ &  inter($s_1,\ldots,s_r$) & $s_1 \cap \ldots \cap s_r$ \\
Difference     & 2 & diff($s,t$) & $s \setminus t$ \\
Symmetric difference    & $r \geq 2$ & sdiff($s_1,\ldots,s_r$) & $s_1 \Delta \ldots \Delta s_r$ \\
Convex Hull & 1 & hull($s$) & $\{i : \min s \leq i \leq \max s \}$ \\
\midrule
\multicolumn{2}{c}{ } \\
\multicolumn{4}{l}{\textcolor{dred}{Relational (set operands, except for the 1st operand of the operator \texttt{in}, and Boolean result)}} \\
\midrule
Membership & 2 & in($x,s$) & $x \in s$ \\
Different from    & 2 &  ne($s,t$) & $s \neq t$ \\
Equal to    & $r \geq 2$ & eq($s_1,\ldots,s_r$) & $s_1 = \ldots = s_r$ \\
Disjoint sets    & 2 &  djoint($s,t$) & $s \cap t = \emptyset$ \\
Strict subset     & 2 & subset($s,t$) & $s \subset t$ \\
Subset or equal to   & 2 & subseq($s,t$) & $s \subseteq t$ \\
Superset or equal to   & 2 & supseq($s,t$) & $s \supseteq t$ \\
Strict superset     & 2 & supset($s,t$) & $s \supset t$ \\
Convexity & 1 & convex($s$) & $s = \{i : \min s \leq i \leq \max s \}$ \\
\bottomrule
\end{tabular}
}
\end{center}
\caption{Operators on sets that can be used to build predicates. \label{tab:semanticss}}
\end{table}

An intensional constraint is defined by an element \xml{intension}, containing an element \xml{function} that describes the functional representation of the predicate, referred to as \bnf{boolExprSet} in the syntax box below, and whose precise syntax is given in Appendix \ref{cha:bnf}. 

\begin{boxsy}
\begin{syntax} 
<intension>
  <function> boolExprSet </function>
</intension>
\end{syntax}
\end{boxsy}

Note that the opening and closing tags for \xml{function} are optional, which gives:

\begin{boxsy}
\begin{syntax} 
<intension> boolExprSet </intension> @\com{Simplified Form}@
\end{syntax}
\end{boxsy}


In the following example, the constraints $c_1$ and $c_2$ are respectively defined by $|s| = 2$ and $t_1 \cup t_2 \subset t_3$, with $s$, $t_1$, $t_2$ and $t_3$ being set variables.

\begin{boxex}
\begin{xcsp} 
<intension id="c1"> 
  eq(card(s),2) 
</intension> 
<intension id="c2"> 
  subset(union(t1,t2),t3) 
</intension> 
\end{xcsp}
\end{boxex}

\subsection{Comparison-based Constraints}

\subsubsection{Constraint \gb{allDifferent}}\label{ctr:allDifferentSet}

Such constraints are similar to those described earlier for integer variables, but set variables/values simply replace integer variables/values. 

\begin{boxsy}
\begin{syntax} 
<allDifferent>
  <list> (setVar wspace)2+ </list>
  [<except> setVal+ </except>]
</allDifferent> 
\end{syntax}
\end{boxsy}

Note that the opening and closing tags for \xml{list} are optional, if \xml{except} is not present, which gives:

\begin{boxsy}
\begin{syntax} 
<allDifferent> (setVar wspace)2+ </allDifferent> @\com{Simplified Form}@
\end{syntax}
\end{boxsy}

\begin{boxex}
\begin{xcsp} 
<allDifferent id="c1"> 
  s1 s2 s3 
</allDifferent> 
<allDifferent id="c2"> 
  <list> t1 t2 t3 t4 </list>
  <except> {} </except>
</allDifferent>
\end{xcsp}
\end{boxex}

\subsubsection{Constraint \gb{allEqual}}\label{ctr:allEqualSet}

Such constraints are similar to those described earlier for integer variables, but set variables simply replace integer variables. 

\begin{boxsy}
\begin{syntax} 
<allEqual>
  <list> (setVar wspace)2+ </list>
  [<except> setVal+ </except>]
</allEqual>
\end{syntax}
\end{boxsy}

Note that the opening and closing tags for \xml{list} are optional, if \xml{except} is not present, which gives:

\begin{boxsy}
\begin{syntax} 
<allEqual> (setVar wspace)2+ </allEqual> @\com{Simplified Form}@
\end{syntax}
\end{boxsy}

\begin{boxex}
\begin{xcsp} 
<allEqual id="c"> 
  s1 s2 s3 s4  
</allEqual> 
\end{xcsp}
\end{boxex}

\subsubsection{Constraint \gb{allIntersecting}}\label{ctr:allIntersecting}

The constraint \gb{allIntersecting} ensures that each pair of set variables in \xml{list} intersects in a number of elements that respects a numerical condition.
It captures for example \gb{at\_most1} in \mzinc, with operator le and value 1.

\begin{boxsy}
\begin{syntax} 
<allIntersecting>
   <list> (setVar wspace)2+ </list>
   <condition> "(" operator "," operand ")" </condition> 
</allIntersecting>
\end{syntax}
\end{boxsy}

\begin{boxse}
\begin{semantics}
$\gb{allIntersecting}(S,(\odot,k))$, with $S= \langle s_1,s_2,\ldots\rangle$, iff
   $\forall (i,j) : 1 \leq i < j \leq |S|, |\va{s}_i \cap \va{s}_j| \odot \va{k}$ 
\end{semantics}
\end{boxse}

In the following example, the constraints $c_1$ states that $|s_1 \cap s_2| = k \land |s_1 \cap s_3| = k \land |s_2 \cap s_3| = k$, whereas the constraint $c_2$ states that $|t_1 \cap t_2| \leq 1 \land |t_1 \cap t_3| \leq 1 \land |t_2 \cap t_3| \leq 1$.

\begin{boxex}
\begin{xcsp}  
<allIntersecting id="c1">
   <list> s1 s2 s3 </list>
   <condition> (eq,k) </condition>
</allIntersecting>
<allIntersecting id="c2">
   <list> t1 t2 t3 </list>
   <condition> (le,1) </condition>
</allIntersecting>
 \end{xcsp}
\end{boxex}

We propose to identify two cases of $\gb{allIntersecting}$ over set variables, corresponding to the case where all sets must be disjoint (value \val{disjoint} for \att{case}) and the case where all sets must be overlapping (value \val{overlapping} for \att{case}).
For both cases, the element \xml{condition} becomes implicit.
The two following constraints $c_3$ and $c_4$:

\begin{boxex}
\begin{xcsp}  
<allIntersecting id="c3">
   <list> s1 s2 s3 </list>
   <condition> (gt,0) </condition>
</allIntersecting>
<allIntersecting id="c4">
   <list> t1 t2 t3 t4 </list>
   <condition> (eq,0) </condition>
</allIntersecting>
\end{xcsp}
\end{boxex}

can then be defined by:
\begin{boxex}
\begin{xcsp}  
<allIntersecting id="c3" case="disjoint">
   s1 s2 s3 
</allIntersecting>
<allIntersecting id="c4" case="overlapping">
   t1 t2 t3 t4 
</allIntersecting>
\end{xcsp}
\end{boxex}

\subsubsection{Constraint \gb{ordered}}\label{ctr:orderedSet}

As for the constraint \gb{ordered\ti set} (i.e., the constraint \gb{ordered} lifted to sets), we consider the inclusion order for sets: the operators that can be used are among $\{\subset,\subseteq,\supseteq,\supset\}$.
This gives:

\begin{boxsy}
\begin{syntax} 
<ordered>
  <list> (setVar wspace)2+ </list>
  <@operator@>  "subset" | "subseq" | "supseq" | "supset" </@operator@>
</ordered> 
\end{syntax}
\end{boxsy}

\begin{boxse}
\begin{semantics}
$\gb{ordered}(S,\odot)$, with $S=\langle s_1,s_2,\ldots\rangle$ and $\odot \in \{\subset,\subseteq,\supseteq,\supset\}$, iff 
  $\forall i : 1 \leq i < |S|, \va{s}_i \odot \va{s}_{i+1}$
\end{semantics}
\end{boxse}

This captures \gb{increasing} (\gb{subsetEq} in Choco3) and \gb{decreasing}, as illustrated below respectively with constraints $c_1$ and $c_2$.

\begin{boxex}
\begin{xcsp}  
<ordered id="c1"> 
  <list> s1 s2 s3 </list>
  <operator> subseq </operator>
</ordered>
<ordered id="c2"> 
  <list> t1 t2 t3 t4 t5 </list>
  <operator> supseq </operator>
</ordered>
 \end{xcsp}
\end{boxex}
 
A second form of \gb{ordered} uses an underlying ordering based either on bounds or on indicator (characteristic) functions.
In the former case, the ordering $\preceq_{bnd}$ is defined as: 
\begin{quote}
$s \preceq_{bnd} t$ iff\\
$~ ~ \max(s) \leq \min(t)$.
\end{quote}
In the latter case, the ordering $\preceq_{ind}$ is defined as follows:  
\begin{quote}
$s \preceq_{ind} t$ iff\\
$~ ~ s = \emptyset \lor t \neq \emptyset \land (\min(s) < \min(t) \lor \min(s) = \min(t) \land s \setminus \{\min(s)\} \preceq_{ind} t \setminus \{\min(t)\})$
\end{quote}

The attribute \att{order} is required here, with either the value \val{bnd} or the value \val{ind}.

\begin{boxsy}
\begin{syntax} 
<ordered order="bnd|ind">
  <list> (setVar wspace)2+ </list>
  <@operator@> "lt" | "le" | "ge" | "gt" </@operator@>
</ordered> 
\end{syntax}
\end{boxsy}

For the semantics, $\gb{ordered}^{bnd}$ and $\gb{ordered}^{ind}$ correspond to versions of \gb{ordered}, with \att{order} set to \val{bnd} and \val{ind}, respectively.

\begin{boxse}
\begin{semantics}
$\gb{ordered}^{bnd}(S,\odot)$, with $S=\langle s_1,s_2,\ldots\rangle$ and $\odot \in \{<,\leq,\geq,>\}$, iff 
  $\forall i : 1 \leq i < |S|, \max(\va{s}_i) \odot \min(\va{s}_{i+1})$
$\gb{ordered}^{ind}(S,\odot)$, with $S=\langle s_1,s_2,\ldots\rangle$ and $\odot \in \{<,\leq,\geq,>\}$, iff 
  $\forall i : 1 \leq i < |S|, \va{s}_i \odot_{ind} \va{s}_{i+1}$
\end{semantics}
\end{boxse}

The constraint with type \val{bnd} captures \gb{sequence} of Gecode.

The constraint \gb{ordered} over set variables can be of course lifted to other structures like lists and sets.
We show this on lists (sequences) of set variables, capturing \gb{lex} defined in \mzinc.
The lexicographic order can rely on $\preceq_{bnd}$ or $\preceq_{ind}$.
For example, for $\preceq_{ind}$, we have: 
\begin{quote}
$S=\langle s_1,s_2,\ldots,s_k\rangle \preceq_{ind} T=\langle t_1,t_2,\ldots t_k\rangle$ iff \\
$~ ~ S=\langle \rangle \lor T \neq \langle \rangle \land (s_1 \prec_{ind} t_1 \lor s_1 = t_1 \land \langle s_2,\ldots,s_k\rangle \preceq_{ind} \langle t_2,\ldots t_k\rangle)$.
\end{quote}

\begin{boxsy}
\begin{syntax} 
<ordered order="bnd|ind">
  (<list> (setVar wspace)+ </list>)2+
  <@operator@> "lt" | "le" | "ge" | "gt" </@operator@>
</ordered>
\end{syntax}
\end{boxsy}


\begin{boxse}
\begin{semantics}
$\gb{ordered}^{ind}(\ns{S},\odot)$, with $\ns{S}=\langle S_1,S_2,\ldots \rangle$ and $\odot \in \{<, \leq, \geq,>\}$, iff 
  $\forall i : 1 \leq i < |\ns{S}|$, $\va{S}_i \odot_{ind} \va{S}_{i+1}$ 

@{\em Prerequisite}@: $|\ns{S}| > 1$ 
\end{semantics}
\end{boxse}


\subsection{Counting Constraints}

\subsubsection{Constraint \gb{sum}}\label{ctr:sumSet}

The constraint \gb{sum} ensures that the sum of coefficients in \xml{coeffs} indexes by values present in \xml{index} respects a numerical condition.

\begin{boxsy}
\begin{syntax} 
<sum>
   <index> setVar </index>
   <coeffs [startIndex="integer"]> (intVal wspace)+ </coeffs>
   <condition> "(" operator "," operand ")" </condition>
</sum>
\end{syntax}
\end{boxsy}

\begin{boxse}
\begin{semantics}
$\gb{sum}(s,C,(\odot,k))$, with $C=\langle c_1,c_2,\ldots\rangle$, iff
  $\sum \{c_i : 1 \leq i \leq |C| \land i \in \va{s}\} \odot \va{k}$ 
\end{semantics}
\end{boxse}

This constraint captures \gb{sum\_set} in the catalog, \gb{sum} in Choco3 and \gb{weights} in Gecode.

\begin{boxex}
\begin{xcsp}  
<sum>
   <index> s </index>
   <coeffs> 4 2 3 2 7 </coeffs>
   <condition> (gt,v) </condition>
</sum>
 \end{xcsp}
\end{boxex}

\subsubsection{Constraint \gb{count}}\label{ctr:countSet}

The constraint \gb{count} ensures that the number of set variables in \xml{list} which are assigned a set among those in \xml{values} respects a numerical condition.

\begin{boxsy}
\begin{syntax} 
<count>
   <list> (setVar wspace)2+ </list>
   <values> (setVal wspace)+ | (setVar wspace)+ </values> 
   <condition> "(" operator "," operand ")" </condition>
</count>
\end{syntax}
\end{boxsy}

To simplify, we assume for the semantics that $T$ is a set of set values (and not variables).

\begin{boxse}
\begin{semantics}
$\gb{count}(S,T,(\odot,k))$, with $S=\langle s_1,s_2,\ldots \rangle$, $T=\langle t_1,t_2,\ldots \rangle$, iff 
  $|\{i : 1 \leq i \leq |S| \land \va{s}_i \in T\}| \odot \va{k}$ 
\end{semantics}
\end{boxse}

\begin{boxex}
\begin{xcsp}  
<count>
   <list> s1 s2 s3 s4 </list>
   <values> t </values>
   <condition> (eq,2) </condition>
</count>
\end{xcsp}
\end{boxex}

The constraint \gb{count} captures \gb{atLeast}, \gb{atMost} and \gb{exactly}.
It also captures \gb{nEmpty} of Choco3, by using \verb?<values> set() </values>?.

\subsubsection{Constraint \gb{range}}\label{ctr:range}

The constraint \gb{range} \cite{BHHKW_range,BHHKW_rangeCPAIOR} holds iff the set of values taken by variables of \xml{list} at indices given by \xml{index} is exactly the set \xml{image}.  
The optional attribute \att{startIndex} gives the number used for indexing the first variable in \xml{list} (0, by default). 

\begin{boxsy}
\begin{syntax}   
<range>
   <list [startIndex="integer"]> (intVar wspace)2+ </list>
   <index> setVal | setVar </index>
   <image> setVal | setVar </image>
</range>
 \end{syntax}
\end{boxsy} 

\begin{boxse}
\begin{semantics}
$\gb{range}(X,s,t)$, with $X= \langle x_1,x_2,\ldots\rangle$, iff
  $\{\va{x}_i : 1 \leq i \leq |X| \land i \in \va{s}\} = \va{t}$ 
\end{semantics}
\end{boxse}

In the following example, the constraints $c_1$ and $c_2$, taken together, permit to represent \gb{nValues}($\{x_0,x_1,x_2,x_3,x_4,x_5\},y$); see \cite{BHHKW_range}.

\begin{boxex}
\begin{xcsp}  
<range id="c1">
   <list> x0 x1 x2 x3 x4 x5 </list>
   <index> {0,1,2,3,4,5} </index>
   <image> t </image>
</range>
<intension id="c2"> eq(y,card(t)) </intension>
 \end{xcsp}
\end{boxex}

\subsubsection{Constraint \gb{roots}}\label{ctr:roots}

The constraint \gb{roots} \cite{BHHKW_range,BHHKW_roots} holds iff \xml{index} represents exactly the set of indices of variables in \xml{list} which are assigned a value in \xml{image}.
The optional attribute \att{startIndex} gives the number used for indexing the first variable in \xml{list} (0, by default). 

\begin{boxsy}
\begin{syntax}   
<roots>
   <list [startIndex="integer"]> (intVar wspace)2+ </list>
   <index> setVal | setVar </index>
   <image> setVal | setVar </image>
</roots>
 \end{syntax}
\end{boxsy} 

\begin{boxse}
\begin{semantics}
$\gb{roots}(X,s,t)$, with $X= \langle x_1,x_2,\ldots \rangle$, iff
  $\{i : 1 \leq i \leq |X| \land \va{x}_i \in \va{t}\} = \va{s}$ 
\end{semantics}
\end{boxse}

In the following example, the constraints $c_1$ and $c_2$, taken together, permit to represent \gb{among}(2,$\{x_0,x_1,x_2,x_3,x_4,x_5\},\{0,1\}$); see \cite{BHHKW_range}.

\begin{boxex}
\begin{xcsp}  
<roots id="c1">
   <list> x0 x1 x2 x3 x4 x5 </list>
   <index> s </index>
   <image> {0,1} </image>
</roots>
<intension id="c2"> eq(2,card(s)) </intension>
 \end{xcsp}
\end{boxex}

\subsection{Connection Constraints}

\subsubsection{Constraint \gb{element}}\label{ctr:elementSet}

The constraint \gb{element} ensures that \xml{value} is element of \xml{list}, i.e., equal to one set among those assigned to the set variables of \xml{list}.
The optional element \xml{index} gives the index (position) of one occurrence of \xml{value} inside \xml{list}.
The optional attribute \att{startIndex} gives the number used for indexing the first variable in \xml{list} (0, by default). 

\begin{boxsy}
\begin{syntax} 
<element>
  <list [startIndex="integer"]> (setVar wspace)2+ </list>
  [<index> intVar </index>]
  <value> setVal | setVar </value>
</element>
\end{syntax}
\end{boxsy}

\begin{boxse}
\begin{semantics}
$\gb{element}(S,t)$, with $S=\langle s_1,s_2,\ldots\rangle$, iff 
  $\exists i : 1 \leq i \leq |S| \land \va{s}_{i} = \va{t}$  
$\gb{element}(S,i,t)$, with $S=\langle s_1,s_2,\ldots\rangle$, iff 
  $\va{s}_{\va{i}} = \va{t}$  
\end{semantics}
\end{boxse}

The first form (without element \xml{index}) captures \gb{member} in \choco.

\begin{boxex}
\begin{xcsp}  
<element id="c1">
  <list> s1 s2 s3 </list>
  <index> i </index>
  <value> t </value>
</element>
<element id="c2">
  <list> t1 t2 t3 t4 </list>
  <value> t5 </value>
</element>
\end{xcsp}
\end{boxex}
 
\subsubsection{Constraint \gb{channel}}\label{ctr:channelSet}

The constraint \gb{channel}, in its first form, sometimes called \gb{inverse\_set} or \gb{inverse} in the literature, ensures that two lists of set variables represent inverse functions.
Any value in the set of a list must be within the index range of the other list.
For each list, the optional attribute \att{startIndex} gives the number used for indexing the first variable in this list (0, by default). 

\begin{boxsy}
\begin{syntax} 
<channel>
  <list [startIndex="integer"]> (setVar wspace)2+ </list>
  <list [startIndex="integer"]> (setVar wspace)2+ </list>
</channel>
\end{syntax}
\end{boxsy}

For the semantics, starting indexes are assumed to be equal to 1.

\begin{boxse}
\begin{semantics}
$\gb{channel}(S,T)$, with $S=\langle s_1,s_2,\ldots \rangle$ and $T=\langle t_1,t_2,\ldots \rangle$, iff 
  $\forall i : 1 \leq i \leq |S|, j \in \va{s}_i \Leftrightarrow i \in \va{t}_j$
\end{semantics}
\end{boxse}

A second form of the constraint \gb{channel}, called \gb{int\_set\_channel} in \mzinc, links a set of integer variables to a set of set variables.

\begin{boxsy}
\begin{syntax} 
<channel>
  <list [startIndex="integer"]> (intVar wspace)+ </list>
  <list [startIndex="integer"]> (setVar wspace)+ </list>
</channel>
\end{syntax}
\end{boxsy}

\begin{boxse}
\begin{semantics}
$\gb{channel}(X,S)$, with $X=\langle x_1,x_2,\ldots\rangle$ and $S=\langle s_1,s_2,\ldots\rangle$, iff 
  $\forall i : 1 \leq i \leq |X|, \va{x}_i = j \Leftrightarrow i \in \va{s}_j$
\end{semantics}
\end{boxse}

A final form, called \gb{bool\_channel} in \choco, is obtained by considering a list of 0/1 variables to be channeled with a set variable.
It constrains this list to be a representation of the set. 

\begin{boxsy}
\begin{syntax} 
<channel>
  <list [startIndex="integer"]> (@{\textsl{\textcolor{dblue}{01Var}}@ wspace)2+ </list>
  <value> setVar </value>
</channel>
\end{syntax}
\end{boxsy}

\begin{boxse}
\begin{semantics}
$\gb{channel}(X,s)$, with $X=\{x_1,x_2,\ldots\}$, iff 
  $\forall i : 1 \leq i \leq |X|, \va{x}_i = 1 \Leftrightarrow i \in \va{s}$
\end{semantics}
\end{boxse}

\subsubsection{Constraint \gb{precedence}}\label{ctr:precedenceSet}

The constraint \gb{precedence} ensures that if a set variable $s_j$ of \xml{list} contains the $i+1th$ value of \xml{values}, but not the $ith$ value, then another set variable of \xml{list}, that precedes $s_j$, contains the $ith$ value of \xml{values}, but not the $i+1th$ value.
The optional attribute \att{covered} indicates whether each value of \xml{values} must belong to at least one set variable of \xml{list} (\val{false}, by default).

\begin{boxsy}
\begin{syntax} 
<precedence>
   <list> (setVar wspace)2+ </list>
   <values [covered="boolean"]> (intVal wspace)2+ </values>
</precedence>
\end{syntax}
\end{boxsy}

For the semantics, $V^{\nm{cv}}$ means that \att{covered} is \val{true}, and we note $v \in \va{S}$ iff there exists a set variable $s_j \in S$ such that $v \in \va{s}_j$.

\begin{boxse}
\begin{semantics}
$\gb{precedence}(S,V)$, with $S=\langle s_1,s_2,\ldots\rangle$ and $V=\langle v_1,v_2,\ldots\rangle$,  iff  
  $\forall i : 1 \leq i < |V|, v_{i+1} \in \va{S} \Rightarrow  v_{i} \in \va{S}$
  $\forall i : 1 \leq i < |V| \land v_{i+1} \in \va{S}, \min\{j : 1 \leq j \leq |S| \land v_i \in \va{s}_j\} < \min\{j :  1 \leq j \leq |S| \land v_{i+1} \in \va{s}_j\}$
$\gb{precedence}(S,V^{\nm{cv}})$ iff  $\gb{precedence}(S,V) \wedge v_{|V|} \in \va{S}$
\end{semantics}
\end{boxse}

\begin{boxex}
\begin{xcsp}  
<precedence>
   <list> s1 s2 s3 s4 </list>
   <values> 4 0 </values>
</precedence>
 \end{xcsp}
\end{boxex}
 
The constraint \gb{precedence} captures \gb{set\_value\_precede} in the \cat. 

\subsubsection{Constraint \gb{partition}}\label{ctr:partition}

The constraint \gb{partition} ensures that \xml{list} represents a partition of \xml{value}.

\begin{boxsy}
\begin{syntax} 
<partition>
  <list> (setVar wspace)2+ </list> 
  <value> setVar </value>
</partition>
\end{syntax}
\end{boxsy}

\begin{boxse}
\begin{semantics}
$\gb{partition}(S,t)$, with $S=\langle s_1,s_2,\ldots\rangle$, iff 
  $\va{t} = \cup_{i=1}^{|S|} \va{s}_i \land \forall (i,j) : 1 \leq i < j \leq |S|, \va{s}_i \cap \va{s}_j = \emptyset$
\end{semantics}
\end{boxse}

\section{Constraints over Graph Variables}

In this section, we present constraints that are defined on graph variables. We have:
\begin{enumerate}
\item \gb{circuit}
\item \gb{nCircuits}
\item \gb{path}
\item \gb{nPaths}
\item \gb{arbo}
\item \gb{nArbos}
\item \gb{nCliques}
\end{enumerate}

\subsection{Constraint \gb{circuit}}\label{ctr:circuitGraph}

This form of the constraint \gb{circuit} ensures that the value taken by the (directed or undirected) graph variable in \xml{graph} represents a circuit (cycle).
The optional element \xml{size} indicates that the circuit must be of a given size (number of nodes).
Consequently, in that case, and only in that case, the circuit is not required to cover all nodes that are present in the upper bound of the graph variable.

\begin{boxsy}
\begin{syntax} 
<circuit>
   <graph> graphVar </graph>
   [<size> intVal | intVar </size>]
</circuit>
 \end{syntax}
\end{boxsy}

For the semantics, $n_{max}$ denotes the number of nodes in the upper bound $g_{max}$ of $g$.

\begin{boxse}
\begin{semantics}
$\gb{circuit}(g)$ iff  
  $\va{g}$ is a circuit of size $n_{max}$
$\gb{circuit}(g,s)$ iff  
  $\va{g}$ is a circuit of size $\va{s} > 1$
\end{semantics}
\end{boxse}

\subsection{Constraint \gb{nCircuits}}\label{ctr:nCircuitsGraph}

For several circuits, one must specify a numerical condition.

\begin{boxsy}
\begin{syntax} 
<nCircuits>
   <graph> graphVar </graph>
   <condition> "(" operator "," operand ")" </condition>
</nCircuits>
 \end{syntax}
\end{boxsy}

\begin{boxse}
\begin{semantics}
$\gb{nCircuits}(g,(\odot,k))$, iff  
  $\va{g}$ represents $p$ node-disjoint circuits with $p \odot \va{k}$
\end{semantics}
\end{boxse}

\subsection{Constraint \gb{path}}\label{ctr:pathGraph}

The constraint \gb{path} ensures that the value taken by the (directed or undirected) graph variable in \xml{graph} represents a path from the node specified in \xml{start} to the node specified in \xml{final}.
The optional element \att{size} indicates that the path must be of a given size (number of nodes).
Consequently, in that case,  and only in that case, the path is not required to cover all nodes that are present in the upper bound of the graph variable.

\begin{boxsy}
\begin{syntax} 
<path>
   <graph> graphVar </graph>
   <start> intVal | intVar </start>
   <final> intVal | intVar </final>
   [<size> intVal | intVar </size>]
</path>
 \end{syntax}
\end{boxsy}

\begin{boxse}
\begin{semantics}
$\gb{path}(g,s,e)$ iff  
  $\va{g}$ is a path from $\va{s}$ to $\va{e}$ of size $n_{max}$
$\gb{path}(g,s,e,z)$ iff  
  $\va{g}$ is a path from $\va{s}$ to $\va{e}$ of size $\va{z} > 1$
\end{semantics}
\end{boxse}

\subsection{Constraint \gb{nPaths}}\label{ctr:nPathsGraph}

For several paths, one must specify a numerical condition.

\begin{boxsy}
\begin{syntax} 
<nPaths>
   <graph> graphVar </graph>
   <condition> "(" operator "," operand ")" </condition>
</nPaths>
 \end{syntax}
\end{boxsy}

\begin{boxse}
\begin{semantics}
$\gb{nPaths}(g,(\odot,k))$, iff  
  $\va{g}$ forms $p$ paths s.t. $p \odot \va{k}$ 
\end{semantics}
\end{boxse}

\subsection{Constraint \gb{arbo}}\label{ctr:arbo}

The constraint \gb{arbo} ensures that the value taken by the directed (respectively, undirected) graph variable in \xml{graph} represents an arborescence (respectively, tree) whose root is specified by \xml{root}.
The optional element \att{size} indicates that the arborescence must be of a given size (number of nodes).
Consequently, in that case, the arborescence is not required to cover all nodes that are present in the upper bound of the graph variable.

\begin{boxsy}
\begin{syntax} 
<arbo>
   <graph> graphVar </graph>
   <root> intVal | intVar </root>
   [<size> intVal | intVar </size>]
</arbo>
 \end{syntax}
\end{boxsy}

\begin{boxse}
\begin{semantics}
$\gb{arbo}(g,r)$ iff  
  $\va{g}$ is an arborescence of root $\va{r}$ and size $n_{max}$
$\gb{arbo}(g,r,s)$ iff  
  $\va{g}$ is an arborescence of root $\va{r}$ and size $\va{s} > 1$
\end{semantics}
\end{boxse}

\subsection{Constraint \gb{nArbos}}\label{ctr:nArbos}

\begin{boxsy}
\begin{syntax} 
<nArbos>
   <graph> graphVar </graph>
   <condition> "(" operator "," operand ")" </condition>
</nArbos>
 \end{syntax}
\end{boxsy}

\begin{boxse}
\begin{semantics}
$\gb{nArbos}(g,(\odot,k))$, with $\odot \in \{<,\leq,>,\geq,=,\neq\}$, iff  
  $\va{g}$ represents $p$ arborescences s.t. $p \odot \va{k}$ 
\end{semantics}
\end{boxse}


\subsection{Constraint \gb{nCliques}}\label{ctr:nCliques}

The constraint \gb{nCliques}, called \gb{nclique} in \cite{F_graph},  ensures that the value taken by the graph variable in \xml{graph} represents a set of cliques whose cardinality must respect a numerical condition.

\begin{boxsy}
\begin{syntax} 
<nCliques>
   <graph> graphVar </graph>
   <condition> "(" operator "," operand ")" </condition>
</nCliques>
 \end{syntax}
\end{boxsy}

\begin{boxse}
\begin{semantics}
$\gb{nCliques}(g,(\odot,k))$, with $\odot \in \{<,\leq,>,\geq,=,\neq\}$, iff  
  $\va{g}$  represents $p$ cliques s.t. $p \odot \va{k}$
\end{semantics}
\end{boxse}
\end{xl}

\begin{xl}
\chapter{\textcolor{gray!95}{Constraints over Continuous Variables}}\label{cha:real}

In this chapter, we introduce constraints over continuous variables, namely, real and qualitative variables.
We shall use the term \bnf{realVar} and \bnf{qualVar} to denote a real variable and a qualitative variable, respectively.
These terms as well as related ones, like for example, \bnf{realVal}, are defined in Appendix \ref{cha:bnf}.

\section{Constraints over Real Variables}

Reasoning over real variables is typically performed by means of intensional arithmetic constraints and sum (linear) constraints.
However, in the future, we might introduce a few global constraints adapted to real variables.

\subsection{Constraint \gb{intension}}\label{ctr:intensionReal}

A constraint \gb{intension} on real variables is built similarly to a constraint \gb{intension} defined on integer variables.
Simply, some additional operators are available. They are given by Table \ref{tab:semanticsf} that does show the specific operators that can be used to build predicates with real operands.
Of course, it is possible to combine such operators with those presented for integers (and Booleans) in Table \ref{tab:semanticsi}.

An intensional constraint is defined by an element \xml{intension}, containing an element \xml{function} that describes the functional representation of the predicate, referred to as \bnf{boolExprReal} in the syntax box above, and whose precise syntax is given in Appendix \ref{cha:bnf}. 

\begin{boxsy}
\begin{syntax} 
<intension>
  <function> boolExprReal </function>
</intension>
\end{syntax}
\end{boxsy}

Note that the opening and closing tags for \xml{function} are optional, which gives:

\begin{boxsy}
\begin{syntax} 
<intension> boolExprReal </intension> @\com{Simplified Form}@
\end{syntax}
\end{boxsy}

\begin{table}[ht]
\begin{center}
{\footnotesize
\begin{tabular}{cccc} 
\rowcolor{v2lgray}{} {\textcolor{dred}{\bf Operation}} &  {\textcolor{dred}{\bf Arity}} &  {\textcolor{dred}{\bf Syntax}} &  {\textcolor{dred}{\bf Semantics}} \\
\multicolumn{2}{c}{ } \\
\multicolumn{4}{l}{\textcolor{dred}{Constants}} \\
\midrule
Pi   & 0 & PI & $\pi$ \\
Euler's constant & 0 & E & $e$ \\
\midrule
\multicolumn{2}{c}{ } \\
\multicolumn{4}{l}{\textcolor{dred}{Arithmetic}} \\
\midrule
Real Division     & 2 & fdiv($x,y$) & $x/y$ \\
Real Remainder   & 2 & fmod($x,y$) & $x\%y$ \\
\midrule
\multicolumn{2}{c}{ } \\
\multicolumn{4}{l}{\textcolor{dred}{Functional}} \\
\midrule
Square root    & 1 & sqrt($x$) & $\sqrt{x}$ \\
$nth$ root    & 2 & nroot($x,n$) & $\sqrt[n]{x}$ \\
Natural exponential & 1 & exp($x$) & $e^x$ \\
Natural Logarithm & 1 & ln($x$) & $\log_e x$ \\
Logarithm to base & 2 & log($x,b)$ & $\log_b x$ \\ 
Sine  & 1 & sin($x$) & $\sin x$ \\
Cosine  & 1 & cos($x$) & $\cos x$ \\
Tangent  & 1 & tan($x$) & $\tan x$ \\
Arcsine  & 1 & asin($x$) & $\arcsin x$ \\
Arccosine  & 1 & acos($x$) & $\arccos x$ \\
Arctangent  & 1 & atan($x$) & $\arctan x$ \\
Hyperbolic sine  & 1 & sinh($x$) & $\sinh x$ \\
Hyperbolic cosine  & 1 & cosh($x$) & $\cosh x$ \\
Hyperbolic tangent  & 1 & tanh($x$) & $\tanh x$ \\
\bottomrule
\end{tabular}
}
\end{center}
\caption{Operators on reals that can be used to build predicates. \label{tab:semanticsf}}
\end{table}

In the following example, the constraints $c_1$ and $c_2$ are respectively defined by $v = \sqrt{w}$ and $x \geq 3 \ln y + z^4$, with $v,w,x,y,z$ assumed to be real variables.

\begin{boxex}
\begin{xcsp} 
<intension id="c1"> 
  eq(v,sqrt(w))
</intension> 
<intension id="c2"> 
  ge(x,add(mul(3,ln(y)),pow(z,4)))
</intension> 
\end{xcsp}
\end{boxex}


\subsection{Constraint \gb{sum}}\label{ctr:sumReal}

The constraint \gb{sum} is the most important constraint in the context of real computation.
When the optional element \xml{coeffs} is missing, it is assumed that all coefficients are equal to 1.
Of course, real values, variables and intervals are expected in \xml{condition}.

\begin{boxsy}
\begin{syntax} 
<sum>
   <list> (realVar wspace)2+ </list>
   [<coeffs> (realVal wspace)+ </coeffs>]
   <condition> "(" operator "," (realVal | realVar | realIntvl) ")" </condition>
</sum>
\end{syntax}
\end{boxsy}

\begin{boxse}
\begin{semantics}
$\gb{sum}(X,C,(\odot,k))$, with $X=\langle x_1,x_2,\ldots\rangle$ and $C=\langle c_1,c_2,\ldots\rangle$,  iff
  $(\sum_{i=1}^{|X|} c_i \times \va{x}_i) \odot \va{k}$ 

$\gbc{Prerequisite}: |X| = |C|$
\end{semantics}
\end{boxse}

In the following example, the constraint expresses $1.5x_0 + 2x_1 + 3.2x_2 > y$. 

\begin{boxex}
\begin{xcsp}  
<sum>
   <list> x0 x1 x2 </list>
   <coeffs> 1.5 2 3.2 </coeffs>
   <condition> (gt,y) </condition>
</sum>
 \end{xcsp}
\end{boxex}

\section{Constraints over Qualitative Variables}

Qualitative Spatial Temporal Reasoning (QSTR) deals with qualitative calculi (also called algebras).
A qualitative calculus is defined from a finite set $\B$ of base relations on a certain domain.
In this section, we introduce some constraints over some important qualitative calculi.
In the future, more algebras might be introduced.

\subsection{Constraint \gb{interval}}\label{ctr:interval}

Interval Algebra, also called Allen's calculus \cite{A_interval} handles temporal entities that represent intervals on the rational line.
The set of base relations of this calculus is:
\begin{quote}
 $B_{int} = \{eq,p,pi,m,mi,o,oi,s,si,d,di,f,fi\}$ 
\end{quote}
where for example $m$ stands for \verb!meets!.

The constraint \gb{interval} allows us to express qualitative information between two intervals.

\begin{boxsy}
\begin{syntax} 
<interval>
   <scope> qualVar wspace qualVar </scope>
   <relation> (iaBaseRelation wspace)*  </relation>
</interval>
\end{syntax}
\end{boxsy}

\begin{boxse}
\begin{semantics}
$\gb{interval}(x,y,R)$, with $R \subseteq B_{int}$,  iff
  $\exists b \in R : (\va{x},\va{y}) \in b$ 
\end{semantics}
\end{boxse}

In the following example, the constraint expresses that either $x$ must overlap $y$ or $x$ must meet $y$.

\begin{boxex}
\begin{xcsp}  
<interval>
  <scope> x y </scope>
  <relation> o m </relation>
</interval>
\end{xcsp}
\end{boxex}

\subsection{Constraint \gb{point}}\label{ctr:point}

Point Algebra (PA) is a simple qualitative algebra defined for time points.
The set of base relations of this calculus is:
\begin{quote}
 $B_{point} = \{b,eq,a\}$ 
\end{quote}
where for example $b$ stands for \verb!before!.

The constraint \gb{point} allows us to express qualitative information between two points.

\begin{boxsy}
\begin{syntax} 
<point>
  <scope> qualVar wspace qualVar </scope>
  <relation> (paBaseRelation wspace)*  </relation>
</point>
\end{syntax}
\end{boxsy}

\begin{boxse}
\begin{semantics}
$\gb{point}(x,y,R)$, with $R \subseteq B_{point}$,  iff
  $\exists b \in R : (\va{x},\va{y}) \in b$ 
\end{semantics}
\end{boxse}

In the following example, the constraint expresses that either $x$ must be before or after $y$.

\begin{boxex}
\begin{xcsp}  
<point>
  <scope> x y </scope>
  <relation> b a </relation>
</point
\end{xcsp}
\end{boxex}

\subsection{Constraint \gb{rcc8}}\label{ctr:rcc8}

RCC8 \cite{RCC_spatial} is a region connection calculus for reasoning about regions in Euclidean space.
The set of base relations of this calculus is:
\begin{quote}
 $B_{rcc8} = \{dc,ec,eq,po,tpp,tppi,nttp,ntppi\}$ 
\end{quote}
where for example $dc$ stands for \verb!disconnected!.

The constraint \gb{region} allows us to express qualitative information between two regions.
For RCC8, the attribute \att{type} is required, and its value must be set to \val{rcc8}.

\begin{boxsy}
\begin{syntax} 
<region type="rcc8">
  <scope> qualVar wspace qualVar </scope>
  <relation> (rcc8BaseRelation wspace)*  </relation>
</region>
\end{syntax}
\end{boxsy}

\begin{boxse}
\begin{semantics}
$\gb{region}^{rcc8}(x,y,R)$, with $R \subseteq B_{rcc8}$,  iff
  $\exists b \in R : (\va{x},\va{y}) \in b$ 
\end{semantics}
\end{boxse}

In the following example, the constraint expresses that either $x$ must be disconnected or externally connected with $y$.

\begin{boxex}
\begin{xcsp}  
<region type="rcc8">
  <scope> x y </scope>
  <relation> dc ec </relation>
</region>
\end{xcsp}
\end{boxex}

\subsection{Constraint \gb{dbd}}\label{ctr:dbd}

For Temporal Constraint Satisfaction \cite{DMP_temporal}, variables represent time points and temporal information is represented by a set of unary and binary constraints, each specifying a set of permitted intervals.

For this framework, we only need to introduce a type of constraints, called \gb{dbd} constraints, for disjunctive binary difference constraints (as in \cite{K_temporal}).

\begin{boxsy}
\begin{syntax} 
<dbd>
  <scope> qualVar (wspace qualVar)* </scope>
  <intervals> (realIntvl wspace)+  </intervals>
</dbd>
\end{syntax}
\end{boxsy}

Below, we give the semantics for unary and binary \gb{dbd} constraints.

\begin{boxse}
\begin{semantics}
$\gb{dbd}(x,I)$, with $I=\langle l_1..u_1,l_2..u_2,\ldots \rangle$,  iff
  $\exists i : 1 \leq i \leq |I| \land \va{x} \in l_i..u_i$
$\gb{dbd}(x,y,I)$, with $I=\langle l_1..u_1,l_2..u_2,\ldots \rangle$,  iff
  $\exists i : 1 \leq i \leq |I| \land \va{y} - \va{x} \in l_i..u_i$
\end{semantics}
\end{boxse}

In the following example, the constraint expresses that either $x_2-x_1$ must be in $[30,40]$ or in $[60,+infinity[$.

\begin{boxex}
\begin{xcsp}  
<dbd>
  <scope> x1 x2 </scope>
  <intervals> [30,40] [60,+infinity[ </intervals>
</dbd>
\end{xcsp}
\end{boxex}
\end{xl}

\part{Advanced Forms of Constraints}

\begin{xl}
In \x3, it is possible to build advanced forms of (basic) constraints by means of lifting, restriction, sliding, logical combination
and relaxation mechanisms. In this part, we introduce them.
\end{xl}
\begin{xc}
In \x3-core, it is possible to build a few advanced forms of (basic) constraints by means of lifting ans sliding mechanisms. In this part, we introduce them.
\end{xc}

\begin{xl}
\chapter{\textcolor{gray!95}{Lifted and Restricted Forms of Constraints}}\label{cha:lifted}

In this chapter, we introduce two mechanisms that allow us to extend the scope of constraints defined in \x3.
First, we show how it is possible to lift constraints over lists, sets, multisets, and also matrices, in a quite generic and natural way.
Second, we show how classical restrictions on constraints can be posed by operating a simple attribute.
These two mechanisms offer flexibility and extensibility, without paying the price of introducing many new XML elements.
\end{xl}
\begin{xc}
\chapter{\textcolor{gray!95}{Lifted Forms of Constraints}}\label{cha:lifted}

In this chapter, we show how some constraints can be lifted to lists and matrices.
\end{xc}

\begin{xl}
\section{Constraints lifted to Lists, Sets and Multisets}

Many constraints, introduced earlier on integer variables can be extended to lists (tuples), sets and multisets.
In \x3, this is quite easy to handle: replace, when appropriate, each integer variable of a list by an element \xml{list}, or replace each of them by an element \xml{set}, or even replace each of them by an element \xml{mset}.
The semantics, initially given for a sequence of variables, is naturally extended to apply to a sequence of lists of variables, a sequence of sets of variables, and a sequence of multisets of variables.
The semantics must handle now tuples of values, sets of values, and multisets of values, since:
\begin{itemize}
\item The values assigned to the variables of an element \xml{list} represent a tuple of values. For example, if we have \verb;<list> x1 x2 x3 </list>; and the instantiation $x_1=1,x_2=0,x_3=1$, we deal with the tuple $\langle 1,0,1\rangle$.
\item The values assigned to the variables of an element \xml{set} represent a set of values. For example, if we have \verb;<set> x1 x2 x3 </set>; and the instantiation $x_1=1,x_2=0,x_3=1$, we deal with the set $\{0,1\}$.
\item The values assigned to the variables of an element \xml{mset} represent a multiset of values. For example, if we have \verb;<mset> x1 x2 x3 </mset>; and the instantiation $x_1=1,x_2=0,x_3=1$, we deal with the multiset $\{\!\{0,1,1\}\!\}$.
\end{itemize}

A lifting operation always applies to an element \xml{list} conceived to contain integer variables.
After all variables have been replaced by lists, sets or tuples, the opening and closing tags for the initial element \xml{list} are no more required.
This is the reason we shall never represent them.

In this section, we show this approach for the most popular ``basic'' constraints. 
More specifically, we present the following constraints lifted to lists, sets and multisets:
\begin{itemize}
\item \gb{allDifferent}
\item \gb{allEqual}
\item \gb{allDistant}
\item \gb{ordered}
\item \gb{allIncomparable}
\item \gb{nValues}
\item \gb{sort}
\end{itemize}
\end{xl}
\begin{xc}
\section{Constraints lifted to Lists}
  
Many constraints, introduced earlier on integer variables can be extended to lists (tuples).
In \x3, this is quite easy to handle: replace, when appropriate, each integer variable of a list by an element \xml{list}.
The semantics, initially given for a sequence of variables, is naturally extended to apply to a sequence of lists of variables.
The semantics must handle now tuples of values.
\begin{itemize}
\item The values assigned to the variables of an element \xml{list} represent a tuple of values. For example, if we have \verb;<list> x1 x2 x3 </list>; and the instantiation $x_1=1,x_2=0,x_3=1$, we deal with the tuple $\langle 1,0,1\rangle$.
\end{itemize}

A lifting operation always applies to an element \xml{list} conceived to contain integer variables.
After all variables have been replaced by lists, the opening and closing tags for the initial element \xml{list} are no more required.
This is the reason we shall never represent them.

In this section, we show this approach for the most popular ``basic'' constraints. 
More specifically, for \x3-core, we present the following constraints lifted to lists:
\begin{itemize}
\item \gb{allDifferent}
\item \gb{ordered}
\end{itemize}
\end{xc}

Because these constraints are defined on several sequences (vectors) of variables, they admit a parameter $\ns{X}=\langle X_1,X_2,\ldots \rangle$, with $X_1=\langle x_{1,1},x_{1,2},\ldots \rangle, X_2=\langle x_{2,1},x_{2,2},\ldots \rangle, \ldots$
If $X=\langle x_1,x_2,\ldots,x_p\rangle$ then $|X|=p$, and:
\begin{itemize}
\item $\va{X}$ denotes $\langle \va{x}_1,\va{x}_2,\ldots,\va{x}_p \rangle$, the tuple of values obtained when considering an arbitrary instantiation of $X$. 
\begin{xl}
\item $\{ \va{X} \}$ denotes $\{ \va{x}_i : 1 \leq i \leq p\}$, the set of values obtained when considering an arbitrary instantiation of $X$.
\item $\{\!\{ \va{X} \}\!\}$ denotes $\{\!\{ \va{x}_i : 1 \leq i \leq p\}\!\}$, the multiset of values obtained when considering an arbitrary instantiation of $X$.
\end{xl}
\end{itemize}

\begin{xl}
Although we do not introduce new constraint types (names) in \x3, in order to clarify the textual description (in particular, for the semantics), we shall refer in the text to the versions of a constraint $\gb{ctr}$, lifted to lists, sets and multisets, by $\gb{ctr \ti list}$, $\gb{ctr \ti set}$ and $\gb{ctr \ti mset}$, respectively.
For example, we shall refer in the text to $\gb{allDifferent \ti list}$, $\gb{allDifferent \ti set}$ and $\gb{allDifferent \ti mset}$ when considering lifted versions of $\gb{allDifferent}$.

\subsection{Lifted Constraints \gb{allDifferent}}

The constraint \gb{allDifferent}, introduced earlier on integer variables, can be naturally extended to lists (tuples), sets and multisets \cite{QW_beyond}.

\subsubsection{On lists (tuples)}\label{ctr:allDifferentList} 
\end{xl}
\begin{xc}
Although we do not introduce new constraint types (names) in \x3, in order to clarify the textual description (in particular, for the semantics), we shall refer in the text to the versions of a constraint $\gb{ctr}$, lifted to lists, by $\gb{ctr \ti list}$.
For example, we shall refer in the text to $\gb{allDifferent \ti list}$, when considering the version of $\gb{allDifferent}$ lifted to lists.

\subsection{Constraint $\gb{allDifferent \ti list}$}\label{ctr:allDifferentList} 

The constraint \gb{allDifferent}, introduced earlier on integer variables, can be naturally extended to lists (tuples) \cite{QW_beyond}.
\end{xc}

If \gb{allDifferent} admits as parameter several lists of integer variables, then the constraint ensures that the tuple of values taken by variables of the first element \xml{list} is different from the tuple of values taken by variables of the second element \xml{list}.
If more than two elements \xml{list} are given, all tuples must be different. 
A variant enforces tuples to take distinct values, except those that are assigned to some specified tuples (often, the single tuple containing only 0), specified in the optional element \xml{except}. 

\begin{boxsy}
\begin{syntax} 
<allDifferent>
  (<list> (intVar wspace)2+ </list>)2+
  [<except> ("(" intVal ("," intVal)+ ")")+ </except>]
</allDifferent>
\end{syntax}
\end{boxsy}

As explained in the introduction of this section, in the text below, $\gb{allDifferent \ti list}$ refers to \gb{allDifferent} defined over several lists of integer variables.

\begin{boxse}
\begin{semantics}
$\gb{allDifferent \ti list}(\ns{X},E)$, with $\ns{X}=\langle X_1, X_2,\ldots \rangle$, $E$ the set of discarded tuples, iff 
  $\forall (i,j) : 1 \leq i < j \leq |\ns{X}|, \va{X}_i \neq \va{X}_j \lor \va{X}_i \in E \lor \va{X}_j \in E$ 
$\gb{allDifferent \ti list}(\ns{X})$ iff $\gb{allDifferent \ti list}(\ns{X},\emptyset)$ 

$\gbc{Prerequisite}: |\ns{X}| \geq 2 \land \forall i : 1 \leq i < |\ns{X}|, |X_i| = |X_{i+1}| \geq 2 \land \forall \tau \in E, |\tau|=|X_1|$
\end{semantics}
\end{boxse}

\begin{boxex}
\begin{xcsp}  
<allDifferent id="c1">
   <list> x1 x2 x3 x4 </list>
   <list> y1 y2 y3 y4 </list>
</allDifferent>
<allDifferent id="c2">
   <list> v1 v2 v3 v4 </list>
   <list> w1 w2 w3 w4 </list>
   <list> z1 z2 z3 z4 </list>
   <except> (0,0,0,0) </except>
</allDifferent>
\end{xcsp}
\end{boxex}

Constraints captured by \gb{allDifferent\ti list} from the \cat:
\begin{itemize}
\item \gb{lex\_different}  
\item \gb{lex\_alldifferent} 
\item \gb{lex\_alldifferent\_except\_0}
\end{itemize}

\begin{xl}
\subsubsection{On sets}\label{ctr:allDifferentSet} 

If \gb{allDifferent} admits several sets of integer variables as parameter, then the constraint ensures that the set of values taken by variables of the first element \xml{set} is different from the set of values taken by variables of the second element \xml{set}.
If more than two elements \xml{set} are given, all sets must be different. 
A variant enforces sets of values to be different, except those that are assigned to some specified sets, specified in the optional element \xml{except}. 

\begin{boxsy}
\begin{syntax} 
<allDifferent>
  (<set> (intVar wspace)+ </set>)2+
  [<except> ("{" intVal ("," intVal)* "}")+ </except>]
</allDifferent>
\end{syntax}
\end{boxsy}

As explained in the introduction of this section, in the text below, $\gb{allDifferent \ti set}$ refers to \gb{allDifferent} defined over several sets of integer variables.

\begin{boxse}
\begin{semantics}
$\gb{allDifferent \ti set}(\ns{X},E)$, with $\ns{X}=\langle X_1,X_2,\ldots \rangle$, $E$ the set of discarded sets, iff 
  $\forall (i,j) : 1 \leq i < j \leq |\ns{X}|, \{\va{X}_i\} \neq \{\va{X}_j\} \lor \{\va{X}_i\} \in E \lor \{\va{X}_j\} \in E$
$\gb{allDifferent \ti set}(\ns{X})$ iff $\gb{allDifferent \ti set}(\ns{X},\emptyset)$ 

$\gbc{Prerequisite}: |\ns{X}| \geq 2 \land \forall i : 1 \leq i \leq |\ns{X}|, |X_i| \geq 1$
\end{semantics}
\end{boxse}

\begin{boxex}
\begin{xcsp}  
<allDifferent id="c1">
   <set> x1 x2 x3 x4 </set>
   <set> y1 y2 y3 </set>
</allDifferent>
<allDifferent id="c2">
   <set> v1 v2 v3 </set>
   <set> w1 w2 w3 w4 </set>
   <set> z1 z2 </set>
   <except> {0,1} </except>
</allDifferent>
\end{xcsp}
\end{boxex}

\subsubsection{On multisets}\label{ctr:allDifferentMset} 

If \gb{allDifferent} admits several multisets of integer variables as parameter, then the constraint ensures that the multiset of values taken by variables of the first element \xml{mset} is different from the multiset of values taken by variables of the second element \xml{mset}.
If more than two elements \xml{mset} are given, all multisets must be different. 
A variant enforces multisets of values to be different, except those that correspond some specified multisets, specified in the optional element \xml{except}. 

\begin{boxsy}
\begin{syntax} 
<allDifferent>
  (<mset> (intVar wspace)2+ </mset>)2+
  [<except> ("{{" intVal ("," intVal)+ "}}")+ </except>]
</allDifferent>
\end{syntax}
\end{boxsy}

As explained in the introduction of this section, in the text below, $\gb{allDifferent \ti mset}$ refers to \gb{allDifferent} defined over several multisets of integer variables.

\begin{boxse}
\begin{semantics}
$\gb{allDifferent \ti mset}(\ns{X},E)$, with $\ns{X}=\langle X_1,X_2,\ldots \rangle$, $E$ the set of discarded msets, iff 
  $\forall (i,j) : 1 \leq i < j \leq |\ns{X}|, \{\!\{\va{X}_i\}\!\} \neq \{\!\{\va{X}_j\}\!\} \lor \{\!\{\va{X}_i\}\!\} \in E \lor \{\!\{\va{X}_j\}\!\} \in E$  
$\gb{allDifferent \ti mset}(\ns{X})$ iff $\gb{allDifferent \ti mset}(\ns{X},\emptyset)$ 

$\gbc{Prerequisite}: |\ns{X}| \geq 2 \land \forall i : 1 \leq i < |\ns{X}|, |X_i| = |X_{i+1}| \geq 2 \land \forall M \in E, |M|=|X_1|$
\end{semantics}
\end{boxse}


\begin{boxex}
\begin{xcsp}  
<allDifferent id="c1">
   <mset> x1 x2 x3 x4 </mset>
   <mset> y1 y2 y3 y4 </mset>
</allDifferent>
<allDifferent id="c2">
   <mset> v1 v2 v3 v4 </mset>
   <mset> w1 w2 w3 w4 </mset>
   <mset> z1 z2 z3 z4 </mset>
   <except> {{0,0,0,0}} </except>
</allDifferent>
\end{xcsp}
\end{boxex}
\end{xl}

\begin{xl}
\subsection{Lifted Constraints \gb{allEqual}}\label{sec:allEqualLifted}

The versions of \gb{allEqual} lifted to lists, sets and multisets are defined similarly to those presented above for \gb{allDifferent}.
The semantics below, are introduced for the base cases (i.e., without \xml{except}).

\subsubsection{On lists (tuples)}\label{ctr:allEqualList} 

\begin{boxsy}
\begin{syntax} 
<allEqual>
  (<list> (intVar wspace)2+ </list>)2+
  [<except> ("(" intVal ("," intVal)+ ")")+ </except>]
</allEqual>
\end{syntax}
\end{boxsy}

\begin{boxse}
\begin{semantics}
$\gb{allEqual \ti list}(\ns{X})$, with $\ns{X}=\langle X_1,X_2,\ldots \rangle$, iff 
  $\forall (i,j) : 1 \leq i < j \leq |\ns{X}|, \va{X}_i = \va{X}_j$ 

$\gbc{Prerequisite}: |\ns{X}| \geq 2 \land \forall i : 1 \leq i < |\ns{X}|, |X_i| = |X_{i+1}| \geq 2$
\end{semantics}
\end{boxse}

Constraints captured by \gb{allEqual\ti list} from the \cat: \gb{lex\_equal}

\subsubsection{On sets}\label{ctr:allEqualSet} 

\begin{boxsy}
\begin{syntax} 
<allEqual>
  (<set> (intVar wspace)+ </set>)2+
  [<except> ("{" intVal ("," intVal)* "}")+ </except>]
</allEqual>
\end{syntax}
\end{boxsy}

\begin{boxse}
\begin{semantics}
$\gb{allEqual \ti set}(\ns{X})$, with $\ns{X}=\langle X_1,X_2,\ldots \rangle$, iff 
  $\forall (i,j) : 1 \leq i < j \leq |\ns{X}|, \{\va{X}_i\} = \{\va{X}_j\}$

$\gbc{Prerequisite}: |\ns{X}| \geq 2 \land \forall i : 1 \leq i \leq |\ns{X}|, |X_i| \geq 1$
\end{semantics}
\end{boxse}

\subsubsection{On multisets}\label{ctr:allEqualMset} 

\begin{boxsy}
\begin{syntax} 
<allEqual>
  (<mset> (intVar wspace)2+ </mset>)2+
  [<except> ("{{" intVal ("," intVal)+ "}}")+ </except>]
</allEqual>
\end{syntax}
\end{boxsy}

\begin{boxse}
\begin{semantics}
$\gb{allEqual \ti mset}(\ns{X})$, with $\ns{X}=\langle X_1,X_2,\ldots \rangle$, iff 
  $\forall (i,j) : 1 \leq i < j \leq |\ns{X}|, \{\!\{\va{X}_i\}\!\} = \{\!\{\va{X}_j\}\!\}$ 

$\gbc{Prerequisite}: |\ns{X}| \geq 2 \land \forall i : 1 \leq i < |\ns{X}|, |X_i| = |X_{i+1}| \geq 2$
\end{semantics}
\end{boxse}

Constraints captured by \gb{allEqual\ti mset} from the \cat: \gb{same}

\subsection{Lifted Constraints \gb{allDistant}}

Similarly to \gb{allDifferent} and \gb{allEqual}, \gb{allDistant} can be lifted to lists, sets and multisets.
Of course, we have to clarify the distances that are used, when considering pairs of lists, sets and multisets.
By default, we shall use the following distances:
\begin{itemize}
\item Hamming distance for lists: $\nm{dist_{H}}(X,Y)=|\{i : 1 \leq i \leq |X| \land x_i \neq y_i\}|$ with $X=\langle x_1,x_2,\ldots \rangle$, $Y=\langle y_1,y_2,\ldots \rangle$ and $|X|=|Y|$.
\item infimum distance for sets (of integers): $\nm{dist_{inf}}(X,Y)=\min \{|x - y| : x \in X \land y \in Y\}$
\item Manhattan distance for multisets: $\nm{dist_{M}}(X,Y)=\sum_{a \in X \cup Y} |\nu_X(a) - \nu_Y(a)|$ with $\nu_Z(c)$ denoting the multiplicity (number of occurrences) of value $c$ in the multiset $Z$.
\end{itemize}


\subsubsection{On lists (tuples)}\label{ctr:allDistantList}

\begin{boxsy}
\begin{syntax} 
<allDistant>
  (<list> (intVar wspace)2+ </list>)2+
  <condition> "(" operator "," operand ")" </condition>
</allDistant>
\end{syntax}
\end{boxsy}

\begin{boxse}
\begin{semantics}
$\gb{allDistant \ti list}(\ns{X},(\odot,k))$, with $\ns{X}=\langle X_1,X_2,\ldots \rangle$, iff 
  $\forall (i,j) : 1 \leq i < j \leq |\ns{X}|, dist_H(\va{X}_i,\va{X}_j) \odot \va{k}$

$\gbc{Prerequisite}: |\ns{X}| \geq 2 \land \forall i : 1 \leq i < |\ns{X}|, |X_i| = |X_{i+1}| \geq 2$
\end{semantics}
\end{boxse}

Constraints captured by \gb{allDistant\ti list} from the \cat:
\begin{itemize}
\item \gb{differ\_from\_at\_least\_k\_pos},  \gb{differ\_from\_at\_most\_k\_pos}
\item \gb{differ\_from\_exactly\_k\_pos} 
\item \gb{all\_differ\_from\_at\_least\_k\_pos}, \gb{all\_differ\_from\_at\_most\_k\_pos} 
\item \gb{all\_differ\_from\_exactly\_k\_pos} 
\end{itemize}

\subsubsection{On sets}\label{ctr:allDistantSet}

\begin{boxsy}
\begin{syntax} 
<allDistant>
  (<set> (intVar wspace)+ </set>)2+
  <condition> "(" operator "," operand ")" </condition>
</allDistant>
\end{syntax}
\end{boxsy}

\begin{boxse}
\begin{semantics}
$\gb{allDistant \ti set}(\ns{X},(\odot,k))$, with $\ns{X}=\langle X_1,X_2,\ldots \rangle$, iff 
  $\forall (i,j) : 1 \leq i < j \leq |\ns{X}|, dist_{inf}(\{\va{X}_i\},\{\va{X}_j\}) \odot \va{k}$

$\gbc{Prerequisite}: |\ns{X}| \geq 2 \land \forall i : 1 \leq i \leq |\ns{X}|, |X_i| \geq 1$
\end{semantics}
\end{boxse}

We propose to identify two types of $\gb{allDistant \ti set}$, corresponding to the case where all sets must be disjoint (value \val{disjoint} for \att{case}) and the case where all sets must be overlapping (value \val{overlapping} for \att{case}).
For both cases, the element \xml{condition} becomes implicit.
The two following constraints $c_1$ and $c_2$:

\begin{boxex}
\begin{xcsp}  
<allDistant id="c1">
   <set> x1 x2 x3 x4 </set>
   <set> y1 y2 y3 y4 </set>
   <set> z1 z2 z3 </set>
   <condition> (gt,0) </condition>
</allDistant>
<allDistant id="c2">
   <set> v1 v2 v3 </set>
   <set> v4 v5 v6 v7 </set>
   <set> v8 v9 </set>
   <condition> (eq,0) </condition>
</allDistant>
\end{xcsp}
\end{boxex}

can then be defined by:
\begin{boxex}
\begin{xcsp}  
<allDistant id="c1" case="disjoint">
   <set> x1 x2 x3 x4 </set>
   <set> y1 y2 y3 y4 </set>
   <set> z1 z2 z3 </set>
</allDistant>
<allDistant id="c2" case="overlapping">
   <set> v1 v2 v3 </set>
   <set> v4 v5 v6 v7 </set>
   <set> v8 v9 </set>
</allDistant>
\end{xcsp}
\end{boxex}

\subsubsection{On multisets}\label{ctr:allDistantMset}

\begin{boxsy}
\begin{syntax} 
<allDistant>
  (<mset> (intVar wspace)2+ </mset>)2+
  <condition> "(" operator "," operand ")" </condition>
</allDistant>
\end{syntax}
\end{boxsy}

\begin{boxse}
\begin{semantics}
$\gb{allDistant \ti mset}(\ns{X},(\odot,k))$, with $\ns{X}=\langle X_1,X_2,\ldots \rangle$, iff 
  $\forall (i,j) : 1 \leq i < j \leq |\ns{X}|, dist_{M}(\{\!\{\va{X}_i\}\!\},\{\!\{\va{X}_j\}\!\}) \odot \va{k}$

$\gbc{Prerequisite}: |\ns{X}| \geq 2 \land \forall i : 1 \leq i < |\ns{X}|, |X_i| = |X_{i+1}| \geq 2$
\end{semantics}
\end{boxse}

\begin{remark}
In the future, we project to deal with other distances by introducing an attribute \att{distance}, while fixing the terminology (values that can be used for \att{distance}).
\end{remark}
\end{xl}

\begin{xl}
\subsection{Lifted Constraints \gb{ordered} (\gb{lex} on lists)}

The constraint \gb{ordered} can be naturally lifted to lists, sets and multisets.
As for \gb{allDistant}, we have to clarify the (total or partial) orders that are used, when comparing lists, sets and multisets.
By default, we shall use the following orders:
\begin{itemize}
\item lexicographic order $\leq_{lex}$ for lists;
\item inclusion order $\subseteq$ for sets;
\item multiplicity inclusion order  $\subseteq$ for multisets: $X \subseteq Y$ iff $\forall a \in X, \nu_X(a) \leq \nu_Y(a)$, with $\nu_Z(c)$ denoting the multiplicity (number of occurrences) of value $c$ in the multiset $Z$.
\end{itemize}

\subsubsection{On lists (tuples)}\label{ctr:orderedList}
\end{xl}
\begin{xc}
\subsection{Constraint \gb{lex} ($\gb{ordered \ti list}$)}\label{ctr:orderedList}

The constraint \gb{ordered} can be naturally lifted to lists.
When comparing lists (tuples), we shall use the lexicographic order $\leq_{lex}$.
\end{xc}

Because this constraint is very popular, it is allowed to use \gb{lex}, instead of \gb{ordered} over lists of integer variables.
The constraint \gb{lex}, see \cite{CB_revisiting,FHKMW_global}, ensures that the tuple formed by the values assigned to the variables of the first element \xml{list} is related to the tuple formed by the values assigned to the variables of the second element \xml{list} with respect to the operator specified in \xml{operator}.
If more than two elements \xml{list} are given, the entire sequence of tuples must be ordered; this captures then \gb{lexChain} \cite{CB_arc}.
Note that, since Version 3.2, it is possible to have lists (vectors) with constants.

\begin{boxsy}
\begin{syntax} 
<lex>
  (<list> (intVar wspace)2+ | (intVal wspace)2+ </list>)2+
  <@operator@> "lt" | "le" | "ge" | "gt" </@operator@>
</lex>
\end{syntax}
\end{boxsy}


\begin{boxse}
\begin{semantics}
$\gb{lex}(\ns{X},\odot)$, with $\ns{X}=\langle X_1,X_2,\ldots \rangle$ and $\odot \in \{<_{lex}, \leq_{lex}, \geq_{lex}, >_{lex}\}$, iff 
  $\forall i : 1 \leq i < |\ns{X}|, \va{X}_i \odot \va{X}_{i+1}$ 

$\gbc{Prerequisite}: |\ns{X}| \geq 2 \land \forall i : 1 \leq i < |\ns{X}|, |X_i| = |X_{i+1}| \geq 2$
\end{semantics}
\end{boxse}

In the following example, the constraint $c_1$ states that $\langle x_1,x_2,x_3,x_4 \rangle \leq_{lex} \langle y_1,y_2,y_3,y_4 \rangle$, whereas $c_2$ states that $\langle z_1,z_2,z_3 \rangle >_{lex} \langle z_4,z_5,z_6 \rangle >_{lex} \langle z_7,z_8,z_9 \rangle$.

\begin{boxex}
\begin{xcsp}  
<lex id="c1">
  <list> x1 x2 x3 x4 </list>
  <list> y1 y2 y3 y4 </list>
  <operator> le </operator>
</lex>
<lex id="c2">
  <list> z1 z2 z3 </list>
  <list> z4 z5 z6 </list>
  <list> z7 z8 z9 </list>
  <operator> gt </operator>
</lex>
 \end{xcsp}
\end{boxex}
 
The next constrraint $c_3$ allows us to compare a list of variables with a spcified (constant) tuple, stating that $\langle x_1,x_2,x_3,x_4 \rangle \leq_{lex} \langle 2,3,0,1 \rangle$.

\begin{boxex}
\begin{xcsp}  
<lex id="c3">
  <list> x1 x2 x3 x4 </list>
  <list> 2 3 0 1 </list>
  <operator> le </operator>
</lex>
\end{xcsp}
\end{boxex}



Constraints captured by \gb{lex} from the \cat:
\begin{itemize}
\item \gb{lex\_between} 
\item \gb{lex\_greater}, \gb{lex\_greatereq}
\item \gb{lex\_less}, \gb{lex\_lesseq} 
\item \gb{lex\_chain\_greater}, \gb{lex\_chain\_greatereq}
\item \gb{lex\_chain\_less}, \gb{lex\_chain\_lesseq}
\end{itemize}

\begin{xl}
\subsubsection{On sets}\label{ctr:orderedSet}

The constraint ensures that the set of values taken by variables of the first element \xml{set} is related to the set of values taken by the variables of the second element \xml{set}, with respect to a relational set operator.
If more than two elements \xml{set} are given, the entire sequence of sets must be ordered.

\begin{boxsy}
\begin{syntax} 
<ordered>
  (<set> (intVar wspace)+ </set>)2+
  <@operator@> "subset" | "subseq" | "supseq" | "supset" </@operator@>
</ordered>
\end{syntax}
\end{boxsy}

\begin{boxse}
\begin{semantics}
$\gb{ordered\ti set}(\ns{X},\odot)$, with $\ns{X}=\{X_1,X_2,\ldots\}$ and $\odot \in \{\subset,\subseteq,\supseteq,\supset\}$, iff 
  $\forall i : 1 \leq i < |\ns{X}|, \{\va{X}_i\} \odot \{\va{X}_{i+1}\}$  

$\gbc{Prerequisite}: |\ns{X}| \geq 2 \land \forall i : 1 \leq i \leq |\ns{X}|, |X_i| \geq 1$
\end{semantics}
\end{boxse}

This captures \gb{uses} \cite{BHHKW_range}.

\subsubsection{On multisets}\label{ctr:orderedMset}

\begin{boxsy}
\begin{syntax} 
<ordered>
  (<mset> (intVar wspace)2+ </mset>)2+
  <@operator@> "subset" | "subseq" | "supseq" | "supset" </@operator@>
</ordered>
\end{syntax}
\end{boxsy}

\begin{boxse}
\begin{semantics}
$\gb{ordered\ti mset}(\ns{X},\odot)$, with $\ns{X}=\{X_1,X_2,\ldots\}$ and $\odot \in \{\subset,\subseteq,\supseteq,\supset\}$, iff 
  $\forall i : 1 \leq i < |\ns{X}|, \{\!\{\va{X}_i\}\!\} \odot \{\!\{\va{X}_{i+1}\}\!\}$  

$\gbc{Prerequisite}: |\ns{X}| \geq 2 \land \forall i : 1 \leq i < |\ns{X}|, |X_i| = |X_{i+1}| \geq 2$
\end{semantics}
\end{boxse}

This captures \gb{usedBy} \cite{BKT_same}.

\begin{remark}
In the future, we project to deal with other orders by introducing an attribute \att{order}, while fixing the terminology (values that can be used for \att{order}).
\end{remark}
\end{xl}

\begin{xl}
\subsection{Lifted Constraints \gb{allIncomparable}}

The constraint \gb{allIncomparable} is not defined on integer variables because $\N$ is totally ordered, but it is meaningful on lists, sets and multisets.
This is why it is only introduced in this chapter.
As for \gb{ordered}, we have to clarify the partial orders that are used, when comparing lists, sets and multisets.
By default, we shall use the following partial orders:
\begin{itemize}
\item product order (and not lexicographic order) for lists:  $X \leq_{prod} Y$ iff $\forall i : 1 \leq i \leq |X|, x_i \leq y_i$ with $X=\langle x_1,x_2,\ldots\rangle$, $Y=\langle y_1,y_2,\ldots\rangle$ and $|X|=|Y|$;
\item inclusion order $\subseteq$ for sets;
\item multiplicity inclusion order  $\subseteq$ for multisets: $X \subseteq Y$ iff $\forall a \in X, \nu_X(a) \leq \nu_Y(a)$, with $\nu_Z(c)$ denoting the multiplicity (number of occurrences) of value $c$ in the multiset $Z$.
\end{itemize}

\subsubsection{On lists (tuples)}\label{ctr:allIncomparableList}

All elements \xml{list} must correspond to tuples that are all incomparable (for the product order).

\begin{boxsy}
\begin{syntax} 
<allIncomparable>
  (<list> (intVar wspace)2+ </list>)2+
</allIncomparable>
\end{syntax}
\end{boxsy}

\begin{boxse}
\begin{semantics}
$\gb{allIncomparable\ti list}(\ns{X})$, with $\ns{X}=\langle X_1,X_2,\ldots \rangle$,  iff
  $\forall (i,j) : 1 \leq i < j \leq |\ns{X}|, \exists (k,l) : 1 \leq k < l \leq |X_i| \land \va{x}_{i,k} > \va{x}_{j,k} \land \va{x}_{i,l} < \va{x}_{j,l} $ 

$\gbc{Prerequisite}: |\ns{X}| \geq 2 \land \forall i : 1 \leq i < |\ns{X}|, |X_i| = |X_{i+1}| \geq 2$
\end{semantics}
\end{boxse}

The constraints captured by \gb{allIncomparable\ti list} from the \cat are: \gb{incomparable} and \gb{all\_incomparable}

\subsubsection{On sets}\label{ctr:allIncomparableSet}

All elements \xml{set} must be incomparable (for inclusion).

\begin{boxsy}
\begin{syntax} 
<allIncomparable>
  (<set> (intVar wspace)+ </set>)2+
</allIncomparable>
\end{syntax}
\end{boxsy}

\begin{boxse}
\begin{semantics}
$\gb{allIncomparable\ti set}(\ns{X})$, with $\ns{X}=\langle X_1,X_2,\ldots \rangle$,  iff
  $\forall (i,j) : 1 \leq i < j \leq |\ns{X}|, \{\va{X}_i\} \not\subseteq \{\va{X}_j\} \land \{\va{X}_j\} \not\subseteq \{\va{X}_i\}$ 

$\gbc{Prerequisite}: |\ns{X}| \geq 2 \land \forall i : 1 \leq i \leq |\ns{X}|, |X_i| \geq 1$
\end{semantics}
\end{boxse}

\subsubsection{On multisets}\label{ctr:allIncomparableMset}

All elements \xml{mset} must be incomparable (for multiplicity inclusion).

\begin{boxsy}
\begin{syntax} 
<allIncomparable>
  (<mset> (intVar wspace)2+ </mset>)2+
</allIncomparable>
\end{syntax}
\end{boxsy}

\begin{boxse}
\begin{semantics}
$\gb{allIncomparable\ti mset}(\ns{X})$, with $\ns{X}=\langle X_1,X_2,\ldots \rangle$,  iff
  $\forall (i,j) : 1 \leq i < j \leq |\ns{X}|, \{\!\{\va{X}_i\}\!\} \not\subseteq \{\!\{\va{X}_j\}\!\} \land \{\!\{\va{X}_j\}\!\} \not\subseteq \{\!\{\va{X}_i\}\!\}$ 

$\gbc{Prerequisite}:  |\ns{X}| \geq 2 \land \forall i : 1 \leq i < |\ns{X}|, |X_i| = |X_{i+1}| \geq 2$
\end{semantics}
\end{boxse}

\begin{remark}
In the future, we project to deal with other partial orders by introducing an attribute \att{order}, while fixing the terminology (values that can be used for \att{order}).
\end{remark}
\end{xl}

\begin{xl}
\subsection{Lifted Constraints \gb{nValues}}

The constraint \gb{nValues} can be naturally lifted to lists, sets and multisets; the term ``value'' being understood successively as a list, a set or a multiset.

\subsubsection{On lists}\label{ctr:nValuesList}

This constraint, sometimes called \gb{nVectors}, ensures that the number of distinct tuples taken by variables of different elements \xml{list} (of same size) must respect a numerical condition.
A variant enforces tuples to take distinct values, except those that are assigned to some specified tuples (often, the single tuple containing only 0), specified in the optional element \xml{except}. 

\begin{boxsy}
\begin{syntax} 
<nValues>
   (<list> (intVar wspace)2+ </list>)2+
   [<except> ("(" intVal ("," intVal)+ ")")+ </except>]
   <condition> "(" operator "," operand ")" </condition>
</nValues>
\end{syntax}
\end{boxsy}

For the semantics, $E$ denotes the set of tuples that must be discarded.

\begin{boxse}
\begin{semantics}
$\gb{nValues\ti list}(\ns{X},E,(\odot,k))$, with $\ns{X}=\langle X_1,X_2,\ldots \rangle$, iff 
  $|\{\va{X}_i : 1 \leq i \leq |\ns{X}|\} \setminus E| \odot \va{k}$ 
$\gb{nValues}(\ns{X},(\odot,k))$ iff $\gb{nValues}(\ns{X},\emptyset,(\odot,k))$

$\gbc{Prerequisite}: |\ns{X}| \geq 2 \land \forall i : 1 \leq i < |\ns{X}|, |X_i| = |X_{i+1}| \geq 2 \land \forall \tau \in E, |\tau|=|X_1|$
\end{semantics}
\end{boxse}

\begin{boxex}
\begin{xcsp}  
<nValues id="c">
   <list> x1 x2 x3 </list> 
   <list> y1 y2 y3 </list> 
   <list> z1 z2 z3 </list>
   <condition> (eq,2) </condition>
</nValues>
\end{xcsp}
\end{boxex}
 
Constraints captured by \gb{nValues\ti list} from the \cat:
\begin{itemize}
\item \gb{atleast\_nvector}, \gb{atmost\_nvector} 
\item \gb{nvector}, \gb{nvectors}
\end{itemize}

\subsubsection{On sets}\label{ctr:nValuesSet}

\begin{boxsy}
\begin{syntax} 
<nValues>
   (<set> (intVar wspace)+ </set>)2+
   [<except> ("{" intVal ("," intVal)* "}")+ </except>]
   <condition> "(" operator "," operand ")" </condition>
</nValues>
\end{syntax}
\end{boxsy}

\begin{boxse}
\begin{semantics}
$\gb{nValues\ti set}(\ns{X},E,(\odot,k))$, with $\ns{X}=\langle X_1,X_2,\ldots \rangle$, iff 
  $|\{ \{\va{X}_i\} : 1 \leq i \leq |\ns{X}|\} \setminus E| \odot \va{k}$ 
$\gb{nValues\ti set}(\ns{X},(\odot,k))$ iff $\gb{nValues}(\ns{X},\emptyset,(\odot,k))$

$\gbc{Prerequisite}: |\ns{X}| \geq 2 \land \forall i : 1 \leq i \leq |\ns{X}|, |X_i| \geq 1$
\end{semantics}
\end{boxse}

\subsubsection{On multisets}\label{ctr:nValuesMset}

\begin{boxsy}
\begin{syntax} 
<nValues>
   (<mset>  (intVar wspace)2+ </mset>)2+
   [<except> ("{{" intVal ("," intVal)+ "}}")+ </except>]
   <condition> "(" operator "," operand ")" </condition>
</nValues>
\end{syntax}
\end{boxsy}

\begin{boxse}
\begin{semantics}
$\gb{nValues\ti mset}(\ns{X},E,(\odot,k))$, with $\ns{X}=\langle X_1,X_2,\ldots \rangle$, iff 
  $|\{ \{\!\{\va{X}_i\}\!\} : 1 \leq i \leq |\ns{X}|\} \setminus E| \odot \va{k}$ 
$\gb{nValues\ti mset}(\ns{X},(\odot,k))$ iff $\gb{nValues}(\ns{X},\emptyset,(\odot,k))$

$\gbc{Prerequisite}: |\ns{X}| \geq 2 \land \forall i : 1 \leq i < |\ns{X}|, |X_i| = |X_{i+1}| \geq 2 \land \forall M \in E, |M|=|X_1|$
\end{semantics}
\end{boxse}
\end{xl}



\section{Constraints lifted to Matrices}

Some constraints, introduced earlier on list(s) of integer variables, can naturally be extended to matrices of variables.
This means that such constraints restraint both row lists and column lists of variables.
The principle is to replace the element \xml{list} of the basic constraint by an element \xml{matrix}.
\begin{xl}However, lifting constraints over matrices is not always purely automatic, as we shall show with \gb{cardinality}.
\end{xl}

As in the previous section, although we do not introduce new constraint types (names) in \x3, we shall refer in the text to the matrix version of a constraint $\gb{ctr}$ by $\gb{ctr \ti matrix}$, in order to clarify the textual description (in particular, for the semantics).
For example, we shall refer in the text to $\gb{allDifferent \ti matrix}$ when considering the matrix version of $\gb{allDifferent}$.

In this section, we present the following constraints defined on matrices of integer variables:
\begin{enumerate}
\item \gb{allDifferent\ti matrix}
\item \gb{ordered\ti matrix (\gb{lex2})}
\item \gb{element\ti matrix}
\begin{xl}\item \gb{cardinality\ti matrix}
\end{xl}
\end{enumerate}

These constraints are defined on matrices of variables.
So, they admit a parameter $\ns{M}=[ X_1,X_2,\ldots,X_n ]$, with $X_1=\langle x_{1,1},x_{1,2},\ldots,x_{1,m} \rangle, X_2=\langle x_{2,1},x_{2,2},\ldots,x_{2,m} \rangle, \ldots$, given by an element \xml{matrix}, assuming here a matrix of size $n \times m$. 
Note here that we use square brackets ("[" and "]") to delimit matrices, in order to distinguish them from lists of lists (where angle brackets are used).
We use the following notations below:
\begin{itemize}
\item $\forall i : 1 \leq i \leq n, \ns{M}[i] = X_i$ denotes the $ith$ row of $\ns{M}$
\item $\forall j : 1 \leq i \leq m, \ns{M}^T[j] = \langle x_{i,j} : 1 \leq i \leq |\ns{M}| \rangle$ denotes the $jth$ column of $\ns{M}$
\end{itemize}

\subsection{Constraint \gb{allDifferent\ti matrix}}\label{ctr:allDifferentMatrix}

The constraint \gb{allDifferent\ti matrix}, called \gb{alldiffmatrix} in \cite{R_globalsurvey} and in JaCoP, ensures that the values taken by variables on each row and on each column of a matrix are all different.
A variant, called allDifferentExcept\ti matrix enforces variables (on each row and each column) to take distinct values, except those that are assigned to some specified values (often, the single value 0). This is the reason why we introduce an optional element \xml{except}.

\begin{boxsy}
\begin{syntax} 
<allDifferent>
  <matrix> ("(" intVar ("," intVar)+ ")")2+ </matrix>
  [<except> (intVal wspace)+ </except>]
</allDifferent>
\end{syntax}
\end{boxsy}

\begin{boxse}
\begin{semantics}
$\gb{allDifferent\ti matrix}(\ns{M},E)$, with $\ns{M}=[X_1,X_2,\ldots,X_n]$, iff 
  $\forall i : 1 \leq i \leq n, \gb{allDifferent}(\ns{M}[i],E)$
  $\forall j : 1 \leq j \leq m, \gb{allDifferent}(\ns{M}^T[j],E)$

$\gb{allDifferent\ti matrix}(\ns{M})$ iff $\gb{allDifferent\ti matrix}(\ns{M},\emptyset)$
  
$\gbc{Prerequisite}: \forall i : 1 \leq i \leq n, |X_i| = m$
\end{semantics}
\end{boxse}

\begin{boxex}
\begin{xcsp}  
<allDifferent>
  <matrix> 
    (x1,x2,x3,x4,x5)
    (y1,y2,y3,y4,y5)
    (z1,z2,z3,z4,z5)
  </matrix>
</allDifferent>
\end{xcsp}
\end{boxex}

Remember that we can use compact forms of arrays in the element \xml{matrix}, as indicated in Section \ref{sub:compactForms}.

\subsection{Constraint \gb{ordered\ti matrix} (\gb{lex2})}\label{ctr:orderedMatrix}

The constraint \gb{ordered\ti matrix}, that can be called \gb{lex\ti matrix} too, corresponds to \gb{lex2} in the literature \cite{FFHKMPW_breaking}.
It ensures that, for a given matrix of variables, both adjacent rows and adjacent columns are lexicographically ordered.
For the syntax, we can use \gb{lex} instead of \gb{ordered}, as for \gb{ordered\ti list}. 

\begin{boxsy}
\begin{syntax} 
<lex>
  <matrix> ("(" intVar ("," intVar)+ ")")2+ </matrix>
  <@operator@> "lt" | "le" | "ge" | "gt" </@operator@>
</lex>
\end{syntax}
\end{boxsy}

\begin{boxse}
\begin{semantics}
$\gb{lex\ti matrix}(\ns{M},\odot)$, with $\ns{M}=[X_1,X_2,\ldots X_n]$ and $\odot = \{<,\leq,\geq,>\}$, iff 
  $\gb{lex}(\langle \ns{M}[1],\ldots,\ns{M}[n] \rangle,\odot)$
  $\gb{lex}(\langle \ns{M}^T[1],\ldots,\ns{M}^T[m] \rangle,\odot)$
\end{semantics}
\end{boxse}

In the following example, the constraint states that:
\begin{itemize}
\item $\langle z_1,z_2,z_3 \rangle \leq_{lex} \langle z_4,z_5,z_6 \rangle \leq_{lex} \langle z_7,z_8,z_9 \rangle$ 
\item $\langle z_1,z_4,z_7 \rangle \leq_{lex} \langle z_2,z_5,z_8 \rangle \leq_{lex} \langle z_3,z_6,z_9 \rangle$.
\end{itemize}

\begin{boxex}
\begin{xcsp}  
<lex>
  <matrix> 
    (z1,z2,z3)
    (z4,z5,z6)
    (z7,z8,z9)
  </matrix>
  <operator> le </operator>
</lex>
 \end{xcsp}
\end{boxex}
 
Remember that we can use compact forms of arrays in the element \xml{matrix}, as indicated in Section \ref{sub:compactForms}.

\subsection{Constraint \gb{element\ti matrix}}\label{ctr:elementMatrix}

The constraint \gb{element\ti matrix} has been introduced in CHIP and called \gb{element\_matrix} in the \cat.
Note here that we need to put two variables in \xml{index} because we need two indexes to designate a variable in the matrix.
The optional attributes \att{startRowIndex} and \att{startColIndex} respectively give the numbers used for indexing the first variable in each row and each column of \xml{matrix} (0, by default). 
For \x3-core, the number used for indexing the first variable in each row and each column of \xml{matrix} is necessarily 0
Note that the matrix can contain either integer variables or integer values.

\begin{boxsy}
\begin{syntax} 
<element>
  <matrix [startRowIndex="integer] [startColIndex="integer"] >
    ("(" intVar ("," intVar)+ ")")2+ | ("(" intVal ("," intVal)+ ")")2+
  </matrix>
  <index> intVar wspace intVar </index>
  <condition> "(" operator "," operand ")" </condition> 
</element>
\end{syntax}
\end{boxsy}

\begin{boxse}
\begin{semantics}
$\gb{element\ti matrix}(\ns{M},\langle i,j \rangle,(\odot,k))$, with $\ns{M}=[ \langle x_{0,0}, x_{0,1},\ldots\rangle, \langle x_{1,0}, x_{1,1},\ldots\rangle, \ldots ]$, iff 
  $\va{x}_{\va{i},\va{j}} \odot \va{k}$ 
\end{semantics}
\end{boxse}

In the following example, $x$ is a two-dimensional array of variables, and $i$ and $j$ are two stand-alone variables.

\begin{boxex}
\begin{xcsp}  
<element>
  <matrix> 
    x[][]
  </matrix>
  <index> i j </index>
  <condition> (eq,5) </condition>
</element>
\end{xcsp}
\end{boxex}

Remember that we can use compact forms of arrays in the element \xml{matrix}, as indicated in Section \ref{sub:compactForms}.
In the example, above, we use a compact form.

A {\em deprecated form} of \gb{element\ti matrix} is obtained by replacing, when the operator is 'eq', the element \xml{condition} by an element \xml{value}.
As an illustration, the deprecated form of the last constraint, introduced earlier, is:

\begin{boxex}
\begin{xcsp}  
<element>
  <matrix> 
    x[][]
  </matrix>
  <index> i j </index>
  <value> 5 </value>
</element>
\end{xcsp}
\end{boxex}

\begin{xl}
\subsection{Constraint \gb{cardinality\ti matrix}}\label{ctr:cardinalityMatrix}

The constraint \gb{cardinality\ti matrix}, see \cite{RG_matrix}, ensures a constraint \gb{cardinality} on each row and each column.
For managing cardinality of values, elements \xml{rowOccurs} and \xml{colOccurs} are introduced.
Below, we only describe the variant with cardinality variables. 

\begin{boxsy}
\begin{syntax} 
<cardinality>
  <matrix> ("(" intVar ("," intVar)+ ")")2+ </matrix>
  <values> (intVal wspace)+ </values>
  <rowOccurs> ("(" intVar ("," intVar)+ ")")2+ </rowOccurs>
  <colOccurs> ("(" intVar ("," intVar)+ ")")2+ </colOccurs>
</cardinality>
\end{syntax}
\end{boxsy}

\begin{boxse}
\begin{semantics}
$\gb{cardinality\ti matrix}(\ns{M},V,R,C)$, with $\ns{M}=[X_1,X_2,\ldots,X_n]$, $R=\langle R_1,R_2,\ldots,R_n\rangle$ $C=\langle C_1,C_2,\ldots,C_m\rangle$, iff
  $\forall i : 1 \leq i \leq n, \gb{cardinality}(\ns{M}[i],V,R[i])$
  $\forall j : 1 \leq j \leq m, \gb{cardinality}(\ns{M}^T[j],V,C[j])$

$\gbc{Prerequisite}: \forall i : 1 \leq i \leq n, |X_i| = m \land \forall i : 1 \leq i \leq n, |R_i| = |V| \land \forall j : 1 \leq j \leq m, |C_j| = |V|$
\end{semantics}
\end{boxse}

\begin{boxex}
\begin{xcsp}  
<cardinality id="c">
  <matrix> 
    (x1,x2,x3,x4)
    (y1,y2,y3,y4)
    (z1,z2,z3,z4)
  </matrix>
  <values> 0 1 </values>
  <rowOccurs> (r10,r11)(r20,r21)(r30,r31) </rowOccurs>
  <colOccurs> (c10,c11)(c20,c21)(c30,c31)(c40,c41) </colOccurs>
</cardinality>
\end{xcsp}
\end{boxex}

Remember that we can use compact forms of arrays in the element \xml{matrix}, as indicated in Section \ref{sub:compactForms}.
\end{xl}

\begin{xl}
\section{Restricted Constraints}\label{sec:restricted}

A few global constraints are defined as specific restricted cases of other ones; restricting means here hardening.
For example, \gb{allDifferentSymmetric} and \gb{increasingNValues} are restricted versions of \gb{allDifferent} and \gb{nValues}, respectively.
Of course, one can always simply post two distinct constraints instead of a combined restricted one in order to to get something equivalent.
However, it is useful to keep partly the model's structure and to inform solvers that it is possible to treat combinations of constraints as a single constraint over the same list of variables. 
That is why we define an attribute \att{restriction}, which can be added to any element containing a list of variables, i.e., any element \xml{list}.
This attribute can be assigned either a restriction or a list of restrictions (whitespace as separator), chosen among the following list:
\begin{itemize}
\item \val{allDifferent}
\item \val{increasing}
\item \val{strictlyIncreasing}
\item \val{decreasing}
\item \val{strictlyDecreasing}
\item \val{irreflexive}
\item \val{symmetric}
\item \val{convex}
\end{itemize}

When applied to an element \xml{list}, inside any \x3 constraint element \gb{ctr}, we obtain the following syntax:
Note that a possible alternative is to use the meta-constraint \gb{and}, but the solution we present here is simpler and more compact.

\begin{boxsy}
\begin{syntax} 
<ctr>
   ...
   <list restriction="restrictionList"> ... </list>
   ...
</ctr>
\end{syntax}
\end{boxsy}

Because there is usually no ambiguity in the way a restriction applies, a constraint \gb{ctr}, with restriction \gb{res}, will be denoted by  \gb{ctr$\tr$res}, as for example \gb{allDifferent$\tr$symmetric}.
The semantics of each restriction type is defined below. 
For simplicity when introducing the semantics, we assume below that the value of the attribute \att{startIndex}, associated with the list (sequence) of variables $X$, is equal to 1 (although it is 0, by default).

\begin{boxse}
\begin{semantics}
$\gb{allDifferent}(X)$, with $X=\langle x_1,x_2,\ldots \rangle$, iff 
  $\forall (i,j) : 1 \leq i < j \leq |X|,  \va{x}_i \neq \va{x}_j$
$\gb{increasing}(X)$, with $X=\langle x_1,x_2,\ldots \rangle$, iff 
  $\forall i : 1 \leq i < |X|,  \va{x}_i \leq \va{x}_{i+1}$
$\gb{strictlyIncreasing}(X)$, with $X=\langle x_1,x_2,\ldots \rangle$, iff 
  $\forall i : 1 \leq i < |X|,  \va{x}_i < \va{x}_{i+1}$
$\gb{decreasing}(X)$, with $X=\langle x_1,x_2,\ldots \rangle$, iff 
  $\forall i : 1 \leq i \leq |X|,  \va{x}_i \geq \va{x}_{i+1}$
$\gb{strictlyDecreasing}(X)$, with $X=\langle x_1,x_2,\ldots \rangle$, iff 
  $\forall i : 1 \leq i \leq |X|,  \va{x}_i > \va{x}_{i+1}$
$\gb{irreflexive}(X)$, with $X=\langle x_1,x_2,\ldots \rangle$, iff 
  $\forall i : 1 \leq i \leq |X|, \va{x}_i \neq i$
$\gb{symmetric}(X)$, with $X=\langle x_1,x_2,\ldots \rangle$, iff 
  $\forall (i,j) : 1 \leq i < j \leq |X|,  \va{x}_i = j \Leftrightarrow \va{x}_j = i$
$\gb{convex}(X)$, with $X=\langle x_1,x_2,\ldots \rangle$, iff 
  $\forall v : \min(\va{X}) \leq v \leq max(\va{X}), v \in \va{X}$
\end{semantics}
\end{boxse}

\begin{remark}
If the value of the attribute \att{startIndex} associated with the list $X$ is denoted by $\nm{start}$, then the conditions expressed for \gb{irreflexive} and \gb{symmetric} must be of the form $\va{x}_i \neq i-1+\nm{start}$.
\end{remark}

\subsection{Constraint \gb{allDifferent$\tr$symmetric}}\label{ctr:allDifferentSymmetric}

The constraint \gb{allDifferent$\tr$symmetric} \cite{R_symmetric} is a classical restriction of \gb{allDifferent}.
Not only all variables must take different values, but they must also be either grouped by pairs or left alone.
It is thus implicit that the variables cannot be symbolic since values must correspond to indexes.
If the $i^{th}$ variable is assigned the value representing element $j$ with $i \neq j$, then the $j^{th}$ variable must be assigned the value representing element $i$.
The constraint \gb{allDifferent$\tr$symmetric} is expressed by setting the attribute \att{restriction} of \xml{list} to the value \val{symmetric}.

\begin{boxse}
\begin{semantics}
$\gb{allDifferent\tr symmetric}(X)$, with $X=\langle x_1,x_2,\ldots \rangle$, iff
  $\gb{allDifferent}(X) \land \gb{symmetric}(X)$
\end{semantics}
\end{boxse}

\begin{boxex}
\begin{xcsp} 
<allDifferent> 
  <list restriction="symmetric"> x1 x2 x3 x4 x5 </list>
</allDifferent> 
\end{xcsp}
\end{boxex}

\subsection{Constraint \gb{allDifferent$\tr$symmetric+irreflexive}}

The constraint \gb{allDifferent$\tr$symmetric} can be further restricted by forcing irreflexivity.
This is called \gb{oneFactor} in \cite{HMT_global}.
By forbidding every variable to be assigned the value representing its own index, all variables are necessarily grouped by pairs.
The constraint \gb{allDifferent$\tr$symmetric+irreflexive} is expressed by setting the attribute \att{restriction} of \xml{list} to the value \val{symmetric irreflexive}.

\begin{boxse}
\begin{semantics}
$\gb{allDifferent\tr symmetric+irreflexive}(X)$, with $X=\langle x_1,x_2,\ldots \rangle$, iff
  $\gb{allDifferent\tr symmetric}(X) \land \gb{irreflexive}(X)$
\end{semantics}
\end{boxse}

\begin{boxex}
\begin{xcsp} 
<allDifferent> 
  <list restriction="symmetric irreflexive"> X[] </list>
</allDifferent> 
\end{xcsp}
\end{boxex}

\subsection{Constraint \gb{allDifferent$\tr$convex}}

The constraint \gb{allDifferent$\tr$convex} is called \gb{alldifferent\_consecutive\_values} in the \cat.
Not only all variables must take different values, but they must also be consecutive.
The constraint \gb{allDifferent$\tr$convex} is expressed by setting the attribute \att{restriction} of \xml{list} to the value \val{convex}.

\begin{boxse}
\begin{semantics}
$\gb{allDifferent\tr convex}(X)$, with $X=\langle x_1,x_2,\ldots \rangle$, iff
  $\gb{allDifferent}(X) \land \gb{convex}(X)$
\end{semantics}
\end{boxse}

\begin{boxex}
\begin{xcsp} 
<allDifferent> 
  <list restriction="convex"> x1 x2 x3 x4 </list>
</allDifferent> 
\end{xcsp}
\end{boxex}

\subsection{Constraint \gb{nValues$\tr$increasing}}\label{ctr:nValuesIncreasing}

The constraint \gb{nValues$\tr$increasing} \cite{BHLP_increasing} combines \gb{nValues} and $\gb{ordered\st increasing}$.
It is expressed by setting the attribute \att{restriction} of \xml{list} to the value \val{increasing}.

\begin{boxse}
\begin{semantics}
$\gb{nValues\tr increasing}(X,(\odot,k))$, with $X=\langle x_1,x_2,\ldots \rangle$, iff 
  $\gb{nValues}(X,(\odot,k)) \land \gb{increasing}(X)$
\end{semantics}
\end{boxse}

In the following example, the constraint $c$ enforces the array of variables $X$ to be in increasing order and to take exactly 2 different values.

\begin{boxex}
\begin{xcsp} 
<nValues id="c">
   <list restriction="increasing"> X[] </list>
   <condition> (eq,2) </condition>
</nValues>
\end{xcsp}
\end{boxex}

\subsection{Constraint \gb{cardinality$\tr$increasing}}

The constraint \gb{cardinality$\tr$increasing}, capturing \gb{increasing\_global\_cardinality} in the \cat, is the conjunction of \gb{cardinality} and \gb{ordered} (with operator $\leq$).
It is obtained by adding \val{increasing} as restriction to the element \xml{list} of the \x3 constraint element \xml{cardinality}.

\begin{boxse}
\begin{semantics}
$\gb{cardinality\tr increasing}(X,V,O)$, with $X=\langle x_1,x_2,\ldots \rangle$, iff 
  $\gb{cardinality}(X,V,O) \land \gb{ordered}(X,\leq)$
\end{semantics}
\end{boxse}

Below, the constraint $c$ enforces the variables $x_1,x_2,x_3,x_4$ to be in increasing order and to be assigned between 1 and 3 occurrences of value 2 and between 1 and 3 occurrences of value 4.
Note that the values taken by these variables do not necessarily have to be in $\{2,4\}$ since the attribute \att{closed} is set to \val{false}.

\begin{boxex}
\begin{xcsp} 
<cardinality id="c">
  <list restriction="increasing"> x1 x2 x3 x4 </list>
  <values closed="false"> 2 4 </values>
  <occurs> 1..3 1..3 </occurs>
</cardinality>
\end{xcsp}
\end{boxex}

\subsection{Constraint \gb{permutation$\tr$increasing} (\gb{sort})}\label{ctr:sort}

The constraint \gb{permutation$\tr$increasing}, often called \gb{sort} in the literature, is the conjunction of \gb{permutation} and \gb{ordered} (with operator $\leq$).
It is obtained by adding \val{increasing} as restriction to the second element \xml{list} of the \x3 constraint element \xml{permutation}.

\begin{boxse}
\begin{semantics}
$\gb{permutation\tr increasing}(X,Y)$, with $X=\langle x_1,x_2,\ldots\rangle$ and $Y=\langle y_1,y_2,\ldots\rangle$, iff 
  $\gb{permutation}(X,Y) \land \gb{ordered}(Y,\leq)$
$\gb{permutation\tr increasing}(X,Y,M)$ iff 
  $\gb{permutation}(X,Y,M)) \land \gb{ordered}(Y,\leq)$

$\gbc{Prerequisite}: |X|=|Y|=|M| > 1$
\end{semantics}
\end{boxse}

\begin{boxex}
\begin{xcsp}  
<permutation id="c1">
   <list> x1 x2 x3 x4 </list>
   <list restriction="increasing"> y1 y2 y3 y4 </list>
</permutation>
<permutation id="c2">
   <list> v1 v2 v3 </list>
   <list restriction="increasing"> v4 v5 v6 </list>
   <mapping> m1 m2 m3 </mapping>
<permutation>
 \end{xcsp}
\end{boxex}
 
\subsection{Constraint \gb{sumCosts$\tr$allDifferent}}

The constraint \gb{sumCosts$\tr$allDifferent} \cite{FLM_cost}, also called minimumWeightAllDifferent in the literature, is the conjunction of \gb{sumCosts} and \gb{allDifferent}.
It is obtained by adding \val{allDifferent} as restriction to the element \xml{list} of the \x3 constraint element \xml{sumCosts}.

\begin{boxex}
\begin{xcsp}  
<sumCosts id="c">
   <list restriction="allDifferent"> y1 y2 y3 </list> 
   <costMatrix> 
     (10,0,5)   // costs for y1
     (0,5,0)    // costs for y2
     (5,10,0)   // costs for y3
   </costMatrix>
   <condition> (eq,z) </condition>
</sumCosts>
\end{xcsp}
\end{boxex} 
\end{xl}

\chapter{\textcolor{gray!95}{Meta-Constraints}}\label{cha:meta}

\begin{xl}
In this chapter, we present general mechanisms that can be used to combine constraints.
Sometimes in the literature, they are called meta-constraints. 
We have:
\begin{itemize}
\item sliding mechanisms over sequences of variables: \gb{slide} and \gb{seqbin}
\item logical mechanisms over constraints: \gb{and}, \gb{or}, \gb{not}, \gb{xor}, \gb{iff}, \gb{ifThen}, \gb{ifThenElse}
\end{itemize}

They are referred to by \bnfX{metaConstraint} in Chapter \ref{cha:intro}.

\begin{remark}
Note that all meta-constraints introduced in this chapter (\gb{slide}, \gb{seqbin}, \gb{and}, \gb{or}, \gb{not}, \gb{xor}, \gb{iff}, \gb{ifThen}, \gb{ifThenElse}) can be reified as they are all considered as single constraints; for reification, see Section \ref{sec:reification}.
\end{remark}
\end{xl}
\begin{xc}
In the literature, there exist general mechanisms that can be used to combine constraints.
Sometimes, they are called meta-constraints. 
In \x3-core, we only consider the sliding mechanism over sequences of variables: \gb{slide}.
\end{xc}

\section{Meta-Constraint \gb{slide}}\label{ctr:slide}

A general mechanism, or meta-constraint, that is useful to post constraints on sequences of variables is \gb{slide} \cite{BHHKW_slide}.
The scheme \gb{slide} ensures that a given constraint is enforced all along a sequence of variables.
To represent such sliding constraints in \x3, we simply build an element \xml{slide} containing a constraint template (for example, one for \xml{extension} or \xml{intension}) to indicate the abstract (parameterized) form of the constraint to be slided, preceded by an element \xml{list} that indicates the sequence of variables on which the constraint must slide.
Constraint templates are described in Section \ref{sec:group}, and possible expressions of \bnfX{constraint} are given in Appendix \ref{cha:bnf}.
The attribute \att{circular} of \xml{slide} is optional (\val{false}, by default); when set to \val {true}, the constraint is slided circularly. 
The attribute \att{offset} of \xml{list} is optional (value 1, by default); it permits, when sliding, to skip more than just one variable of the sequence, capturing \gb{slide$_j$} in \cite{BHHKW_slide}.

\begin{boxsy}
\begin{syntax} 
<slide [circular="boolean"]>
  <list [offset="integer"]> (intVar wspace)2+ </list>
  <constraint.../> @\com{constraint template, i.e., constraint involving parameters}@
</slide>
\end{syntax}
\end{boxsy}

For the semantics, we consider that $\gb{ctr}(\%0,\ldots,\%q-1)$ denotes the template of the constraint \gb{ctr} of arity $q$, and that $\gb{slide}^{circ}$ means the circular form of \gb{slide} (i.e., with \att{circular}=\val{true} in \x3).

\begin{boxse}
\begin{semantics}
$\gb{slide}(X,\gb{ctr}(\%0,\ldots,\%q-1))$, with $X=\langle x_0,x_1,\ldots \rangle$, iff  
  $\forall i : 0 \leq i \leq |X|-q, \gb{ctr}(x_i,x_{i+1},\ldots,x_{i+q-1})$ 
$\gb{slide}(X,\nm{os},\gb{ctr}(\%0,\ldots,\%q-1))$, with an offset $os$, iff  
  $\forall i : 0 \leq i \leq (|X|-q)/\nm{os}, \gb{ctr}(x_{i \times \nm{os}},x_{i \times \nm{os}+1},\ldots,x_{i \times \nm{os}+q-1})$ 
$\gb{slide}^{circ}(X,\gb{ctr}(\%0,\ldots,\%q-1))$ iff 
  $\forall i : 0 \leq i \leq |X|-q+1, \gb{ctr}(x_i,x_{i+1}\ldots,x_{(i+q-1)\%|X|})$ 
\end{semantics}
\end{boxse}

In the following example, $c_1$ is the constraint $x_1 + x_2 = x_3$  $\wedge$  $x_2 + x_3 = x_4$, $c_2$ is the circular sliding table constraint $(y_1,y_2) \in T \wedge (y_2,y_3) \in T \wedge (y_3,y_4) \in T \wedge (y_4,y_1) \in T$ with $T=\{(a,a),(a,c),(b,b),(c,a),(c,b)\}$ and $c_3$ is the sliding $\neq$ constraint $w_1 \neq z_1 \wedge w_2 \neq z_2 \wedge w_3 \neq z_3$, with offset $2$.

\begin{boxex}
\begin{xcsp}  
<slide id="c1">
  <list> x1 x2 x3 x4 </list>
  <intension> eq(add(%0,%1),%2) </intension>
</slide>
<slide id="c2" circular="true">
  <list> y1 y2 y3 y4 </list>
  <extension>
    <list> %0 %1 </list>
    <supports> (a,a)(a,c)(b,b)(c,a)(c,b) </supports>
  </extension>
</slide>
<slide id="c3">
  <list offset="2"> w1 z1 w2 z2 w3 z3 </list>
  <intension> ne(%0,%1) </intension>
</slide>
 \end{xcsp}
\end{boxex}

\begin{xl}
In some cases, it may be more practical to handle more than one sliding list.
This is the reason we can put several successive elements \xml{list}.
In that case, variables are collected from list to list in the order they are put (all variables of a list are considered before starting with the next list).
The number of variables to be collected from one list at each iteration is given by the optional attribute \att{collect} (value 1, by default).
The attribute \att{offset} can be associated independently with each list.

The general syntax is: 
\begin{boxsy}
\begin{syntax} 
<slide>
  (<list [offset="integer"] [collect="integer"]> (intVar wspace)+ </list>)2+
   <constraint.../> @\com{constraint template, i.e., constraint involving parameters}@
</slide>
\end{syntax}
\end{boxsy}

As an illustration, the following constraint $c_4$ corresponds to $x_1+y_1 = z_1 \land x_2+y_2 = z_2 \land x_3+y_3 = z_3$.

\begin{boxex}
\begin{xcsp}  
<slide id="c4">
  <list offset="2" collect="2"> x1 y1 x2 y2 x3 y3 </list>
  <list> z1 z2 z3 </list>
  <intension> eq(add(%0,%1),%2) </intension>
</slide>
 \end{xcsp}
\end{boxex}
\end{xl}

\begin{remark}
Note that \gb{slide}
\begin{itemize}
\begin{xl}\item can be relaxed/softened, obtaining then \gb{cardPath}; see Section \ref{ctr:cardpath};
\end{xl}
\item cannot be a descendant of (i.e., involved in) an element \xml{group}, \xml{slide} or \xml{seqbin}. 
\end{itemize}
\end{remark}

In \x3-core, the constraint template must be of form \gb{intension} or \gb{extension}.

\begin{xl}
\section{Some Classical Uses of \gb{slide}}

\subsection{Constraint \gb{sequence}}\label{ctr:sequence}

The constraint \gb{sequence}, see \cite{BC_chip,R_globalsurvey}, also called \gb{among\_seq} in \cite{BC_revisiting}, enforces a set of \gb{count} constraints over a sequence of variables. 
Although it can be represented by \gb{slide}, we introduce this specific constraint because it is often used. 
The arity of sliding constraints is given by the attribute \att{window}.

\begin{boxsy}
\begin{syntax} 
<sequence>
  <list window="integer"> (intVar wspace)2+ </list>
  <values> (intVal wspace)+ </values>
  <condition> "(" operator "," operand ")" </condition>
</sequence>
\end{syntax}
\end{boxsy}

For the semantics below, the arity of sliding constraints (i.e., the window size) is given by $q$, the set of values by $V$ and the condition by $(\odot,k)$ (classically, corresponding to an interval  $l..u$ of possible cumulated occurrences).

\begin{boxse}
\begin{semantics}
$\gb{sequence}(X,q,V,(\odot,k)$, with $X=\langle x_1,x_2,\ldots \rangle$, iff  
  $\forall i : 1 \leq i \leq |X|-q+1, \gb{count}(\langle x_{i+j} \in X : 0 \leq j < q \rangle,V,(\odot,k))$
\end{semantics}
\end{boxse}

The following constraint 
\begin{quote}
$\gb{sequence}(\langle x_1,x_2,x_3,x_4,x_5 \rangle,3,\{0,2,4\},(\in,0..1))$ 
\end{quote}
is then equivalent to:
\begin{quote}
$~ ~ \gb{count}(\langle x_1,x_2,x_3 \rangle,\{0,2,4\},(\in,0..1))$ \\
$\land\; \gb{count}(\langle x_2,x_3,x_4 \rangle,\{0,2,4\},(\in,0..1))$ \\
$\land\; \gb{count}(\langle x_3,x_4,x_5 \rangle,\{0,2,4\},(\in,0..1))$
\end{quote}

This gives in \x3:

\begin{boxex}
\begin{xcsp}  
<sequence id="c">
  <list window="3"> x1 x2 x3 x4 x5 </list>
  <values> 0 2 4 </values>
  <condition> (in,0..1) </condition>
</sequence>
\end{xcsp}
\end{boxex}

and with \gb{slide}:

\begin{boxex}
\begin{xcsp}  
<slide id="c">
  <list> x1 x2 x3 x4 x5 </list>
  <count>
     <list> %0 %1 %2 </list>
     <values> 0 2 4 </values>
     <condition> (in,0..1) </condition>
  </count>
</slide>
\end{xcsp}
\end{boxex}

\subsection{Constraint \gb{slidingSum}}\label{ctr:slidingSum}

The constraint \gb{slidingSum} \cite{BC_revisiting}, enforces a set of \gb{sum} constraints over a sequence of variables.

For the semantics below, the arity of sliding constraints is given by $q$, and the interval of possible values for the sum by $l..u$.

\begin{boxse}
\begin{semantics}
$\gb{slidingSum}(X,q,l..u)$, with $X=\langle x_1,x_2,\ldots \rangle$, iff  
  $\forall i : 1 \leq i \leq |X|-q+1, l \leq \sum_{j = i}^{i+q-1} \va{x}_j \leq u$
\end{semantics}
\end{boxse}

In the following example, $c$ is the sliding sum $1 \leq x_1 + x_2 +x_3 \leq 3 \wedge 1 \leq x_2 + x_3 + x_4 \leq 3 \wedge 1 \leq x_3 + x_4 +x_5 \leq 3$.

\begin{boxex}
\begin{xcsp}  
<slide id="c">
  <list> x1 x2 x3 x4 x5 </list>
  <sum>
     <list> %0 %1 %2 </list>
     <condition> (in,1..3) </condition>
  </sum>
</slide>
 \end{xcsp}
\end{boxex}

\subsection{Constraints \gb{change} and \gb{smooth}}

These two constraints are sliding constraints.
However, as they also integrate relaxation, we introduce them in Chapter \ref{cha:cost}.

\section{Meta-Constraint \gb{seqbin}}\label{ctr:seqbin}

The meta-constraint \gb{seqbin} \cite{PBL_seqbin,KNW_seqbin} ensures that a binary constraint holds down a sequence of variables, and counts how many times another binary constraint is violated.

\begin{boxsy}
\begin{syntax} 
<seqbin>
  <list> (intVar wspace)2+ </list>
  <constraint.../> @\com{template of the hard binary constraint}@
  <constraint.../> @\com{template of the soft binary constraint}@
  <@number@> intVal | intVar </@number@>
</seqbin>
\end{syntax}
\end{boxsy}

The attribute \att{violable} is required (with value \val{true}) for the soft binary constraint.

\begin{boxse}
\begin{semantics}
$\gb{seqbin}(X,\gb{ctr^{h}}(\%0,\%1),\gb{ctr^s}(\%0,\%1),z)$, with $X=\langle x_1,x_2,\ldots \rangle$, iff 
  $\forall i : 1 \leq i < |X|, \gb{ctr^h}(x_i,x_{i+1})$
  $|\{i : 1 \leq i < |X| \land \lnot \gb{ctr^s}(x_i,x_{i+1})\}| = \va{z}$
\end{semantics}
\end{boxse}

\begin{remark}
Note that in our definition, we do not add 1 to $\va{z}$ as proposed in \cite{KNW_seqbin}, and implicitly defined in \cite{PBL_seqbin}. 
We do believe that it can be easily handled by solvers.
\end{remark}

\begin{boxex}
\begin{xcsp}  
<seqbin id="c1">
  <list> x1 x2 x3 x4 x5 x6 x7 </list>
  <intension> ne(%0,%1) </intension>
  <intension violable="true"> lt(%0,%1) </intension>
  <number> y </number>
</seqbin>
<seqbin id="c2">
  <list> w1 w2 w3 w4 w5 w6 </list>
  <extension> 
    <list> %0 %1 </list>
    <supports> (0,0)(0,1)(0,3)(1,1)(1,2)(2,3)(3,0) </supports>
  </extension>
  <intension violable="true"> gt(%0,%1) </intension>
  <number> z </number>
</seqbin>
\end{xcsp}
\end{boxex}

\begin{remark}
Note that \gb{seqbin} cannot be a descendant of (i.e., involved in) an element \xml{group}, \xml{slide} or \xml{seqbin}.
\end{remark}

\section{Meta-Constraint \gb{and}}\label{ctr:and}

Sometimes, it may be interesting to combine logically constraints \cite{L_arc,L_practical}.  
The meta-constraint \gb{and} ensures the conjunction of a set of constraints (and possibly, meta-constraints) that are put together inside a same element.
This may be useful for obtaining a stronger filtering and/or dealing with reification.
Below, \bnfX{constraint} and \bnfX{metaConstraint} represent any constraint and meta-constraint introduced in \x3 (see Appendix \ref{cha:bnf} for an exhaustive list). 

\begin{boxsy}
\begin{syntax} 
<and>
   (<constraint.../> | <metaConstraint.../>)2+
</and>
\end{syntax}
\end{boxsy}

\begin{boxse}
\begin{semantics}
$\gb{and}(\gb{ctr}_1,\gb{ctr}_2,\ldots,\gb{ctr}_k)$, where $\gb{ctr}_1,\gb{ctr}_2,\ldots,\gb{ctr}_k$ are $k$ (meta-)constraints, iff 
  $\gb{ctr}_1 \land \gb{ctr}_2 \ldots \land \gb{ctr}_k$
\end{semantics}
\end{boxse}

The following example shows a constraint $c$ that represents $c_1 \land c_2$.

\begin{boxex}
\begin{xcsp}  
<and id="c">
  <intension id="c1"> eq(x,add(y,z)) </intension>
  <extension id="c2">
    <list> x z </list>
    <supports> (0,1)(1,3)(1,2)(1,3)(2,0)(2,2)(3,1) </supports>
  </extension>
</and>
\end{xcsp}
\end{boxex}

\begin{remark}
Note that \gb{and} can be relaxed/softened, obtaining then \gb{soft\ti and}; see Section \ref{ctr:softAnd}.
\end{remark}

\section{Some Classical Uses of \gb{and}}\label{sec:andUse}

Some constraints conjunction are classical.

\subsection{Constraint \gb{gen-sequence}}

A general form of \gb{sequence}, with name \gb{gen-sequence}, has been proposed in \cite{HPRS_revisiting}.
It allows specifying the precise sequences of variables on which a constraint \gb{count} must hold; so, this is no more exactly a sliding constraint.
Each such \gb{count} constraint is defined by a window (interval $p..q$, giving the indexes of the variables of the consecutive variables of the window) and a range (interval $l..u$ giving the constraining bounds for the variables of the window).

\begin{boxse}
\begin{semantics}
$\gb{gen\ti sequence}(X,W,V,R)$, with $X=\langle x_1,x_2,\ldots \rangle$, $W=\langle p_1..q_1,p_2..q_2,\ldots \rangle$, $R=\langle l_1..u_1,l_2..u_2,\ldots \rangle$, iff $\forall i : 1 \leq i \leq |W|$, $\gb{count}(\langle x_j \in X : p_i \leq j \leq q_i \rangle,V,(\in,l_i..u_i))$

@{\em Prerequisite}@: $|W| = |R|$
\end{semantics}
\end{boxse}

The constraint 
\begin{quote}
$\gb{gen\ti sequence}(\langle x_0,x_1,x_2,x_3,x_4 \rangle, \langle 0..2,2..4 \rangle,\{0,2,4\},\langle 0..1,1..2 \rangle)$ 
\end{quote}
is equivalent to:
\begin{quote}
$\gb{count}(\langle x_0,x_1,x_2 \rangle,\{0,2,4\},(\in,0..1)) \,\land\; \gb{count}(\langle x_2,x_3,x_4 \rangle,\{0,2,4\},(\in,1..2))$
\end{quote}

In \x3, we simply combine the two constraints by \gb{and}.

\begin{boxex}
\begin{xcsp}  
<and>
  <count>
     <list> x0 x1 x2 </list>
     <values> 0 2 4 </values>
     <condition> (in,0..1) </condition>
  </count>
  <count>
     <list> x2 x3 x4 </list>
     <values> 0 2 4 </values>
     <condition> (in,1..2) </condition>
  </count>
</slide>
\end{xcsp}
\end{boxex}

\subsection{Constraint \gb{gsc}}\label{ctr:gsc}

The global sequencing constraint (\gb{gsc}) \cite{RP_gsc} combines \gb{sequence} with \gb{cardinality}. 
An example is given below.

\begin{boxex}
\begin{xcsp} 
<and id ="c">
  <sequence>
    <list id="X" window="3"> x1 x2 x3 x4 x5 </list>
    <values> 0 2 4 </values>
    <condition> (in,0..1) </condition>
  </sequence>
  <cardinality>
    <list as="X" />
    <values> 0 1 </values>
    <occurs> 0..2 2..3 </occurs>
  </cardinality>
</and>
\end{xcsp}
\end{boxex}

\subsection{Constraint \gb{cardinalityWithCosts}}\label{ctr:gcccosts}

The constraint \gb{cardinalityWithCosts} \cite{R_cost} combines \gb{cardinality} with \gb{sumCosts}.
An example is given below.

\begin{boxex}
\begin{xcsp}  
<and>
   <cardinality>
      <list id="X"> x1 x2 x3 x4 x5 </list>
      <values> 0 1 2 </values>
      <occurs> y1 y2 y3 </occurs>
   </cardinality>
   <sumCosts>
      <list as="X" /> 
      <values @\violet{cost}@="5"> 0 2 </values>
      <values @\violet{cost}@="0"> default </values>
      <condition> (eq,z) </condition>
   </sumCosts>
</and>
\end{xcsp}
\end{boxex}

\subsection{Constraints \gb{costRegular} and \gb{multicostRegular}}\label{ctr:costRegular}

The constraint \gb{costRegular} \cite{DPR_cost} combines \gb{regular} with \gb{sumCosts}.
An example is given below (variables are assumed to have binary domains).

\begin{boxex}
\begin{xcsp}  
<and>
   <regular>
      <list id="B"> b1 b2 b3 b4 </list>
      <transitions> 
        (a,0,a)(a,1,b)(b,0,c)(b,1,d)(c,0,d)
      </transitions>
      <start> a </start>
      <final> d </final>
   </regular>
   <sumCosts>
      <list as="B" /> 
      <costMatrix> (10,0)(0,5)(5,2)(5,0) </costMatrix>
      <condition> (le,z) </condition>
    </sumCosts>
</and>
\end{xcsp}
\end{boxex}

It is possible, by introducing several constraints \gb{sumCosts} to represent \gb{multicostregular} \cite{MD_sequencing}.

\section{Meta-Constraint \gb{or}}\label{ctr:or}

The meta-constraint \gb{or} ensures the disjunction of a set of constraints that are put together inside a same element. 
This may be useful for modeling and/or dealing with reification.

\begin{boxsy}
\begin{syntax} 
<or>
  (<constraint.../> | <metaConstraint.../>)2+
</or>
\end{syntax}
\end{boxsy}

\begin{boxse}
\begin{semantics}
$\gb{or}(\gb{ctr}_1,\gb{ctr}_2,\ldots,\gb{ctr}_k)$, where $\gb{ctr}_1,\gb{ctr}_2,\ldots,\gb{ctr}_k$ are $k$ (meta-)constraints, iff 
  $\gb{ctr}_1 \lor \gb{ctr}_2 \ldots \lor \gb{ctr}_k$
\end{semantics}
\end{boxse}

The following example shows a constraint $c$ that represents $c_1 \lor c_2$.

\begin{boxex}
\begin{xcsp}  
<or id="c">
  <intension id="c1"> eq(x,add(y,z)) </intension>
  <extension id="c2">
    <list> x z </list>
    <supports> (0,1)(1,3)(1,2)(1,3)(2,0)(2,2)(3,1) </supports>
  </extension>
</or>
\end{xcsp}
\end{boxex}

\section{Meta-Constraint \gb{not}}\label{ctr:not}

The meta-constraint \gb{not} ensures the negation of a constraint that is put inside the element. 
This may be useful for modeling and/or dealing with reification.

\begin{boxsy}
\begin{syntax} 
<not>
  <constraint.../> | <metaConstraint.../>
</not>
\end{syntax}
\end{boxsy}

\begin{boxse}
\begin{semantics}
$\gb{not}(\gb{ctr})$, where $\gb{ctr}$ is a (meta-)constraint, iff 
  $\lnot \gb{ctr}$
\end{semantics}
\end{boxse}

The meta-constraint \gb{not} can be used to represent well-known global constraints such as \gb{notAllEqual} ensuring that at least one of the involved variables in the constraint \gb{allEqual} must differ from the other ones.\label{ctr:notAllEqual}

\begin{boxex}
\begin{xcsp}
<not>
  <allEqual> 
    x1 x2 x3 x4 x5
  </allEqual>
</not> 
\end{xcsp}
\end{boxex}

Actually, note that it is possible to represent more directly \gb{notAllEqual} with \gb{nValues}.

\section{Meta-Constraint \gb{xor}}\label{ctr:xor}

The meta-constraint \gb{xor} ensures that an odd number of constraints put inside the element is satisfied.
This may be useful for modeling and/or dealing with reification.

\begin{boxsy}
\begin{syntax} 
<xor>
  <constraint.../> | <metaConstraint.../>
</xor>
\end{syntax}
\end{boxsy}

\begin{boxse}
\begin{semantics}
$\gb{xor}(\gb{ctr}_1,\gb{ctr}_2,\ldots,\gb{ctr}_k)$, where $\gb{ctr}_1,\gb{ctr}_2,\ldots,\gb{ctr}_k$ are $k$ (meta-)constraints, iff 
  $\gb{ctr}_1 \oplus \gb{ctr}_2 \ldots \oplus \gb{ctr}_k$
\end{semantics}
\end{boxse}

\section{Meta-Constraint \gb{iff}}\label{ctr:iff}

The meta-constraint \gb{iff} ensures that all constraints put inside the element are equally satisfied or dis-satisfied.
This may be useful for modeling and/or dealing with reification.

\begin{boxsy}
\begin{syntax} 
<iff>
  <constraint.../> | <metaConstraint.../>
</iff>
\end{syntax}
\end{boxsy}

\begin{boxse}
\begin{semantics}
$\gb{iff}(\gb{ctr}_1,\gb{ctr}_2,\ldots,\gb{ctr}_k)$, where $\gb{ctr}_1,\gb{ctr}_2,\ldots,\gb{ctr}_k$ are $k$ (meta-)constraints, iff 
  $\gb{ctr}_1 = \gb{ctr}_2 \ldots = \gb{ctr}_k$
\end{semantics}
\end{boxse}

\section{Meta-Constraint \gb{ifThen}}\label{ctr:ifThen}

The meta-constraint \gb{ifThen} involves two (meta-)constraints.
If the first specified one is satisfied, the second one must be satisfied.
Below, \bnfX{constraint} and \bnfX{metaConstraint} represent any constraint and meta-constraint introduced in \x3 (see Appendix \ref{cha:bnf} for an exhaustive list). 

\begin{boxsy}
\begin{syntax} 
<ifThen>
  (<constraint.../> | <metaConstraint.../>) @\com{Condition Part}@ 
  (<constraint.../> | <metaConstraint.../>) @\com{Then Part}@ 
</ifThen>
\end{syntax}
\end{boxsy}

\begin{boxse}
\begin{semantics}
$\gb{ifThen}(\gb{ctr}_1,\gb{ctr}_2)$, where $\gb{ctr}_1$ and $\gb{ctr}_2$ are two (meta-)constraints, iff 
  $\gb{ctr}_1 \Rightarrow \gb{ctr}_2 $
\end{semantics}
\end{boxse}

The following example shows a meta-constraint \gb{ifThen} that denotes
\begin{quote}
  $x=10 \Rightarrow allDifferent(y)$
\end{quote}
where $y$ is an array of variables.

\begin{boxex}
\begin{xcsp}  
<ifThen>
  <intension> eq(x,10) </intension>
  <allDifferent> y[] </allDifferent>
</ifThen>
\end{xcsp}
\end{boxex}

\section{Meta-Constraint \gb{ifThenElse}}\label{ctr:ifThenElse}

The meta-constraint \gb{ifThenElse} involves three (meta-)constraints.
If the first specified one is satisfied, the second one must be satisfied.
Otherwise, the third one must be satisfied.

\begin{boxsy}
\begin{syntax} 
<ifThenElse>
  (<constraint.../> | <metaConstraint.../>) @\com{Condition Part}@ 
  (<constraint.../> | <metaConstraint.../>) @\com{Then Part}@
  (<constraint.../> | <metaConstraint.../>) @\com{Else Part}@ 
</ifThenElse>
\end{syntax}
\end{boxsy}

\begin{boxse}
\begin{semantics}
$\gb{ifThenElse}(\gb{ctr}_1,\gb{ctr}_2,\gb{ctr}_3)$, where $\gb{ctr}_1$, $\gb{ctr}_2$ and $\gb{ctr}_3$ are 3 (meta-)constraints, iff 
  $(\gb{ctr}_1 \land \gb{ctr}_2) \lor ( \lnot \gb{ctr}_1 \land \gb{ctr}_3)$
\end{semantics}
\end{boxse}

The following example shows a meta-constraint \gb{ifThenElse} that denotes
\begin{quote}
  $(x=10 \land allDifferent(y)) \lor (x \neq 10 \land allEqual(y))$
\end{quote}
where $y$ is an array of variables.

\begin{boxex}
\begin{xcsp}  
<ifThenElse>
  <intension> eq(x,10) </intension>
  <allDifferent> y[] </allDifferent>
  <allEqual> y[] </allEqual>
</ifThenElse>
\end{xcsp}
\end{boxex}
\end{xl}

\begin{xl}
\chapter{\textcolor{gray!95}{Soft Constraints}}\label{cha:cost}

It is sometimes useful, or even necessary, to express preferences or costs when modeling problems.
Actually, many works in the literature focus on the concept of costs that can be associated with certain instantiations of variables inside constraints.
There are two main ways of integrating costs:

\begin{itemize}
\item by {\em restricting} constraints, this way, additionally ensuring that some introduced costs are within certain limits. 
\item by {\em relaxing} constraints, this way, permitting a certain degree of violation with respect to the original forms of constraints;
\end{itemize}

In Chapters \ref{cha:lifted} and \ref{cha:meta}, it was shown how to build restricted constraints, either with the attribute \att{restriction} or with the meta-constraint \gb{and} that enables the integration of the cost-related constraint \gb{sumCosts}.

In this chapter, we focus on relaxation, and in this context, we introduce cost-based {\em soft} constraints.
There are two main ways\footnote{We shall discuss other frameworks for expressing preferences/costs in Chapter \ref{cha:frameworks}.} of handling (cost-based) soft constraints.
We can deal with hard constraints that are relaxed by integrating cost variables, obtaining so-called {\em relaxed constraints} hereafter; see \cite{RPBP_original,RPBP_meta,BP_cost}.
Or we can deal with {\em cost functions} that are also called {\em weighted} constraints in the litterature (framework WCSP); see \cite{L_node,LS_solving,MRS_soft,CGSSZW_soft}. 
After introducing relaxed constraints and cost functions in Section \ref{sec:costsyn}, we describe simple relaxation in Section \ref{sec:simpleRelaxation}.
In Sections \ref{sec:costgen} and \ref{sec:costglob}, complex relaxation for generic constraints and global constraints is presented.
Related information concerning WCSP can be found in Section \ref{sec:wcsp} of Chapter \ref{cha:frameworks}.

Although we do not introduce new constraint types (names) in \x3, in order to clarify the textual description (and, make semantics unambiguous), we shall refer in the text to a soft variant of a constraint $\gb{ctr}$ by $\gb{soft\ti ctr}$.
For example, we shall refer in the text to $\gb{soft\ti allDifferent}$ when considering a soft version of $\gb{allDifferent}$.

\section{Relaxed Constraints and Cost Functions}\label{sec:costsyn}

A soft constraint in \x3 is an XML constraint element with an attribute \att{type} set to the value \val{soft}.
A soft constraint in \x3 is either a relaxed constraint or a cost function.

\paragraph{Relaxed Constraints.}
In \x3, a relaxed constraint is a constraint explicitly integrating a cost component.
This means that the XML constraint element contains an element \xml{cost}, which is besides necessarily the last child of the constraint. 
This new element \xml{cost} can contain a numerical condition (similarly to an element \xml{condition}), which typically involves a cost variable as operand: the actual cost of an instantiation is related to the value of the cost variable with respect to the specified relational operator.  The element \xml{cost} can also simply contain an integer variable: in that case, the value of this variable represents exactly the cost of the constraint, and we have \verb!intVar! which is equivalent to \verb!"(eq," intVar ")"!.
As we shall see in the next sections, some optional attributes in mutual exclusion, called \att{violationCost}, \att{defaultCost} and \att{violationMeasure}, can also be specifically introduced. 

The syntax is as follows:
\begin{boxsy}
\begin{syntax} 
<constraint type="soft" [violationCost="integer" | 
                         defaultCost="integer" |
                         violationMeasure="measureType"]>  
  ... 
  <cost> "(" operator "," operand ")" | intVar </cost>
</constraint>
\end{syntax}
\end{boxsy}   

\begin{remark}
It is important to note that a relaxed constraint remains a hard constraint.
Either it is satisfied because the cost condition holds, or it is not.
\end{remark}

\paragraph{Cost Functions.}
In \x3, a cost function is defined similarly to a relaxed constraint, except that it does not integrate the cost component (element \xml{cost}).
A cost function is not a constraint since it returns integer values and not Boolean values.

\begin{remark}
It is possible to mix hard constraints with relaxed constraints and cost functions.
However, when an \x3 instance involves at least one cost function, the attribute \att{type} for \xml{instance} must be given the value \val{WCSP} and no objectives must be present (because the objective function is implicitly defined).     
\end{remark}

Many illustrations (especially showing the difference between the use of relaxed constraints and cost functions) are given in the next sections.

\section{Simple Relaxation}\label{sec:simpleRelaxation}

The simplest relaxation mechanism consists in associating a fixed integer cost $q$ with some constraints.
The cost of the constraint is 0 when the constraint is satisfied, and $q$ otherwise.
To deal with simple relaxation, which can be applied to any types of constraints, it suffices to introduce the attribute \att{violationCost} whose value must be an integer value.
   
As an illustration, let us consider a very basic problem: ``Lucy, Mary, and Paul want to align in one row for taking a photo. Some of them have preferences next to whom they want to stand: Lucy wants to stand at the left of Mary and Paul also wants to stand at the left of Mary.''.
Not satisfying Lucy's preference costs 3 while not satisfying Paul's preference costs 2.
The objective here is to minimize the sum of violation costs.

\paragraph{Illustration with Relaxed Constraints.}
First, let us represent this problem with relaxed constraints.
We have three variables for denoting the positions (numbered 1, 2 and 3 from left to right) of Lucy, Mary and Paul in the row.
We have a hard constraint \gb{allDifferent} and two relaxed constraints (note that we need to introduce two variables for representing the violation costs associated with the preferences expressed by Lucy and Paul).
Finally, we have an explicit objective: minimizing the sum of the violation costs.

\begin{boxex}
\begin{xcsp}
<instance format="XCSP3" type="COP">
  <variables>
    <var id="lucy"> 1..3 </var>
    <var id="mary"> 1..3 </var>
    <var id="paul"> 1..3 </var>
    <var id="z1"> 0 3 </var>
    <var id="z2"> 0 2 </var>
   </variables>
  <constraints>
    <allDifferent> lucy mary paul </allDifferent>
    <intension type="soft" violationCost="3"> 
      <function> eq(lucy,sub(mary,1)) </function>
      <cost> z1 </cost>
    </intension>
    <intension type="soft" violationCost="2"> 
      <function> eq(paul,sub(mary,1)) </function>
      <cost> z2 </cost>
    </intension>
  </constraints>
  <objectives>
    <minimize type="sum"> z1 z2 </minimize>
  </objectives>
</instance>
\end{xcsp} 
\end{boxex}
 
\paragraph{Illustration with Cost Functions.}
Now, let us represent this problem with cost functions.
We still have three variables for denoting the positions (numbered 1, 2 and 3 from left to right) of Lucy, Mary and Paul in the row.
We have a hard constraint \gb{allDifferent} and two cost functions for representing the preferences expressed by Lucy and Paul.
Note that we do not need to introduce cost variables and to define an explicit objective (WCSP is discussed with more details in Section \ref{sec:wcsp}).

\begin{boxex}
\begin{xcsp}
<instance format="XCSP3" type="WCSP">
  <variables>
    <var id="lucy"> 1..3 </var>
    <var id="mary"> 1..3 </var>
    <var id="paul"> 1..3 </var>
   </variables>
  <constraints>
    <allDifferent> lucy mary paul </allDifferent>
    <intension type="soft" violationCost="3"> 
      eq(lucy,sub(mary,1)) 
    </intension>
    <intension type="soft" violationCost="2"> 
      eq(paul,add(lucy,1)) 
    </intension>
  </constraints>
</instance>
\end{xcsp} 
\end{boxex}

\section{Complex Relaxation of Generic Constraints}\label{sec:costgen}

In this section, we show how to build complex relaxation of the generic constraint types \xml{intension} and \xml{extension}. 
This relaxation form is more general than the simple one described in Section \ref{sec:simpleRelaxation} (but dedicated to these two special constraint types).
Note that for these two constraint types, we need to modify the way these constraints are classically built: an integer function replaces a Boolean function (predicate), and elements \xml{tuple} replace elements \xml{supports} and \xml{conflicts}, respectively.

\subsection{Constraint \gb{soft\ti intension}}

When relaxing an intensional constraint, one has first to replace in \xml{function} the predicate expression (that returns either 0, standing for $\nm{false}$, or 1, standing for $\nm{true}$) by an integer functional expression (that may return any integer, and should return 0 when the original constraint is satisfied), which is referred to as \bnf{intExpr} in the syntax boxes below, and whose precise syntax is given in Appendix \ref{cha:bnf}.

\paragraph{Constraint \gb{soft\ti intension} as a Relaxed Constraint.}\label{ctr:softIntension}
We have to introduce an element \xml{cost}, which gives the following syntax:

\begin{boxsy}
\begin{syntax} 
<intension type="soft">
   <function> intExpr </function>
   <cost> "(" operator "," operand ")"  | intVar </cost>
</intension>
\end{syntax}
\end{boxsy}

Below, $F$ denotes a function (functional expression) with $r$ formal parameters (not shown here, for simplicity), $X=\langle x_1,x_2,\ldots,x_r \rangle$ a sequence of $r$ variables, and $F(\va{x}_1,\va{x}_2,\ldots,\va{x}_r)$ the value returned by function $F$ for a given instantiation of variables in $X$. The numerical cost condition is represented by $(\odot,z)$.

\begin{boxse}
\begin{semantics}
$\gb{soft\ti intension}(F,X,(\odot,z))$, with $X=\langle x_1,x_2,\ldots,x_r \rangle$, iff 
  $F(\va{x}_1,\va{x}_2,\ldots,\va{x}_r) \odot \va{z}$ 
\end{semantics}
\end{boxse}

For example, suppose that we want to use $|x - y|$ as cost for the relaxed form of constraint $x = y$, and make this value equal to a cost variable $z$.
We obtain constraint $c_1$ below.
Now, suppose that we want to use $(v - w)^2$ as cost for the relaxed form of constraint $v \leq w$. 
Note that we need to be careful about ensuring that the cost is 0 when the constraint is satisfied.
Here, the cost must be less than or equal to the value of a cost variable $u$.
We obtain constraint $c_2$.

\begin{boxex}
\begin{xcsp} 
<intension id="c1" type="soft">
   <function> dist(x,y) </function>
   <cost> z </cost>
</intension>
<intension id="c2" type="soft">
   <function> if(le(v,w),0,sqr(sub(v,w))) </function>
   <cost> (le,u) </cost>
</intension>
\end{xcsp}
\end{boxex}

\paragraph{Constraint \gb{soft\ti intension} as a Cost Function.}\label{ctr:wIntension}
The syntax is given by:

\begin{boxsy}
\begin{syntax} 
<intension type="soft">
   <function> intExpr </function>
</intension>
\end{syntax}
\end{boxsy}

The opening and closing tags of \xml{function} are optional, which gives:

\begin{boxsy}
\begin{syntax} 
<intension type="soft"> intExpr </intension> @\com{Simplified Form}@
\end{syntax}
\end{boxsy}

For example, suppose that we want to use $|x - y|$ as cost function.
We obtain constraint $c_1$ below.
Now, suppose that we want to use $(v - w)^2$ as cost function when $v > w$. 
Again, note that we need to be careful about ensuring that the cost is 0 when the constraint is satisfied.
We obtain constraint $c_2$.

\begin{boxex}
\begin{xcsp} 
<intension id="c1" type="soft">
   dist(x,y) 
</intension>
<intension id="c2" type="soft">
   if(le(v,w),0,sqr(sub(v,w))) 
</intension>
\end{xcsp}
\end{boxex}


\subsection{Constraint \gb{soft\ti extension}}

When relaxing an extensional constraint, one has first to refine the enumeration of tuples, by indicating the cost for each of them.
The elements \xml{supports} and \xml{conflicts} are then replaced by a sequence of elements \xml{tuples}, where each element \xml{tuples} has a required attribute \att{cost} that gives the common cost of all tuples contained inside the element.
It is necessary to use the attribute \att{defaultCost} of \xml{extension} to specify the cost of  all implicit tuples (i.e., those not explicitly listed). 

\paragraph{Constraint \gb{soft\ti extension} as a Relaxed Constraint.}\label{ctr:softExtension}
As usual, we have to introduce an element \xml{cost}.
This gives the following syntax for extensional constraints of arity greater than or equal to 2:

\begin{boxsy}
\begin{syntax} 
<extension type="soft" defaultCost="integer">
   <list> (intVar wspace)2+ </list>
   (<tuples cost="integer"> ("(" intVal ("," intVal)+ ")")+ </tuples>)+
   <cost> "(" operator "," operand ")" | intVar </cost> 
</extension>
\end{syntax}
\end{boxsy}

\begin{boxse}
\begin{semantics}
$\gb{soft\ti extension}(X,\ns{T},(\odot,z))$, with $X=\langle x_1,x_2,\ldots,x_r\rangle$, $\ns{T} = \langle T_1,T_2,\ldots\rangle$ where each $T_i$ is a set of tuples of cost $cost(T_i)$, iff 
  $\langle \va{x}_1,\va{x}_2,\ldots,\va{x}_r\rangle \in T_i \Rightarrow cost(T_i) \odot \va{z}$ 

@{\em Prerequisite}@: 
  $\forall i : 1 \leq i \leq |\ns{T}|, \forall \tau \in T_i, |\tau|=|X|$
  $\forall (i,j) : 1 \leq i < j \leq |\ns{T}|, T_i \cap T_j = \emptyset$
  $\forall \tau \in dom(x_1) \times dom(x_2) \times \ldots \times dom(x_r), \exists i : 1 \leq i \leq |\ns{T}| \land \tau \in T_i$
\end{semantics}
\end{boxse}

Assume that we have a relaxed extensional constraint involving three variables $x_1,x_2,x_3$, each of them with domain $\{1,2\}$.
Assume that the cost for tuples $\langle 1,2,2 \rangle$, $\langle 2,1,2 \rangle$ and $\langle 2,2,1 \rangle$ is 10, the cost for tuples $\langle 1,1,2 \rangle$ and $\langle 1,1,3 \rangle$ is 5, and the cost for all other tuples --that is, $\langle 1,1,1 \rangle$, $\langle 1,2,1 \rangle$, $\langle 2,1,1 \rangle$, $\langle 2,2,2 \rangle$-- is 0. 
If the cost of an instantiation must be given by a cost variable $z$, we write:

\begin{boxex}
\begin{xcsp}
<extension type="soft" defaultCost="0">
  <list> x1 x2 x3 </list>
  <tuples @\violet{cost}@="10"> (1,2,2)(2,1,2)(2,2,1) </tuples>
  <tuples @\violet{cost}@="5"> (1,1,2)(1,1,3) </tuples>
  <cost> z </cost>
</extension>
 \end{xcsp}
\end{boxex} 

For a unary constraint, we obtain:

\begin{boxsy}
\begin{syntax} 
<extension type="soft" defaultCost="integer">
   <list> intVar </list>
   (<tuples cost="integer"> ((intVal | intIntvl) wspace)+ </tuples>)+
   <cost> "(" operator "," operand ")" | intVar </cost> 
</extension>
\end{syntax}
\end{boxsy}


\paragraph{Constraint \gb{soft\ti extension} as a Cost Function.}\label{ctr:wExtension} 
We have the following syntax for cost functions of arity greater than or equal to 2:

\begin{boxsy}
\begin{syntax} 
<extension type="soft" defaultCost="integer">
   <list> (intVar wspace)2+ </list>
   (<tuples cost="integer"> ("(" intVal ("," intVal)+ ")")+ </tuples>)+
</extension>
\end{syntax}
\end{boxsy}

The previous example gives under the form of a cost function:

\begin{boxex}
\begin{xcsp}
<extension type="soft" defaultCost="0">
  <list> x1 x2 x3 </list>
  <tuples @\violet{cost}@="10"> (1,2,2)(2,1,2)(2,2,1) </tuples>
  <tuples @\violet{cost}@="5"> (1,1,2)(1,1,3) </tuples>
</extension>
 \end{xcsp}
\end{boxex} 


For a unary cost function, we obtain:

\begin{boxsy}
\begin{syntax} 
<extension type="soft" defaultCost="integer">
   <list> intVar </list>
   (<tuples cost="integer"> ((intVal | intIntvl) wspace)+ </tuples>)+
</extension>
\end{syntax}
\end{boxsy}

\section{Complex Relaxation of Global Constraints}\label{sec:costglob}

In this section, we show how to build complex relaxation of (some) global constraints. 
This relaxation form is more general than the simple one described in Section \ref{sec:simpleRelaxation}.
The way global constraints are defined (i.e., their intern elements or parameters) remains the same, contrary to what we have seen for \xml{intension} and \xml{extension} in Section \ref{sec:costgen}.

For global constraints, the attribute \att{violationMeasure}, when present, indicates the violation measure used to assess the cost of the constraints.
Currently, there are four pre-defined values, but in the future, other values might be introduced too.
A default violation measure is considered when this attribute is not present.

It is then possible to soften global constraints by referring to some known violation measures \cite{PRB_specific,HPR_global}.
A violation measure $\mu$ is simply a cost function that guarantees that cost 0 is associated with, and only with, any tuple that fully satisfies the constraint.
Two following violation measures are general-purpose:
\begin{itemize}
\item \val{var}: it measures the number of variables that need to change their values in order to satisfy the constraint
\item \val{dec}: it measures the number of violated constraints in the binary decomposition of the constraint
\end{itemize}

\begin{remark}
In the future, we might introduce an additional attribute \att{violationParameters} to provide some information needed by some violation measures. 
For example, the violation measure \val{var} can be refined \cite{BP_cost}  by indicating the subset of variables that are allowed to change their values, when measuring.
\end{remark}

\begin{remark}
The measure \val{dec}, although a general scheme, cannot always be used: for some global constraints, no natural binary decomposition is known.
\end{remark}

Let us say a few words about the semantics, when considering relaxed global constraints.
For the semantics, we consider that $\gb{soft\ti ctr}^{\mu}$ is the relaxed form of the global constraint $\gb{ctr}$ of scope $X$, considering the violation measure $\mu$.
We also consider that $\mu$ applied to the constraint $\gb{ctr}$ returns the cost for any given instantiation of $X$, and that $(\odot,z)$ represents the numerical cost condition.

\begin{boxse}
\begin{semantics}
$\gb{soft\ti ctr}^{\mu}(X,(\odot,z))$, with $X=\langle x_1,x_2,\ldots,x_r\rangle$, iff 
  $\mu(\gb{ctr}(\va{x}_1,\va{x}_2,\ldots,\va{x}_r)) \odot z$ 
\end{semantics}
\end{boxse}

Finally, when several constraints are at stake, the attribute \att{violationMeasure} is usually absent, implicitly referring to the number of constraints that are violated in a set.
In the next subsections, we only focus on relaxed global constraints.
For global cost functions, it suffices to remove systematically the element \xml{cost}.

\subsection{Constraint \gb{soft\ti and} (Cardinality Operator)}\label{ctr:softAnd}

The cardinality operator \cite{HD_cardinality}, which must not be confused with the cardinality constraint, connects a cost variable with a set (conjunction) of constraints: the value of the cost variable is the number of violated constraints in the set.
In \x3, we just need to relax the element \xml{and}; the violation measure is implicit: it counts the number of violated constraints.
Note that we use a violation measure, and not a satisfaction measure; it is immediate to pass from one form to the other.

\begin{boxsy}
\begin{syntax} 
<and type="soft">
   (<constraint.../> | <metaConstraint.../>)2+
   <cost> "(" operator "," operand ")" | intVar </cost> 
</and>
\end{syntax}
\end{boxsy}

\begin{boxse}
\begin{semantics}
$\gb{soft\ti and}(\gb{ctr}_1,\gb{ctr}_2,\ldots,\gb{ctr}_k,(\odot,z))$, iff  
  $|\{i : 1 \leq i \leq k \land \lnot \gb{ctr}_i\}| \odot z$
\end{semantics}
\end{boxse}

An example is given below for a conjunction of three constraints.

\begin{boxex}
\begin{xcsp}
<and type="soft">
  <intension> eq(w,add(x,y)) </intension>
  <extension>
    <list> v w </list>
    <supports> (0,1)(1,3)(1,2)(1,3)(2,0)(2,2)(3,1) </supports>
  </extension>
  <allDifferent> v w x y </allDifferent> 
  <cost> z </cost>
</and>
\end{xcsp}
\end{boxex}

\subsection{Constraint \gb{soft\ti slide} (\gb{cardPath})}\label{ctr:softSlide}\label{ctr:cardpath}

A general scheme, or meta-constraint, that is useful to post constraints on sequences of variables is \gb{cardPath} \cite{BC_revisiting}.
In \x3, we just need to relax the element \xml{slide}; the violation measure is implicit: it counts the number of violated constraints.

\begin{boxsy}
\begin{syntax} 
<slide [circular="boolean"] type="soft">
  <list [offset="integer"]> (intVar wspace)+ </list>
  <constraint.../> @\com{constraint template, i.e., constraint involving parameters}@
  <cost> "(" operator "," operand ")" | intVar </cost>
</slide>
\end{syntax}
\end{boxsy}

\begin{boxse}
\begin{semantics}
$\gb{soft\ti slide}(X,\gb{ctr}(\%1,\ldots,\%q),(\odot,z))$, with $X=\langle x_1,x_2,\ldots \rangle$ iff  
  $|\{i : 1 \leq i \leq |X|-q+1 \land \lnot \gb{ctr}(\va{x}_i,\ldots,\va{x}_{i+q-1})| \odot \va{z}$
\end{semantics}
\end{boxse}

We first illustrate \gb{soft\ti slide} with the constraint \gb{change} that constrains the number of times an inequality (or another relational operator) holds over a sequence of variables.

\begin{boxse}
\begin{semantics}
$\gb{change}(X,\odot,z)$, with $X=\langle x_1,x_2,\ldots \rangle$ and $\odot \in \{<,\leq,>,\geq,=,\neq\}$, iff  
  $|\{i : 1 \leq i < |X| \land \va{x}_i \odot \va{x}_{i+1}\}| = \va{z}$
\end{semantics}
\end{boxse}

Considering that we use a violation measure, and not a satisfaction measure, one can indirectly represent \gb{change}, as in the following example (where $z$ is the complementary value of the one given in the semantics).

\begin{boxex}
\begin{xcsp}  
<slide id="c1" type="soft">
  <list> x1 x2 x3 x4 x5 </list>
  <intension> ne(%0,%1) </intension>
  <cost> z </cost>
</slide>
 \end{xcsp}
\end{boxex}

We now illustrate \gb{soft\ti slide} with the constraint \gb{smooth} that constrains the number of times a binary constraint comparing a distance between two successive variables and a specified limit holds over a sequence of variables.

\begin{boxse}
\begin{semantics}
$\gb{smooth}(X,l,z)$, with $X=\langle x_1,x_2,\ldots \rangle$, iff  
  $|\{i : 1 \leq i < |X| \land |\va{x}_i - \va{x}_{i+1}| > l\}| = \va{z}$
\end{semantics}
\end{boxse}

Considering that we use a violation measure, and not a satisfaction measure, one can indirectly represent \gb{smooth}, as in the following example.

\begin{boxex}
\begin{xcsp}  
<slide id="c2" type="soft">
  <list> y1 y2 y3 y4 </list>
  <intension> gt(dist(%0,%1),2) </intension>
  <cost> w </cost>
</slide>
 \end{xcsp}
\end{boxex}

Finally, we illustrate \gb{soft\ti slide} with the constraint \gb{softSlidingSum} \cite{BC_revisiting}.

\begin{boxse}
\begin{semantics}
$\gb{softSlidingSum}(X,q,l..u,\nm{min}..\nm{max})$, with $X=\langle x_1,x_2,\ldots \rangle$, iff  
  $|\{ i : 1 \leq i \leq |X|-q+1 \land l \leq \sum_{j = i}^{i+q-1} \va{x}_j \leq u\}| \in \nm{min}..\nm{max}$
\end{semantics}
\end{boxse}

In the following example, $c_3$ is the \x3 form of the relaxed sliding sum $1 \leq x_1 + x_2 +x_3 \leq 3 \wedge 1 \leq x_2 + x_3 + x_4 \leq 3 \wedge 1 \leq x_3 + x_4 +x_5 \leq 3$ with the constraint that 2 or 3 constraints must be satisfied.
Note that the range is $0..1$, the complementary of $2..3$. 

\begin{boxex}
\begin{xcsp}
<slide id="c3" type="soft">
  <list> x1 x2 x3 x4 x5 </list>
  <sum> 
    <list> %0 %1 %2 </list>
    <condition> (le,3) </condition>
  </sum>
  <cost> (in,0..1) </cost>
</slide>
\end{xcsp}
\end{boxex}

\subsection{Constraint \gb{soft\ti allDifferent}}\label{ctr:softAllDifferent}

For the relaxed version \cite{PRB_specific,HK_global} of \gb{allDifferent}, we can use the two general-purpose violation measures \val{var} and \val{dec}.
We only give here the syntax for the relaxed basic variant (and without \xml{except}) of \gb{allDifferent}.

\begin{boxsy}
\begin{syntax} 
<allDifferent type="soft" violationMeasure="var|dec">
  <list> (intVar wspace)2+ </list>
  <cost> "(" operator "," operand ")" | intVar </cost>
</allDifferent> 
\end{syntax}
\end{boxsy}

For the measure \val{dec}, the semantics is:

\begin{boxse}
\begin{semantics}
$\gb{soft\ti allDifferent}^{dec}(X,(\odot,z))$, with $X=\langle x_1,x_2,\ldots\rangle$, $\odot \in \{<,\leq,>,\geq,=,\neq\}$, iff 
 $|\{(i,j) : 1 \leq i < j \leq |X| \land \va{x}_i = \va{x}_j\}| \odot \va{z}$
\end{semantics}
\end{boxse}

An example is given below:

\begin{boxex}
\begin{xcsp}
<allDifferent type="soft" violationMeasure="dec"> 
  <list> x1 x2 x3 x4 x5 </list>
  <cost> (le,z) </cost>
</allDifferent>
\end{xcsp}
\end{boxex}

\subsection{Constraint \gb{soft\ti cardinality}}\label{ctr:softCardinality}

For the soft version of \gb{cardinality}, we can use \val{var} as well as the measure \val{val} defined in \cite{HPR_global}.

\begin{boxsy}
\begin{syntax} 
<cardinality type="soft" violationMeasure="var|val">
   <list> (intVar wspace)2+ </list>
   <values [closed="boolean"]> (intVal wspace)+ | (intVar wspace)+ </values>
   <occurs> (intVal wspace)+ | (intVar wspace)+ | (intIntvl wspace)+ </occurs>
   <cost> "(" operator "," operand ")" | intVar </cost>
</cardinality>
\end{syntax}
\end{boxsy}

\subsection{Constraint \gb{soft\ti regular}}\label{ctr:softRegular}

For the relaxed version of \gb{regular}, we can use \val{var} as well as the measure \val{edit} defined in  \cite{HPR_global}.

\begin{boxsy}
\begin{syntax} 
<regular type="soft" violationMeasure="var|edit"> 
  <list> (intVar wspace)+ </list>
  <transitions> ("(" state "," intVal "," state ")")+ </transitions>
  <start> state </start>
  <final> (state wspace)+ </final>
  <cost> "(" operator "," operand ")" | intVar </cost>
</regular>
\end{syntax}
\end{boxsy}

\subsection{Constraint \gb{soft\ti permutation} (\gb{same})}\label{ctr:softPermutation}

For the relaxed version of \gb{permutation} (\gb{same}), we can use \val{var} \cite{HPR_global}, while discarding the optional element \xml{mapping} of the hard version.

\begin{boxsy}
\begin{syntax} 
<permutation type="soft" violationMeasure="var">
  <list> (intVar wspace)2+ </list>
  <list [startIndex="integer"]> (intVar wspace)2+ </list>
  <cost> "(" operator "," operand ")" | intVar </cost>
</permutation>
\end{syntax}
\end{boxsy}

\section{Summary}

Soft (cost-based) constraints in \x3 can be managed by means of either relaxed constraints or cost functions.
On the one hand, it is important to note that relaxed constraints must be understood and treated as hard constraints.
Simply, they (typically) involve cost variables which can possibly (i.e., not necessarily) be used for optimization.  
On the other hand, cost functions necessarily imply an implicit optimization task that consists in miminiming the sum of the constraint costs.
Currently, in \x3, it is not possible to have both a cost function and an explicit objective function (although extensions in the future can be envisioned).
Importantly, when a cost function is present, the type of the instance is necessarily \val{WCSP}; see Section \ref{sec:wcsp}.

Finally, a soft constraint is easily identifiable: its attribute \att{type} is given the value \val{soft}.
It is also easy to determine whether we have a relaxed constraint or a cost function: just check the presence of an element \xml{cost} (as last child).
A simple relaxation will necessarily involve the atribute \att{violationCost}.
\end{xl}

\part{Groups, Frameworks and Annotations}

\begin{xl}
\chapter{\textcolor{gray!95}{Groups, Blocks, Reification, Views and Aliases}}\label{cha:groups}
\end{xl}
\begin{xc}
\chapter{\textcolor{gray!95}{Groups, Blocks, Views and Aliases}}\label{cha:groups}
\end{xc}

Several important features of \x3 are introduced in this chapter.
In particular, groups of constraints are an essential mechanism to preserve the structure of the problem instances.

\section{Constraint Templates and Groups}\label{sec:group}

A constraint template is a kind of constraint abstraction, that is to say, an element representing a constraint in \x3 where 
some formal parameters are present (typically, for representing missing values and/or variables).
Such parameters are denoted by the symbol $\%$ followed by a parameter index. 
A constraint template has $p \geq 1$ parameters(s), with indices going from $0$ to $p-1$, and of course, a parameter can appear several times.

For example, here is a constraint template, with three parameters, for the element \gb{intension}:

\begin{boxex}
\begin{xcsp} 
<intension> eq(add(%0,%1),%2) </intension>
\end{xcsp}
\end{boxex}

and another one, with two parameters, for the element \gb{extension}:

\begin{boxex}
\begin{xcsp} 
<extension>
  <list> %0 %1 </list>
  <supports> (1,2)(2,1)(2,3)(3,1)(3,2)(3,3) </supports>
</extension>
\end{xcsp}
\end{boxex}

A constraint template must be used in a context that permits to provide the actual parameters, or arguments.
A first possibility is within constraints \gb{slide}\begin{xl} and \gb{seqbin}\end{xl}, where the arguments are automatically given by the variables of a sequence (sliding effect).
A second possibility is to build an element \xml{group} whose role is to encapsulate a constraint template followed by a sequence of elements \xml{args}.
\begin{xl}Currently, the constraint template \bnfX{constraint} is either put directly inside \xml{group}, or put indirectly inside \xml{group} through the meta-constraint \xml{not}\footnote{We might extend the possibilities of building groups, in the future.}.
\end{xl}
Each element \xml{args} must contain as many arguments as the number of parameters in the constraint template (using whitespace as separator). 
Of course, the first argument in \xml{args} corresponds to $\%0$, the second one to $\%1$, and so on.

\begin{remark}
Do note that a constraint template can only be found in elements \xml{slide}\begin{xl}, \xml{seqbin}\end{xl} and \xml{group}, and that no two such elements can be related by an ancestor-descendant relationship.
\end{remark}   

Considering constraints over integer variables, an argument can be any integer functional expression, referred to as \bnf{intExpr} in the syntax box below (its precise syntax is given in Appendix \ref{cha:bnf}).
When parsing, a solver has to replace the formal parameters of the predicate template by the corresponding arguments to build the constraint.

As a formal parameter, in the context of a group, it is also possible to use $\%...$ that stands for a variable number of arguments. 
If $\%i$ is the formal parameter with the highest index $i$ present in the template, then $\%...$ will be replaced by the sequence (whitespace as separator) of arguments that come after the one associated with $\%i$. 
If $\%...$ is the only formal parameter present in the template, then $\%...$ will be replaced by the full sequence of arguments.

\begin{remark}
It is currently forbidden to use $\%...$ inside elements \xml{slide}\begin{xl} and \xml{seqbin}\end{xl}.
Also, note that  $\%...$ can never occur in a functional expression (for example, \verb!add(x,%...)! is clearly invalid)..  
\end{remark} 

The syntax for the element \xml{group}, admitting an optional attribute \att{id}, is: 

\begin{xl}
\begin{boxsy}
\begin{syntax}  
<group [id="identifier"]>
   <not> <constraint.../> </not> | <constraint.../> @\com{constraint template}@ 
   (<args> (intExpr wspace)+ </args>)2+
</group>
\end{syntax}
\end{boxsy} 
\end{xl}
\begin{xc}
\begin{boxsy}
\begin{syntax}  
<group [id="identifier"]>
   <constraint.../> @\com{constraint template}@ 
   (<args> (intExpr wspace)+ </args>)2+
</group>
\end{syntax}
\end{boxsy} 
\end{xc}

To summarize, an element \xml{group} defines a group of constraints sharing the same constraint template. 
This is equivalent to posting as many constraints as the number of elements \xml{args} inside \xml{group}.
When the attribute \att{id} is specified for a group, we can consider that the id of each constraint is given by the id of the group followed by [i] where $i$ denotes the position (starting at 0) of the constraint inside the group.
At this point, it should be clear that (syntactic) groups are useful.
First, they permit to partially preserve the structure of the problem.
Second, there is no need to parse several times the constraint template.
Third, the representation is made more compact.

Let us illustrate this important concept of syntactic groups of constraints.
The following group of constraints is equivalent to post:
\begin{itemize}
\item g[0]: $x_0+x_1=x_2$
\item g[1]: $x_3+x_4=x_5$
\item g[2]: $x_6+x_7=x_8$
\end{itemize}

\begin{boxex}
\begin{xcsp} 
<group id="g">
  <intension> eq(add(%0,%1),%2) </intension>
  <args> x0 x1 x2 </args> 
  <args> x3 x4 x5 </args> 
  <args> x6 x7 x8 </args>
</group> 
\end{xcsp}
\end{boxex}

With $T=\{(1,2),(2,1),(2,3),(3,1),(3,2),(3,3)\}$, the following group of constraints is equivalent to post:
\begin{itemize}
\item h[0]: $(w,x) \in T$ 
\item h[1]: $(w,z) \in T$ 
\item h[2]: $(x,y) \in T$ 
\end{itemize}

\begin{boxex}
\begin{xcsp} 
<group id="h">  
  <extension>
    <list> %0 %1 </list>
    <supports> (1,2)(2,1)(2,3)(3,1)(3,2)(3,3) </supports>
  </extension>
  <args> w x </args>
  <args> w z </args>
  <args> x y </args>
</group>
\end{xcsp}
\end{boxex}


Now, we give the \x3 formulation for the 3-order instance of the Latin Square problem (fill an $n \times n$ array with $n$ different symbols, such that each symbol occurs exactly once in each row and exactly once in each column).

\begin{boxex}
\begin{xcsp} 
<instance format="XCSP3" type="CSP">
  <variables>
    <array id="x" size="[3][3]"> 1..3 </array>
  </variables>
  <constraints>
    <group>  
      <allDifferent> %0 %1 %2 </allDifferent>
      <args> x[0][0] x[0][1] x[0][2] </args>
      <args> x[1][0] x[1][1] x[1][2] </args>
      <args> x[2][0] x[2][1] x[2][2] </args>
      <args> x[0][0] x[1][0] x[2][0] </args>
      <args> x[0][1] x[1][1] x[2][1] </args>
      <args> x[0][2] x[1][2] x[2][2] </args>
    </group>
  </constraints>
</instance>
\end{xcsp}
\end{boxex} 

Note that we can use shorthands for lists of variables taken from arrays, and also the formal parameter $\%...$, as illustrated by:

\begin{boxex}
\begin{xcsp} 
<instance format="XCSP3" type="CSP">
  <variables>
    <array id="x" size="[3][3]"> 1..3 </array>
  </variables>
  <constraints>
    <group>  
      <allDifferent> %... </allDifferent>
      <args> x[0][] </args>  
      <args> x[1][] </args>  
      <args> x[2][] </args>
      <args> x[][0] </args>
      <args> x[][1] </args>
      <args> x[][2] </args>
    </group>
  </constraints>
</instance>
\end{xcsp}
\end{boxex}

Of course, in this very special context, a group is not very useful as it is possible to use the constraint \gb{allDifferent$\ti$matrix}, leading to:

\begin{boxex}
\begin{xcsp} 
<instance format="XCSP3" type="CSP">
  <variables>
    <array id="x" size="[3][3]"> 1..3 </array>
  </variables>
  <constraints>
    <allDifferent>  
      <matrix> x[][] </matrix>
    </allDifferent>
  </constraints>
</instance>
\end{xcsp}
\end{boxex} 

As a last illustration, we give the \x3 formulation for the 3-order instance of the Magic Square problem.
Note that, we cannot replace the element \xml{group} by a more compact representation.

\begin{boxex}
\begin{xcsp} 
<instance format="XCSP3" type="CSP">
  <variables>
    <array id="x" size="[3][3]"> 1..9 </array>
  </variables>
  <constraints>
    <allDifferent> x[][] </allDifferent>
    <group>
      <sum>
        <list> %... </list>
        <condition> (eq,15) </condition>
      </sum>
      <args> x[0][] </args>
      <args> x[1][] </args>
      <args> x[2][] </args>
      <args> x[][0] </args>
      <args> x[][1] </args>
      <args> x[][2] </args>
      <args> x[0][0] x[1][1] x[2][2] </args>
      <args> x[2][0] x[1][1] x[0][2] </args>
    </group> 
  </constraints>
</instance>
\end{xcsp}
\end{boxex}

\begin{remark}
Note that an element \xml{group}
\begin{itemize}
\begin{xl}\item cannot be reified, as it is considered as a set of constraints, and not a single one;
\item cannot be relaxed/softened, for the same reason;
\end{xl}
\item can only be a child of \xml{constraints} or a child of an element \xml{block}; hence, it cannot be a descendant of (i.e., involved in) an element \xml{group}\begin{xl}, \xml{seqbin}\end{xl} or \xml{slide}.
\end{itemize}
\end{remark}

\section{Blocks and Classes}\label{sec:blocks}

We have just seen that we can declare syntactic group of constraints, i.e. groups of constraints built from the same template.
But it may be interesting to identify blocks of constraints that are linked together semantically.
For example, in a problem model, one may declare a set of constraints that corresponds to clues (typically, for a game), a set of symmetry breaking constraints, a set of constraints related to week-ends when scheduling the shifts of nurses, and so on.
This kind of information can be useful for users and solvers.
Sometimes, one might also emphasize that there exist some links between variables and constraints.

In \x3, there are two complementary ways of managing semantic groups of constraints (and variables): blocks and classes.
A block is represented by an XML element whereas a class is represented by an XML attribute.
To declare a block of constraints, you just need to introduce an element \xml{block}, with an optional attribute \att{id}.
Each block may contain several constraints, meta-constraints, groups of constraints and intern blocks.
Most of the times, a block will be tagged by one or more classes (a class is simply an identifier), just by introducing the attribute \att{class} as in HTML.

\begin{boxsy}
\begin{syntax}  
<block [id="identifier"] [class="(identifier wspace)+"]>
   (<constraint.../> | <metaConstraint.../> | <group.../> | <block.../>)+ 
</block>
\end{syntax}
\end{boxsy} 

Predefined classes are:
\begin{itemize}
\item \val{clues}, used for identifying clues or hints usually given for a game,
\item \val{symmetry-breaking}, used for identifying elements that are introduced for breaking some symmetries,
\item \val{redundant-constraints}, used for identifying redundant (implied) constraints,
\item \val{nogoods}, used for identifying elements related to nogood recording
\end{itemize}
Other predefined values might be proposed later.
It is also possible to introduce user-defined classes (i.e., any arbitrary identifier), making this approach very flexible.

An an illustration, we give the skeleton of an element \xml{constraints} that contains several blocks.
A first block contains the constraints corresponding to some clues (for example, the initial values of a Sudoku grid).
A second block introduces some symmetry breaking constraints (\gb{lex}).
A third block introduces some redundant constraints.
And finally, the two last blocks refer to constraints related to the management of two different weeks; by introducing blocks here, 
we can associate a note (short comment) with them. 

\begin{boxex}
\begin{xcsp}
<constraints>
  <block class="clues">
    <intension> ... </intension> 
    <intension> ... </intension>
    ...
  </block>
  <block class="symmetry-breaking">
    <lex> ... </lex> 
    <lex> ... </lex>
    ...
  </block>
  <block class="redundant-constraints"> ... </block>
  <block note="Management of first week"> ... </block>
  <block note="Management of second week"> ... </block>
</constraints>
\end{xcsp}
\end{boxex}

\begin{remark}
The class \val{redundant-constraints} permits to identify implied constraints that are usually posted in order to improve the solving process. Because they are properly identified (by means of the attribute \att{class}), a solver can be easily asked to discard them, so as to compare its behavior when solving an instance with and without the redundant constraints.  Note that this facility can be used with any value of \att{class}.
\end{remark}

\begin{xl}
\begin{remark}
The attribute \att{type} of \xml{block}, introduced in previous specifications, is deprecated.
\end{remark}
\end{xl}

\begin{xl}
\section{Reification}\label{sec:reification}

Reification of a (hard) constraint $c$ means associating a 0/1 variable $x_c$ with $c$ such that $x_c$ denotes the truth value of $c$.
Reification can be stated as: 
\begin{quote}
$x_c \Leftrightarrow c$
\end{quote}

Reification of $c$ through $x_c$ means that:
\begin{itemize}
\item $c$ must be true if $x_c$ is set to 1
\item $\lnot c$ must be true if $x_c$ is set to 0
\item $x_c$ must be set to 1 if $c$ becomes entailed
\item $x_c$ must be set to 0 if $c$ becomes disentailed 
\end{itemize}

Half reification uses implication instead of equivalence. 
Half reification can be stated as: 
\begin{quote}
$x_c \Rightarrow c$
\end{quote}

Half reification of $c$ orientated from $x_c$ means that:
\begin{itemize}
\item $c$ must be true if $x_c$ is set to 1
\item $x_c$ must be set to 0 if $c$ becomes disentailed 
\end{itemize}

Half reification can also be stated as: 
\begin{quote}
$x_c \Leftarrow c$
\end{quote}

Half reification of $c$ orientated towards $x_c$ means that:
\begin{itemize}
\item $\lnot c$ must be true if $x_c$ is set to 0
\item $x_c$ must be set to 1 if $c$ becomes entailed
\end{itemize}

In \x3, to reify a constraint, you just need to specify an attribute \att{reifiedBy} whose value is the id of a 0/1 variable, to be associated with the \x3 element defining this constraint.
For half reification, we use either the attribute \att{hreifiedFrom} (for $x_c \Rightarrow c$) or the attribute \att{hreifiedTo} (for $x_c \Leftarrow c$).

As an example, taken from \cite{FSS_half}, let us suppose that you need to express the following constraint:
\begin{quote}
$x \leq 4 \Rightarrow t[x] \times y \geq 6$
\end{quote}
which requires that if $x \leq 4$ then the value at the $xth$ position of array $t$ (assumed here to be $[2,5,3,1,4]$) multiplied by $y$ must be at least 6.
By introducing variables $b_1$, $b_2$ for reification and variable $z$ for handling array indexing, we obtain:

\begin{quote}
$c_1$: $b_1 \Leftrightarrow x \leq 4$ \\
$c_2$: $\gb{Element}(t,x,z)$ \\
$c_3$: $b_2 \Leftrightarrow z \times x \geq 6$ \\
$c_4$: $ b_1 \Rightarrow b_2$
\end{quote}

This gives in \x3 form:

\begin{boxex}
\begin{xcsp}
<instance format="XCSP3" type="CSP">
  <variables>
    <var id="x"> 0..10 </var>
    <var id="y"> 0..10 </var>
    <var id="z"> 1..5 </var>
    <var id="b1"> 0 1 </var>
    <var id="b2"> 0 1 </var>
  </variables>   
  <constraints>
    <intension id="c1" reifiedBy="b1"> le(x,4) </intension>
    <element id="c2">
       <list> 2 5 3 1 4 </list>
       <index> x </index>
       <value> z </value>
    </element>
    <intension id="c3" reifiedBy="b2"> ge(mul(z,x),6) </intension>
    <intension id="c4"> imp(b1,b2) </intension>
  </constraints>
</instance>
\end{xcsp}
\end{boxex}

Another example taken from \cite{FSS_half} is:
\begin{quote}
$x > 4 \vee \gb{allDifferent}(x,y,y-x)$
\end{quote}

With reification, we should have a propagator for the negation of \gb{allDifferent}.
With half reification, we obtain:

\begin{quote}
$c_1$: $b_1 \Rightarrow x > 4$ \\
$c_2$: $z = y-x$ \\
$c_3$: $b_2 \Rightarrow \gb{allDifferent}(x,y,z)$ \\
$c_4$: $ b_1 \vee b_2$
\end{quote}

This gives in \x3 form:

\begin{boxex}
\begin{xcsp}
<instance format="XCSP3" type="CSP">
  <variables>
    <var id="x"> 0..10 </var>
    <var id="y"> 0..10 </var> 
    <var id="z"> -10..10 </var>
    <var id="b1"> 0 1 </var>
    <var id="b2"> 0 1 </var>
  </variables>   
  <constraints>
    <intension id="c1" hreifiedFrom="b1"> gt(x,4) </intension>
    <intension id="c2"> eq(z,sub(y,x)) </intension>
    <allDifferent id="c3" hreifiedFrom="b2"> x y z </allDifferent>
    <intension id="c4"> @or@(b1,b2) </intension>
  </constraints>
</instance>
\end{xcsp}
\end{boxex}

Note that it is also possible to post a constraint \gb{allDifferent} containing \verb;x y sub(y,x);, as shown in the next section.
\end{xl}

\section{Generalized Forms (Views)}\label{sec:views}

\x3 allows the user to express generalized forms of constraints (and objectives), typically by permitting the use of expressions where a variable is usually expected.
Some constraint solvers are able to cope directly with such formulations through the concept of views \cite{ST_views}.
A variable view is a kind of adaptor that performs some transformations when accessing the variable it abstracts over.
For example, suppose that you have a filtering algorithm (propagator) for the constraint \texttt{allDifferent}.
Can you directly deal with $\texttt{allDifferent}(x_1+1,x_2+2,x_3+3)$, or must you introduce intermediate variables?
In \x3, we let the possibility of generating instances that are ``view-compatible''.
As a classical illustration, let us consider the 8-queens problem instance.
We just need 8 variables $x_0$, $x_1$, \ldots, $x_7$ and ensure that the values of all $x_i$ variables, the values of all $x_i -i$, and the values of all $x_i + i$ must be pairwise different.
This leads to a \x3 instance composed of an array of 8 variables and three  $\texttt{allDifferent}$ constraints.

\begin{boxex}
\begin{xcsp}
<instance format="XCSP3" type="CSP">
  <variables>
    <array id="x" size="[8]"> 1..8 </array>
  </variables>   
  <constraints>
    <allDifferent id="rows">
      x[0] x[1] x[2] x[3] x[4] x[5] x[6] x[7] 
    </allDifferent>
    <allDifferent id="diag1">
      add(x[0],0) add(x[1],1) add(x[2],2) add(x[3],3) 
      add(x[4],4) add(x[5],5) add(x[6],6) add(x[7],7)
    </allDifferent>
    <allDifferent id="diag2">
      sub(x[0],0) sub(x[1],1) sub(x[2],2) sub(x[3],3) 
      sub(x[4],4) sub(x[5],5) sub(x[6],6) sub(x[7],7)
    </allDifferent>
  </constraints>
</instance>
\end{xcsp}
\end{boxex}


\section{Aliases (attribute \att{as})}\label{sec:as}

In some cases, it is not possible to avoid some redundancy (similar contents of elements), even when using the mechanisms described above.
This is the reason why we have a mechanism of aliases.
It is implemented by an attribute \att{as} that can be used by any element, anywhere in the document, to refer to the content of another element of the document.
The semantics is the following: the content of an element \xml{elt} with attribute  \att{as} set to value \val{idOther} is defined as being the content of the element \xml{eltOther} in the document with attribute \att{id} set to value \val{idOther}. 
There are a few restrictions:
\begin{itemize}
\item the element  \xml{elt} must not contain anything of its own, 
\item the element  \xml{eltOther} must precede \xml{elt} in the document,
\item the element  \xml{eltOther} cannot contain an element equipped with the attribute \att{id}.
\item the element  \xml{eltOther} cannot be specified an attribute \att{as} (no allowed transitivity)
\end{itemize}

Let us illustrate this.
Suppose that an instance must contain variables $x_0, x_1,x_2,x_3$, with domain $\{a,b,c\}$ for $x_0$ and $x_2$, and domain $\{a,b,c,d,e\}$ for $x_1$ and $x_3$.
If for some reasons, you want to preserve the names of the variables, you have to write:

\begin{boxex}
\begin{xcsp}
<variables>
  <var id="x0"> a b c </var>
  <var id="x1"> a b c d e </var>
  <var id="x2"> a b c </var>
  <var id="x3"> a b c d e </var>
</variables>
\end{xcsp}
\end{boxex}

By using the attribute \att{as}, you obtain the following equivalent non-redundant form:

\begin{boxex}
\begin{xcsp}
<variables>
  <var id="x0"> a b c </var>
  <var id="x1"> a b c d e </var>
  <var id="x2" as="x0" />
  <var id="x3" as="x1" /> 
</variables>
\end{xcsp}
\end{boxex}

\begin{xl}
Of course, this can be applied to any kind of elements.
For example, if for some reason, you describe twice the same set of tuples in the document, like for example :

\begin{boxex}
\begin{xcsp}
<constraints>
  ...
  <supports> (a,b)(a,c)(b,a)(b,c)(c,a)(c,c) </supports>
  ... 
  <conflicts> (a,b)(a,c)(b,a)(b,c)(c,a)(c,c) </conflicts>
  ...
</constraints>
\end{xcsp}
\end{boxex}

then, you can simply write (note that we have introduced here the attribute \att{id} that is optional for certain elements):

\begin{boxex}
\begin{xcsp}
<constraints>
  ...
  <supports id="tps"> (a,b)(a,c)(b,a)(b,c)(c,a)(c,c) </supports>
  ... 
  <conflicts as="tps" /> 
  ...
</constraints>
\end{xcsp}
\end{boxex}
\end{xl}
\begin{xc}
In \x3-core, it is only authorized to use aliases with elements \xml{var} and \xml{array}.
\end{xc}

\chapter{\textcolor{gray!95}{Frameworks}}\label{cha:frameworks}

\begin{xl}
  In this chapter, we show how to define instances in various CP frameworks.
\end{xl}
\begin{xc}
In this chapter, we introduce the two frameworks of \x3-core: CSP and COP.
\end{xc}

\section{Dealing with Satisfaction (CSP)}

A discrete  {\em Constraint Network} (CN) $P$ is a pair $(\mathscr X, \mathscr C)$ where $\mathscr X$ denotes a finite set of variables and $\mathscr C$ denotes a finite set of constraints.
A CN is also called a CSP instance.

To define a CSP instance, in \x3, you have to:
\begin{itemize}
\item set the attribute \att{type} of \xml{instance} to \val{CSP};
\item enumerate variables (at least one) within \xml{variables}; 
\item enumerate constraints within \xml{constraints};
\end{itemize}


The syntax is as follows:

\begin{boxsy}
\begin{syntax} 
<instance format="XCSP3" type="CSP">
  <variables>
    (<var.../> | <array.../>)+ 
  </variables>
  <constraints>
    (<constraint.../> | <metaConstraint.../> | <group.../> | <block.../>)* 
  </constraints>
  [<annotations.../>]
</instance>
\end{syntax}
\end{boxsy}

As an illustration, here is a way to represent the instance \texttt{langford-2-04}; see \href{https://www.csplib.org/Problems/prob024}{CSPLib}.

\begin{boxex}
\begin{xcsp}
<instance format="XCSP3" type="CSP">
  <variables>
    <array id="x" size="[2][4]"> 0..7 </array>
  </variables>
  <constraints>
    <allDifferent> x[][] </allDifferent>
    <group>
      <intension> eq(%0,add(%1,%2)) </intension>
      <args> x[1][0] x[0][0] 2 </args>
      <args> x[1][1] x[0][1] 3 </args>
      <args> x[1][2] x[0][2] 4 </args>
      <args> x[1][3] x[0][3] 5 </args>
    </group>
  </constraints>
</instance>
\end{xcsp}
\end{boxex}

\section{Dealing with Optimization (COP)}

A COP instance is defined by a set of variables $\mathscr X$, a set of constraints $\mathscr C$, as for a CN, together with a set of objective functions $\mathscr O$.
Mono-objective optimization is when only one objective function is present in $\mathscr O$. Otherwise, this is multi-objective optimization. 

To define a COP instance, in \x3, you have to:
\begin{itemize}
\item set the attribute \att{type} of \xml{instance} to \val{COP}; 
\item enumerate variables (at least one) within \xml{variables};
\item enumerate constraints (if any) within \xml{constraints};
\item enumerate objectives (at least one) within \xml{objectives}.
\end{itemize}

The syntax is as follows:

\begin{boxsy}
\begin{syntax} 
<instance format="XCSP3" type="COP">
  <variables>
    (<var.../> | <array.../>)+ 
  </variables>
  <constraints>
    (<constraint.../> | <metaConstraint.../> | <group.../> | <block.../>)* 
  </constraints>
  <objectives [combination="combinationType"]> 
    (<minimize.../> | <maximize.../>)+ 
  </objectives>
  [<annotations.../>]
</instance>
\end{syntax}
\end{boxsy}

As an illustration, let us consider the Coins problem: what is the minimum number of coins that allows one to pay exactly any price $p$ smaller than one euro \cite{A_principles}.
Here, we consider the instance of this problem for $p=83$.

\begin{boxex}
\begin{xcsp}
<instance format="XCSP3" type="COP">
  <variables>
    <var id="c1"> 1..100 </var>
    <var id="c2"> 1..50 </var>
    <var id="c5"> 1..20 </var>
    <var id="c10"> 1..10 </var>
    <var id="c20"> 1..5 </var>
    <var id="c50"> 1..2 </var>
  </variables>
  <constraints>
    <sum>
      <list> c1 c2 c5 c10 c20 c50 </list>
      <coeffs> 1 2 5 10 20 50 </coeffs>
      <condition> (eq,83) </condition>
    </sum>
  </constraints>
  <objectives>
    <minimize type="sum"> c1 c2 c5 c10 c20 c50 </minimize>
  </objectives>
</instance>
\end{xcsp}
\end{boxex}

In \x3-core, only mono-objective optimization is authorized.

\begin{remark}
MaxCSP is an optimization problem, which consists in satisfying the maximum number of constraints of a given CN.
Typically, you may be interested in that problem, when CSP instances are over-constrained, i.e., without any solution.
In the format, we do not need to refer to MaxCSP: simply feed your solver with (unsatisfiable) CSP instances, and activate MaxCSP solving.
\end{remark}

\begin{xl}
\section{Dealing with Preferences through Soft Constraints} 

The classical CSP framework can be extended by introducing valuations to be associated with constraint tuples \cite{BMRSVF_semiring}, making it possible to express preferences.
We show below how two main specializations of the Valued Constraint Satisfaction Problem (VCSP) are represented in \x3. 

\subsection{WCSP}\label{sec:wcsp}

WCSP (Weighted CSP) is an extension to CSP that relies on a specific valuation structure $S(k) = ([0,\ldots,k],\oplus,\geq)$ where:
\begin{itemize}
\item $k \in [1,\ldots,\infty]$ is either a strictly positive natural or infinity,
\item $[0,1,\ldots,k]$ is the set of naturals less than or equal to $k$,
\item $\oplus$ is the sum over the valuation structure defined as: $a \oplus b = min\{k,a+b\}$,
\item $\geq$ is the standard  order among naturals. 
\end{itemize}

A Weighted Constraint Network (WCN), or WCSP instance, is defined by a valuation structure $S(k)$, a set of variables, and a set of weighted constraints, also called cost functions (see Chapter \ref{cha:cost}).
A WCN is also known as a CFN (Cost Function Network).

Some examples of cost functions, taken from Chapter \ref{cha:cost} are given below:
 
\begin{boxex}
\begin{xcsp}
<extension type="soft" defaultCost="0">
  <list> x1 x2 x3 </list>
  <tuples @\violet{cost}@="10"> (1,2,2)(2,1,2)(2,2,1) </tuples>
  <tuples @\violet{cost}@="5"> (1,1,2)(1,1,3) </tuples>
</extension>
<intension type="soft" violationCost="3"> 
      eq(lucy,sub(mary,1)) 
</intension>
<intension type="soft"> dist(x,y) </intension>
<allDifferent type="soft" violationMeasure="dec"> 
  x1 x2 x3 x4 x5 
</allDifferent>
\end{xcsp}
\end{boxex} 

\begin{remark} 
When representing a WCSP instance, it is possible to refer to hard constraints.
For such constraints, an allowed tuple has a cost of $0$ while a disallowed tuple has a cost equal to the value of $k$.
\end{remark}


To define a CFN (WCSP instance), in \x3, you have to;
\begin{itemize}
\item set the attribute \att{type} of \xml{instance} to \val{WCSP}, 
\item enumerate variables (at least one) inside \xml{variables};
\item optionally set a value to the optional attribute \att{ub} of the element \xml{constraints}; this value represents $k$ in the WCSP valuation structure and must be an integer greater than or equal to 1, or the special value \val{+infinity}; the default value for \att{ub} is \val{+infinity};
\item optionally set a value to the optional attribute \att{lb} of the element \xml{constraints}. If present, this value must be an integer greater than or equal to 0, and less than or equal to the value of \att{ub}. It represents a constant cost that must be added to the other costs, sometimes called the $0$-ary constraint of the WCSP framework (e.g. see \cite{LS_quest}). Of course, if it is not present, it is assumed to be equal to $0$;
\item enumerate hard constraints, relaxed constraints and cost functions (at least, one) inside \xml{constraints};
\end{itemize}

The syntax is as follows:

\begin{boxsy}
\begin{syntax} 
<instance format="XCSP3" type="WCSP">
  <variables>
    (<var.../> | <array.../>)+ 
  </variables>
  <constraints [lb="integer"] [ub="integer|+infinity"]>
    (<constraint .../> | 
     <constraint type="soft".../> | 
     <group.../> | 
     <block.../>)* 
  </constraints>
  [<annotations.../>]
</instance>
\end{syntax}
\end{boxsy}

In the simple example below, we have a CFN with two variables, two unary extensional constraints and one binary extensional constraint.
Note that the forbidden cost limit is 5 (value of \att{ub}).

\begin{boxex}
\begin{xcsp}
<instance format="XCSP3"  type="WCSP">
  <variables>
    <var id="x"> 0..3 </var>  
    <var id="y"> 0..3 </var>  
  </variables>
  <constraints @\violet{ub}@="5">
     <extension type="soft" defaultCost="0"> 
       <list> x </list>
       <tuples @\violet{cost}@="1"> 1 3 </tuples>
    </extension>
    <extension type="soft" defaultCost="0"> 
       <list> y </list>
       <tuples @\violet{cost}@="1"> 1 2 </tuples>
    </extension>
    <extension type="soft" defaultCost="0">
      <list> x y </list>
      <tuples @\violet{cost}@="5"> 
        (0,0)(0,1)(1,0)(1,1)(1,2)(2,1)(2,2)(2,3)(3,2)(3,3) 
      </tuples>
    </extension>
  </constraints>
</instance>
 \end{xcsp}
\end{boxex} 

\subsection{FCSP}

A fuzzy constraint network (FCN) is composed of a set of variables and a set of fuzzy constraints.
Each fuzzy constraint represents a fuzzy relation on its scope: it associates a value in $[0,1]$ (either a rational, a decimal or one of the integers 0 and 1), called membership degree, with each constraint tuple $\tau$, indicating to what extent $\tau$ belongs to the relation and therefore satisfies the constraint.
The membership degree of a tuple gives us the preference for that tuple. In fuzzy constraints, preference 1 is the best one and preference 0 the worst one.

In \x3, a fuzzy constraint is systematically represented by an XML element whose attribute \att{type} is set to \val{fuzzy}.
Below, we show how to represent fuzzy constraints, either in extensional form, or in intensional form.

\paragraph{Fuzzy Constraints in Extension.}\label{ctr:fExtension} For representing such a constraint, an element \xml{extension}  with the attribute \att{type} set to \val{fuzzy} is used, containing an element \xml{list} and a sequence of elements \xml{tuples}. Each element \xml{tuples} has a required attribute \att{degree} that gives the common membership degree of all tuples contained inside the element.
It is possible to use the optional attribute \att{defaultDegree} of \xml{extension} to specify the membership degree of all implicit tuples (i.e., those not explicitly listed). 

This gives the following syntax for fuzzy extensional constraints of arity greater than or equal to 2:

\begin{boxsy}
\begin{syntax} 
<extension type="fuzzy" [defaultDegree="number"]>
   <list> (intVar wspace)2+ </list>
   (<tuples degree="number"> ("(" intVal ("," intVal)+ ")")+ </tuples>)+
</extension>
\end{syntax}
\end{boxsy}

An example is given below.

\begin{boxex}
\begin{xcsp}
<extension id="c1" type="fuzzy" defaultDegree="0">
  <list> w x y z </list>
  <tuples degree="0.1"> (1,2,3,4)(2,1,2,3)(3,1,3,4) </tuples>
  <tuples degree="0.6"> (1,1,3,4)(2,2,2,2) </tuples>
</extension>
 \end{xcsp}
\end{boxex}

For a fuzzy unary constraint, we obtain:

\begin{boxsy}
\begin{syntax} 
<extension type="fuzzy" [defaultDegree="number"]>
   <list> intVar </list>
   (<tuples degree="number"> ((intVal | intIntvl) wspace)+ </tuples>)+
</extension>
\end{syntax}
\end{boxsy}

\paragraph{Fuzzy Constraints in Intension.}\label{ctr:fIntension}
For representing such a constraint, we need an element \xml{intension} with the attribute \att{type} set to \val{fuzzy}. 
It contains a functional expression representing a fuzzy relation returning a real value (comprised between 0 and 1).

\begin{boxsy}
\begin{syntax} 
<intension type="fuzzy">
   <function> realExpr </function>
</intension>
\end{syntax}
\end{boxsy}

The opening and closing tags of \xml{function} are optional, which gives:

\begin{boxsy}
\begin{syntax} 
<intension type="fuzzy"> realExpr </intension> @\com{Simplified Form}@
\end{syntax}
\end{boxsy}

An example is given below.

\begin{boxex}
\begin{xcsp}
<intension id="c1" type="fuzzy">
  if(eq(x,y),0.5,0) 
</intension>
 \end{xcsp}
\end{boxex} 

To define a FCSP instance, in \x3, you have to;
\begin{itemize}
\item set the attribute \att{type} of \xml{instance} to \val{FCSP}, 
\item enumerate variables (at least one) within \xml{variables};
\item enumerate fuzzy constraints within \xml{constraints};
\end{itemize}

The syntax is as follows:

\begin{boxsy}
\begin{syntax} 
<instance format="XCSP3" type="FCSP">
  <variables>
    (<var.../> | <array.../>)+ 
  </variables>
  <constraints>
    (<constraint type="fuzzy".../> | <group.../> | <block.../>)* 
  </constraints>
  [<annotations.../>]
</instance>
\end{syntax}
\end{boxsy}

As an illustration, let us consider the \x3 representation of the problem instance described in \cite{F_thesis}, page 110.

\begin{boxex}
\begin{xcsp}
<instance format="XCSP3" type="FCSP">
  <variables>
    <var id="x"> 0..7 </var> 
    <var id="y"> 0..7 </var> 
    <var id="z"> 0..7 </var>
  </variables>
  <constraints>
    <extension id="c0" type="fuzzy" defaultDegree="0.25">
      <list> x </list>
      <tuples degree="1"> 4 </tuples>
      <tuples degree="0.75"> 3 5 </tuples>
    </extension>
    <extension id="c1" type="fuzzy" defaultDegree="0.5">
      <list> y </list>
      <tuples degree="1"> 3 4 </tuples>
    </extension>
    <extension id="c2" type="fuzzy" defaultDegree="0">
      <list> z </list>
      <tuples degree="1"> 2 </tuples>
      <tuples degree="0.75"> 1 3 </tuples>
    </extension>
    <intension id="c3" type="fuzzy">
      if(eq(add(x,y,z),7),1,0) 
    </intension>
  </constraints>
</instance>
\end{xcsp}
\end{boxex}

\section{Dealing with Quantified Variables}

Two frameworks for dealing with quantification of variables have been introduced in the literature.
The former, QCSP$^{+}$, is an extension of CSP, and the latter, QCOP$^{+}$, is an extension of COP.  

\subsection{QCSP($^{+}$)}

The Quantified Constraint Satisfaction Problem (QCSP) is an extension of CSP in which variables may be quantified universally or existentially\footnote{This is a revised version of the XCSP 2.1 proposal made initially for QCSP and QCSP$^{+}$ by M. Benedetti, A. Lallouet and J. Vautard.}. 
A QCSP instance corresponds to a sequence of quantified variables, called prefix, followed by a conjunction of constraints.
QCSP and its semantics were introduced  in \cite{BM_beyond}.
QCSP$^{+}$ is an extension of QCSP, introduced in \cite{BLV_qcsp} to overcome some difficulties that may occur when modeling real problems with classical QCSP.
From a logic viewpoint, an instance of QCSP+ is a formula in which (i) quantification scopes of alternate type are nested one inside the other, (ii) the quantification in each scope is \emph{restricted} by a CSP called \emph{restriction} or \emph{precondition}, and (iii) a CSP to be satisfied, called goal, is attached to the innermost scope.  An example with 4 scopes is:


\begin{eqnarray}
& & \forall X_{1} ~~ (L^{\forall}_{1}(X_{1}) \rightarrow \nonumber\\
& & ~~~~~~~ \exists Y_{1} ~~ ( L^{\exists}_{1}(X_{1},Y_{1}) \wedge \nonumber\\
& & ~~~~~~~ ~~~~~~~ \forall X_{2} ~~ (L^{\forall}_{2}(X_{1},Y_{1},
X_{2}) \rightarrow \nonumber\\
& & ~~~~~~~ ~~~~~~~ ~~~~~~~ \exists Y_{2} ~~ (L^{\exists}_{2}
(X_{1},Y_{1},X_{2},Y_{2}) \wedge ~~ G(X_{1},X_{2},Y_{1},Y_{2})) \nonumber\\
& & ~~~~~~~ ~~~~~~~ ) \nonumber\\
& & ~~~~~~~ ) \nonumber\\
& & ) \label{ex3}
\end{eqnarray}

\noindent where $X_1$, $X_2$, $Y_1$, and $Y_2$ are in general sets of variables, and each $L^{Q}_i$ is a conjunction of constraints.
A more compact and readable syntax for QCSP$^+$ employs square braces to enclose restrictions. An example with 3 scopes is as follows:
$$\small \forall X_1 [L_1^\forall (X_1)]\ \exists Y_1 [L_1^\exists (X_1,Y_1)]\ \forall X_2 [L_2^\forall (X_1,Y_1,X_2)]\ \ G(X_1,Y_1,X_2)$$

\noindent which reads ``for all values of $X_1$ which satisfy the constraints $L^{\forall}_{1}(X_{1})$, there exists a value for $Y_1$ that satisfies $L^{\exists}_{1}(X_{1},Y_{1})$ and is  such that for all values for $X_2$ which satisfy $L^{\forall}_{2}(X_{1},Y_{1},X_{2})$, the goal  $G(X_{1},X_{2},Y_{1})$ is satisfied''. 

A standard QCSP can be viewed as a particular case of QCSP$^{+}$ in which all quantifications are unrestricted, i.e. all the CPSs $L^{Q}_i$ are empty.

\medskip
To define a QCSP/QCSP$^{+}$ instance, in \x3, you have to;
\begin{itemize}
\item set the attribute \att{type} of \xml{instance} to \val{QCSP} or \val{QCSP+};
\item enumerate variables (at least one) within \xml{variables};
\item enumerate constraints within \xml{constraints};
\item handle quantification in an element \xml{quantification}.
\end{itemize}

\paragraph{Handling Quantification.} The element \xml{quantification} provides an ordered list of quantification blocks.
Notice that the order in which quantification blocks are listed inside
this XML element provides key information, as it specifies the
left-to-right order of (restricted) quantifications associated with
the QCSP/QCSP$^{+}$ instance.  Each block of a QCSP/QCSP$^{+}$
instance is represented by either an element \xml{forall} or an element \xml{exists}, which gives the kind of quantification for all the variables in the block.
For QCSP, the list of variables of a block is given directly inside elements \xml{forall} and  \xml{exists}.
However, for QCSP$^{+}$, if restrictions are present, we put the ids of restrictions (restricting constraints) inside an element \xml{ctrs}, and put the list of variables inside an element \xml{vars}; note that restrictions are put with other constraints in \xml{constraints}.
At least one variable must be present in each block, and the order in which variables are listed is not relevant.

\begin{remark}
Each variable can be mentioned in \emph{at most one} block.
Each variable must be mentioned in \emph{at least} one block; this means the problem is \emph{closed}, i.e. no free variables are allowed\footnote{This restriction may be relaxed by future formalizations of open QCSPs.};
\end{remark}

For QCSP, the syntax is as follows:

\begin{boxsy}
\begin{syntax} 
<instance format="XCSP3" type="QCSP">
  <variables>
    (<var.../> | <array.../>)+ 
  </variables>
  <constraints>
    (<constraint.../> | <metaConstraint.../> | <group.../> | <block.../>)* 
  </constraints>
  <quantification>
    (<exists> (intVar wspace)+ <exists> | <forall> (intVar wspace)+ <forall>)+
  </quantification>
  [<annotations.../>]
</instance>
\end{syntax}
\end{boxsy}

Let $w$, $x$, $y$ and $z$ be four variables, whose domains are $\{1,2,3,4\}$.
An \x3 encoding of the QCSP: $$\exists w,x ~ ~ \forall y  ~~ \exists z ~~~ w + x = y + z, ~~y \neq z$$ is given by:

\begin{boxex}
\begin{xcsp}
<instance format="XCSP3" type="QCSP"> 
  <variables> 
    <var id="w"> 1..4 </var>
    <var id="x"> 1..4 </var>
    <var id="y"> 1..4 </var>
    <var id="z"> 1..4 </var>
  </variables> 
  <constraints> 
    <intension> eq(add(w,x),add(y,z)) </intension> 
    <intension> ne(y,z)  </intension> 
  </constraints> 
  <quantification>
    <exists> w x </exists>
    <forall> y </forall>
    <exists> z </exists>    
  </quantification>
</instance> 
\end{xcsp}
\end{boxex} 

Recall that, in \x3, an \bnf{intVar} corresponds to the id of a variable declared in \xml{variables}. Similarly, an \bnf{intCtr} corresponds to the id of a constraint declared in \xml{constraints}.
For QCSP$^{+}$, the syntax is then as follows:

\begin{boxsy}
\begin{syntax} 
<instance format="XCSP3" type="QCSP+">
  <variables>
    (<var.../> | <array.../>)+ 
  </variables>
  <constraints>
    (<constraint.../> | <metaConstraint.../> | <group.../> | <block.../>)* 
  </constraints>
  <quantification>
    ( <exists> 
        <vars> (intVar wspace)+ <vars>
        [<ctrs> (intCtr wspace)+ </ctrs>] 
      </exists>    
    | 
      <forall> 
        <vars> (intVar wspace)+ <vars>
        [<ctrs> (intCtr wspace)+ </ctrs>] 
      <forall>
    )+
  </quantification>
  [<annotations.../>]
</instance>
\end{syntax}
\end{boxsy}

An \x3 encoding of the QCSP$^{+}$: {\small $$\small\exists w,x [w+x< 8, w-x>2] ~\forall y [w \neq y, x \neq y] ~\exists z [z < w-y] ~ w + x = y + z$$} is given by:

\begin{boxex}
\begin{xcsp}
<instance format="XCSP3" type="QCSP+"> 
  <variables> 
    <var id="w"> 1..4 </var>
    <var id="x"> 1..4 </var>
    <var id="y"> 1..4 </var>
    <var id="z"> 1..4 </var>
  </variables> 
  <constraints> 
    <intension id="r1a"> lt(add(w,x),8) </intension> 
    <intension id="r1b"> gt(sub(w,x),2) </intension> 
    <intension id="r2a"> ne(w,y) </intension> 
    <intension id="r2b"> ne(x,y) </intension> 
    <intension id="r3"> gt(sub(w,y),z) </intension>  
    <intension id="goal"> eq(add(w,x),add(y,z)) </intension> 
  </constraints> 
  <quantification>
    <exists>
      <vars> w x </vars> 
      <ctrs> r1a r1b </ctrs> 
    </exists>
    <forall> 
      <vars> y </vars>
      <ctrs> r2a r2b </ctrs> 
    </forall>
    <exists>
      <vars> z </vars> 
      <ctrs> r3 </ctrs>
    </exists> 
  </quantification>
</instance> 
\end{xcsp}
\end{boxex} 

Note here that we have to specify an id for each restricting constraint in order to be able to reference them. 

\subsection{QCOP($^{+}$)}

QCOP($^{+}$), Quantified Constraint Optimization problem, is a framework \cite{BLV_qcop} that allows us to formally express preferences over QCSP($^{+}$) strategies.
A QCOP($^{+}$) instance is obtained from a QCSP($^{+}$) instance by adding preferences and aggregates to the quantification blocks.
For aggregation, we need aggregate ids (which look like local variables but cannot be part of constraints) and aggregate functions.
Possible aggregate functions are:
\begin{itemize}
\item \val{sum}, 
\item \val{product}
\item \val{average}
\item \val{deviation}
\item \val{median}
\item \val{count}
\end{itemize}

Compared to QCSP($^{+}$), we must add information to elements \xml{exists} and \xml{forall} for dealing with optimization.
For an element \xml{exists}, we need an optimization condition, which is represented by an element \xml{minimize} or an element \xml{maximize}, the content of which is a variable id or an aggregate id.
However, when no element \xml{minimize} or \xml{maximize} is present for an element \xml{exists}, this implicitly corresponds to the atom {\em any}  \cite{BLV_qcop}. 
For an element \xml{forall}, we must add a sequence of one or several elements \xml{aggregate}, each of them with a required attribute \att{id} that gives the aggregate identifier.
The content of the element is an expression of the form $f(x)$ where $f$ denotes an aggregate function and $x$ denotes a variable id or an aggregate id.    

To define a QCOP/QCOP$^{+}$ instance, in \x3, you have to:
\begin{itemize}
\item set the attribute \att{type} of \xml{instance} to \val{QCOP} or \val{QCOP+};
\item enumerate variables (at least one) within \xml{variables};
\item enumerate constraints (at least one) within \xml{constraints};
\item handle quantification in an element \xml{quantification}.
\end{itemize}

As an illustration, here is the \x3 representation of the toy problem introduced in \cite{BLV_qcop}, page 472.
We assume here that the array $A$, given as an input to the instance is $\{12,5,6,9,4,3,13,10,12,5\}$

\begin{boxex}
\begin{xcsp}
<instance format="XCSP3" type="QCOP+"> 
  <variables> 
    <var id="x"> 0 1 </var>
    <var id="i"> 0..9 </var>
    <var id="z"> 0..+infinity </var>
  </variables> 
  <constraints> 
    <intension id="res"> eq(mod(i,2),x) </intension> 
    <element id="goal">
      <list> 12 5 6 9 4 3 13 10 12 5 </list>
      <index> i </index> 
      <value> z </value>
    </element>
  </constraints>  
  <quantification>
    <exists>
      <vars> x </vars> 
      <minimize> s </minimize>
    </exists>
    <forall> 
      <vars> i </vars>
      <ctrs> res </ctrs>
      <aggregate id="s"> @sum@(z) </aggregate>
    </forall>
    <exists>
      <vars> z </vars> 
    </exists> 
  </quantification>
</instance> 
\end{xcsp}
\end{boxex}

\section{Stochastic Constraint Reasoning}

Two frameworks have been introduced for dealing with uncontrollable variables: SCSP (Stochastic Constraint Satisfaction Problem) and SCOP (Stochastic Constraint Optimization Problem).

\subsection{SCSP}

To define a SCSP instance, in \x3, you have to:
\begin{itemize}
\item set the attribute \att{type} of \xml{instance} to \val{SCSP};
\item enumerate variables (at least one) within \xml{variables};
\item enumerate constraints (at least one) within \xml{constraints}; either there is an attribute \att{threshold} associated with \xml{constraints} or an attribute \att{threshold} associated with each constraint element present in \xml{constraints}.
\item handle stages with an element \xml{stages}, which alternately contains elements \xml{decision} and \xml{stochastic}; in each of these elements, we find a sequence of variable ids.  
\end{itemize}

The syntax is:

\begin{boxsy}
\begin{syntax} 
<instance format="XCSP3" type="SCSP">
  <variables>
    (<var.../> | <array.../>)+ 
  </variables>
  <constraints [threshold="number"]>
    (<constraint [threshold="number"].../> | <group.../> | <block.../>)* 
  </constraints>
  <stages>
    (<decision> (intVar wspace)+ <decision> | 
     <stochastic> (intVar wspace)+ <stochastic>)+
  </stages>
  [<annotations.../>]
</instance>
\end{syntax}
\end{boxsy}

Here is an illustration taken from \cite{W_stochastic} about modeling a simple $m$
quarter production planning problem. In each quarter, we will sell
between 100 and 105 copies of a book. To keep customers happy,
we want to satisfy demand in all $m$ quarters with 80\% probability.
At the start of each quarter, we decide how many books to print for that quarter. 
There are $m$ decision variables, $x_i$ representing production in each quarter. 
There are also $m$ stochastic variables, $y_i$ representing demand in each quarter. These take values between 100 and 105 with
equal probability. There is a constraint to ensure 1st quarter production meets 1st quarter demand:
$x_1 \geq y_1$.
There is also a constraint to ensure 2nd quarter production meets 2nd
quarter demand plus any unsatisfied demand or less any stock:
$x_2 \geq y_2 +(y_1 - x_1)$.
And there is a constraint to ensure $j$th ($j \geq 2$) quarter production 
meets $j$th quarter demand plus any unsatisfied demand or less any stock:
$x_j \geq y_j + \sum_{i=1}^{j-1}(y_i - x_i)$
We must satisfy these constraints with a threshold probability of $0.8$.

For our illustration, with $m=2$, we obtain:

\begin{boxex}
\begin{xcsp}
<instance format="XCSP3" type="SCSP"> 
  <variables> 
    <var id="x1"> 100..105 </var>
    <var id="x2"> 100..105 </var>
    <var id="y1" type="stochastic"> 100..105:1/6 </var>
    <var id="y2" type="stochastic"> 100..105:1/6 </var>
  </variables> 
  <constraints threshold="0.8">  
    <intension> ge(x1,y1) </intension> 
    <intension> ge(x2,add(y2,sub(y1,x1)) </intension> 
  </constraints> 
  <stages>
    <decision> x1 </decision>
    <stochastic> y1 </stochastic>
    <decision> x2 </decision>
    <stochastic> y2 </stochastic>
  </stages>
</instance> 
\end{xcsp}
\end{boxex}

Suppose now that there is a specific threshold $0.7$ and $0.9$ for the two constraints.
The elements \xml{constraints} become:

\begin{boxex}
\begin{xcsp}
  <constraints>  
    <intension threshold="0.7"> 
      ge(x1,y1) 
    </intension> 
    <intension threshold="0.9"> 
      ge(x2,add(y2,sub(y1,x1)) 
    </intension> 
  </constraints> 
\end{xcsp}
\end{boxex}

\subsection{SCOP}

To define a SCOP instance, in \x3, you have to:
\begin{itemize}
\item set the attribute \att{type} of \xml{instance} to \val{SCOP};
\item enumerate variables (at least one) within \xml{variables};
\item enumerate constraints (at least one) within \xml{constraints};
\item handle stages with an element \xml{stages};
\item enumerate objectives (at least one) within \xml{objectives};
\end{itemize}


\section{Qualitative Spatial Temporal Reasoning}

Qualitative Spatial Temporal Reasoning (QSTR) deals with qualitative calculi.
A qualitative calculus is defined from a finite set $\B$ of base relations on a domain $\D$.  
The elements of $\D$ represent temporal or spatial entities, and the elements of $\B$ represent all possible configurations between two entities. 
$\B$ is a set that satisfies the following properties \cite{LR_what}: $\B$ forms a partition of $\D \times \D$, $\B$ contains the identity relation ${\mathsf{Id}}$, and $\B$ is closed under the converse operation ($^{-1}$). 
A (complex) relation is the union of some base relations, but it is customary to represent a relation as the set of base relations contained in it.
Hence, the set $2^{\B}$ represents the set of relations of the qualitative calculus. 


A QSTR instance is a pair composed of a set of variables and a set of constraints. 
Each variable represents a spatial/temporal entity of the system that is modeled. 
Each constraint represents a set of acceptable qualitative configurations between two variables and is defined by a relation.

\subsection{Interval Calculus}

A well known temporal qualitative formalism is the Interval Algebra, also called Allen's calculus \cite{A_interval}.
The domain $\DINT$ of this calculus is the set $\{(x^-,x^+) \in {\Q \times \Q} : x^-<x^+\}$ since temporal entities are intervals of the rational line.
The set $\BINT$ of this calculus is the set $\{eq,p,pi,m,mi,o,oi,s,si,d,di,f,fi\}$ of thirteen binary relations representing all orderings of the four endpoints of two intervals. 
For example, $m=\{((x^-,x^+),(y^-,y^+)) \in \DINT \times \DINT : x^+=y^- \}$. 


To define a QSTR instance based on the Interval Algebra, in \x3, you have to:
\begin{itemize}
\item set the attribute \att{type} of \xml{instance} to \val{QSTR}, 
\item enumerate interval variables (at least one) within \xml{variables};
\item enumerate interval constraints within \xml{constraints}.
\end{itemize}

The syntax is:

\begin{boxsy}
\begin{syntax} 
<instance format="XCSP3" type="QSTR">
  <variables>
    (<var type="interval".../> | <array type="interval".../>)+ 
  </variables>
  <constraints> <interval.../>* </constraints>
  [<annotations.../>]
</instance>
\end{syntax}
\end{boxsy}

\begin{remark}
Although not indicated above, groups and blocks of interval constraints can also be used.
\end{remark}

\begin{center}
\begin{tikzpicture}[>=stealth]
  \tikzstyle{snode}=[draw,circle,minimum size=6mm,color=dred]
  \tikzstyle{sedge}=[draw,->,>=latex]
  \tikzstyle{slabel}=[midway,scale=1]
  \node[snode] (0) at (-2,2) {$v_0$};
  \node[snode] (1) at (2,2) {$v_1$};
  \node[snode] (2) at (0,0) {$v_2$};
  \node[snode] (3) at (0,-2.2) {$v_3$};
  \path[sedge] (0) edge node[slabel,above]{$\{di,m,s\}$} (1);
  \path[sedge] (1) edge node[slabel,near start,left]{$\{o,oi\}$} (2);
  \path[sedge] (0) edge node[slabel,left]{$\{m,s,fi\}$} (3);
  \path[sedge] (2) edge node[slabel, near start]{$\{d,o,fi\}$} (3);
\end{tikzpicture}
\end{center}

As seen in Chapter \ref{cha:real}, an interval constraint is defined by an element \xml{interval} that contains an element \xml{scope} and an element \xml{relation}. 
An illustration is given for the qualitative constraint network depicted above. 
Its \x3 encoding is given by:

\begin{boxex}
\begin{xcsp}
<instance format="XCSP3" type="QSTR">
  <variables>
    <var id="v0" type="interval"/>
    <var id="v1" type="interval"/>
    <var id="v2" type="interval"/>
    <var id="v3" type="interval"/>
  </variables>   
  <constraints>
    <interval>
      <scope> v0 v1 </scope>
      <relation> di m s </relation>
    </interval> 
    <interval>
      <scope> v0 v3 </scope>
      <relation> m s fi </relation>
    </interval>
    <interval>
      <scope> v1 v2 </scope>
      <relation> o oi </relation>
    </interval>
    <interval>
      <scope> v2 v3 </scope>
      <relation> d o fi </relation>
    </interval>
  </constraints>
</instance>
\end{xcsp}
\end{boxex}



\subsection{Point Calculus}

Point Algebra (PA) is a simple qualitative algebra defined for time points. 

To define a QSTR instance based on the Point Algebra, in \x3, you have to:
\begin{itemize}
\item set the attribute \att{type} of \xml{instance} to \val{QSTR}, 
\item enumerate point variables (at least one) within \xml{variables};
\item enumerate point constraints within \xml{constraints}.
\end{itemize}

The syntax is:

\begin{boxsy}
\begin{syntax} 
<instance format="XCSP3" type="QSTR">
  <variables>
    (<var type="point".../> | <array type="point".../>)+ 
  </variables>
  <constraints> <point.../>* </constraints>
  [<annotations.../>]
</instance>
\end{syntax}
\end{boxsy}

\begin{remark}
Although not indicated above, groups and blocks of point constraints can also be used.
\end{remark}


As seen in Chapter \ref{cha:real}, a point constraint is defined by an element \xml{point} that contains an element \xml{scope} and an element \xml{relation}. 
An illustration is given by:

\begin{boxex}
\begin{xcsp}
<instance format="XCSP3" type="QSTR">
  <variables>
    <var id="x" type="point"/>
    <var id="y" type="point"/>
    <var id="z" type="point"/>
  </variables>   
  <constraints>
    <point>
      <scope> x y </scope>
      <relation> b a </relation>
    </point>
    <point>
      <scope> x z </scope>
      <relation> b eq </relation>
    </point>
    <point>
      <scope> y z </scope>
      <relation> a </relation>
    </point>
  </constraints>
</instance>
\end{xcsp}
\end{boxex}

\subsection{Region Connection Calculus}

RCC8 consists of 8 basic relations that are possible between two regions:

\begin{itemize}
\item disconnected (DC)
\item externally connected (EC)
\item equal (EQ)
\item partially overlapping (PO)
\item tangential proper part (TPP)
\item tangential proper part inverse (TPPi)
\item non-tangential proper part (NTPP)
\item non-tangential proper part inverse (NTPPi)
\end{itemize}

So, as before, to define a QSTR instance based on the Region Connection Calculus 8, in \x3, you have to:
\begin{itemize}
\item set the attribute \att{type} of \xml{instance} to \val{QSTR}, 
\item enumerate region variables (at least one) within \xml{variables};
\item enumerate rcc8 constraints within \xml{constraints}.
\end{itemize}

As seen in Chapter \ref{cha:real}, a rcc8 constraint is defined by an element \xml{region} with attribute \att{type} set to \val{rcc8} and containing an element \xml{scope} and an element \xml{relation}. 



\subsection{TCSP}

A well-known framework for time reasoning is TCSP (Temporal Constraint Satisfaction Problem) \cite{DMP_temporal}.
In this framework, variables represent time points and temporal information is represented by a set of unary and binary constraints, each specifying a set of permitted intervals.

For this framework, as seen in Chapter \ref{cha:real}, we only need to introduce a type of constraints, called \gb{dbd} constraints, for disjunctive binary difference constraints (as in \cite{K_temporal}).
A unary \gb{dbd} constraint on variable $x$ has the form:
\begin{quote}
$a_1 \leq x \leq b_1 \lor \ldots \lor a_p \leq x \leq b_p$
\end{quote}
A binary \gb{dbd} constraint on variables $x$ and $y$ has the form:
\begin{quote}
$a_1 \leq y-x \leq b_1 \lor \ldots \lor a_p \leq y-x \leq b_p$
\end{quote}
where $x$ and $y$ are real variables representing time points and $a_1, \ldots, a_p, b_1, \ldots, b_p$ are real numbers denoting the bounds of $p$ considered intervals.

In \x3, a \gb{dbd} constraint is defined by an element \xml{dbd} that contains an element \xml{scope} and an element \xml{intervals}.
When a scope contains two variables $x$ $y$ in that order, it means that the difference is $y-x$.

To define a TCSP instance, in \x3, you have to;
\begin{itemize}
\item set the attribute \att{type} of \xml{instance} to \val{TCSP}, 
\item enumerate point variables (at least one) within \xml{variables};
\item enumerate dbd constraints within \xml{constraints}.
\end{itemize}

The syntax is:

\begin{boxsy}
\begin{syntax} 
<instance format="XCSP3" type="TCSP">
  <variables>
    (<var type="point".../> | <array type="point".../>)+ 
  </variables>
  <constraints> <dbd.../>* </constraints>
  [<annotations.../>]
</instance>
\end{syntax}
\end{boxsy}

Here is an example taken from \cite{DMP_temporal}.
John goes to work either by car (30-40 minutes) or by bus (at least 60 minutes). 
Fred goes to work either by car (20-30 minutes) or in a car pool (40-50 minutes). 
Today John left home between 7:10 and 7:20, and Fred arrived at work between 8:00 and 8:10. 
We also know that John arrived at work about 10-20 minutes after Fred left home.

Following \cite{K_temporal}, we can introduce the following variables: $x_0$ for a special time point denoting the “beginning of time” (7:00 in our case), $x_1$ for the time at which John left home, $x_2$ for the time at which John arrived at work, $x_3$ for the time at which Fred left home, and $x_4$ for the time at which Fred arrived at work.
We then obtain:

\begin{boxex}
\begin{xcsp}
<instance format="XCSP3" type="TCSP">
  <variables>
    <var id="x0" type="point"/> 
    <var id="x1" type="point"/>
    <var id="x2" type="point"/> 
    <var id="x3" type="point"/> 
    <var id="x4" type="point"/>
  </variables>   
  <constraints>
    <dbd> 
      <scope> x0 x1 </scope>
      <intervals> [10,20] </intervals> 
    </dbd>
    <dbd> 
      <scope> x1 x2 </scope>
      <intervals> [30,40] [60,+infinity[ </intervals> 
    </dbd>
    <dbd> 
      <scope> x3 x4 </scope>
      <intervals> [20,30] [40,50] </intervals> 
    </dbd>
    <dbd> 
      <scope> x3 x2 </scope>
      <intervals> [10,20] </intervals> 
    </dbd>
    <dbd> 
      <scope> x0 x4 </scope>
      <intervals> [60,70] </intervals> 
    </dbd>
  </constraints>
</instance>
\end{xcsp}
\end{boxex}

\section{Continuous Constraint Solving}

\subsection{NCSP}

Numerical CSP instances are defined as CSP instances, but involve real variables and constraints. 

To define a NCSP instance, in \x3, you have to:
\begin{itemize}
\item set the attribute \att{type} of \xml{instance} to \val{NCSP};
\item enumerate variables (at least one) within \xml{variables};
\item enumerate constraints within \xml{constraints};
\end{itemize}

The syntax is as follows:

\begin{boxsy}
\begin{syntax} 
<instance format="XCSP3" type="NCSP">
  <variables>
    (<var type="real".../> | <array type="real".../>)+ 
  </variables>
  <constraints> (<intension.../> | <sum.../>)* </constraints>
  [<annotations.../>]
</instance>
\end{syntax}
\end{boxsy}

\begin{boxex}
\begin{xcsp}
<instance format="XCSP3" type="NCSP"> 
  <variables> 
    <var id="x" type="real"> [-4,4] </var>
    <var id="y" type="real"> [-4,4] </var>
  </variables> 
  <constraints> 
    <intension> eq(sub(y,pow(x,2)),0) </intension> 
    <intension> eq(sub(y,add(x,1)),0) </intension> 
  </constraints> 
</instance> 
\end{xcsp}
\end{boxex}

\subsection{NCOP}

Numerical COP instances are defined as COP instances, but involve real variables and constraints.

To define a NCOP instance, in \x3, you have to:
\begin{itemize}
\item set the attribute \att{type} of \xml{instance} to \val{NCOP}  
\item enumerate variables (at least one) within \xml{variables};
\item enumerate constraints within \xml{constraints};
\item enumerate objectives (at least one) within \xml{objectives}.
\end{itemize}

The syntax is as follows:

\begin{boxsy}
\begin{syntax} 
<instance format="XCSP3" type="NCOP">
  <variables>
    (<var type="real".../> | <array type="real".../>)+ 
  </variables>
  <constraints> (<intension.../> | <sum.../>)* </constraints>
  <objectives [combination="combinationType"]> 
    (<minimize.../> | <maximize.../>)+ 
  </objectives>
  [<annotations.../>]
</instance>
\end{syntax}
\end{boxsy}



\section{Distributed Constraint Reasoning}

Distributed Constraint Reasoning has been studied for both satisfaction and optimization. 

\subsection{DisCSP}

DisCSP (Distributed CSP) instance, is defined as a classical CN $P=(\mathscr X,\mathscr C)$, together with a set of $p$ agents $\mathscr A =\{a_1,a_2,\ldots,a_p\}$.
Each agent $a \in \mathscr A$ controls a proper subset of variables $vars(a)$ of $\mathscr X$, and knows a subset of constraints $ctrs(a)$ of $\mathscr C$.
The sets of controlled variables of all agents form a partition of $\mathscr X$. 
When an agent $a_i$ can directly send messages to another agent $a_j$, $a_j$ is said to be reachable from $a_i$.

To describe the way agents are defined and interact, we add to \xml{instance} an element \xml{agents} that contains a sequence of elements \xml{agent}.
Each element \xml{agent} has an optional attribute \att{id} and contains an element \xml{vars}, an element \xml{ctrs} and a sequence of elements \xml{comm}.
Note that, as described below, some of these intern elements can be omitted in certain cases.
They are defined as follows:

\begin{itemize}
\item The element \xml{vars} contains the list of variables, given by their ids, controlled by the agent.
\item The element \xml{ctrs} contains the list of constraints, given by their ids, known by the agent. 
\item When there is no (special) communication cost between agents, there is only one element \xml{comm} that contains the list of agents, given by their ids, reachable from the agent.
\item When there are specific communication costs between agents, there is a sequence of elements \xml{comm}, each one that contains the list of agents, given by their ids, reachable from the agent with a cost given by an attribute \att{cost}.
\end{itemize}
 
Note that the introduced elements \xml{comm}, taken together, allows us to describe the topology of the communication graph/network.

There are special cases where some elements can be omitted.
It depends on the general configuration of agents, which is characterized by three (optional) attributes associated with the element \xml{agents}. 
The optional attribute \att{varsModel} characterizes the partition of variables between agents. When there is a bijection between $\mathscr X$ and $\mathscr A$, i.e., each agent controls exactly one distinct variable, the value of \att{varsModel} is set to \val{bijection}.
The optional attribute \att{ctrsModel} characterizes the knowledge of agents on constraints. When every agent $a$ exactly knows the set of constraints involving at least one variable controlled by the agent, i.e., $ctrs(a) = \{ c \in \mathscr C \mid scp(c) \cap vars(a) \neq \emptyset\}$, the value of \att{ctrsModel} is set to \val{TKC}, which stands for Totally Known Constraints.
The optional attribute \att{commModel} characterizes the way agents can communicate, i.e. what is the topology of the communication network.
When the communication graph is the same as the constraint (primal) graph, the value of \att{commModel} is set to \val{scope}.

Now, here are the possible simplifications according to the value of these three attributes.
First, for any element \xml{agent}, the sequence of elements \xml{comm} can be empty when there is no (special) communication cost between agents and when the value of the attribute \att{commModel} is \val{scope}, because it is possible to infer the communication graph in that case.
Similarly,  for any element \xml{agent}, the element \xml{ctrs}  can be omitted when the value of \att{ctrsModel} is \val{TKC}.
Finally,  when \att{varsModel} has value \val{bijection}, \att{ctrsModel} has value \val{TKC} and \att{commModel} has value \val{scope}, we can avoid to introduce all intern elements \xml{agent}.

To define a DisCSP instance, in \x3, you have to;
\begin{itemize}
\item set the attribute \att{type} of \xml{instance} to \val{DisCSP}, 
\item enumerate variables (at least one) within \xml{variables};
\item enumerate constraints (at least one) within \xml{constraints}; 
\item enumerate agents within \xml{agents}.
\end{itemize}

An illustration is given below:

\begin{boxex}
\begin{xcsp}
<instance format="XCSP3" type="DisCSP">
  <variables>
    <var id="w"> red green blue </var>
    <var id="x"> red green blue </var>
    <var id="y"> red green blue </var>
    <var id="z"> red green blue </var>
  </variables>   
  <constraints>
    <intension id="c1"> ne(w,x) </intension>
    <intension id="c2"> ne(w,y) </intension>
    <intension id="c3"> ne(w,z) </intension>
    <intension id="c4"> ne(x,z) </intension>
    <intension id="c5"> ne(y,z) </intension>
  </constraints>
  <agents>
    <agent id="a1">
      <vars> w </vars>
      <ctrs> c1 c2 c3 </ctrs>
      <comm> a2 a3 a4 </comm> 
    </agent>
    <agent id="a2">
      <vars> x </vars>
      <ctrs> c1 c4 </ctrs>
      <comm> a1 a4 </comm> 
    </agent>
    <agent id="a3">
      <vars> y </vars>
      <ctrs> c2 c5 </ctrs>
      <comm> a1 a4 </comm> 
    </agent>
    <agent id="a4">
      <vars> z </vars>
      <ctrs> c3 c4 c5 </ctrs>
      <comm> a1 a2 a3 </comm> 
    </agent>
  </agents>
</instance>
\end{xcsp}
\end{boxex}

Interestingly enough, for the element \xml{agents}, we can use the following abridged version:

\begin{boxex}
\begin{xcsp}
    <agents varsModel="bijection" 
            ctrsModel="TKC" 
            commModel="scope" />
\end{xcsp}
\end{boxex}

Let us suppose now that we only consider three agents, one controlling the variables $w$ and $x$, and the two others controlling the variables $y$ and $z$, respectively.
We keep a TKC model, and consider that the communication cost between any pair of agents is $10$, except between the first and third agents, where it is $100$.
We then obtain the following element \xml{agents}:

\begin{boxex}
\begin{xcsp}
  <agents ctrsModel="TKC" commModel="scope">
    <agent id="a1">
      <vars> w x </vars>
      <ctrs> c1 c2 c3 c4 </ctrs>
      <comm @\violet{cost}@="10"> a2 </comm>
      <comm @\violet{cost}@="100"> a3 </comm>
    </agent>
    <agent id="a2">
      <vars> y </vars>
      <ctrs> c2 c5 </ctrs>
      <comm @\violet{cost}@="10"> a1 a3 </comm> 
    </agent>
    <agent id="a3">
      <vars> z </vars>
      <ctrs> c3 c4 c5 </ctrs>
      <comm @\violet{cost}@="100"> a1 </comm>
      <comm @\violet{cost}@="10"> a2 </comm>
    </agent>
  </agents>
\end{xcsp}
\end{boxex}

Because the model is TKC, we could discard all elements \xml{ctrs}.

\subsection{DisWCSP (DCOP)}

A DisWCSP (Distributed WCSP) instance, often called DCOP in the literature, is defined as a classical WCN $P=(\mathscr X,\mathscr C,k)$, together with a set of $p$ agents $\mathscr A =\{a_1,a_2,\ldots,a_p\}$.
The way the agents are defined are exactly as for DisCSP.

To define a DisWCSP instance, in \x3, you have to;
\begin{itemize}
\item set the attribute \att{type} of \xml{instance} to \val{DisWCSP}, 
\item enumerate variables (at least one) within \xml{variables};
\item enumerate weighted constraints (at least one) within \xml{constraints}; 
\item enumerate agents within \xml{agents}.
\end{itemize}
\end{xl}

\chapter{\textcolor{gray!95}{Annotations}}\label{cha:annotations}

\begin{xc}
In \x3, it is possible to insert an element \xml{annotations} to express search and filtering advice.
However, in \x3-core, we currently only consider the possibility of indicating the set of decision variables.

\begin{boxsy}
\begin{syntax} 
<annotations>
  [<decision> (intVar wspace)+ </decision>]
</annotations>
\end{syntax}
\end{boxsy}

One can then indicate the variables that the solver should branch in, so-called branching or decision variables, by means of the element \xml{decision}, inside \xml{annotations}.
\end{xc}
\begin{xl}
In \x3, it is possible to insert an element \xml{annotations} to express search and filtering advice.
In this chapter, we present this while considering integer variables only (i.e., CSP and COP frameworks), but of course, what follows can be directly adapted to other frameworks. 

Recall that, in \x3, an \bnf{intVar} corresponds to the id of a variable declared in \xml{variables}. Similarly, an \bnf{intCtr} corresponds to the id of a constraint declared in \xml{constraints}.
Besides, all BNF non-terminals such as \bnf{filteringType}, \bnf{consistencyType}, \bnf{branchingType} and \bnf{restartsType} are defined in Appendix \ref{cha:bnf}.
The general syntax for \xml{annotations} is as follows:

\begin{boxsy}
\begin{syntax} 
<annotations>
  [<decision> (intVar wspace)+ </decision>]
  [<output> (intVar wspace)+ </output>]
  [<varHeuristic.../>]
  [<valHeuristic.../>]
  (<filtering type="filteringType"> (intCtr wspace)+ </filtering>)*
  [<filtering type="filteringType" />]
  [<prepro consistency="consistencyType" />]
  [<search [consistency="consistencyType"] [branching="branchingType"] />
  [<restarts type="restartsType" cutoff="unsignedInteger" [factor="decimal"] />]
</annotations>
\end{syntax}
\end{boxsy}

\section{Annotations about Variables}

It is possible to associate annotations with variables by introducing a few elements inside \xml{annotations}.
First, one can indicate the variables that the solver should branch in, so-called branching or decision variables, and the variables that are considered relevant when outputting solutions.
This is made possible by introducing an element \xml{decision} and an element \xml{output} inside \xml{annotations}.
For example, below, we identify variables $x_0$, $x_1$ and $x_2$ as being three decision variables, and variables $y$ and $z$ as being those that should be displayed whenever a solution is found. 

\begin{boxex}
\begin{xcsp}
<annotations>
  <decision> x0 x1 x2 </decision>
  <output> y z </output>
</annotations>
\end{xcsp}
\end{boxex}


It may be interesting to indicate which heuristic(s) should be followed by a solver.
This is the role of the elements \xml{varHeuristic} and \xml{valHeuristic}.
For variable ordering, one can introduce several elements inside \xml{varHeuristic}, chosen among \xml{static}, \xml{random}, \xml{min} and \xml{max}.
An element \xml{static} indicates in which exact order variables contained in it must be selected by the solver.
An element \xml{random} indicates that variables contained in it should be selected at random.
An element \xml{min} or \xml{max} has a required attribute \att{type} whose value indicates the type of variable ordering heuristic, based on one or several criteria, that must be followed for the variables it contains.
Basic types, based on elementary criteria, are currently:

\begin{itemize}
\item \val{lexico}. 
\item \val{dom} \cite{HE_FC}. 
\item \val{deg} \cite{U_algorithm}, \val{ddeg} and \val{wdeg} \cite{BHLS_boosting}
\item \val{impact} \cite{G_dual,R_impact}
\item \val{activity} \cite{MV_activity}
\end{itemize}

It is possible to combine such criteria with the operator ``/'', thus considering ratios, and also with the operator ``+'' to determine how to break ties.

We can then use complex types, as for example:

\begin{itemize}
\item \val{dom/deg}, \val{dom/ddeg} and \val{dom/wdeg}
\item \val{dom+ddeg}
\end{itemize}

For example, \val{dom+ddeg} is known as the Brelaz heuristic \cite{B_brelaz,S_brelaz}.

Note that the BNF syntax for all possible values of \att{type} is given by \bnf{varhType} in Appendix \ref{cha:bnf}.
The syntax for \xml{varHeuristic} is then:
\begin{boxsy}
\begin{syntax} 
<varHeuristic [lc="unsignedInteger"]>
  (   <static> (intVar wspace)+ </static>
    | <random> (intVar wspace)+ </random>
    | <min type="varhType"> (intVar wspace)+ </min>
    | <max type="varhType"> (intVar wspace)+ </max>
  )*
  [<random/> | <min type="varhType"/> | <max type="varhType"/>] @\com{default}@
</varHeuristic>
\end{syntax}
\end{boxsy}

As you can see, there may be several elements inside \xml{varHeuristic}, including one in last position that may contain nothing.
Actually, when there are several elements inside \xml{varHeuristic}, it means that the variables of the first element should be selected first (using the semantics of the element), then the variables of the second element should be selected, and so on.
Finally, when there is an element at last position without any content, it means that it applies to all remaining variables (i.e., those not explicitly listed in previous elements): this is the default heuristic.
By means of the attribute \att{lc}, it is also possible to indicate that last conflict reasoning \cite{LSTV_reasonning} should be employed: the value of the attribute indicates the maximal size ($k$) of the testing set.

If the advice is simply to use {\tt min dom/wdeg} over the entire constraint network, you will write:

\begin{boxex}
\begin{xcsp}
<varHeuristic>
  <min type="dom/wdeg"/> 
</varHeuristic>
\end{xcsp}
\end{boxex}

In another example, below, we assume that the solver should select variables $x_0$, $x_7$ and $x_{12}$ first (in that order), then variables $x_{10}$, $x_{20}$ in an order that depends on their domain sizes, and finally all other variables by considering at each decision step the variable with the greatest weighted degree.
On our example, it is also advised to use $lc(2)$.

\begin{boxex}
\begin{xcsp}
<varHeuristic lc="2">
  <static> x0 x7 x12 </static>
  <min type="dom"> x10 x20 </min>
  <max type="wdeg"/> 
</varHeuristic>
\end{xcsp}
\end{boxex}

For value ordering, one can introduce several elements inside \xml{valHeuristic}, chosen among \xml{static}, \xml{random}, \xml{min} and \xml{max}.
An element \xml{static} indicates in which order values for the variables contained in it should be selected by the solver. The order for values is given by the attribute \att{order}.
An element \xml{random} indicates that values for the variables contained in it should be selected randomly.
An element \xml{min} or \xml{max} has a required attribute \att{type} whose value indicates the type of value ordering heuristic that must be followed for the variables it contains.
Currently, possible values of \att{type} (i.e., values of \bnf{valhType} defined in Appendix \ref{cha:bnf}) are:

\begin{itemize}
\item \val{conflicts} 
\item \val{value}. 
\end{itemize}

Here, \nn{conflicts} refers to the number of conflicts with neighbors \cite{MJPL_minimizing,FD_value}.

The syntax for \xml{valHeuristic} is then:
\begin{boxsy}
\begin{syntax} 
<valHeuristic>
  (   <static order="(intVal wspace)+"> (intVar wspace)+ </static>
    | <random> (intVar wspace)+ </random>
    | <min type="valhType"> (intVar wspace)+ </min>
    | <max type="valhType"> (intVar wspace)+ </max>
  )*
  [<random /> | <min type="valhType"/> | <max type="valhType"/>]  @\com{default}@
</valHeuristic>
\end{syntax}
\end{boxsy}

If the advice is simply to use {\tt min value} over the entire constraint network, you will write:

\begin{boxex}
\begin{xcsp}
<valHeuristic>
  <min type="value"/> 
</valHeuristic>
\end{xcsp}
\end{boxex}

Another example is given below.
Whenever $x$ is selected, the solver must choose the first valid value in the order given by \att{order}. 
Whenever $y$ and $z$ are selected, the solver must select the value that has the smallest number of conflicts (with its neighbors).
Finally, whenever any other variable is selected by the solver, the greatest value in the domain of the selected variable must be chosen.

\begin{boxex}
\begin{xcsp}
<valHeuristic>
  <static order = "4 5 3 6 2 1"> x </static>
  <min type="conflicts"> y z </min>
  <max type="value"/>
</valHeuristic>
\end{xcsp}
\end{boxex}

\section{Annotations about Constraints}

It is possible to associate annotations with constraints by introducing a few elements inside \xml{annotations}.
First, one can indicate the level of filtering that the solver should try to enforce on specified constraints by using elements \xml{filtering} with the \att{type} attribute set to the wished value.
Currently, there are five possible values for \att{type}: 
\begin{itemize}
\item \val{boundsZ} for integer bounds propagation. See Bounds(Z) in \cite{CHLS_finite}.
\item \val{boundsD} for a stronger integer bounds propagation. See Bounds(D) in \cite{CHLS_finite}.
\item \val{boundsR} for real bounds propagation. See Bounds(R) in \cite{CHLS_finite}.
\item \val{AC} for (Generalized) Arc Consistency. This value is valid for both binary and non-binary constraints (when it is usually called GAC).
\item \val{FC} for Forward Checking. In that case, the element \xml{filtering} has an optional attribute \att{delay} that indicates the number of unassigned variables under which AC should be enforced (since FC is a partial form of AC). 
\end{itemize}

When an element \xml{filtering} contains nothing, it must be the last element, meaning that it applies to all remaining constraints (i.e., those not explicitly listed in previous elements).

An example is given below.
Here, the solver should enforce Bounds(Z) on constraints $c_0$, $c_1$ and $c_2$, Bounds(D) on constraints $c_3$ and $c_4$ and full (generalized) arc consistency on all other constraints.

\begin{boxex}
\begin{xcsp}
<annotations>
  <filtering type="boundsZ"> c0 c1 c2 </filtering>
  <filtering type="boundsD"> c3 c4 </filtering>
  <filtering type="AC"/> 
</annotations>
\end{xcsp}
\end{boxex}

\section{Annotations about Preprocessing and Search}

In some cases, it may be useful to indicate the overall level of consistency that the solve should enforce at preprocessing (i.e, before search) and/or during search; for this purpose, we introduce elements \xml{prepro} and \xml{search}.
Currently, this level of consistency can be any value among: 
\begin{itemize}
\item \val{FC} for Forward Checking
\item \val{BC} for Bounds Consistency
\item \val{AC} for (Generalized) Arc Consistency (i.e., for both binary and non-binary constraints)
\item \val{SAC} for Singleton Arc Consistency
\item \val{FPWC} for Full Pairwise Consistency
\item \val{PC} for Path Consistency
\item \val{CDC} for Conservative Dual Consistency
\item \val{FDAC} For Full Directional Arc Consistency
\item \val{EDAC} for Existential Directional Arc Consistency
\item \val{VAC} for Virtual Arc Consistency
\end{itemize}

To record this information, we use an attribute \att{consistency} whose value is one among those cited just above.
For search, we can also indicate the kind of branching that the solver should use, by means of the attribute \att{branching} whose value must currently be one among \val{2-way} and \val{d-way}.
Finally, we can give relevant information about restarts.
We just have to introduce an element \xml{restarts} with two attributes \att{type} and \att{cutoff}.
Currently, possible values for \att{type} are:
\begin{itemize}
\item \val{luby}
\item \val{geometric}
\end{itemize}
The attribute \att{cutoff} indicates the number of conflicts that must be encountered before restarting the first time.
When the progression is geometric, there is an additional attribute \att{factor}.

An example is given below.

\begin{boxex}
\begin{xcsp}
<annotations>
  <prepro consistency="SAC"/>
  <search consistency="AC" branching="2-way" />
  <restarts type="geometric" cutoff="10" factor="1.1"/>
</annotations>
\end{xcsp}
\end{boxex}
\end{xl}



\chapter*{Acknowledgments}

This work has been supported by both CNRS and OSEO (BPI France) within the ISI project 'Pajero', and the project CPER Data from the region “Hauts-de-France”.
This work still benefits from the support of the National Research Agency under France 2030, MAIA Project ANR-22-EXES-0009.


\begin{appendices}

\chapter{KeyWords}

\x3 keywords are:
\begin{quote}
\item neg, abs, add, sub, mul, div, mod, sqr, pow, min, max, dist, lt, le, ge, gt, ne, eq, set, in, not, and, or, xor, iff, imp, if, card, union, inter, diff, sdiff, hull, djoint, subset, subseq, supseq, supset, convex, PI, E, fdiv, fmod, sqrt, nroot, exp, ln, log, sin, cos, tan, asin, acos, atan, sinh, cosh, tanh, others 
\end{quote}

\bigskip
\noindent There are some restrictions about the identifiers that can be used in \x3:
\begin{itemize}
\item the set of keywords, the set of symbolic values and the set of \att{id} values (i.e., values of attributes \att{id}) must all be disjoint, 
\item it is not permitted to have two attributes \att{id} with the same value,
\item it is not permitted to have an attribute \att{id} of an element \xml{array} that corresponds to a prefix of any other attribute \att{id}, 
\item it is not permitted to have an attribute \att{id} of an element \xml{group} that corresponds to a prefix of any other attribute \att{id}. 
\end{itemize}

Recall that characters ``[`` and ``]'' are not allowed in identifiers.
If you need to build a sub-network (new file) with a selection of variables (from arrays) and constraints (from groups),
it is recommended to adopt the following usage: replace each occurrence of the form ``[i]'' by ``\_i''. 
For example, \verb!x[0][3]! becomes \verb!x_0_3! and \verb!g[0]! becomes \verb!g_0!. 


\chapter{Syntax}\label{cha:bnf}

The syntax given in this document combines two languages:

\begin{itemize}
\item XML for describing the main elements and attributes of \x3 (high level description)
\item BNF for describing the textual contents of \x3 elements and attributes (low level description)  
\end{itemize}

At some places, we need to postpone the description of some XML elements.
In that case, we just employ \norX{elt} to stand for an XML element of name \verb!elt!, whose description is given elsewhere.
This roughly corresponds to the notion of XML non-terminal.
For example, because, as seen below, \verb!|!, parentheses and \verb!+! respectively stand for alternation, grouping and repetition (at least one), for indicating that the element \xml{variables} can contain some elements \xml{var} and \xml{array}, we just write: 

\begin{syntax} 
<variables>
  (<var.../> | <array.../>)+ 
</variables>
\end{syntax}

The ``XML non-terminals'' \norX{var} and \norX{array} must then be precisely described at another place.
For example, for \norX{var}, we can use the following piece of \x3 syntax, where one can observe that we have an XML element \xml{var} with two attributes \att{id} and \att{type} (optional).
The value of the attribute \att{id} as well as the content of the element \xml{var} is defined using BNF, with BNF non-terminals written in dark blue italic form. 
The non-terminals, referred to all along the document, are defined in this section.

\begin{syntax} 
<var id="identifier" [type="@integer@"]>
  ((intVal | intIntvl) wspace)*
</var>
\end{syntax}

In many situations (as for our example above), the content of an XML text-only element corresponds to a list of basic data (values, variables, ...) with whitespace as separator.
In such situations, the whitespace that follows the last object of the list must always be considered as optional.

More generally, the following rule applies for \x3:

\begin{remark}
In \x3, leading and trailing whitespace are tolerated, but not required, in any XML text-only element.
\end{remark}

We use a variant of Backus-Naur Form (BNF) defined as follows:

\begin{itemize}
\item a non-terminal definition is preceded by \verb!::=!
\item terminals are between quotation marks as e.g., \verb!"CSP"!
\item non-terminals are composed of alphanumeric characters (and the character "\_"), and so, do not contain any white space character, as e.g., \verb!frameworkType!. When such non-terminals are mixed with XML notation (in the document), they are written in dark blue italic form as e.g., \bnf{frameworkType}
\item concatenation is given by any non-empty sequence of white space characters
\item alternatives are separated by a vertical bar as, e.g., \verb!"+" | "-"! 
\item square brackets are used for surrounding optional elements, as, e.g., \verb!["+" | "-"]! 
\item an element followed by * can occur 0 or any number of times
\item an element followed by + must occur at least 1 time 
\item an element followed by n+ must occur at least n times
\item parentheses are used for grouping (often used with *, + and n+)
\end{itemize}

The syntax is given for fully expanded \x3 code, meaning that compact lists of array variables (such as $x[]$) are not handled.

\begin{xl}
The basic BNF non-terminals used in this document are:

\begin{verbatim}
wspace ::= (" " | "\t" | "\n" | "\r")+

digit ::= "0".."9" 
unsignedInteger  ::=  digit+
integer ::= ["+" | "-"] unsignedInteger 
decimal ::= integer "." unsignedInteger 
rational ::= integer "/" unsignedInteger 
number ::= integer | decimal | rational 
   
boolean ::= "false" | "true" 
intSet ::= "set(" [integer ("," integer)*] ")"

letter ::= "a".."z" | "A".."Z"  
identifier ::= letter (letter | digit | "_" )*
\end{verbatim}

The basic data type are (note that we include many aliases so as to facilitate readability):

\begin{verbatim}
indexing ::= ("[" unsignedInteger "]")+
variable ::= identifier [indexing] 

01Var ::= variable
intVar ::= variable
symVar ::= variable
realVar ::= variable
setVar ::= variable
graphVar ::= variable
qualVar ::= variable

intVal ::= integer
realVal ::= number
intIntvl ::= (integer | "-infinity") ".." (integer | "+infinity")
realIntvl ::= ("]" | "[") (number | "-infinity") "," 
   (number | "+infinity") ("]" | "[")
setVal ::= "{" [integer ("," integer)*] "}"
proba ::= 
    unsignedInteger "." unsignedInteger 
  | unsignedInteger "/" unsignedInteger 
  | 0 | 1

operator ::= "lt" | "le" | "ge" | "gt" | "ne" | "eq" | "in" | "notin" 
operand  ::= intVal | intVar | intIntvl | intSet

intValShort ::= intVal | "*" 
intValCompressed ::= intVal | "*" | setVal

state ::= identifier
symbol ::= identifier
intCtr ::= identifier

dimensions ::= indexing      
\end{verbatim}

Some types are defined as:

\begin{verbatim}
frameworkType ::= 
    "CSP" | "COP" | "WCSP" | "FCSP" | "QCSP" | "QCSP+" | "QCOP" | "QCOP+" 
  | "SCSP" | "SCOP" | "QSTR" | "TCSP" 
  | "NCSP" | "NCOP" | "DisCSP" | "DisWCSP" 

varType ::= 
     "integer" | "symbolic" | "real" | "set" | "symbolic set" 
  |  "undirected graph" | "directed graph" | "stochastic" 
  |  "symbolic stochastic" | "point" | "interval" | "region"

orderedType ::= 
     "increasing" | "strictlyIncreasing" 
  |  "decreasing" | "strictlyDecreasing" 

rankType ::= "any" | "first" | "last"

iaBaseRelation ::= 
     "eq" | "p" | "pi" | "m" | "mi" | "o" | "oi" 
  |  "s" | "si" | "d" | "di" | "f" | "fi"

paBaseRelation ::= "b" | "eq" | "a"

rcc8BaseRelation ::= 
     "dc" | "ec" | "eq" | "po" 
  | "tpp" | "tppi" | "nttp" | "nttpi"

measureType ::= "var" | "dec" | "val" | "edit"

combinationType ::= "lexico" | "pareto"

blockType ::=
    "clues" | "symmetry-breaking" | "redundant-constraints" | "nogoods"

varhType ::=
    "lexico" | "dom" | "deg" | "ddeg" | "wdeg" | "immpact" | "activity"
  | varhType "/" varhType 
  | varhType "+" varhType

valhType ::= "conflicts" | "value"

filteringType ::= "boundsZ" | "boundsD" | "boundsR" | "AC" | "FC"

consistencyType ::=
    "FC" | "BC" | "AC" | "SAC" | "FPWC" | "PC" | "CDC" | "FDAC" | "EDAC" | "VAC"

branchingType ::= "2-way" | "d-way" 

restartsType ::= "luby" | "geometric" 
\end{verbatim}

The grammar used to build predicate expressions with {\bf integer variables} is:


\begin{verbatim}
   intExpr ::= 
       boolExpr | intVar | integer 
     | "neg(" intExpr ")" 
     | "abs(" intExpr ") 
     | "add(" intExpr ("," intExpr)+ ")"   
     | "sub(" intExpr "," intExpr ")" 
     | "mul(" intExpr ("," intExpr)+ ")"
     | "div(" intExpr "," intExpr ")" 
     | "mod(" intExpr "," intExpr ")"
     | "sqr(" intExpr ")" 
     | "pow(" intExpr "," intExpr ")"
     | "min(" intExpr ("," intExpr)+ ")" 
     | "max(" intExpr ("," intExpr)+ ")" 
     | "dist(" intExpr "," intExpr ")"
     | "if(" boolExpr "," intExpr "," intExpr ")"
       
   boolExpr ::=
       01Var 
     | "lt(" intExpr "," intExpr ")"
     | "le(" intExpr "," intExpr ")"
     | "ge(" intExpr "," intExpr ")"
     | "gt(" intExpr "," intExpr ")"
     | "ne(" intExpr "," intExpr ")"
     | "eq(" intExpr ("," intExpr)+ ")"
     | "in(" intExpr "," intSet ")"
     | "not(" boolExpr ")"   
     | "and(" boolExpr ("," boolExpr)+ ")" 
     | "or(" boolExpr ("," boolExpr)+ ")" 
     | "xor(" boolExpr ("," boolExpr)+ ")" 
     | "iff(" boolExpr ("," boolExpr)+ ")" 
     | "imp(" boolExpr "," boolExpr ")"
\end{verbatim}

The grammar used to build predicate expressions with {\bf integer variables and set variables} is:

\begin{verbatim}
   intExprSet ::= 
        01Var | intVar | integer 
     | "neg(" intExprSet ")" 
     | "abs(" intExprSet ") 
     | "add(" intExprSet ("," intExprSet)+ ")"   
     | "sub(" intExprSet "," intExprSet ")" 
     | "mul(" intExprSet ("," intExprSet)+ ")"
     | "div(" intExprSet "," intExprSet ")" 
     | "mod(" intExprSet "," intExprSet ")"
     | "sqr(" intExprSet ")" 
     | "pow(" intExprSet "," intExprSet ")"
     | "min(" intExprSet ("," intExprSet)+ ")" 
     | "max(" intExprSet ("," intExprSet)+ ")" 
     | "dist(" intExprSet "," intExprSet ")"
     | "if(" boolExprSet "," intExprSet "," intExprSet ")"
     | "card(" setExpr ")" 
     | "min(" setExpr ")" 
     | "max(" setExpr ")"

   setExpr ::= 
       setVar | "set(" [intExprSet ("," intExprSet)*] ")"
     | "union(" setExpr ("," setExpr)+ ")" 
     | "inter(" setExpr ("," setExpr)+ ")" 
     | "diff(" setExpr "," setExpr ")" 
     | "sdiff(" setExpr ("," setExpr)+ ")" 
     | "hull(" setExpr ")" 
 
   boolExprSet ::=
       01Var 
     | "lt(" intExprSet "," intExprSet ")"
     | "le(" intExprSet "," intExprSet ")"
     | "ge(" intExprSet "," intExprSet ")"
     | "gt(" intExprSet "," intExprSet ")"
     | "ne(" intExprSet "," intExprSet ")"
     | "eq(" intExprSet ("," intExprSet)+ ")"
     | "in(" intExprSet "," setExpr ")"  
     | "ne(" setExpr "," setExpr ")" 
     | "eq(" setExpr ("," setExpr)+ ")"
     | "not(" boolExprSet ")"   
     | "and(" boolExprSet ("," boolExprSet)+ ")" 
     | "or(" boolExprSet ("," boolExprSet)+ ")" 
     | "xor(" boolExprSet ("," boolExprSet)+ ")" 
     | "iff(" boolExprSet ("," boolExprSet)+ ")" 
     | "imp(" boolExprSet "," boolExprSet ")"
     | "djoint(" setExpr "," setExpr ")"
     | "subset(" setExpr "," setExpr ")"
     | "subseq(" setExpr "," setExpr ")"
     | "supseq(" setExpr "," setExpr ")" 
     | "supset(" setExpr "," setExpr ")"
     | "convex(" setExpr ")" 
\end{verbatim}

The grammar used to build predicate expressions with {\bf integer variables and real variables} is:

\begin{verbatim}
  realExpr ::= 
       01Var | intVar | realVar | number | PI | E
     | "neg(" realExpr ")" 
     | "abs(" realExpr ") 
     | "add(" realExpr ("," realExpr)+ ")"   
     | "sub(" realExpr "," realExpr ")" 
     | "mul(" realExpr ("," realExpr)+ ")"
     | "fdiv(" realExpr "," realExpr ")" 
     | "fmod(" realExpr "," realExpr ")"
     | "min(" realExpr ("," realExpr)+ ")" 
     | "max(" realExpr ("," realExpr)+ ")" 
     | "sqr(" realExpr ")" 
     | "pow(" realExpr "," realExpr ")"
     | "sqrt(" realExpr ")"
     | "nroot(" realExpr "," integer ")"
     | "exp(" realExpr ")" 
     | "ln(" realExpr ")" 
     | "log(" realExpr "," integer ")"
     | "sin(" realExpr ")" 
     | "cos(" realExpr ")" 
     | "tan(" realExpr ")" 
     | "asin(" realExpr ")" 
     | "acos(" realExpr ")" 
     | "atan(" realExpr ")" 
     | "sinh(" realExpr ")" 
     | "cosh(" realExpr ")"  
     | "tanh(" realExpr ")" 
     | "if(" boolExprReal "," realExpr "," realExpr ")"
       
   boolExprReal ::=
       01Var 
     | "lt(" realExpr "," realExpr ")"
     | "le(" realExpr "," realExpr ")"
     | "ge(" realExpr "," realExpr ")"
     | "gt(" realExpr "," realExpr ")"
     | "ne(" realExpr "," realExpr ")"
     | "eq(" realExpr ("," realExpr)+ ")"
     | "not(" boolExprReal ")"   
     | "and(" boolExprReal ("," boolExprReal)+ ")" 
     | "or(" boolExprReal ("," boolExprReal)+ ")" 
     | "xor(" boolExprReal ("," boolExprReal)+ ")" 
     | "iff(" boolExprReal ("," boolExprReal)+ ")" 
     | "imp(" boolExprReal "," boolExprReal ")"
\end{verbatim}

Finally, we need some non-terminal tokens to be used for XML elements.

\begin{verbatim}
constraint ::=
    "extension" | "intension" 
  | "regular" | "grammar" | "mdd" 
  | "allDifferent" | "allEqual" | "allDistant" | "ordered" | "allIncomparable"
  | "sum" | "count" | "nValues" | "cardinality" | "balance" | "spread" | "deviation" | 
    "sumCosts" | "sequence" 
  | "maximum" | "minimum" | | "maximumArg" | "minimumArg"
  | "element" | "channel" | "precedence" | "permutation"  
  | "stretch" | "noOverlap" | "cumulative" | "binPacking" | "knapsack" | "flow"
  | "circuit" | "nCircuits" | "path" | "nPaths" | "tree" | "nTrees" 
  | "clause" | "instantiation" 
  | "allIntersecting" | "range" | "roots" | "partition"
  | "arbo" | "nArbos" | "nCliques"
  | "adhoc"

metaConstraint ::=
  "slide" | "seqbin" | "and" | "or" | "not" | "ifThen" | "ifThenElse"  
\end{verbatim}
\end{xl}
\begin{xc}
The basic BNF non-terminals used in this document are:

\begin{verbatim}
wspace ::= (" " | "\t" | "\n" | "\r")+

digit ::= "0".."9" 
unsignedInteger  ::=  digit+
integer ::= ["+" | "-"] unsignedInteger 
number ::= integer 
   
boolean ::= "false" | "true" 
intSet ::= "set(" [integer ("," integer)*] ")"

letter ::= "a".."z" | "A".."Z"  
identifier ::= letter (letter | digit | "_" )*
\end{verbatim}

The basic data type are (note that we include many aliases so as to facilitate readability):

\begin{verbatim}
indexing ::= ("[" unsignedInteger "]")+
variable ::= identifier [indexing] 

01Var ::= variable
intVar ::= variable
symVar ::= variable

intVal ::= integer
intIntvl ::= (integer) ".." (integer)
setVal ::= "{" [integer ("," integer)*] "}"

operator ::= "lt" | "le" | "ge" | "gt" | "ne" | "eq" | "in" | "notin" 
operand  ::= intVal | intVar | intIntvl | intSet

intValShort ::= intVal | "*" 
intValCompressed ::= intVal | "*" | setVal

state ::= identifier
symbol ::= identifier
intCtr ::= identifier

dimensions ::= indexing      
\end{verbatim}

Some types are defined as:

\begin{verbatim}
frameworkType ::= 
    "CSP" | "COP" 

varType ::= 
     "integer" | "symbolic" 

orderedType ::= 
     "increasing" | "strictlyIncreasing" 
  |  "decreasing" | "strictlyDecreasing" 

blockType ::=
    "clues" | "symmetry-breaking" | "redundant-constraints" | "nogoods"

varhType ::=
    "lexico" | "dom" | "deg" | "ddeg" | "wdeg" | "immpact" | "activity"
  | varhType "/" varhType 
  | varhType "+" varhType

valhType ::= "conflicts" | "value"

filteringType ::= "boundsZ" | "boundsD" | "boundsR" | "AC" | "FC"

consistencyType ::=
    "FC" | "BC" | "AC" | "SAC" | "FPWC" | "PC" | "CDC" | "FDAC" | "EDAC" | "VAC"

branchingType ::= "2-way" | "d-way" 

restartsType ::= "luby" | "geometric" 
\end{verbatim}

The grammar used to build predicate expressions with {\bf integer variables} is:


\begin{verbatim}
   intExpr ::= 
       boolExpr | intVar | integer 
     | "neg(" intExpr ")" 
     | "abs(" intExpr ") 
     | "add(" intExpr ("," intExpr)+ ")"   
     | "sub(" intExpr "," intExpr ")" 
     | "mul(" intExpr ("," intExpr)+ ")"
     | "div(" intExpr "," intExpr ")" 
     | "mod(" intExpr "," intExpr ")"
     | "sqr(" intExpr ")" 
     | "pow(" intExpr "," intExpr ")"
     | "min(" intExpr ("," intExpr)+ ")" 
     | "max(" intExpr ("," intExpr)+ ")" 
     | "dist(" intExpr "," intExpr ")"
     | "if(" boolExpr "," intExpr "," intExpr ")"
       
   boolExpr ::=
       01Var 
     | "lt(" intExpr "," intExpr ")"
     | "le(" intExpr "," intExpr ")"
     | "ge(" intExpr "," intExpr ")"
     | "gt(" intExpr "," intExpr ")"
     | "ne(" intExpr "," intExpr ")"
     | "eq(" intExpr ("," intExpr)+ ")"
     | "in(" intExpr "," intSet ")"
     | "not(" boolExpr ")"   
     | "and(" boolExpr ("," boolExpr)+ ")" 
     | "or(" boolExpr ("," boolExpr)+ ")" 
     | "xor(" boolExpr ("," boolExpr)+ ")" 
     | "iff(" boolExpr ("," boolExpr)+ ")" 
     | "imp(" boolExpr "," boolExpr ")"
\end{verbatim}

Finally, we need some non-terminal tokens to be used for XML elements.

\begin{verbatim}
constraint ::=
    "extension" | "intension" 
  | "regular" | | "mdd" 
  | "allDifferent" | "allEqual" | "ordered" 
  | "sum" | "count" | "nValues" | "cardinality" 
  | "maximum" | "minimum" | | "maximumArg" | "minimumArg"
  | "element" | "channel" | "precedence" 
  | "noOverlap" | "cumulative" 
  | "circuit" 
  | "instantiation" 

metaConstraint ::=
  "slide" 
\end{verbatim}
\end{xc}

\chapter{Index of Constraints}\label{cha:indctr}

\begin{xl}
\begin{center}
\begin{longtable}{cccc}
  \rowcolor{v2lgray}{} Constraint &  Type & Page  & Remark \\
\gb{adhoc} & & \pageref{ctr:adhoc} & Arbitrary constraint \\
\gb{allDifferent} & Integer & \pageref{ctr:allDifferent} & \\
\gb{allDifferent\ti list}  & Integer & \pageref{ctr:allDifferentList} & \\
\gb{allDifferent\ti set}  & Integer & \pageref{ctr:allDifferentSet} & \\
\gb{allDifferent\ti mset}  & Integer & \pageref{ctr:allDifferentMset} & \\
\gb{allDifferent\ti matrix} & Integer & \pageref{ctr:allDifferentMatrix} & \\
\gb{allDifferent} & Set & \pageref{ctr:allDifferentSet} & \\
\gb{allDifferent$\tr$symmetric} & Integer & \pageref{ctr:allDifferentSymmetric} & restriction \\
\gb{allDistant} & Integer & \pageref{ctr:allDistant} & \\
\gb{allDistant\ti list}  & Integer & \pageref{ctr:allDistantList} & \\
\gb{allDistant\ti set}  & Integer & \pageref{ctr:allDistantSet} & \\
\gb{allDistant\ti mset}  & Integer & \pageref{ctr:allDistantMset} & \\
\gb{allEqual} & Integer & \pageref{ctr:allEqual} & \\
\gb{allEqual\ti list}  & Integer & \pageref{ctr:allEqualList} & \\
\gb{allEqual\ti set}  & Integer & \pageref{ctr:allEqualSet} & \\
\gb{allEqual\ti mset}  & Integer & \pageref{ctr:allEqualMset} & \\
\gb{allEqual} & Set & \pageref{ctr:allEqualSet} & \\
\gb{allIncomparable\ti list}  & Integer & \pageref{ctr:allIncomparableList} & \\
\gb{allIncomparable\ti set}  & Integer & \pageref{ctr:allIncomparableSet} & \\
\gb{allIncomparable\ti mset}  & Integer & \pageref{ctr:allIncomparableMset} & \\
\gb{allIntersecting} & Set & \pageref{ctr:allIntersecting} & \\
\gb{among} & Integer & see \gb{count} & \\
\gb{and} & Integer & \pageref{ctr:and} & meta \\
\gb{arbo} & Graph & \pageref{ctr:arbo} & \\
\gb{atLeast} & Integer & see \gb{count} & \\
\gb{atMost} & Integer & see \gb{count} & \\
\gb{balance} & Integer & \pageref{ctr:balance} & \\
\gb{binPacking} & Integer & \pageref{ctr:binPacking} & \\
\gb{cardinality} & Integer & \pageref{ctr:cardinality} & \\
\gb{cardinality\ti matrix} & Set & \pageref{ctr:cardinalityMatrix} & \\
\gb{cardinalityWithCosts} & Integer & \pageref{ctr:gcccosts} & construction \\
\gb{cardPath}  & Integer & see \gb{soft\ti slide}  &  meta \\ 
\gb{channel} & Integer & \pageref{ctr:channel} & \\
\gb{channel} & Set & \pageref{ctr:channelSet} & \\
\gb{change}  & Integer & see \gb{soft\ti slide}  &  \\ 
\gb{circuit} & Integer & \pageref{ctr:circuit} & \\
\gb{circuit} & Graph & \pageref{ctr:circuitGraph} & \\
\gb{clause} & Integer & \pageref{ctr:clause} & \\
\gb{costRegular} & Integer & \pageref{ctr:costRegular} & construction \\
\gb{count} & Integer & \pageref{ctr:count} & \\
\gb{count} & Set & \pageref{ctr:countSet} & \\
\gb{cube} & Integer & \pageref{ctr:cube} & \\
\gb{cumulative} & Integer & \pageref{ctr:cumulative} & \\
\gb{cumulatives} & Integer & see \gb{cumulative} & \\
\gb{dbd} & Qualitative & \pageref{ctr:dbd} & \\
\gb{decreasing} & Integer & see \gb{ordered} & \\
\gb{deviation} & Integer & \pageref{ctr:deviation} & \\
\gb{diffn} & Integer & see \gb{noOverlap} & \\
\gb{disjunctive} & Integer & see \gb{noOverlap} & \\
\gb{distribute} & Integer & see \gb{cardinality} & \\
\gb{element} & Integer & \pageref{ctr:element} & \\
\gb{element\ti matrix}  & Integer & \pageref{ctr:elementMatrix} & \\
\gb{element} & Set & \pageref{ctr:elementSet} & \\
\gb{exactly} & Integer & see \gb{count} & \\
\gb{extension} & Integer & \pageref{ctr:extension} & \\
hybrid \gb{extension}  & Integer & see \pageref{ctr:hybrid}  &  \\
\gb{flow} & Integer & \pageref{ctr:flow} & \\
\gb{fuzzy\ti extension} & Integer & \pageref{ctr:fExtension}  &  \\
\gb{fuzzy\ti intension} & Integer & \pageref{ctr:fIntension}  &  \\
\gb{gcc} & Integer & see \gb{cardinality} & \\
\gb{globalCardinality} & Integer & see \gb{cardinality} & \\
\gb{grammar} & Integer & \pageref{ctr:grammar} & \\
\gb{iff} & Integer & \pageref{ctr:iff} & meta \\
\gb{ifThen} & Integer & \pageref{ctr:ifThen} & meta \\
\gb{ifThenElse} & Integer & \pageref{ctr:ifThenElse} & meta \\
\gb{increasing} & Integer & see \gb{ordered} & \\
\gb{instantiation} & Integer & \pageref{ctr:instantiation} & \\
\gb{intension} & Integer & \pageref{ctr:intension} & \\
\gb{intension} & Real & \pageref{ctr:intensionReal} & \\
\gb{intension} & Set & \pageref{ctr:intensionSet} & \\
\gb{interDistance} & Integer & see \gb{allDistant} & \\
\gb{interval} & Qualitative & \pageref{ctr:interval} & \\
\gb{knapsack} & Integer & \pageref{ctr:knapsack} & \\
\gb{lex}  & Integer & see \pageref{ctr:orderedList} & \\
\gb{lex2}  & Integer & see \gb{ordered\ti matrix} & \\
\gb{linear} & Integer & see \gb{sum} & \\
\gb{linear} & Real & see \gb{sum} & \\
\gb{maximum} & Integer & \pageref{ctr:maximum} & \\
\gb{maximumArg} & Integer & \pageref{ctr:maximumArg} & \\
\gb{mdd} & Integer & \pageref{ctr:mdd} & \\
\gb{member} & Integer & see \gb{element} & \\
\gb{minimum} & Integer & \pageref{ctr:minimum} & \\
\gb{minimumArg} & Integer & \pageref{ctr:minimumArg} & \\
\gb{nArbos} & Graph & \pageref{ctr:nArbos} & \\
\gb{nCircuits} & Integer & \pageref{ctr:nCircuits} & \\
\gb{nCircuits} & Graph & \pageref{ctr:nCircuitsGraph} & \\
\gb{nCliques} & Graph & \pageref{ctr:nCliques} & \\
\gb{noOverlap} & Integer & \pageref{ctr:noOverlap} & \\
\gb{not} & Integer & \pageref{ctr:not} & meta \\
\gb{notAllEqual} & Integer & \pageref{ctr:notAllEqual} & construction \\
\gb{nPaths} & Integer & \pageref{ctr:nPaths} & \\
\gb{nPaths} & Graph & \pageref{ctr:nPathsGraph} & \\
\gb{nTrees} & Integer & \pageref{ctr:nTrees} & \\
\gb{nValues} & Integer & \pageref{ctr:nValues} & \\
\gb{nValues\ti list}  & Integer & \pageref{ctr:nValuesList} & \\
\gb{nValues\ti set}  & Integer & \pageref{ctr:nValuesSet} & \\
\gb{nValues\ti mset}  & Integer & \pageref{ctr:nValuesMset} & \\
\gb{nValues$\tr$increasing} & Integer & \pageref{ctr:nValuesIncreasing} & restriction \\
\gb{or} & Integer & \pageref{ctr:or} & meta \\
\gb{ordered} & Integer & \pageref{ctr:ordered} & \\
\gb{ordered\ti list}  & Integer & \pageref{ctr:orderedList} & \\
\gb{ordered\ti set}  & Integer & \pageref{ctr:orderedSet} & \\
\gb{ordered\ti mset}  & Integer & \pageref{ctr:orderedMset} & \\
\gb{ordered\ti matrix}  & Integer & \pageref{ctr:orderedMatrix} & \\
\gb{ordered} & Set & \pageref{ctr:orderedSet} & \\
\gb{partition} & Set & \pageref{ctr:partition} & \\
\gb{path} & Integer & \pageref{ctr:path} & \\
\gb{path} & Graph & \pageref{ctr:pathGraph} & \\
\gb{permutation} & Integer & \pageref{ctr:permutation} & \\
\gb{permutation$\tr$increasing} & Integer & \pageref{ctr:sort} & restriction \\
\gb{point} & Qualitative & \pageref{ctr:point} & \\
\gb{precedence} & Integer & \pageref{ctr:precedence} & \\
\gb{precedence} & Set & \pageref{ctr:precedenceSet} & \\
\gb{range} & Set & \pageref{ctr:range} & \\
\gb{rcc8} & Qualitative & \pageref{ctr:rcc8} & \\
\gb{regular} & Integer & \pageref{ctr:regular} & \\
\gb{roots} & Set & \pageref{ctr:roots} & \\
\gb{same}  & Integer & see \gb{soft\ti permutation}  &  \\ 
\gb{seqbin} & Integer & \pageref{ctr:seqbin} & meta \\
\gb{sequence} & Integer & \pageref{ctr:sequence} & \\
\gb{slide} & Integer & \pageref{ctr:slide} & meta \\
\gb{slidingSum} & Integer & \pageref{ctr:slidingSum} & construction \\
\gb{smooth}  & Integer & see \gb{soft\ti slide}  &  \\ 
\gb{soft\ti allDifferent} & Integer & \pageref{ctr:softAllDifferent}  &  \\
\gb{soft\ti and}  & Integer & \pageref{ctr:softAnd}  &  meta \\ 
\gb{soft\ti cardinality} & Integer & \pageref{ctr:softCardinality}  &  \\
\gb{soft\ti extension} & Integer & \pageref{ctr:softExtension}  &  \\
\gb{soft\ti intension} & Integer & \pageref{ctr:softIntension}  &  \\
\gb{soft\ti permutation} & Integer & \pageref{ctr:softPermutation}  &  \\
\gb{soft\ti regular} & Integer & \pageref{ctr:softRegular}  &  \\
\gb{soft\ti slide}  & Integer & \pageref{ctr:softSlide}  &  meta \\ 
\gb{softSlidingSum}  & Integer & see \gb{soft\ti slide}  &  \\ 
\gb{sort} & Integer & see \gb{permutation$\tr$increasing} & restriction \\
\gb{spread} & Integer & \pageref{ctr:spread} & \\
\gb{stretch} & Integer & \pageref{ctr:stretch} & \\
\gb{strictlyDecreasing} & Integer & see \gb{ordered} & \\
\gb{strictlyIncreasing} & Integer & see \gb{ordered} & \\
\gb{sum} & Integer & \pageref{ctr:sum} & \\
\gb{sum} & Real & \pageref{ctr:sumReal} & \\
\gb{sum} & Set & \pageref{ctr:sumSet} & \\
\gb{sumCosts} & Integer & \pageref{ctr:sumCosts} & \\
\gb{table} & Integer & see \gb{extension} & \\
\gb{tree} & Integer & \pageref{ctr:tree} & \\
\gb{weighted\ti allDifferent} & Integer & \pageref{ctr:softAllDifferent}   &  \\
\gb{weighted\ti cardinality} & Integer & \pageref{ctr:softCardinality}  &  \\
\gb{weighted\ti extension} & Integer & \pageref{ctr:wExtension}  &  \\
\gb{weighted\ti intension} & Integer & \pageref{ctr:wIntension}  &  \\
\gb{weighted\ti permutation} & Integer & \pageref{ctr:softPermutation}   &  \\
\gb{weighted\ti regular} & Integer &  \pageref{ctr:softRegular}  &  \\
\gb{xor} & Integer & \pageref{ctr:xor} & meta \\
\end{longtable}
\end{center}
\end{xl}

\begin{xc}
\begin{center}
\begin{longtable}{cccc}
\rowcolor{v2lgray}{} Constraint &  Type & Page  & Remark \\
\gb{allDifferent} & Integer & \pageref{ctr:allDifferent} & \\
\gb{allDifferent\ti list}  & Integer & \pageref{ctr:allDifferentList} & \\
\gb{allDifferent\ti matrix} & Integer & \pageref{ctr:allDifferentMatrix} & \\
\gb{allEqual} & Integer & \pageref{ctr:allEqual} & \\
\gb{among} & Integer & see \gb{count} & \\
\gb{atLeast} & Integer & see \gb{count} & \\
\gb{atMost} & Integer & see \gb{count} & \\
\gb{cardinality} & Integer & \pageref{ctr:cardinality} & \\
\gb{channel} & Integer & \pageref{ctr:channel} & \\
\gb{circuit} & Integer & \pageref{ctr:circuit} & \\
\gb{count} & Integer & \pageref{ctr:count} & \\
\gb{cumulative} & Integer & \pageref{ctr:cumulative} & \\
\gb{decreasing} & Integer & see \gb{ordered} & \\
\gb{diffn} & Integer & see \gb{noOverlap} & \\
\gb{disjunctive} & Integer & see \gb{noOverlap} & \\
\gb{distribute} & Integer & see \gb{cardinality} & \\
\gb{element} & Integer & \pageref{ctr:element} & \\
\gb{element\ti matrix}  & Integer & \pageref{ctr:elementMatrix} & \\
\gb{exactly} & Integer & see \gb{count} & \\
\gb{extension} & Integer & \pageref{ctr:extension} & \\
\gb{gcc} & Integer & see \gb{cardinality} & \\
\gb{globalCardinality} & Integer & see \gb{cardinality} & \\
\gb{increasing} & Integer & see \gb{ordered} & \\
\gb{instantiation} & Integer & \pageref{ctr:instantiation} & \\
\gb{intension} & Integer & \pageref{ctr:intension} & \\
\gb{lex}  & Integer & see \pageref{ctr:orderedList} & \\
\gb{lex2}  & Integer & see \gb{ordered\ti matrix} & \\
\gb{linear} & Integer & see \gb{sum} & \\
\gb{maximum} & Integer & \pageref{ctr:maximum} & \\
\gb{mdd} & Integer & \pageref{ctr:mdd} & \\
\gb{member} & Integer & see \gb{element} & \\
\gb{minimum} & Integer & \pageref{ctr:minimum} & \\
\gb{noOverlap} & Integer & \pageref{ctr:noOverlap} & \\
\gb{nValues} & Integer & \pageref{ctr:nValues} & \\
\gb{ordered} & Integer & \pageref{ctr:ordered} & \\
\gb{ordered\ti list}  & Integer & \pageref{ctr:orderedList} & \\
\gb{ordered\ti matrix}  & Integer & \pageref{ctr:orderedMatrix} & \\
\gb{precedence} & Integer & \pageref{ctr:precedence} & \\
\gb{regular} & Integer & \pageref{ctr:regular} & \\
\gb{slide} & Integer & \pageref{ctr:slide} & meta \\
\gb{strictlyDecreasing} & Integer & see \gb{ordered} & \\
\gb{strictlyIncreasing} & Integer & see \gb{ordered} & \\
\gb{sum} & Integer & \pageref{ctr:sum} & \\
\gb{table} & Integer & see \gb{extension} & \\
\end{longtable}
\end{center}
\end{xc}

\begin{xl}
\chapter{XML and JSON}\label{cha:json}

JSON (JavaScript Object Notation) is another popular language-independent data format. 
We show in this appendix how we can translate \x3 instances from XML to JSON, without loosing any information.
However, note that although JSON allows us to build objects with duplicate keys, JSON decoders might handle those objects differently; we discuss this important issue at the end of this chapter.
We basically adopt and adapt the rules proposed by O'Reilly (\href{https://www.xml.com}{www.xml.com}).

\begin{itemize}
\item Each XML attribute is represented as a name/value property. The name of the property is the name of the attribute preceded by @, and the value of the property is the value of the attribute.
\item Each XML element is represented as a name/value property. The name of the property is the name of the element, and the value of the property is defined as follows:
\begin{itemize}
\item If the element has no attribute and no content, the value is ``null''.
\item If the element has no attribute and only a (non-empty) textual content, the value is the textual content.
\item In all other cases, the value is an object with each XML attribute and each child element represented as a property of the object. Note that a child text node is represented by a property with "\#text" as name and the text as value, except for the following special cases:
\begin{itemize}
\item if the parent node name is ``var'' or ``array'', then the name of the property (for the child text node) is ``domain'',  
\item if the parent node name is ``intension'', then the name of the property (for the child text node) is ``function'',  
\item if the parent node name is ``minimize'' or ``maximize'', then the name of the property (for the child text node) is either ``expression'' (if the objective has a functional form) or ``list'' (if the objective has a specialized form); see forms of objectives in Chapter \ref{cha:objectives}, 
\item if the parent node name is ``allDifferent'', ``allEqual'', ``ordered'', ``channel'', ``circuit'', ``clause'', ``cube'', or ``allIntersecting'', the name of the property (for the child text node) is ``list''.
\end{itemize}
\item Refining the previous rule, any sequence of XML elements of same names, is represented as a property whose name is the name of the elements and the value an array containing the values of these elements in sequence.
\item The order of elements must be preserved.
\item The root element ``instance'' is not represented. We directly give the representation of the content of this element (together with its attributes). So, for a CSP instance, we obtain something like: 
\begin{simplex}
\begin{json}
{
  !\jsn{"@format"}!: "XCSP3",
  !\jsn{"@type"}!: "CSP",
  ... 
}
\end{json}
\end{simplex}
instead of:
\begin{simplex}
\begin{json}
{
  !\jsn{"instance"}!: {
    !\jsn{"@format"}!: "XCSP3",
    !\jsn{"@type"}!: "CSP",
    ... 
  }
}
\end{json}
\end{simplex}

\end{itemize}
\end{itemize}

An illustration of these rules (special cases are not illustrated) is given below:

\begin{simplex}
\begin{small}
\begin{json}
<!\jsn{elt}!/>                     !\jsn{"elt"}!: null  

<!\jsn{elt}!>                      !\jsn{"elt"}!: "text" 
  text
</!\jsn{elt}!>     
                     
<!\jsn{elt}!                       !\jsn{"elt"}!: {
  att="val"                  !\jsn{"@att"}!: "val"
/>                         }

<!\jsn{elt}!                       !\jsn{"elt"}!: {
  att1="val1"                !\jsn{"@att1"}!: "val1", 
  att2="val2"                !\jsn{"@att2"}!: "val2"
/>                         }

<!\jsn{elt}! att="val">            !\jsn{"elt"}!: {
  text                        !\jsn{"@att"}!: "val",
</!\jsn{elt}!>                        !\jsn{"\#text"}!: "text"
                           }

<!\jsn{elt}!>                      !\jsn{"elt"}!: {
  <!\jsn{a}!> text1 </!\jsn{a}!>               !\jsn{"a"}!: "text1",
  <!\jsn{b}!> text2 </!\jsn{b}!>               !\jsn{"b"}!: "text2"
</!\jsn{elt}!>                     }

<!\jsn{elt}!>                      !\jsn{"elt"}!: {
  <!\jsn{a}!> text1 </!\jsn{a}!>               !\jsn{"a"}!: [                                  
  <!\jsn{a}!> text2 </!\jsn{a}!>                  "text1",       
</!\jsn{elt}!>                            "text2"
                               ]
                           }

<!\jsn{elt}!>                      !\jsn{"elt"}!: {
  <!\jsn{a}!> text1 </!\jsn{a}!>               !\jsn{"a"}!: [
  <!\jsn{a}! att="val"> text2 </!\jsn{a}!>       "text1",    
  <!\jsn{b}!> text3 </!\jsn{b}!>                 { !\jsn{"@att"}!: "val", !\jsn{"\#text"}!: "text2" }  
  <!\jsn{a}!> text4 </!\jsn{a}!>               ],
<!\jsn{/elt}!>                         !\jsn{"b"}!: "text3",
                               !\jsn{"a"}!: "text4"
                           }   
\end{json}
\end{small}
\end{simplex}

An XSLT-based converter will be made available soon on our website.

\begin{remark}
JSON is an attractive format. However, in our context of representing CP instances, the reader must be aware of {\bf two possible limitations}, described below.
\end{remark}

First, although it is theoretically valid to build objects with duplicate keys, JSON parsers may behave differently. For example, 
\begin{simplex}
\begin{json}
JSON.parse('{"a": "texta", "b" :"textb", "a": "texta2"}')
\end{json}
\end{simplex}
yields with some parsers:
\begin{simplex}
\begin{json}
{ !\jsn{a}!: "texta2", !\jsn{b}!: "textb" } 
\end{json}
\end{simplex}

Just imagine what you can obtain with the following constraints:
\begin{simplex}
\begin{xcsp}
<constraints>
  <intension> eq(x,0) </intension>
  <extension> 
    <list> t[0] t[1] </list>
    <supports> (2,4)(3,5) </supports>
  </extension>
  <intension> le(y,z) </intension>
</constraints>
\end{xcsp}
\end{simplex}
translated in JSON as:
\begin{simplex}
\begin{json}
!\jsn{"constraints"}!: {
  !\jsn{"intension"}!: "eq(x,0)",
  !\jsn{"extension"}!: {
    !\jsn{"list"}!: "t[0] t[1]",
    !\jsn{"supports"}!: "(2,4)(3,5)"
  },
  !\jsn{"intension"}!: "le(y,z)"
}
\end{json}
\end{simplex}
If the parser you use has not the appropriate behavior for our purpose, you have to transform the JSON file, putting in arrays the contents of similar keys. In our example, this would give:
\begin{simplex}
\begin{json}
!\jsn{"constraints"}!: {
  !\jsn{"intension"}!: ["eq(x,0)", "le(y,z)"],
  !\jsn{"extension"}!: {
    !\jsn{"list"}!: "t[0] t[1]",
    !\jsn{"supports"}!: "(2,4)(3,5)"
  }
}
\end{json}
\end{simplex}

Note that the initial order of constraints cannot be preserved.
Besides, for some frameworks (e.g., SCSP, QCSP), is not possible to proceed that way. 
We have to define other transformation rules for cases like:

\begin{simplex}
\begin{xcsp}
<quantification >
  <exists > w x </exists>
  <forall > y </forall>
  <exists > z </exists>
</quantification>
\end{xcsp}
\end{simplex}
\end{xl}

\chapter{Changelog}\label{cha:versioning}

\begin{xl}
  \begin{itemize}
  \item \x3, version 3.2. Published on August 28, 2024.
    The constraint \gb{lex} has been generalized, allowing the presence of constant lists (tuples); see Section \ref{ctr:orderedList}.
     The constraint \gb{noOverlap} has been slightly modified (letting the possibility of combining integers and variables when specifying lengths); see Section \ref{ctr:precedence}.
    The constraint \gb{allDifferent\ti matrix}, now accepts an element \xml{except}; see Section \ref{ctr:allDifferentMatrix}.
    The meta-constraints \gb{xor} and \gb{iff} have been added; see Section \ref{ctr:xor} and \ref{ctr:iff}.
    Remark \ref{rem:divneg} indicates that integer division with possibly negative operand(s) is strongly discouraged.
    The constraint \gb{adhoc} has been introduced; see Section \ref{ctr:adhoc}.
     
  \item \x3, version 3.1. Published on November 10, 2022.  The constraint \gb{networkFlow} is renamed as \gb{flow}.
    The constraint \gb{smart} has been removed because hybrid (smart) forms of the constraint \gb{extension} are more directly built by simply extending the way tables can be written; see Section \ref{ctr:hybrid}. 
    The constraint \gb{precedence} has been slightly modified; see Section \ref{ctr:precedence}.
    The constraint \gb{allEqual}, as well as its lifted variants on lists, sets and multisets now accepts an element \xml{except}.
    The constraint \gb{knapsack} has been generalized, replacing the parameter \xml{limit} by a parameter \xml{condition}; see Section \ref{ctr:knapsack}.
    Two news forms have been defined for the constraint \gb{binPacking}, based on the parameters \xml{limits} and \xml{loads}; see Section \ref{ctr:binPacking}.
    Besides, \gb{precedence}, \gb{knapsack} and \gb{binPacking} (except its fourth general form) belongs now to \x3-core.
    Variables (instead of integers) can be used in the element \xml{coeffs} of an objective.
    Integer expressions can now be used in the element \xml{coeffs} of a constraint \gb{sum} or in the element \xml{coeffs} of an objective.
    The 'args' forms of \gb{maximum} and \gb{minimum} become specific constraints \gb{maximumArg} and \gb{minimumArg}. 
    Additional places where compact forms of integer sequences can be used are indicated at the end of Chapter \ref{cha:variables}.
    
\item \x3, version 3.0.7. Published on January 18, 2021. The syntax of the constraints \gb{element} and \gb{element\ti matrix} has been generalized: it is now possible to use an element \xml{condition} (instead of \xml{value}).
  It is now possible to use compact forms of integer sequences (in elements \xml{values} of \xml{instantiation} and \xml{coeffs} of \xml{sum}) by writting $v$x$k$ when the integer $v$ occurs $k$ times in sequence.

\item \x3, version 3.0.6. Published on September 1, 2020.
  Licence is specified in Page 2. The variant of \gb{element\ti matrix} where the specified matrix contains integer values has been introduced; see Section \ref{ctr:orderedMatrix}.
Generalized forms (views) are made more explicit for some important cases. The type ``product'' of specialized objectives is deprecated (because, we don't know any useful case); it has been discarded from the text in Chapter \ref{cha:objectives}.
  
\item \x3, version 3.0.5. Published on November 25, 2017.
  The general introduction has been slightly modified; see Section \ref{sec:chain}.
  \x3-core is slightly modified: \gb{circuit} enters \x3-core whereas \gb{stretch} leaves \x3-core; see Section \ref{sec:core}.
  The constraint \gb{ordered} admits now an optional element specifying the minimum distance between two successive variables of the main list; see Section \ref{ctr:ordered}. 
  The default value of the attribute \att{closed} used by the constraint \gb{cardinality} is now \val{false}, instead of \val{true}, see Section \ref{ctr:cardinality}.
  A useful variant of the constraint \gb{element}, where a list of values is handled, has been added; see Section \ref{ctr:element}. 
  A variant of the constraint \gb{channel}, involving two lists of variables of different size, has been added; see Section \ref{ctr:channel}. 
  The meta-constraints \gb{ifThen} and \gb{ifThenElse} have been added; see Section \ref{ctr:ifThen} and \ref{ctr:ifThenElse}. 
  The constraint \gb{smart} has been added.
  Forms of \gb{allDifferent} and \gb{sum}, related to the concept of views, are illustrated at the end of Section \ref{ctr:allDifferent} and \ref{ctr:sum}; these forms can be handled by our tools (modeling API, and parsers).
  Minor simplifications have been performed on annotations (see Chapter \ref{cha:annotations}); currently, decision variables can be handled by our tools (modeling API, and parsers).
  
\item \x3, version 3.0.4. Published on August 20, 2016.
A few modifications about soft constraints have been achieved: see Chapter \ref{cha:cost} and Section \ref{sec:wcsp}.
Simple relaxation is now possible, simplified use of cost variable is authorized, and we have a better integration between relaxed constraints and cost functions.
Constraint \gb{sum} now admits variable coefficients.
The semantics of constraints \gb{circuit}, \gb{path} and \gb{tree} have been slighlty modified.
\item \x3, version 3.0.3. Published on May 20, 2016.
Solutions can now be represented; see Section \ref{sec:solutions}.
Related to solutions, the constraint \gb{instantiation} is introduced; see Section \ref{ctr:instantiation}. It generalizes and replaces the constraint \gb{cube}. 
\x3-core has been slightly updated: it involves now 20 constraints (including the constraint \gb{instantiation} and the meta-constraint \gb{slide}); see Section \ref{sec:core}.
More examples are given in Section \ref{sec:skeleton}, when presenting the skeleton of \x3 instances.
A problem with the semantics of \gb{slide} has been fixed; see Section \ref{ctr:slide}.
\item \x3, version 3.0.2. Published on February 5, 2016. 
\x3 is an intermediate format, as explained in Section \ref{sec:features}.
More information about ``XML versus JSON'' is given in Appendix \ref{cha:json} as well as in Section \ref{sec:json} of Chapter \ref{cha:intro}; limitations of JSON are identified.
New universal attributes are introduced: \att{note} for associating a short comment with any element, and \att{class} for associating either predefined or user-defined tags with any element; this is discussed in Sections \ref{sec:notes} and \ref{sec:blocks}.
\item \x3, version 3.0.1. Published on October 27, 2015. 
Converting \x3 instances from XML to JSON is discussed in a new Appendix as well as in Section \ref{sec:json} of Chapter \ref{cha:intro}.
\x3-core is introduced in Section \ref{sec:core} of Chapter \ref{cha:intro}.
\item \x3, version 3.0.0. First official version. Published on September 1, 2015. 
\item \x3, Release Candidate.  Published on June 22, 2015.
\end{itemize}
\end{xl}

\begin{xc}
  \begin{itemize}
  \item \x3-core, version 3.2.Published on August 28, 2024.
    The constraint \gb{lex} has been generalized, allowing the presence of constant lists (tuples); see Section \ref{ctr:orderedList}.
    The constraint \gb{noOverlap} has been slightly modified (letting the possibility of combining integers and variables when specifying lengths); see Section \ref{ctr:precedence}.
    The constraint \gb{allDifferent\ti matrix}, now accepts an element \xml{except}; see Section \ref{ctr:allDifferentMatrix}.
    Remark \ref{rem:divneg} indicates that integer division with possibly negative operand(s) is strongly discouraged.

    \item \x3-core, version 3.1. Published on November 10, 2022.  
    The constraint \gb{precedence} has been slightly modified.
    The constraint \gb{allEqual}, as well as its lifted variants on lists, sets and multisets now accepts an element \xml{except}.
    Two news forms have been defined for the constraint \gb{binPacking}, based on the parameters \xml{limits} and \xml{loads}.
    Besides, \gb{precedence}, \gb{knapsack} and \gb{binPacking} (except its fourth general form) belongs now to \x3-core.
    Variables (instead of integers) can be used in the element \xml{coeffs} of an objective.
    Integer expressions can now be used in the element \xml{coeffs} of a constraint \gb{sum} or in the element \xml{coeffs} of an objective.
    Additional places where compact forms of integer sequences can be used are indicated at the end of Chapter \ref{cha:variables}.

 \item \x3-core, version 3.0.7. Published on January 18, 2021. The syntax of the constraints \gb{element} and \gb{element\ti matrix} has been generalized: it is now possible to use an element \xml{condition} (instead of \xml{value}).
  It is now possible to use compact forms of integer sequences (in elements \xml{values} of \xml{instantiation} and \xml{coeffs} of \xml{sum}) by writting $v$x$k$ when the integer $v$ occurs $k$ times in sequence.
\item \x3-core, version 3.0.6. Published on September 1, 2020. First version of the document restricted to \x3-core. For changes in previous versions, see Changelog in the main document, i.e., the one giving the specifications of (full) \x3. 
\end{itemize}
\end{xc}

\end{appendices}


\end{document}